\documentclass[Afour,sageh,times]{packages/sagej}
\usepackage{float}

\usepackage{amssymb}% http://ctan.org/pkg/amssymb
\usepackage{pifont}% http://ctan.org/pkg/pifont
\newcommand{\cmark}{\ding{51}}%
\newcommand{\xmark}{\ding{55}}%

\usepackage{textcomp}
\usepackage{xspace}
\usepackage{moreverb,url}
\usepackage{multirow}
\usepackage{indentfirst}
\usepackage{soul}
\usepackage{graphicx}
\usepackage{mathtools}
\usepackage{makecell}
\usepackage[colorlinks,bookmarksopen,bookmarksnumbered,citecolor=red,urlcolor=red]{hyperref}
\usepackage{amsthm}

\usepackage{subcaption} % for subfigure environment
\usepackage{booktabs}

\usepackage[table]{xcolor}

\newcommand{\rl}[1]{{\textcolor{red}{#1}}}

\definecolor{gl}{HTML}{008000}
\newcommand{\gl}[1]{{\textcolor{gl}{#1}}}

\newcommand{\ie}{i.e.,\xspace}
\newcommand{\eg}{e.g.,\xspace}

\usepackage{booktabs}
\makeatletter
\renewcommand\@seccntformat[1]{\csname the#1\endcsname. \;}
\makeatother

\newcommand{\figref}[1]{Fig.~\ref{#1}}

\newcommand\BibTeX{{\rmfamily B\kern-.05em \textsc{i\kern-.025em b}\kern-.08em
T\kern-.1667em\lower.7ex\hbox{E}\kern-.125emX}}

\def\volumeyear{2016}

\begin{document}

\runninghead{Kim et al.}

\title{Commerge: Communication-Efficient, Robust and Fast LiDAR Map Merging Framework for Multi-Robot Coordination in Resource-Constrained Scenarios}

\author{Hogyun Kim\affilnum{1}, Jiwon Choi\affilnum{1}, Juwon Kim\affilnum{1}, Geonmo Yang\affilnum{1}, Seokhwan Jeong\affilnum{1}, \\ Hyungtae Lim\affilnum{2$\dagger$}, and Younggun Cho\affilnum{1$\dagger$}}

\affiliation{\affilnum{1}Dept. of Electrical and Computer Engineering, INHA University, S.Korea \\
\affilnum{2}Laboratory for Information and Decision Systems (LIDS), Massachusetts Institute of Technology, Cambridge, MA, USA}

\corrauth{Younggun Cho, Hyungtae Lim}
\email{yg.cho@inha.ac.kr, shapelim@mit.edu}

\begin{abstract}
By maintaining global consistency across robot teams, multi-robot light detection and ranging~(LiDAR) map merging enables faster exploration and efficient area coverage.
However, map merging requires exchanging massive sensor data between the server and robots, making communication the bottleneck, especially in communication-constrained environments. 
Therefore, we present \textit{Commerge}, a communication-efficient map merging framework that achieves bandwidth reduction through graph-theoretic selective data exchange.
By doing so, our Commerge reduces inter-robot communication by up to 5,000$\times$ while maintaining alignment accuracy.
Our key insight is that only a small subset of carefully selected scans is sufficient for robust map merging. 
We formulate this as a three-stage cascaded optimization problem on an exchange graph, where vertices represent robot keyframes and edges denote candidate inter-robot loops.
Through three cascade stages, we select a sequentially overlapped, balanced-transmission-cost, and geometrically-perceptually optimal scan subset that preserves alignment quality while reducing communication.
Unlike existing approaches that either transmit whole scans, which require GB-scale data exchange, or employ na\"ive downsampling, our approach exchanges only MB-scale data while achieving comparable alignment accuracy.
Extensive evaluation on five public datasets and four in-house datasets covering cave, planetary-analog, indoor, and outdoor campus environments shows up to 99.98\% reduction in data exchange~(\eg from 7,000\,MB to 1.3\,MB on the HeLiPR dataset), while maintaining alignment performance across embedded to desktop platforms. 
Furthermore, we demonstrate that our proposed framework operates on resource-constrained hardware where existing methods fail, enabling real-world multi-robot deployment in bandwidth-limited field operations.
Finally, we validate system robustness under a real-world non-line-of-sight~(NLOS) communication environment with up to 73.3\% dropout rate, as well as under emulated network degradation scenarios including delay, bandwidth limitation, and packet loss, confirming reliable map merging even under severely degraded connectivity.
The supplementary materials are available at \url{https://sparolab.github.io/research/commerge}.
\end{abstract}

\keywords{Communication, Multi-Robot, Map Merging, Data Exchange, Efficiency, Graph Theory}

\maketitle

\newcommand{\robot}{\raisebox{-0.6ex}{\includegraphics[height=2.4ex]{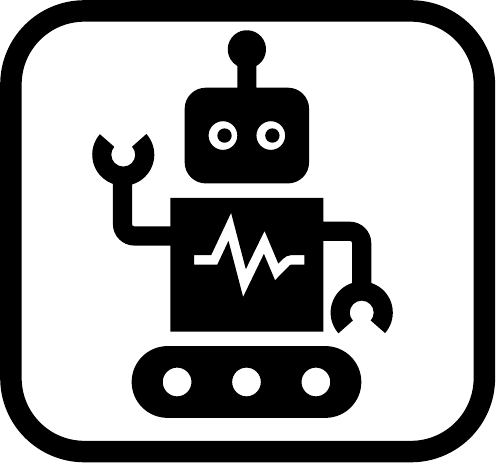}}}
\newcommand{\server}{\raisebox{-0.6ex}{\includegraphics[height=2.4ex]{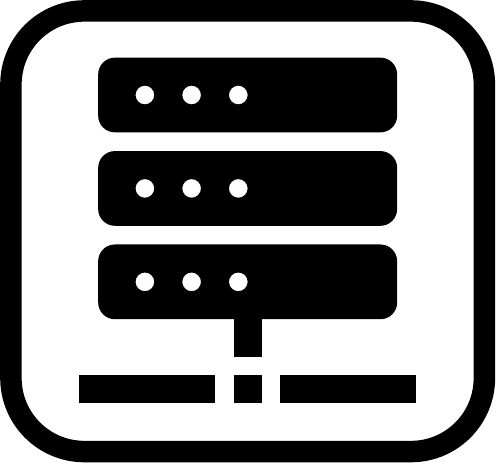}}}
\newcommand{\wifi}{\raisebox{-0.6ex}{\includegraphics[height=2.4ex]{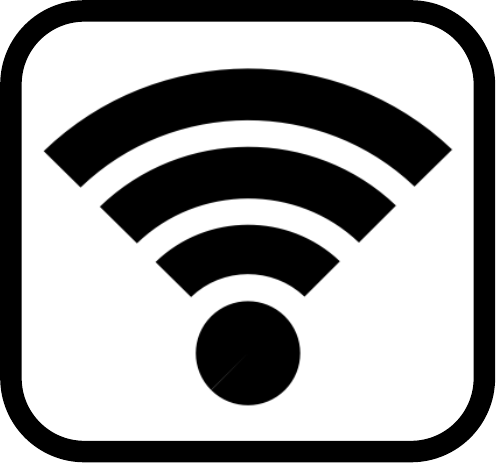}}}
\newcommand{\falpha}{\raisebox{-0.4ex}{\includegraphics[height=2.2ex]{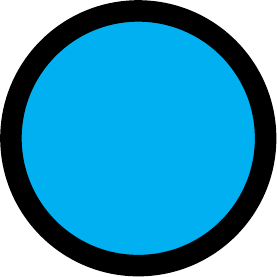}}}
\newcommand{\fbeta}{\raisebox{-0.4ex}{\includegraphics[height=2.2ex]{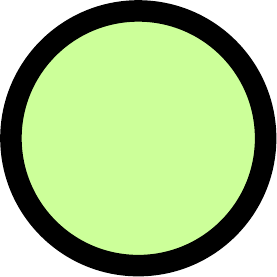}}}
\newcommand{\fedge}{\raisebox{-0.4ex}{\includegraphics[height=2.0ex]{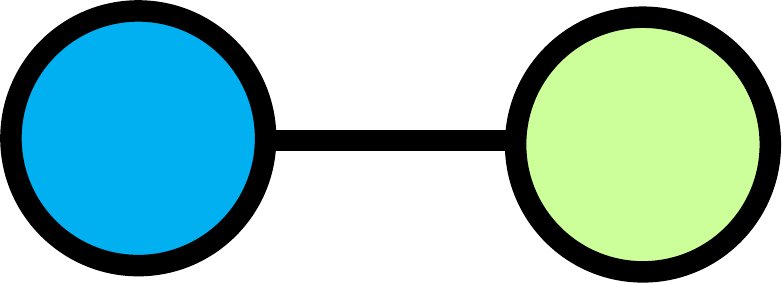}}}
\newcommand{\sedge}{\raisebox{-0.8ex}{\includegraphics[height=2.8ex]{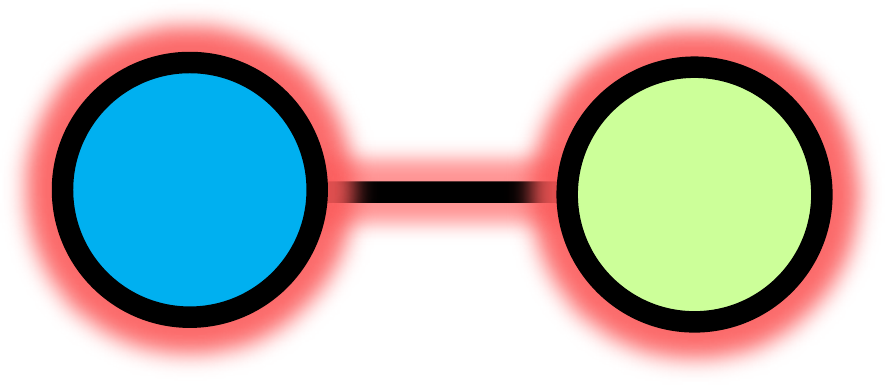}}}

\newcommand{\spin}{\raisebox{-0.6ex}{\includegraphics[height=2.4ex]{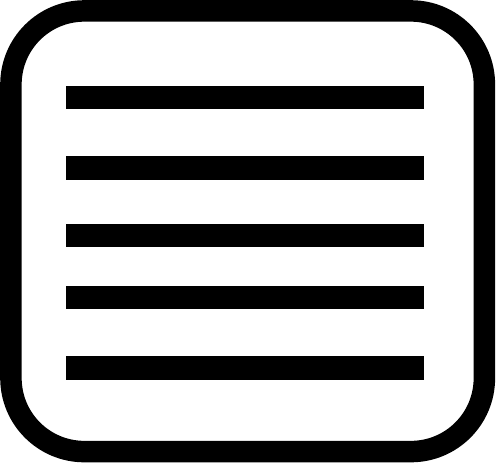}}}
\newcommand{\livox}{\raisebox{-0.6ex}{\includegraphics[height=2.4ex]{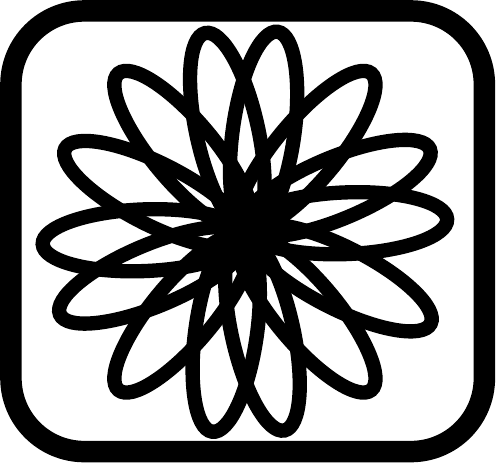}}}

\newcommand{\complete}{\raisebox{-0.6ex}{\includegraphics[height=2.4ex]{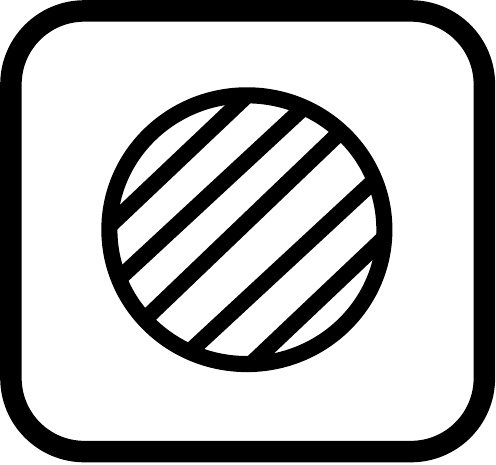}}}
\newcommand{\moderate}{\raisebox{-0.6ex}{\includegraphics[height=2.4ex]{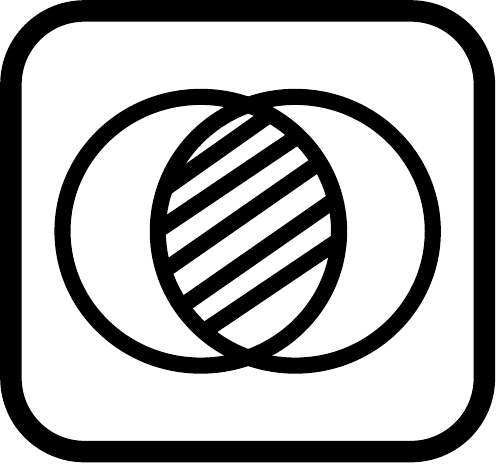}}}
\newcommand{\smalloverlap}{\raisebox{-0.6ex}{\includegraphics[height=2.4ex]{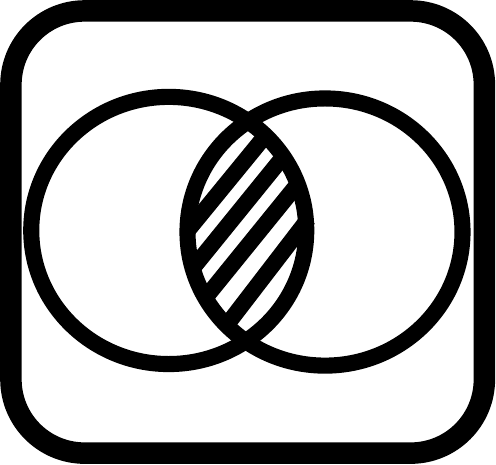}}}

\section{Introduction}
\vspace{-0.1cm}
\label{sec:introduction}
\noindent
Multi-robot light detection and ranging~(LiDAR)-based map merging associates the global descriptors and raw scans of multiple robots to estimate inter-robot relative poses. % as they explore an area
A central entity, known as a server~(or central robot), typically coordinates these frameworks, managing not only global descriptors matching and loop closure detection but also allocating shared resources such as bandwidth, battery power, memory, and computational capacity \citep{burgard2005coordinated, khamis2015multi, madridano2021trajectory, cortes2017coordinated}.
Such map merging frameworks are indispensable in large-scale missions, including faster exploration, frontier allocation, and coverage planning, where maintaining a consistent global frame boosts coverage efficiency and reduces exploration time in the field.
Moreover, as autonomous systems move toward real-time operation in unknown environments, scalable and efficient multi-robot map merging has become a critical capability for the robotics community~\citep{scientific, best2024multi, orr2023multi, zhou2021survey, kim2025diter++, jeong2025marscalib}.

% ========================================================
\begin{figure*}[t]
  \centering
  \vspace{-0.4cm}
  \captionsetup{justification=justified}
  \includegraphics[width=\linewidth, trim=0 0 0 0, clip]{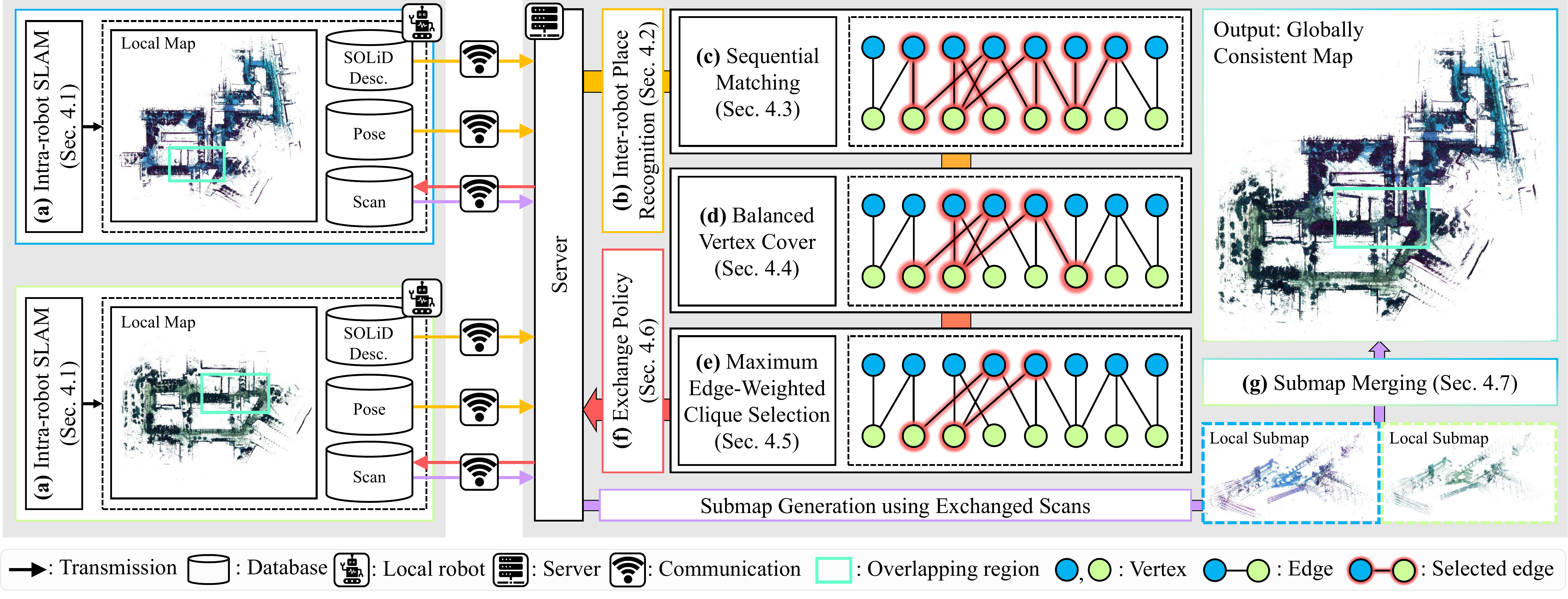}
  \vspace{-6.5mm}
  \caption{Overall pipeline of \textit{Commerge}, a communication-efficient, robust, and fast multi-robot LiDAR map merging framework. 
 ~(a)~Each robot~($\robot$) runs intra-robot SLAM~(Section~\ref{sec:intra_slam}) to generate SOLiD descriptors, poses, and scans, producing local maps. 
 ~(b)~The server~($\server$) receives SOLiD descriptors and poses via communication~($\wifi$) and constructs an affinity matrix for inter-robot place recognition~(Section~\ref{sec:inter_pr}). 
 ~(c)~Sequential matching~(Section~\ref{sec:seq_match}) identifies sequentially overlapped candidate inter-robot links~($\fedge$), forming an exchange graph where $\falpha$ and $\fbeta$ vertices represent keyframes from the two robots. 
 ~(d)~Balanced vertex cover~(Section~\ref{sec:vertex_cover}) selects a balanced set of vertices to cover all candidate edges, ensuring complete coverage with balanced transmission requirements.
 ~(e)~Maximum edge-weighted clique selection~(Section~\ref{sec:edge_clique}) further refines the selection to retain only pairwise consistent loops~($\sedge$), identifying high-overlapped regions that maintain inter-robot consistency with minimal communication overhead.
 ~(f)~The server~($\server$) computes an exchange policy specifying which scans to transmit and sends this policy back to the robots~($\robot$) via communication~($\wifi$)~(Section~\ref{sec:exchange_policy}). 
 ~(g)~Each robot then transmits only the scans incident to the selected edges~($\sedge$), reducing data exchange from GB-scale local maps to MB-scale submaps. The server builds compact submaps from these exchanged scans and performs submap merging~(Section~\ref{sec:map_merging}) to maintain inter-robot alignment.
  As a result, our proposed pipeline achieves communication-efficient, fast, and robust multi-robot map merging by transmitting only partial scans from submaps instead of all scans from local maps, while preserving global consistency.}
\label{fig:main}
\vspace{-3mm}
\end{figure*}
\noindent
% ========================================================

\indent With recent advances in high-bandwidth mobile networks, several LiDAR-based map merging frameworks have demonstrated impressive performance under ideal communication conditions \citep{ltmapper, gil2025ephemerality, wei2024large, zou2024lta, kang2025uni, yin2023automerge}.
However, two major challenges still hinder their practical deployment in the field.
First, the server or each robot must execute computationally intensive operations, such as loop closure detection and descriptor generation, despite being equipped with limited onboard processors. 
Second, the server must exchange and process massive amounts of global descriptors or sensor data, which quickly exhausts bandwidth and drains battery power. 
In larger-scale missions~(\eg more area to explore or more robots), maintaining accurate inter-robot relative pose estimates becomes more challenging under limited resources.

To address these challenges, researchers have explored two mainstream strategies for multi-robot 3D LiDAR map merging: centralized \citep{ltmapper, gil2025ephemerality, wei2024large, zou2024lta, stathoulopoulos2024frame, stathoulopoulos2023frame, kang2025uni, yin2023automerge, damigos2024communication, lim2025multi} and distributed frameworks \citep{ huang2021disco, zhong2023dcl, kim2025skid, he2025ldg, lajoie2025multi, chang2025lidar, lajoie2023swarm, liu2025sparse, chang2022lamp, lajoie20253d}.
Most centralized frameworks aggregate all robot data on a server for global optimization, providing effective coordination. 
However, they require continuous high-bandwidth communication and create scalability bottlenecks as data volume grows with mission duration.
Most distributed frameworks reduce server reliance through peer-to-peer data exchange during robot encounters, offering better scalability. 
However, due to the lack of explicit selection mechanisms, these frameworks often transmit entire data backlogs at each encounter, leading to bursty communications and redundant data exchange.

Although these multi-robot map merging frameworks adopt resource-saving strategies such as downsampling, heuristic pruning, conditional keyframing, or representation compression~\citep{zheng2024real} to mitigate these issues, they primarily reduce data volume without explicitly considering the selection of informative inter-robot loop closures.
Consequently, geometric detail may be sacrificed and valuable correspondences discarded, degrading overall map quality.
Moreover, most works implicitly assume stable and high-speed communication links, overlooking bandwidth and energy constraints as well as practical issues such as scalability and bursty data transmission in real-world operations.
As a result, there is a potential gap between controlled experimental demonstrations and field-deployable or resource-aware multi-robot mapping systems.

To address this gap, we propose \textit{Commerge}, a multi-robot 3D LiDAR map merging framework designed for bandwidth- and resource-constrained environments, with a focus on selective data exchange.
Unlike existing multi-robot map merging frameworks, our proposed method mitigates both scalability limitations and bursty communication patterns by leveraging server-guided global management while transmitting only essential scans through balanced selective data exchange. 
As shown in \figref{fig:main}, our method uses graph-theoretic filtering to extract scans that optimally represent inter-robot relative poses~(\ie \figref{fig:main}(c),~\figref{fig:main}(d), and~\figref{fig:main}(e)).
The selected scans are then aggregated into compact submaps and aligned in a single step by the server's fast and robust coarse-to-fine registration~(\ie \figref{fig:main}(g)).
This design enables server-guided management while drastically reducing massive global descriptors or sensor data exchange and computation.

In summary, our contributions are as follows:
\begin{itemize}
\item We propose Commerge, a system-level and communication-aware multi-robot 3D LiDAR map merging framework that achieves high mapping accuracy with lower communication and computation costs, making it practical for communication-constrained field deployments.

\item We formulate a communication-efficient exchange graph and solve a three-stage cascaded optimization problem~(\ie sequential matching, balanced vertex cover, and maximum edge-weighted clique selection) to select an optimal subset of inter-robot loops while suppressing redundant scan transfers.
Rather than formulating a single globally optimal end-to-end objective, our cascade framework provides complementary structural roles at each stage, namely reliability filtering, balance-aware reduction, and consistency enforcement, while making the next stage computationally tractable.

\item We validate Commerge through extensive experiments on five public datasets and four in-house datasets covering cave, planetary-analog, indoor, and outdoor campus environments, demonstrating substantial bandwidth efficiency and comparable or better mapping accuracy than state-of-the-art map merging approaches. 
Furthermore, we demonstrate system robustness under real-world non-line-of-sight~(NLOS) communication with up to 73.3\% dropout rate, as well as under emulated network degradation scenarios including delay, bandwidth limitation, and packet loss.
\end{itemize}

Building on our previous work~\citep{kim2025skid}, we demonstrate that fast descriptor exchange and robust registration can accelerate multi-robot map merging.
Nevertheless, the overall volume of exchanged descriptors and scans is effectively the same as in centralized systems that ship all data, so bandwidth remains a critical bottleneck in practice.
As a result, bandwidth and hardware resources are still quickly saturated in real-world deployments, especially when mission time increases or network quality fluctuates.
These observations motivate the need for a new framework that achieves robust and fast map merging while drastically reducing data transmission through a communication-efficient module.

\vspace{-2mm}
\section{Related Works}
\noindent
Mobile robots are inherently constrained by the limited resources they can carry onboard, such as processing power and wireless communication modules. 
In multi-robot systems, these constraints become even more critical as teams must coordinate their limited resources while sharing communication bandwidth and computational tasks. 

A key challenge in this setting is not only reducing the volume of transmitted data, but also determining which inter-robot correspondences should be exchanged under resource constraints.
In particular, as the number of potential inter-robot correspondences increases, transmitting all candidate data becomes inefficient and often impractical.

To address this challenge, this section reviews prior work from three perspectives:
(i)~inter-robot association, including loop closure detection~(Section~\ref{sec:loop_closure}) and point cloud registration~(Section~\ref{sec:registration});
(ii)~multi-robot map merging frameworks, covering centralized~(Section~\ref{sec:centralized}) and distributed~(Section~\ref{sec:distributed}) approaches;
and~(iii)~communication-efficient strategies for bandwidth optimization~(Section~\ref{sec:comm_efficient}), with a focus on selective data exchange.

\subsection{Loop Closure Detection for Multi-Robot 3D Point Cloud Map Merging}
\label{sec:loop_closure}
\noindent To construct a globally aligned 3D multi-robot map, robots first perform inter-robot loop closure detection~(\ie place recognition) to identify correspondences between scans collected by different robots.
Traditional descriptors such as Scan Context~\citep{kim2018scan}, LiDAR-Iris~\citep{wang2020lidar}, STD~\citep{yuan2023std}, BTC~\citep{yuan2024btc}, RING++~\citep{xu2023ring++}, SOLiD~\citep{kim2024narrowing}, and OSK~\citep{zhang2024osk} encode geometric features into compact representations, enabling lightweight and robust loop detection.

With the rise of deep learning, numerous researchers have also studied learning-based approaches. For example, deep learning-based descriptors, including PointNetVLAD~\citep{uy2018pointnetvlad}, OverlapTransformer~\citep{ma2022overlaptransformer}, LoGG3D-Net~\citep{vidanapathirana2022logg3d}, CVTNet~\citep{ma2023cvtnet}, BEVPlace++~\citep{luo2025bevplace++}, and 3DEG~\citep{stathoulopoulos20243deg}, further improve viewpoint robustness through learned feature representations, but their larger data sizes and high computational costs limit their use in embedded systems.

While these methods improved matching robustness and recall, they primarily focused on generating candidate correspondences and did not address which inter-robot loop closures should be selected and transmitted under communication constraints.
In multi-robot settings, this distinction becomes critical.
Unlike intra-robot loop closures, where all loop candidates can be locally verified, inter-robot loop detection requires transmitting candidate information across bandwidth-limited communication links.
Therefore, beyond accurate detection, selecting a compact and reliable subset of inter-robot loop closures is essential for scalable and communication-efficient multi-robot map merging.

Additionally, incorrect inter-robot loop closures can propagate errors across all connected robot maps, making robustness a key requirement.
To improve reliability, several approaches leverage sequential consistency among neighboring loop candidates to enforce temporal and spatial coherence~\citep{milford2012seqslam, siam2017fast, oishi2019seqslam++}. 
For instance, AutoMerge~\citep{yin2023automerge} employs SeqSLAM-based window identification followed by clustering and geometric verification to filter unreliable candidates. 
However, such pipelines retain a large number of intermediate candidates throughout the verification process, resulting in substantial computational and communication overhead in multi-robot settings, and highlighting the need for more efficient selection mechanisms.

% ===========================================================
\subsection{Point Cloud Registration For Multi-Robot 3D Point Cloud Map Merging}
% ===========================================================
\label{sec:registration}
\noindent 
Following loop closure detection, point cloud registration estimates the relative pose between the loop pair.
In multi-robot settings, accurate and robust registration is essential, as alignment errors can propagate across multiple robots and degrade the consistency of the global map.

Most multi-robot mapping frameworks adopt local registration methods such as ICP and its variants~\citep{segal2009generalized, pomerleau2013comparing, koide2021voxelized, besl1992method}.
These methods can converge to accurate solutions when good initial alignment is available, but suffer from narrow convergence basins and often fail under large pose discrepancies.

To address this limitation, global registration approaches such as FGR~\citep{zhou2016fast}, Go-ICP~\citep{yang2015go}, and PHASER~\citep{bernreiter2021phaser} search over a wider pose space to mitigate local minima, at the cost of increased computational complexity.
More recent approaches, including learning-based methods such as Predator~\citep{huang2021predator}, BUFFER~\citep{ao2023buffer}, and BUFFER-X~\citep{seo2025buffer}, further improve robustness under challenging conditions by leveraging learned feature correspondences and context-aware matching.
In addition, several place recognition methods such as BEVPlace++~\citep{luo2025bevplace++} and RING++~\citep{xu2023ring++} inherently provide coarse pose estimates without explicit feature matching.
However, most existing registration methods operate at the scan level, focusing on pairwise alignment between individual point clouds.
While computationally efficient, such approaches rely on limited geometric information and are sensitive to noise, partial overlap, and large viewpoint changes.

In contrast, map-level registration aligns accumulated maps composed of multiple scans, providing richer geometric and structural information that enables more robust and accurate alignment.
This is particularly important in multi-robot scenarios, where reliable global map merging requires strong geometric constraints across diverse viewpoints and environments.
However, as the size of the accumulated maps increases, the number of correspondences grows substantially, leading to increased computational cost.

Our prior work, KISS-Matcher~\citep{kiss_matcher}, mitigates this trade-off by integrating feature extraction, correspondence pruning, and efficient pose estimation into a unified framework. 
Nevertheless, as mission scale grows, even efficient map-level registration becomes computationally demanding when operating on full accumulated maps, motivating more compact aggregation strategies for scalable multi-robot mapping.
% ===========================================================

% ===========================================================
\subsection{Centralized Multi-Robot 3D Point Cloud Map Merging Frameworks}
% ===========================================================
\label{sec:centralized}
\noindent 
Centralized frameworks aggregate all global descriptors and scans at a server, which performs global optimization across all robots.
These approaches provide server-managed coordination and global optimization across all robots, enabling comprehensive map integration~\citep{ltmapper, gil2025ephemerality, wei2024large, zou2024lta, stathoulopoulos2024frame, stathoulopoulos2023frame, kang2025uni, yin2023automerge}.
Multi-robot systems built on this paradigm also benefit from inherent redundancy and fault tolerance, as overlapping observations across robots can overcome sensor uncertainties~\citep{burgard2005coordinated}.

However, centralized approaches incur heavy communication and computation costs.
Even with aggressive downsampling or keyframing, the server must handle large volumes of data, and the bandwidth demand grows rapidly with mission duration and data size.
Furthermore, reliance on a single server introduces bottlenecks and potential single points of failure, limiting scalability in real-world deployments where stable network links are rarely guaranteed.

These limitations highlight the need for scalable strategies that preserve the benefits of centralized coordination while mitigating data transmission and computational bottlenecks.
% ===========================================================

\subsection{Distributed Multi-Robot 3D Point Cloud Map Merging Frameworks}
\label{sec:distributed}
\noindent 
Distributed frameworks reduce dependence on a central server by allowing robots to exchange locally acquired data only when they encounter each other~\citep{huang2021disco, zhong2023dcl, kim2025skid, he2025ldg, chang2025lidar}.
This peer-to-peer strategy improves scalability and eliminates the need for continuous high-bandwidth links, making it attractive for multi-robot systems. 

However, since robots cannot communicate when spatially separated, distributed systems require robots to be in proximity for data exchange.
When robots do encounter each other, they often transmit all accumulated keyframes or scans collected since their last meeting, resulting in large and bursty transmissions with substantial redundancy~\citep{schmuck2021covins, xu2024d2slam}.
Furthermore, when encounters are infrequent, teams also have to coordinate rendezvous timing and location, which delays inter-robot corrections and leaves drift unchecked, undermining timely and consistent map alignment~\citep{roy2001collaborative, ko2003practical}.

Recent approaches, such as Multi-Proxy~\citep{wang2026communication}, propose communication-robust distributed mapping frameworks with an asynchronous smoothing backend.
In addition, robots exchange compact proxy representations derived from key geometric features, \eg STD~\citep{yuan2023std} corner points, to reduce communication overhead.
However, as the overlap between robots increases, the number of potential inter-robot loop candidates grows rapidly, and the corresponding scans for each candidate must still be transmitted for verification and alignment, leading to bursty transmissions that can saturate bandwidth in large-scale multi-robot systems.

These limitations suggest that reducing per-message data size alone is insufficient; rather, selectively determining which loop-related scans to transmit within overlapping regions is essential for efficient multi-robot map merging.

\subsection{Communication-Efficient Strategies for Multi-Robot Map Merging Frameworks}
\label{sec:comm_efficient}
\noindent 
Multi-robot map merging often suffers from substantial communication overhead due to the continuous exchange of large scans or detailed descriptors.
Common methods such as uniform downsampling and periodic keyframing have been widely used to reduce communication overhead by decreasing the size of transmitted data.

Recent approaches further optimize keyframe selection to reduce redundancy, such as the minimal subset approach~(MSA)~\citep{stathoulopoulos2025minimal}, which formulates keyframe selection as an optimization problem to retain informative frames while minimizing transmission.
In addition, environment and point cloud compression methods such as Draco~\citep{de2016compression}, geometry-based point cloud compression~(G-PCC)~\citep{graziosi2020overview}, and real-time compressed point cloud coding~(RCPCC)~\citep{cao2025real} reduce bandwidth usage by encoding geometric information into compact forms before sharing across robots.

In particular, \citet{zheng2024real} target real-time environment sharing in multi-robot cooperative systems by compressing environmental point clouds into compact panoramic representations for efficient transmission.
While effective in reducing transmission size and computational overhead, such approaches focus on how to compress and share environment representations, rather than explicitly addressing which inter-robot loop-related data should be retained and transmitted within overlapping regions.

More sophisticated approaches focus on conditional or incremental data exchange.
\citet{cieslewski2017efficient} introduce a distributed visual place recognition scheme in which only a subset of compact query words is initially transmitted, followed by larger keypoint or descriptor packets only when a promising loop candidate is identified.
In their subsequent work~\citep{cieslewski2018data}, a two-stage decentralized map merging pipeline is proposed: robots first broadcast a single NetVLAD descriptor~\citep{arandjelovic2016netvlad} to query potential loops, and exchange detailed data only upon successful verification.
These studies demonstrate that selective and incremental transmission substantially lowers bandwidth consumption while preserving map accuracy.

Building on this foundation, graph-theoretic approaches emerge to optimize data exchange patterns.
Several works~\citep{giamou2018talk, tian2018near, tian2021resource, lajoie2023swarm} introduce the concept of an exchange graph for distributed loop closure detection, in which vertices represent robot poses and edges correspond to potential inter-robot loops.
They formulate the data selection challenge as finding the minimal-cost subset that covers all potential loops, where costs represent practical metrics such as CPU time or network bandwidth.
By solving this as a minimum bipartite vertex-weighted cover, their frameworks determine optimal data exchanges while minimizing transmission overhead.
This approach also enables workload balancing based on available robot resources for resource-aware multi-robot systems.
However, these formulations may lead to unbalanced solutions, where selected vertices are concentrated on a single robot, resulting in asymmetric communication patterns.
Such a communication pattern can overload specific robots while underutilizing others, which is undesirable in real-world multi-robot systems with limited resources.

Furthermore, in practical multi-robot deployments, assigning accurate transmission costs remains challenging because CPU time and wireless bandwidth fluctuate during operation, so costs measured during optimization do not accurately predict actual transmission costs, and continuous remeasurement introduces additional overhead~\citep{lajoie2022towards, kulkarni2019deepchannel, gielis2022critical, formis2023predicting}.
Therefore, reducing data size alone is insufficient. Instead, scalable and efficient multi-robot map merging requires selectively transmitting informative data within overlapping regions while maintaining balanced data exchange between robots.

\newcommand{\stepA}{\textsf{step~(A)}}
\newcommand{\stepB}{\textsf{step~(B)}}
\newcommand{\stepC}{\textsf{step~(C)}}

\newcommand{\alphapose}{\mathbf{T}^\alpha_i}
\newcommand{\alphascan}{\mathbf{P}^\alpha_i}
\newcommand{\alphadesc}{\mathbf{D}^\alpha_i}

\newcommand{\betapose}{\mathbf{T}^\beta_j}
\newcommand{\betascan}{\mathbf{P}^\beta_j}
\newcommand{\betadesc}{\mathbf{D}^\beta_j}

\newcommand{\edge}{\boldsymbol{e}}

\section{Preliminaries}
\begin{figure*}[t]
    \centering
    \captionsetup{justification=justified}
    \captionsetup[subfigure]{justification=centering}
    \includegraphics[trim={0 0 0 0},clip, width=1.94\columnwidth]{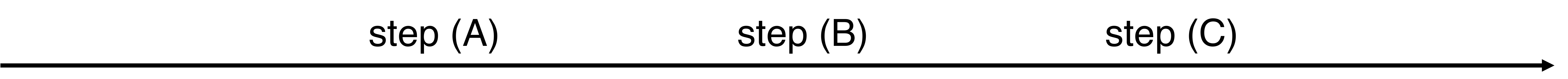}
    \begin{subfigure}{0.55\columnwidth}
        \includegraphics[trim={0 0 0 0},clip, width=\columnwidth]{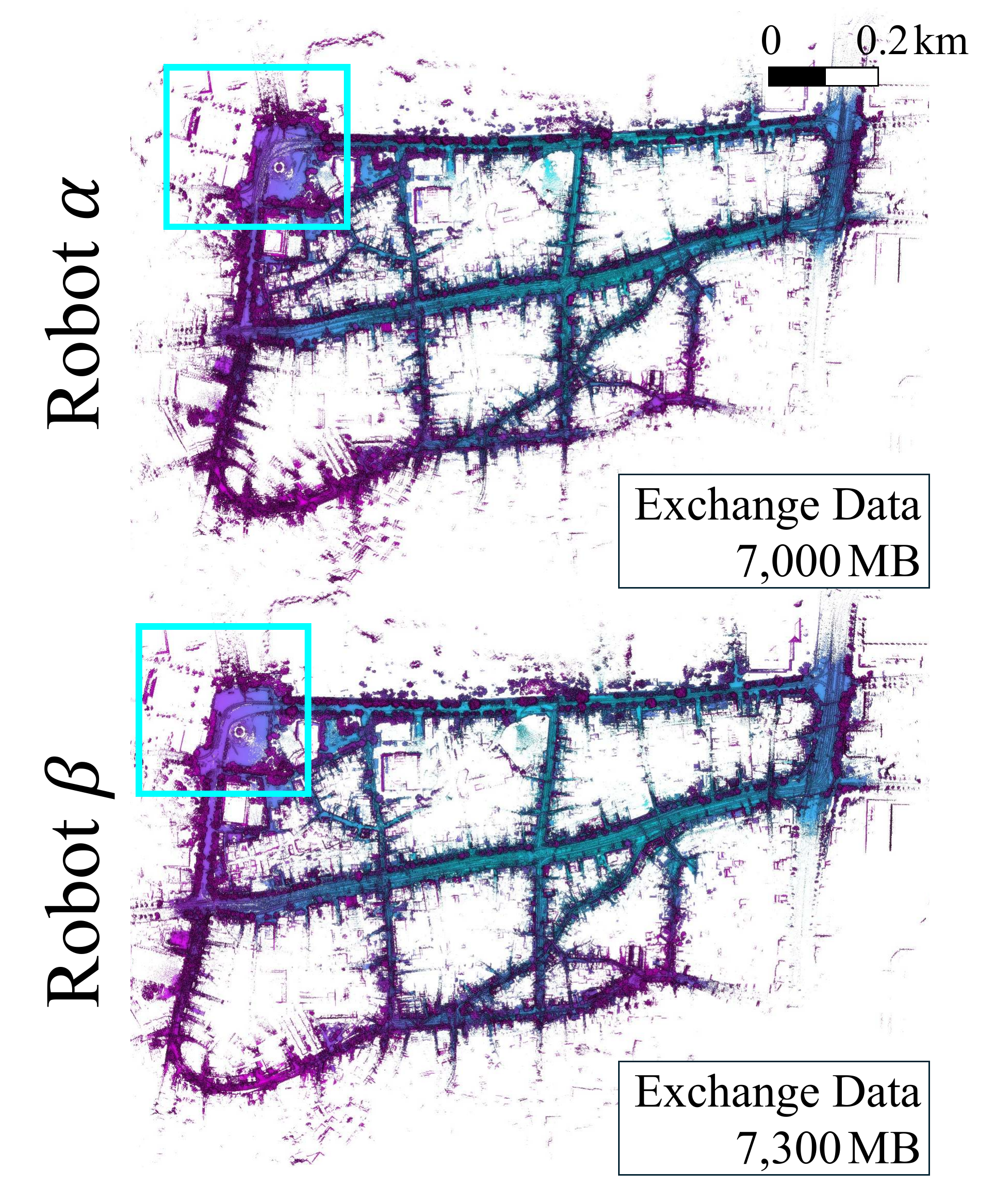}
        \vspace{-1.5em}
        \caption{}
    \end{subfigure}
    \begin{subfigure}{0.4575\columnwidth}
        \includegraphics[trim={0 0 0 0},clip, width=\columnwidth]{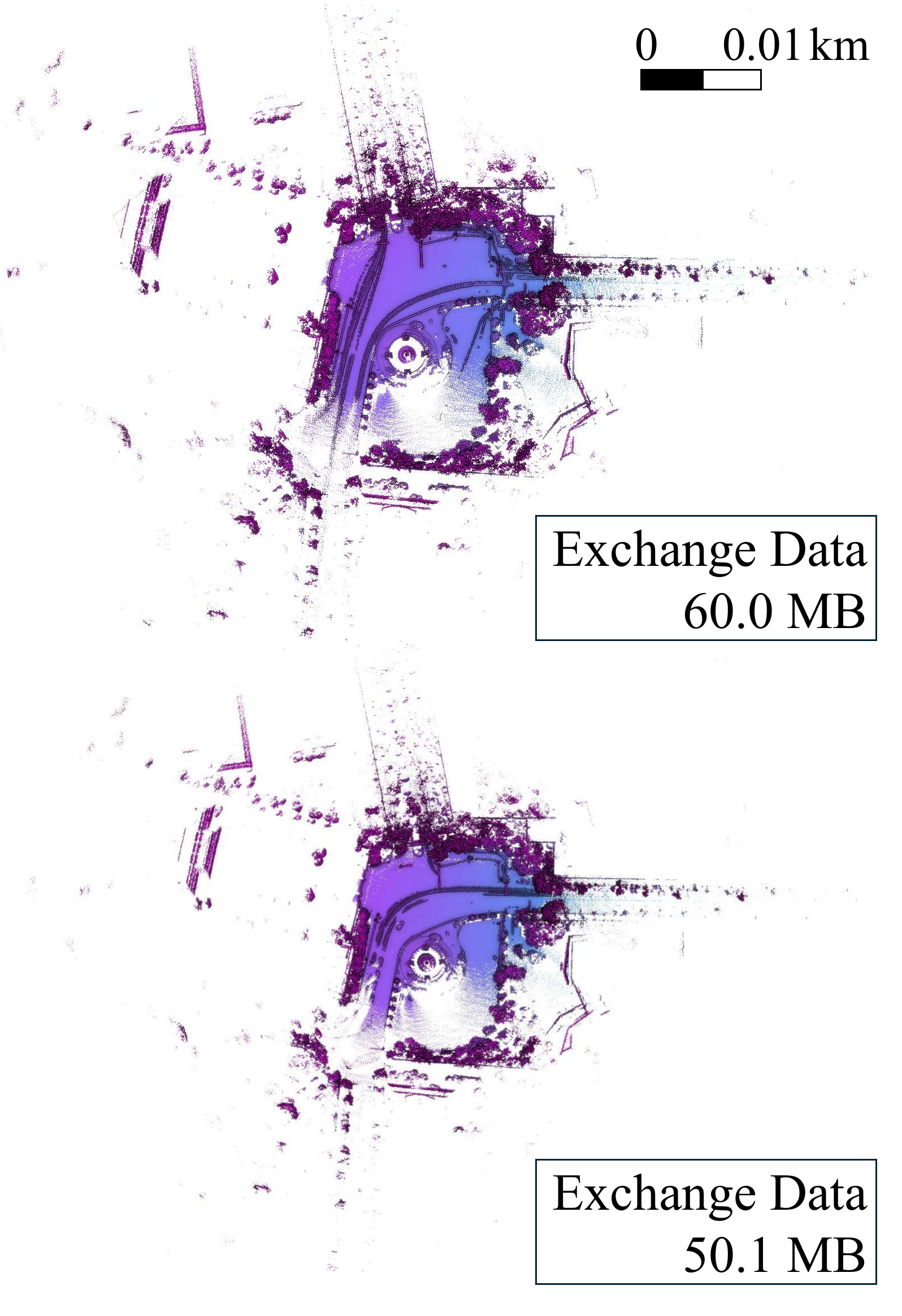}
        \vspace{-1.5em}
        \caption{}
    \end{subfigure}
    \begin{subfigure}{0.47\columnwidth}
        \includegraphics[trim={0 0 0 0},clip, width=\columnwidth]{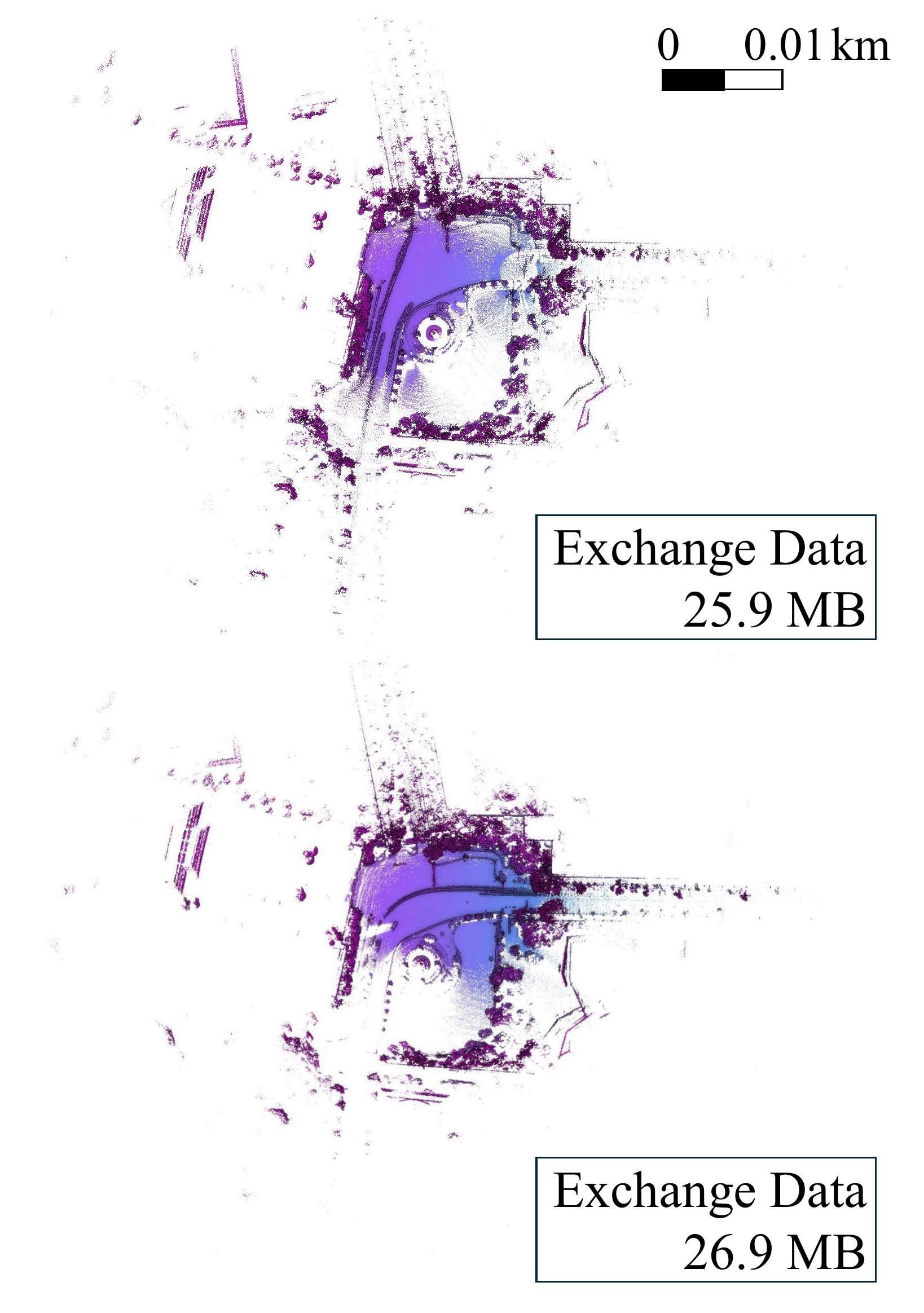}
        \vspace{-1.5em}
        \caption{}
    \end{subfigure}
    \begin{subfigure}{0.4575\columnwidth}
        \includegraphics[trim={0 0 0 0},clip, width=\columnwidth]{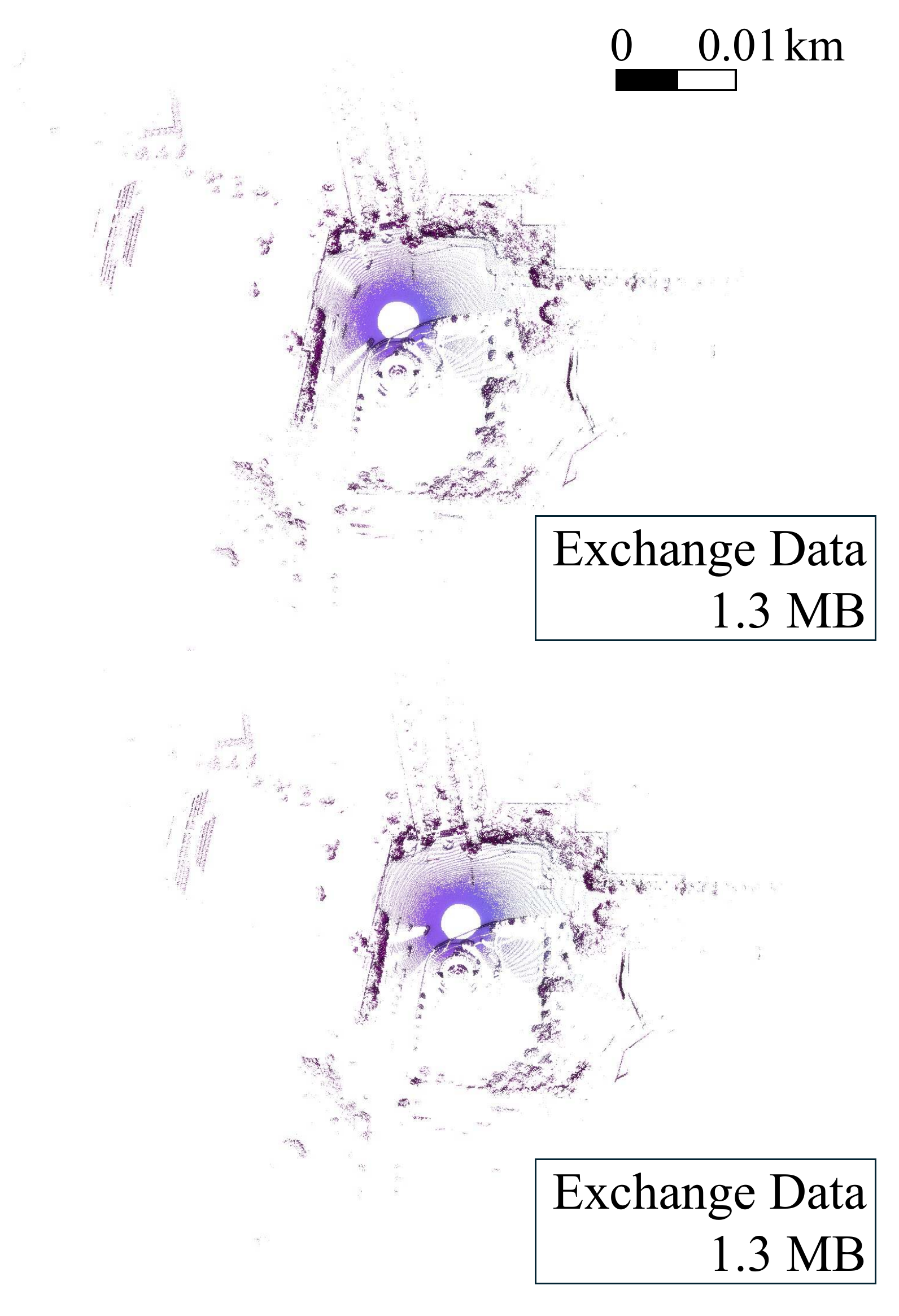}
        \vspace{-1.5em}
        \caption{}
    \end{subfigure}
    \caption{Submaps and associated exchange data sizes across modules in our framework. 
           (a) Without selection, each robot requires approximately 7,000\,MB of exchanged data from intra-robot SLAM. The cyan boxes highlight sequentially overlapping regions that are selected in $\stepA$.
           (b) After sequential matching, \ie $\stepA$, the exchange volume is reduced to 60.0\,MB for robot $\alpha$ and 50.1\,MB for robot $\beta$.
           (c) With balanced vertex cover, \ie $\stepB$, the required data is further reduced to 25.9\,MB and 26.5\,MB, respectively.
           (d) Finally, the maximum edge-weighted clique selection, \ie $\stepC$ reduces the exchanged data to just 1.3\,MB per robot.
           This progressive reduction demonstrates the communication efficiency of our proposed map merging pipeline, achieving an overall 99.98\% reduction in transmission cost (from 7,000\,MB to 1.3\,MB), while still enabling successful submap merging.}  
  \label{fig:figure_main}
  \vspace{-3mm}
\end{figure*}

\noindent In this section, we introduce the fundamental concepts and theoretical foundation of the exchange graph, which serves as the core algorithm of our \textit{Commerge}. 
We first present the notation of variables and then formally define the problem of optimal data selection using the exchange graph.
As shown in \figref{fig:figure_main}, we solve this problem through three cascaded stages: $\stepA$ from \figref{fig:figure_main}(a) to \figref{fig:figure_main}(b), $\stepB$ from \figref{fig:figure_main}(b) to \figref{fig:figure_main}(c), and $\stepC$ from \figref{fig:figure_main}(c) to \figref{fig:figure_main}(d).

% =============================================================
% =============================================================

\subsection{Notation}
\label{sec:notation}
\noindent
We assume robot $\alpha$ provides a set of observation tuples
$\mathcal{V}^\alpha = \{v_i^\alpha\}_{i=1}^{N}$, where each $v_i^\alpha = (\alphapose, \alphascan,\alphadesc)$
for indices $i\in\mathcal{I}^\alpha:=\{1,\dots,N\}$ (resp. robot~$\beta$ provides
$\mathcal{U}^\beta = \{u_j^\beta\}_{j=1}^{M}$ where $u_j^\beta = (\betapose,\betascan,\betadesc)$
for $j\in\mathcal{J}^\beta:=\{1,\dots,M\}$; see Section~\ref{sec:intra_slam}).
Here, $\alphapose$ (resp. $\betapose$) denotes a 6-DoF keyframe pose in $\mathrm{SE}(3)$; for example, $\mathbf{T}^{\alpha}_c = \begin{bmatrix} \mathbf{R}^{\alpha}_c & \mathbf{t}^{\alpha}_c \\ \mathbf{0}^\top & 1 \end{bmatrix} \in \text{SE}(3)$ with $\mathbf{R}^{\alpha}_c\in\mathrm{SO}(3)$ and $\mathbf{t}^{\alpha}_c\in\mathbb{R}^3$ \big(resp. $\mathbf{T}^{\beta}_q = \begin{bmatrix} \mathbf{R}^{\beta}_q & \mathbf{t}^{\beta}_q \\ \mathbf{0}^\top & 1 \end{bmatrix} \in \text{SE}(3)$\big);
$\alphascan\in\mathbb{R}^{L_i\times 3}$ (resp. $\betascan$) is the LiDAR scan with $L_i$ points; and $\alphadesc\in\mathbb{R}^{E}$ (resp. $\betadesc$) is an $E$-dimensional compact global descriptor extracted from $\alphascan$.
Using these observations, we define the exchange graph for robots $\alpha$ and $\beta$ as a bipartite weighted undirected graph as follows:
\begin{equation}
    \mathcal{G}^{\alpha,\beta}
    =
    \Big(
        \mathcal{V}^{\alpha} \cup \mathcal{U}^{\beta},\;
        \mathcal{E}^{\alpha,\beta},\;
        \mathcal{W}^{\alpha,\beta}
    \Big),
    \label{eq:exchange_graph}
\end{equation}
where $\mathcal{E}^{\alpha,\beta}$ is the edge set of inter-robot correspondences, and $\mathcal{W}^{\alpha,\beta}$ contains edge weights.

Next, let $\mathcal{L}\coloneqq\mathcal{I}^\alpha\times\mathcal{J}^\beta$ denote the set of potential inter-robot loop candidates (see Section~\ref{sec:inter_pr}). 
Let $\mathcal{L}_{\text{seq}}\coloneqq\mathcal{I}^\alpha_{\text{seq}}\times\mathcal{J}^\beta_{\text{seq}}$ denote the sequentially matched subset (see Section~\ref{sec:seq_match}), \ie $\mathcal{L}_{\text{seq}}\subseteq\mathcal{L}$.
Here, let the sequentially selected index ranges be $\mathcal{I}^\alpha_{\text{seq}}\subseteq\mathcal{I}^\alpha$ (resp. $\mathcal{J}^\beta_{\text{seq}}\subseteq\mathcal{J}^\beta$), and the vertex sets are as follows:
\begin{equation}
\mathcal{V}^{\alpha}_{\text{seq}}=\{v^{\alpha}_{c}\mid c\in\mathcal{I}^\alpha_{\text{seq}}\},
\qquad
\mathcal{U}^{\beta}_{\text{seq}} =\{u^{\beta}_{q}\mid q\in\mathcal{J}^\beta_{\text{seq}}\},
\end{equation}
where $v^{\alpha}_c$ and $u^{\beta}_q$ denote the $c$-th and $q$-th sequentially matched observations, respectively.
Accordingly, the edge set of the exchange graph is defined as follows:
\begin{equation}
\mathcal{E}^{\alpha,\beta}_{\text{seq}}
=
\Bigl\{\,
\edge^{\alpha,\beta}_{c,q}
=
\{v^{\alpha}_c,u^{\beta}_q\}
\;\Bigm|\;
(c,q)\in\mathcal{L}_{\text{seq}}
\,\Bigr\}.
\label{eq:edge}
\end{equation}
Here, the relative transformation for an edge $\edge^{\alpha,\beta}_{c,q}$ is $\mathbf{T}^{\alpha,\beta}_{c,q} = (\mathbf{T}^{\alpha}_c)^{-1} \mathbf{T}^{\beta}_q = \begin{bmatrix} \mathbf{R}^{\alpha,\beta}_{c,q} & \mathbf{t}^{\alpha,\beta}_{c,q} \\ \mathbf{0}^\top & 1 \end{bmatrix} \in \text{SE}(3)$ with $\mathbf{R}^{\alpha,\beta}_{c,q}\in\mathrm{SO}(3)$ and $\mathbf{t}^{\alpha,\beta}_{c,q}\in\mathbb{R}^3$.
Finally, the weight set is defined as follows:
\begin{equation}
    \mathcal{W}^{\alpha,\beta}_{\text{seq}} =
\Bigl\{
    w^{\alpha,\beta}_{c,q}
    \;\Big|\;
    \edge^{\alpha,\beta}_{c,q} \in \mathcal{E}^{\alpha,\beta}_{\text{seq}}
\Bigr\},
\label{eq:weight_set}
\end{equation}
where each weight is induced by an edge–weighted function, $f_w(\cdot): \edge^{\alpha,\beta}_{c,q} \in \mathcal{E}^{\alpha,\beta}_{\text{seq}} \mapsto \mathbb{R}_{\ge 0}$.
Concretely, each scan is encoded into a compact descriptor via a mapping function, $f_{\text{SOLiD}}(\cdot)$, \eg $\alphadesc = f_{\text{SOLiD}}(\alphascan)$ $\big($resp. $\betadesc = f_{\text{SOLiD}}(\betascan)$$\big)$, and we then compute $w^{\alpha, \beta}_{c, q}$ in \eqref{eq:weight_set} from descriptor similarity as follows:
\begin{equation}
w^{\alpha, \beta}_{c, q} = f_w\big(\edge^{\alpha,\beta}_{c,q}\big)
= f_{\text{sim}}\!\big(\mathbf{D}^{\alpha}_c,\mathbf{D}^{\beta}_q\big).
\label{eq:f_sim}
\end{equation}

As a result, the undirected exchange graph from sequential matching is defined as follows:
\begin{equation}
    \mathcal{G}^{\alpha,\beta}_{\text{seq}}
    =
    \Big(
        \mathcal{V}^{\alpha}_{\text{seq}} \cup \mathcal{U}^{\beta}_{\text{seq}},\;
        \mathcal{E}^{\alpha,\beta}_{\text{seq}},\;
        \mathcal{W}^{\alpha,\beta}_{\text{seq}}
    \Big),
    \label{eq:seq_exchange_graph}
\end{equation}
where $\mathcal{G}^{\alpha,\beta}_{\text{seq}} \subseteq \mathcal{G}^{\alpha,\beta}$. 
In this work, processing all potential loops in \eqref{eq:exchange_graph} would be computationally prohibitive. 
Therefore, we employ a three-stage approach: $\stepA$ identifies sequentially overlapped regions (\ie from \eqref{eq:exchange_graph} to \eqref{eq:seq_exchange_graph}), followed by $\stepB$ and $\stepC$, which refine the selection to identify scans that substantially reduce computational cost while preserving essential inter-robot correspondences.

\subsection{Problem Definition}
\label{sec:problem}
\noindent 
Based on the notations above, our goal is to select an optimal subset of loop closures that are both informative and minimal in size.  
Given the candidate inter-robot loops $\mathcal{L}=\mathcal{I}^\alpha\times\mathcal{J}^\beta$, we select the final subset through three nested stages as follows: $\stepA$~sequential matching, $\stepB$~transmission cost minimization, and $\stepC$~consistency-based filtering.

First, $\stepA$ is defined as follows:
\begin{equation}
\mathcal{L}_{\text{seq}}
:= \operatorname*{arg\,min}_{\mathcal{L}^\prime\subseteq\mathcal{L}} J_{\text{seq}}(\mathcal{L}^\prime),
\label{eq:seq}
\end{equation}
where $J_{\text{seq}}(\cdot)$ finds the partial loop subset $\mathcal{L}_{\text{seq}}$ that best~represents sequential overlap from the whole potential loop set~$\mathcal{L}$.

Second, taking $\mathcal{L}_{\text{seq}}$ in \eqref{eq:seq} as input, $\stepB$ is defined as follows:
\begin{equation}
\mathcal{L}_{\text{V}}
:= \operatorname*{arg\,min}_{\mathcal{L}_{\text{seq}}^\prime\subseteq\mathcal{L}_{\text{seq}}} J_{\text{V}}(\mathcal{L}_{\text{seq}}^\prime).
\label{eq:vertex}
\end{equation}
That is, $J_{\text{V}}(\cdot)$ removes redundant loops from the sequentially overlapped $\mathcal{L}_{\text{seq}}$ to find $\mathcal{L}_{\text{V}}$ with minimal transmission cost.

Third, with $\mathcal{L}_{\text{V}}$ in \eqref{eq:vertex}, $\stepC$ is defined as follows:
\begin{equation}
\mathcal{L}_{\text{K}}
:= \operatorname*{arg\,max}_{\mathcal{L}_{\text{V}}^\prime\subseteq\mathcal{L}_{\text{V}}} J_{\text{K}}(\mathcal{L}_{\text{V}}^\prime).
\label{eq:clique}
\end{equation}
Here, $J_{\text{K}}(\cdot)$ chooses the subset $\mathcal{L}_{\text{K}}$ from the minimal transmission cost loops $\mathcal{L}_{\text{V}}$ to preserve strong pairwise consistency.

From the definitions of $\mathcal{L}_{\text{seq}}$, $\mathcal{L}_{\text{V}}$, and $\mathcal{L}_{\text{K}}$, we obtain the nested selection as follows:
\begin{equation}
\mathcal{L}_{\text{K}} \subseteq \mathcal{L}_{\text{V}} \subseteq \mathcal{L}_{\text{seq}} \subseteq \mathcal{L}.
\end{equation}
We therefore cast loop selection as a three-stage cascaded optimization over the objectives, $J_{\text{seq}}(\cdot)$, $J_{\text{V}}(\cdot)$, and $J_{\text{K}}(\cdot)$.
The resulting solution of this cascade, $\mathcal{L}_{\text{K}}$, is both informative and compact, and it is the key to our Commerge's communication-efficient, fast, and robust multi-robot 3D LiDAR map merging.
\section{Commerge: Communication-Efficient, Robust and Fast Map Merging Framework}
\noindent
In this section, we describe the overall algorithm of our proposed method, \textit{Commerge}, as shown in \figref{fig:main}.
Unlike existing multi-robot map merging frameworks, our proposed method bridges the dichotomy between effective coordination and scalability by leveraging a server-guided global management while transmitting only essential scans through balanced selective data exchange.

The entire process proceeds in seven stages: (i) intra-robot SLAM, (ii) inter-robot place recognition, (iii) inter-robot sequential matching, (iv) balanced vertex cover~(BVC) in exchange graph, (v) maximum edge-weighted clique selection in exchange graph, (vi) exchange policy generation, and (vii) submap-based map merging framework.
From intra-robot SLAM at the local level to map merging at the server level, each robot and the server communicate organically and refine candidate loops incrementally, ensuring that only the most informative and mutually consistent data are exchanged.
% ==================================================================================================================================================================================
\subsection{Intra-Robot SLAM for Local Mapping}
\label{sec:intra_slam}
\noindent
Each robot in the team first estimates its trajectory and builds a local map using a LiDAR-based SLAM system.
Following the notation introduced in Section~\ref{sec:notation}, we denote the keyframe poses of robots $\alpha$ and $\beta$ as $\mathcal{X}^\alpha=\{\alphapose \mid i\in\mathcal{I}^\alpha\}$ and $\mathcal{X}^\beta=\{\mathbf{T}^\beta_j \mid j\in\mathcal{J}^\beta\}$, respectively.
These pose sets correspond to the vertices in our exchange graph formulation.
Without loss of generality, we describe the trajectory estimation process for robot $\alpha$; robot $\beta$ follows an identical procedure.
Then, the intra-robot trajectory estimation is formulated as a maximum a posteriori (MAP) estimation problem as follows:
\begin{equation}
\mathcal{X}^{\alpha*} =
\operatorname*{arg\,max}_{\mathcal{X}^\alpha}
P(\mathcal{X}^\alpha \mid \boldsymbol{z})
\;=\;
\operatorname*{arg\,min}_{\mathcal{X}^\alpha}
\{ -\log P(\mathcal{X}^\alpha \mid \boldsymbol{z}) \},
\label{eq:map}
\end{equation}
where  $\boldsymbol{z}$ represents all LiDAR measurements.
Assuming Gaussian noise on relative pose measurements, \eqref{eq:map} reduces to the pose graph optimization as follows:
\begin{equation}
\begin{aligned}
\mathcal{X}^{\alpha *}
=
\operatorname*{arg\,min}_{\mathcal{X}^\alpha}\;
\sum_{(t,t+1)\in\mathcal{C}^{\text{odom}}_{\text{intra}}}
&\Big\|
(\mathbf{T}^\alpha_t)^{-1}\mathbf{T}^\alpha_{t+1}
\boxminus
\mathbf{T}^{\text{odom}}_{t,t+1}
\Big\|^2_{\boldsymbol{\Omega}^{\text{odom}}_{t,t+1}} \\
\;+\!
\sum_{(a,b)\in\mathcal{C}^{\text{loop}}_{\text{intra}}}
&\Big\|
(\mathbf{T}^\alpha_a)^{-1}\mathbf{T}^\alpha_b
\boxminus
\mathbf{T}^{\text{loop}}_{a,b}
\Big\|^2_{\boldsymbol{\Omega}^{\text{loop}}_{a,b}},
\end{aligned}
\end{equation}
where $\mathcal{C}^{\text{odom}}_{\text{intra}}$ is an odometry constraint set between two temporally consecutive $t$-th and $(t+1)$-th poses of the robot $\alpha$, and $\mathcal{C}^{\text{loop}}_{\text{intra}}$ is a loop constraint set between two non-consecutive $a$-th and $b$-th poses of the robot $\alpha$ (\ie $a+1 \neq b$) corresponding to intra-robot revisits, detected by loop detection.
Here, $\boxminus$ denotes the box-minus operator, the manifold analogue of subtraction, which returns the tangent-space increment that takes $\mathbf{Y}$ to $\mathbf{X}$; specifically, for $\mathbf{X},\mathbf{Y} \in \text{SE}(3)$, $\mathbf{X} \boxminus \mathbf{Y} \coloneqq \log\!\big(\mathbf{Y}^{-1}\mathbf{X}\big) \in \mathfrak{se}(3)\simeq\mathbb{R}^6$ \citep{he2021kalman}.
The measurements $\mathbf{T}^{\text{odom}}_{t,t+1}$ and $\mathbf{T}^{\text{loop}}_{a,b}$ correspond to the observed relative poses from odometry~\citep{xu2022fast} and registration with loop detection, respectively, and each residual is weighted by its associated information matrix, $\boldsymbol{\Omega}^{\text{odom}}_{t,t+1}$ or $\boldsymbol{\Omega}^{\text{loop}}_{a,b}$.
The squared Mahalanobis norm $\| \cdot \|^2_{\boldsymbol{\Omega}}$ accounts for measurement uncertainty, and the optimization seeks the trajectory $\mathcal{X}^{\alpha*}$ that best explains all relative constraints under Gaussian noise assumptions.

\begin{figure}[t]
    \centering
    \captionsetup{justification=justified}
    \captionsetup[subfigure]{justification=centering}
    \begin{subfigure}{0.49\columnwidth}
        \includegraphics[trim={30 0 0 0},clip, width=\columnwidth]{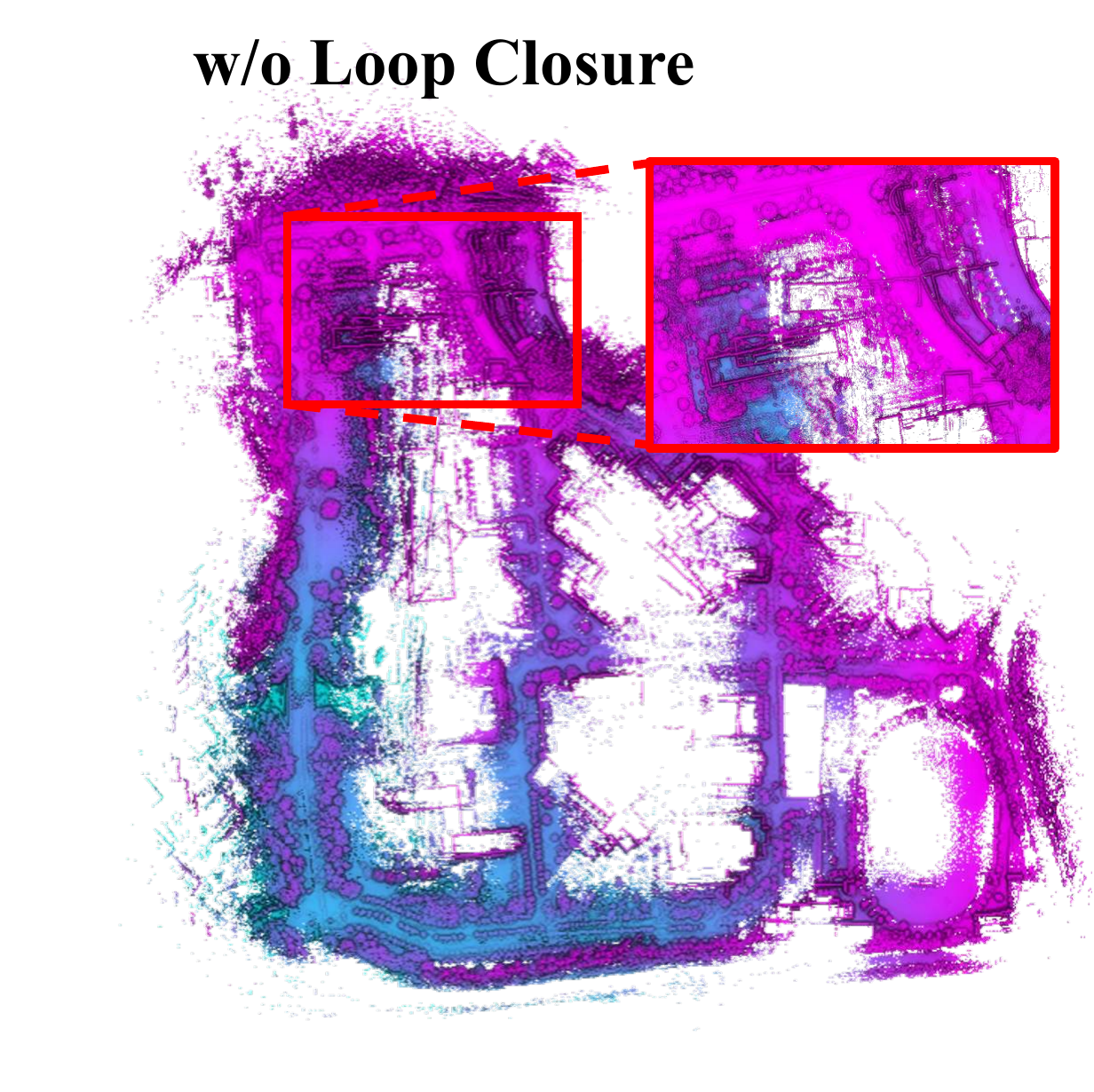}
        \vspace{-1.5em}
        \caption{}
    \end{subfigure}
    \begin{subfigure}{0.49\columnwidth}
        \includegraphics[trim={30 0 0 0},clip, width=\columnwidth]{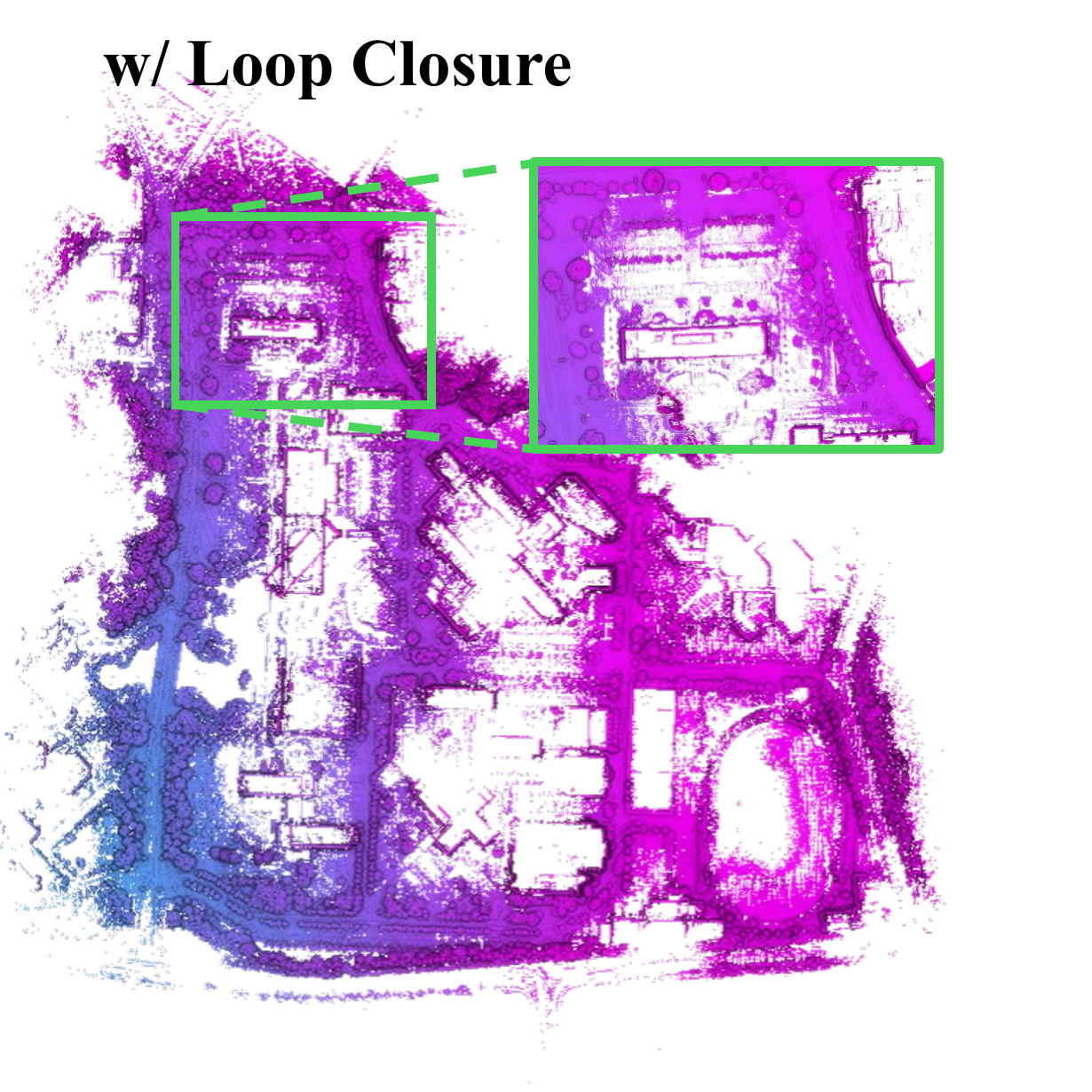}
        \vspace{-1.5em}
        \caption{}
    \end{subfigure}
    \caption{Qualitative comparison of LiDAR map reconstruction on the Mulran dataset, which is acquired by an Ouster OS1-64 sensor \citep{kim2020mulran}.
    (a) Without loop closure. Accumulated drift leads to noticeable misalignments and structural inconsistencies (red box).
    (b) With loop closure. The global map exhibits improved structural alignment and reduced distortion, with well-aligned overlapping regions (green box).  
    This demonstrates that SOLiD-based loop closure effectively maintains global consistency and enables accurate map alignment.}
    \label{fig:drift}
      \vspace{-3mm}
\end{figure}

To correct the accumulated drift in odometry as shown in \figref{fig:drift}(a), we incorporate loop closure constraints, identified through place recognition using our lightweight global descriptor, SOLiD.
As explained in Section~\ref{sec:notation}, for each $i$-th keyframe point cloud, $\mathbf{P}^\alpha_i$, we extract a compact descriptor $\mathbf{D}^\alpha_i$ using the SOLiD generation function, $f_{\text{SOLiD}}(\cdot)$.
For a detailed description of SOLiD, please refer to our previous paper \citep{kim2024narrowing}.

To efficiently retrieve loop closure candidates, we build a Kd-tree over the descriptor set $\mathcal{D}^\alpha=\{\mathbf{D}^\alpha_i\mid i\in\mathcal{I}^\alpha\}$.
Given the $a$-th frame $\mathbf{D}^\alpha_a$ as a query, we search over the index set excluding a short temporal window (to avoid trivial matches) as follows:
\begin{equation}
b \;=\; \operatorname*{arg\,min}_{\,b' \in \mathcal{T}_a}\ \big\| \mathbf{D}^\alpha_a - \mathbf{D}^\alpha_{b'} \big\|_2,
\label{eq:past-nn}
\end{equation}
where $\mathcal{T}_a \coloneqq \big\{\varphi \in \mathcal{I}^\alpha \ \big|\ 1 \le \varphi \le a-\tau_{\text{window}} \big\}$ and $\tau_{\text{window}}\in\mathbb{N}$ is a user-defined parameter.
This ignores indices $\varphi\in\{a-\tau_{\text{window}}+1,\dots,a\}$ (\ie the current frame and the last $\tau_{\text{window}}-1$ past frames).
If $a-\tau_{\text{window}}<1$, then $\mathcal{T}_a=\varnothing$ and no candidate is returned.
A Kd-tree built over $\{\mathbf{D}^\alpha_\varphi\}_{\varphi\in\mathcal{T}^\alpha}$ is used only to accelerate the search over $\mathcal{T}_a$.
We then accept a loop candidate only if the descriptor distance is below a threshold using \eqref{eq:f_sim}, as follows:
\begin{equation}
f_{\text{sim}}(\mathbf{D}^\alpha_a, \mathbf{D}^\alpha_b) = \big\|\mathbf{D}^\alpha_a - \mathbf{D}^\alpha_b\big\|_2 \;\leq\; \tau_{\text{SOLiD}}^{\text{intra}}.
\label{eq:intra-gate}
\end{equation}

While each loop candidate $(a,b)$ is retrieved from descriptor similarity, we verify it geometrically via point-to-point ICP, which estimates the relative poses $\mathbf{R}^{\text{ICP}}_{a,b},\,\mathbf{t}^{\text{ICP}}_{a,b}$ between the associated point clouds $\mathbf{P}^\alpha_a$ and $\mathbf{P}^\alpha_b$ as follows:
\begin{equation}
\mathbf{R}^{\text{ICP}}_{a,b},\,\mathbf{t}^{\text{ICP}}_{a,b} = \operatorname*{arg\,min}_{\substack{\mathbf{R}\in\text{SO}(3)\\ \mathbf{t}\in\mathbb{R}^3}}
\sum_{(\mathbf{p}_\varpi,\mathbf{q}_\varpi)\in \mathcal{C}^{\text{inlier}}_{a,b}}
\Big\|\, (\mathbf{R}\,\mathbf{p}_\varpi + \mathbf{t}) - \mathbf{q}_\varpi \,\Big\|_2^2,
\label{eq:icp}
\end{equation}
where $\mathcal{C}^{\text{inlier}}_{a,b}=\big\{(\mathbf{p}_\varpi,\mathbf{q}_\varpi)\big\}_{\varpi=1}^{N_{\text{iter}}}$ is the inlier correspondence set, which is updated at each ICP iteration, with $N_{\text{iter}}$ pairs $\mathbf{p}_\varpi\in\mathbf{P}^\alpha_a$ and $\mathbf{q}_\varpi\in\mathbf{P}^\alpha_b$.
During each iteration of ICP, a total of $N_{\text{iter}}$ candidate correspondences $\mathcal{C}_{a,b}=\big\{(\mathbf{p}_\varpi,\mathbf{q}_\varpi)\big\}_{\varpi=1}^{N_{\text{iter}}}$ are obtained by Kd-tree-based point-to-point nearest neighbor search on the warped points; the inlier set $\mathcal{C}^{\text{inlier}}_{a,b}\subseteq\mathcal{C}_{a,b}$ is then produced via distance-based thresholding and/or a robust kernel.

We accept the geometric verification if the mean squared residual is below a threshold as follows:
\begin{equation}
\varepsilon^{\text{ICP}}_{a,b}
\coloneqq \frac{1}{N_{\text{inlier}}}
\sum_{(\mathbf{p}_\varpi,\mathbf{q}_\varpi)\in \mathcal{C}^{\text{inlier}}_{a,b}}
\Big\|\, (\mathbf{R}^{\text{ICP}}_{a,b}\,\mathbf{p}_\varpi + \mathbf{t}^{\text{ICP}}_{a,b}) - \mathbf{q}_\varpi \,\Big\|_2^2,
\end{equation}
\begin{equation}
\varepsilon^{\text{ICP}}_{a,b} \;\leq\; \tau_{\text{ICP}}.
\label{eq:icp-accept}
\end{equation}
If \eqref{eq:icp-accept} holds, we set the loop measurement $\mathbf{T}^{\text{loop}}_{a,b}\coloneqq\begin{bmatrix} \mathbf{R}^{\text{ICP}}_{a,b} & \mathbf{t}^{\text{ICP}}_{a,b} \\ \mathbf{0}^\top & 1 \end{bmatrix}$ and add $(a,b)$ to $\mathcal{C}_{\text{intra}}^{\text{loop}}$ with the corresponding information matrix $\boldsymbol{\Omega}^{\text{loop}}_{a,b}$ based on the point-wise residuals.

In summary, each observation tuple consists of a pose, a point cloud, and a global descriptor, formally defined as:
\begin{equation}
v^\alpha_i = \left( \mathbf{T}^{\alpha}_i,\; \mathbf{P}^\alpha_i,\; \mathbf{D}^\alpha_i \right), \quad i = 1, \dots, N,
\label{eq:observation}
\end{equation}
where the SLAM state can be equivalently represented as a triplet of sets, $\mathcal{V}^\alpha= \left( \mathcal{X}^{\alpha*},\; \mathcal{P}^\alpha,\; \mathcal{D}^\alpha \right)$, with $\mathcal{V}^{\alpha} = \{v^{\alpha}_i\}_{i=1}^N$, $\mathcal{X}^{\alpha*} = \{\mathbf{T}^{\alpha}_i\}_{i=1}^N$, $\mathcal{P}^\alpha = \{\mathbf{P}^\alpha_i\}_{i=1}^N$, and $\mathcal{D}^\alpha = \{\mathbf{D}^\alpha_i\}_{i=1}^N$ \big(resp. $\mathcal{U}^\beta = \left( \mathcal{X}^{\beta*},\; \mathcal{P}^\beta,\; \mathcal{D}^\beta \right)$\big).
As illustrated in \figref{fig:drift}(b), our SOLiD-based SLAM demonstrates robust, lightweight, and efficient performance, effectively mitigating drift.
Moreover, it maintains robust performance across diverse LiDAR scan types and configurations; see our previous paper \citep{kim2024narrowing}.

As shown in \figref{fig:main}(a), each robot in the team performs intra-robot SLAM while communicating with the server by transmitting SOLiD descriptor database $\mathcal{D}^\alpha$ (resp. $\mathcal{D}^\beta$) and pose database~$\mathcal{X}^{\alpha*}$ (resp. $\mathcal{X}^{\beta*}$).
The server then generates a selective transmission policy (see Section~\ref{sec:inter_pr}, Section~\ref{sec:seq_match}, Section~\ref{sec:vertex_cover}, and Section~\ref{sec:edge_clique}), instructing robots to send only partial scan subsets (see Section~\ref{sec:exchange_policy}) rather than the entire scan database~$\mathcal{P}^\alpha$~(resp. $\mathcal{P}^\beta$).
% ==================================================================================================================================================================================

% =============================================================
% =============================================================

% ==================================================================================================================================================================================
\subsection{Inter-Robot Place Recognition}
\label{sec:inter_pr}
\noindent
From the received descriptor databases $\mathcal{D}^\alpha$ and $\mathcal{D}^\beta$ from each robot, the server conducts inter-robot place recognition.
This section serves as a preprocessing step for $\stepA$, $\stepB$, and $\stepC$ defined in Section~\ref{sec:problem}, constructing the affinity matrix used to determine the sequentially overlapped subset of loop closures.

Let $\mathcal{R} \in \{1, \dots, r, \dots, N_{\text{robot}}\}$ denote the robot index~(\eg $r = \alpha, \beta$), and define the set of keyframes for robot~$r$ among deployed $N_{\text{robot}}$ robots.
Then, we can let the observations $\mathcal{V}^\alpha = \{ v_i^\alpha \}_{i=1}^{N}$ and $\mathcal{U}^\beta = \{ u_j^\beta \}_{j=1}^{M}$ in \eqref{eq:observation}, where $i$ and $j$ are used to index the keyframes from robot~$\alpha$ and robot~$\beta$, respectively, as presented in Section~\ref{sec:intra_slam}.  

Given the global descriptors sets $\mathcal{D}^\alpha = \{ \mathbf{D}_i^\alpha \}_{i=1}^{N}$ and $\mathcal{D}^\beta = \{ \mathbf{D}_j^\beta \}_{j=1}^{M}$ from observation $\mathcal{V}^\alpha$ and $\mathcal{U}^\beta$ in \eqref{eq:observation}, we measure inter-robot similarity using \eqref{eq:f_sim} as follows:
\begin{equation}
f_\text{sim}(\mathbf{D}_i^\alpha, \mathbf{D}_j^\beta) = \big\| \mathbf{D}_i^\alpha - \mathbf{D}_j^\beta \big\|_2 .
\end{equation}

We collect similarity scores over all index pairs in the complete candidate domain $\mathcal{L}\coloneqq\mathcal{I}^\alpha\times\mathcal{J}^\beta$. 
These scores are arranged in an affinity matrix as follows:
\begin{equation}
\mathbf{A}_{i,j}=f_{\mathrm{sim}}(\mathbf{D}^\alpha_i,\mathbf{D}^\beta_j), \quad \forall (i,j)\in\mathcal{L},
\label{eq:affinity}
\end{equation}
where $\mathbf{A}\in\mathbb{R}^{|\mathcal{I}^\alpha|\times|\mathcal{J}^\beta|}$ is the affinity matrix.
Low values in $\mathbf{A}$ indicate strong place matches and serve as loop closure candidates as shown in \figref{fig:inter_pr}(a).
In subsequent processing stages, we use $\mathbf{A}$ for matrix-based computations, while $\mathcal{L}$ and its subsets represent the corresponding set-theoretic formulation.

Unlike in Section~\ref{sec:intra_slam}, we avoid Kd-tree-based approximate retrieval because sparse or approximate neighbor selection can break the diagonal connectivity patterns in $\mathbf{B}$ that are essential for sequential overlap detection.
Specifically, we need to preserve diagonal connections at positions $(i,j)$, $(i\pm1,j\pm1)$ in the affinity matrix to identify coherent sequential matches.
Instead, leveraging SOLiD's comparably fast computation advantages, we compute the dense affinity through brute force scoring to examine all possible correspondences.
The resulting dense affinity matrix facilitates robust sequential overlap analysis by preserving complete spatial connectivity patterns in Section~\ref{sec:seq_match}.
% ==================================================================================================================================================================================

% ==================================================================================================================================================================================
\subsection{Inter-Robot Sequential Matching}
\label{sec:seq_match}
\noindent
We now provide a detailed procedure of $\stepA$, the first stage of our~cascaded three-stage approach introduced in Section~\ref{sec:problem}. 
This step introduces sequential matching for robust inter-robot place recognition by detecting temporally contiguous correspondence, defining the sequential set $\mathcal{L}_{\text{seq}}$ in \eqref{eq:seq}.

Formally, given $\mathbf{A}$, each of whose elements are defined by~\eqref{eq:affinity}, we first threshold it to obtain a binary matrix $\mathbf{B}\in\{0,1\}^{N\times M}$ whose each element is defined as follows:
\begin{equation}
\mathbf{B}_{i,j} =
\begin{cases}
1 & \text{if } \mathbf{A}_{i,j} \leq \tau_{\text{SOLiD}}^{\text{inter}}, \\
0 & \text{otherwise},
\end{cases}
\label{eq:binary}
\end{equation}
where $\tau_{\text{SOLiD}}^{\text{inter}}$ is a user-defined threshold that filters out low confidence pairs.
Given $\mathbf{B}$, we apply 8-connected component labeling with a structuring element $\mathbf{S} =
\begin{bmatrix}
1 & 1 & 1 \\
1 & 1 & 1 \\
1 & 1 & 1
\end{bmatrix}$ to cluster connected regions.
This algorithm groups neighboring 1-valued cells into clusters. 
Each cluster represents contiguous regions of strong place matches~(similarities lower $\tau_{\text{SOLiD}}^{\text{inter}}$) forming coherent sequences of matches between the trajectories of the two robots.
This produces $N_{\text{label}}$ clusters, where they are assigned a label $\ell \in \{1,\dots,N_{\text{label}}\}$.
Importantly, each cluster contains sequential rows and columns, that is, sequential nodes from robot $\alpha$ and robot $\beta$, where each node contributes to forming at least one inter-robot loop within the cluster.
This inter-robot relationship is leveraged in Section~\ref{sec:vertex_cover}.

\begin{figure}[t]
    \centering
    \captionsetup{justification=justified}
    \captionsetup[subfigure]{justification=centering}
    \begin{subfigure}{0.44\columnwidth}
        \includegraphics[trim={0 0 0 0},clip, width=\columnwidth]{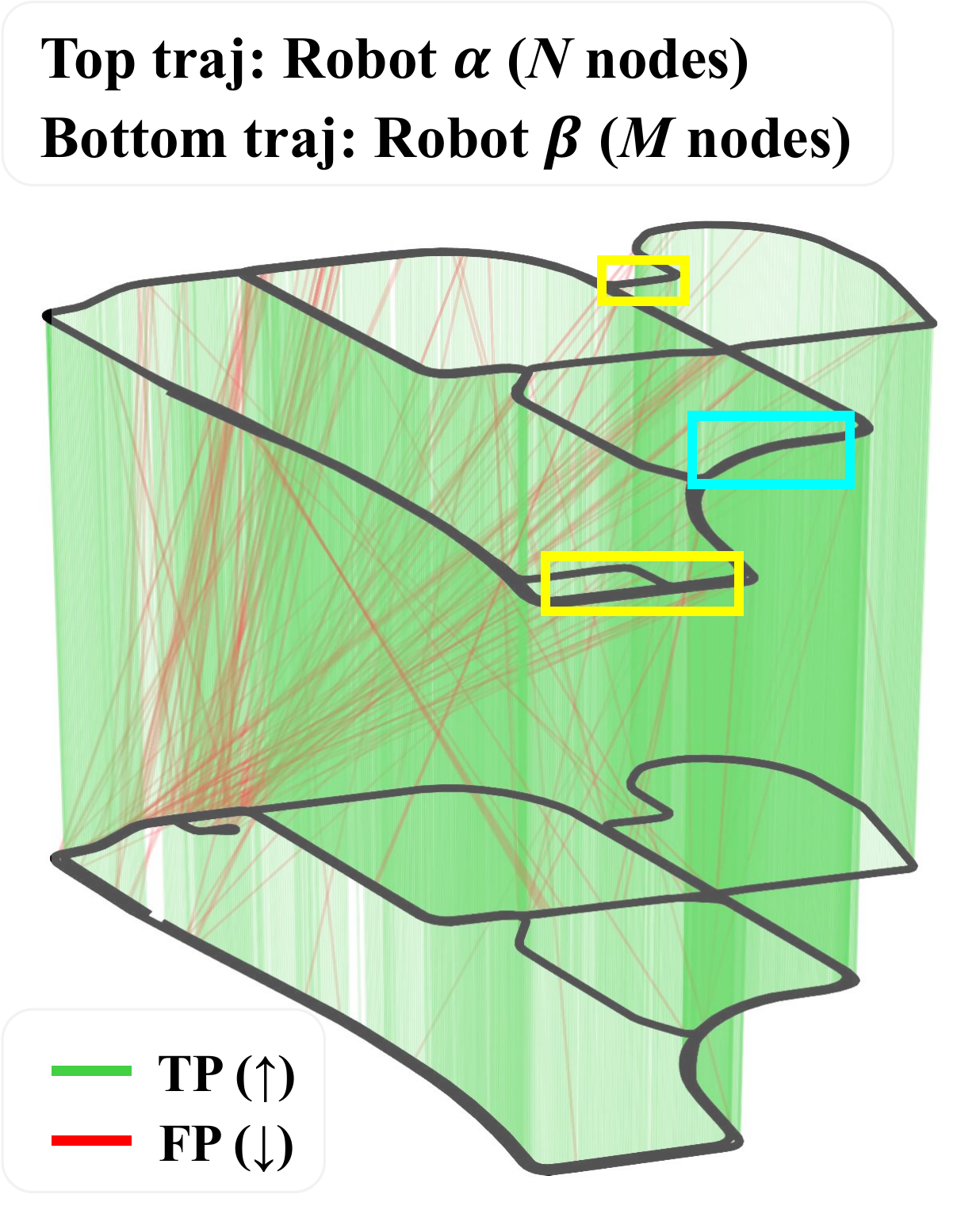}
        \vspace{-1.5em}
        \caption{}
    \end{subfigure}
    \begin{subfigure}{0.54\columnwidth}
        \includegraphics[trim={0 0 0 0},clip, width=\columnwidth]{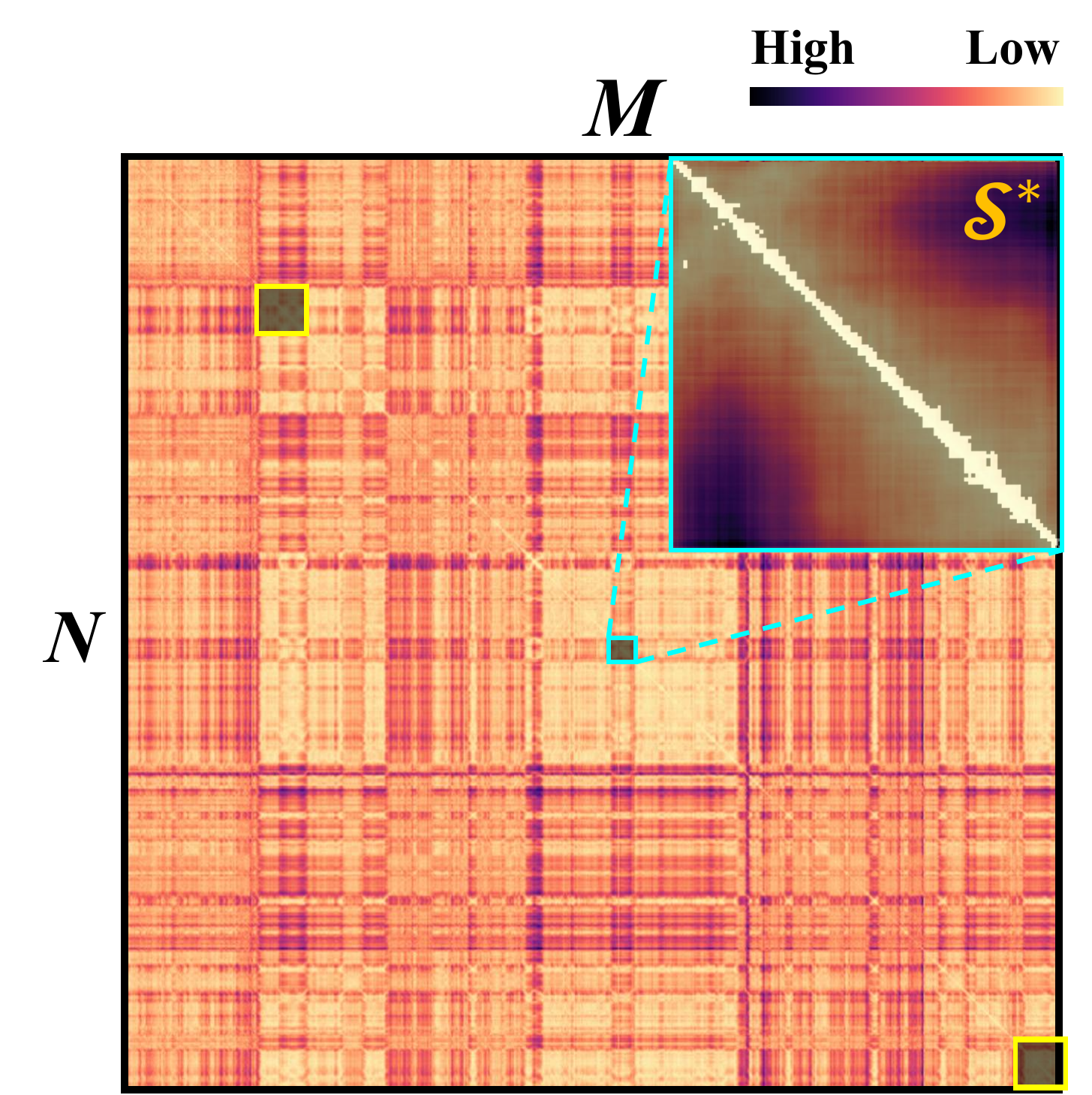}
        \vspace{-1.5em}
        \caption{}
    \end{subfigure}
    \caption{Visualization of inter-robot place recognition results and the corresponding affinity matrix (see Section~\ref{sec:inter_pr}). 
           (a) Detected inter-robot loop closures between trajectories consisting of $N$ nodes (robot $\alpha$) and $M$ nodes (robot $\beta$), where green edges represent true positives and red edges represent false positives. 
           (b) $N \times M$ affinity matrix~$\mathbf{A}$ constructed from SOLiD descriptor distances, where the $N$ columns correspond to robot $\alpha$'s descriptors and the $M$ rows correspond to robot $\beta$'s descriptors. 
           The cyan box highlights a sequentially consistent low distance region, $\mathcal{S}_\ell$, grouped by connected white diagonal patterns $\mathbf{S}$ in binary matrix~$\mathbf{B}$ representing sequential matches that form a coherent cluster.
           This cluster becomes a candidate subset for $\mathcal{S}^*$, whereas the yellow box shows scattered high-distance responses that are excluded during connected component labeling (see Section~\ref{sec:seq_match}).}
  \vspace{-3mm}
  \label{fig:inter_pr}
\end{figure}

Let $\mathcal{S} = \{ \mathcal{S}_\ell \}_{\ell=1}^{N_{\text{label}}}$ denote the set of all labeled clusters.
To select the most informative cluster (\ie $\mathcal{L}_{\text{seq}}$) from $\mathcal{S}$, for each cluster $\mathcal{S}_\ell$, we first compute its rectangular span in the affinity matrix $\mathbf{A}$ as follows:
\begin{equation}
\mathrm{Span}(\mathcal{S}_\ell)
= \big[i^{(\ell)}_{\min},\, i^{(\ell)}_{\max}\big)
\times
\big[j^{(\ell)}_{\min},\, j^{(\ell)}_{\max}\big),
\end{equation}
where $\mathrm{Span}(\cdot)$ is obtained from the minimum and maximum row and column indices over all pairs $(i,j)\in\mathcal{S}_\ell$.
We then score each cluster by its average similarity as follows:
\begin{equation}
f_{\mathrm{avg}}(\mathcal{S}_\ell)
= \frac{1}{|\mathcal{S}_\ell|}\sum_{(i,j)\in\mathcal{S}_\ell}\mathbf{A}_{i,j}.
\end{equation}
Here, $|\cdot|$ denotes the cardinality of the set.
As illustrated in \figref{fig:inter_pr}(b), we pick the best one among $N_{\text{label}}$ clusters whose span area exceeds a threshold $\tau_{\text{area}}$ as follows:
\begin{equation}
\mathcal{S}^*
= \operatorname*{arg\,min}_{\ell\;:\;|\mathrm{Span}(\mathcal{S}_\ell)|\ge \tau_{\text{area}}}
f_{\mathrm{avg}}(\mathcal{S}_\ell).
\label{eq:final_seq}
\end{equation}
Here, $|\mathrm{Span}(\cdot)|$ denotes the area of the rectangular region.
Finally, we set the most informative subset of sequentially matched loop candidates to $\mathcal{L}_{\text{seq}} \coloneqq \mathcal{S}^*$.

% =======================================================
% =======================================================

In summary, from the input $\mathcal{L}$ in \eqref{eq:affinity}, we compute the affinity matrix $\mathbf{A}$. 
Subsequently, using $\mathbf{A}$ as input, we obtain clusters $\mathcal{S}$ through \eqref{eq:binary} and $\mathbf{S}$. 
Finally, we identify the most informative cluster via \eqref{eq:final_seq}. 
This process implements $J_{\text{seq}}(\cdot)$ from \eqref{eq:seq}, taking the input $\mathcal{L}$ and completing the sequential matching stage, fulfilling $\stepA$ of our framework.
Following the index convention in Section~\ref{sec:notation}, we relabel pairs in $\mathcal{L}_{\text{seq}}$ as $(c,q)$ with $c\in\mathcal{I}^{\alpha}_{\text{seq}}$ and $q\in\mathcal{J}^{\beta}_{\text{seq}}$ for robots $\alpha$ and $\beta$, respectively.

% ==================================================================================================================================================================================

% ==================================================================================================================================================================================
\subsection{Balanced Vertex Cover in Exchange Graph}
\label{sec:vertex_cover}
\noindent
Building on the sequential matching results from $\stepA$, we now present $\stepB$ of our cascaded framework explained in Section~\ref{sec:problem}.
Given a set of potential inter-robot loop closures $\mathcal{L}_{\text{seq}}$ obtained from sequential matching in \eqref{eq:final_seq}, the next critical challenge is to select which subset of scans should be transmitted among robots for efficient communication.
We exploit the key property from Section~\ref{sec:seq_match} that every sequential node participates in loop formation. Our sequential matching design explicitly exploits this property to construct a \emph{complete matching} on a selected partition.

In real-world deployments, continuously exchanging complete scans is prohibitively expensive due to bandwidth, energy, and CPU constraints. 
Inspired by \citet{giamou2018talk}, which formulates data exchange as a graph-theoretic optimization problem, we can cast the scan selection problem as a minimum weighted vertex cover (MWVC) on the bipartite exchange graph in \eqref{eq:seq_exchange_graph}: choose a minimum cost subset of vertices so that the selected scans collectively cover all candidate loop edges in \eqref{eq:edge}, thereby maintaining complete spatial coverage while minimizing transmission cost. 

Formally, let $\pi:\mathcal{V}^{\alpha}\cup\mathcal{U}^{\beta}\to\{0,1\}$ indicate whether a scan is transmitted and let $w_x>0$ denote its transmission cost (\eg CPU time or network bandwidth), where $x$ represents any vertex in our exchange graph.
The MWVC is employed as follows:
\begin{equation} \label{eq:mwvc}
\begin{aligned}
\pi^\star = \operatorname*{arg\,min}_{\pi} & \quad \sum_{x\in\mathcal{V}^{\alpha}_{\text{seq}} \cup \, \mathcal{U}^{\beta}_{\text{seq}}} w_x\,\pi(x) \\
\text{s.t.}\quad
& \pi(v^{\alpha}_c)+\pi(u^{\beta}_q)\;\ge\;1,\
\forall\,\{v^{\alpha}_c,u^{\beta}_q\}\in\mathcal{E}^{\alpha,\beta}_{\text{seq}},\\
& \pi(x)\in\{0,1\},\quad \forall\,x\in\mathcal{V}^{\alpha}_{\text{seq}} \cup \, \mathcal{U}^{\beta}_{\text{seq}}.
\end{aligned}
\end{equation}

However, as discussed in Section~\ref{sec:comm_efficient}, assigning the transmission cost $w_x$ reliably is infeasible in practice because CPU time and wireless bandwidth fluctuate during operation~\citep{lajoie2022towards, kulkarni2019deepchannel, gielis2022critical, formis2023predicting}.

To avoid these limitations, we adopt uniform weights $w_x=1$ for all $x$. 
This strategy prioritizes minimizing worst-case system complexity over optimizing for uncertain metrics, making the vertex count a robust proxy for communication cost while simplifying the problem into an~(unweighted) minimum vertex cover (MVC).
Under this assumption, \eqref{eq:mwvc} reduces to the MVC as follows:
\begin{equation} \label{eq:mvc}
\begin{aligned}
\pi^\star = \operatorname*{arg\,min}_{\pi} &  \quad
\sum_{x \in \mathcal{V}^{\alpha}_{\text{seq}} \cup \, \mathcal{U}^{\beta}_{\text{seq}}} \pi(x) \\
\text{s.t.}\quad
& \pi(v^{\alpha}_c)+\pi(u^{\beta}_q)\;\ge\;1,\quad
\forall\,\{v^{\alpha}_c,u^{\beta}_q\}\in\mathcal{E}^{\alpha,\beta}_{\text{seq}}, \\ 
& \pi(x)\in\{0,1\},\quad \forall\,x\in\mathcal{V}^{\alpha}_{\text{seq}} \cup \, \mathcal{U}^{\beta}_{\text{seq}}.
\end{aligned}
\end{equation}
By doing so, this design allows us to leverage fundamental graph-theoretic results.
Recall from Section~\ref{sec:seq_match} that every sequential node from both robots participates in at least one inter-robot loop formation within $\mathcal{L}_{\text{seq}}$, which removes isolated vertices and yields well-formed bipartite components.

Specifically, in our multi-robot setting, inter-robot loop closures are established by pairing keyframes from robot~$\alpha$ and $\beta$, which naturally form a bipartite exchange graph.
For each connected component produced by the sequential matching process (Section~\ref{sec:seq_match}), we consider the two keyframe sets $\mathcal{V}^{\alpha}_{\text{seq}}$ and $\mathcal{U}^{\beta}_{\text{seq}}$ with cardinalities $N_{\text{seq}}$ and $M_{\text{seq}}$, respectively.

Without loss of generality, assume $N_{\text{seq}} \leq M_{\text{seq}}$.
By construction, the sequential matching process ensures that each connected component admits a \emph{complete matching} on the smaller partition.
Under this component-wise selection, the selected partition is fully covered by a matching, and Hall’s condition~\citep{schrijver2003combinatorial} holds for that partition.

Formally, for a bipartite graph $\mathcal{G} = (\mathcal{X} \cup \mathcal{Y}, \mathcal{E})$, there exists a matching that covers all vertices in $\mathcal{X}$ if and only if for every subset $\mathcal{Z} \subseteq \mathcal{X}$, $|N(\mathcal{Z})| \ge |\mathcal{Z}|$, as characterized by Hall's theorem.
Consequently, the maximum matching size for the selected partition equals $N_{\text{MM}} = |\mathcal{X}|$, \ie $N_{\text{MM}} = N_{\text{seq}}$ (resp. $N_{\text{MM}} = M_{\text{seq}}$ when $M_{\text{seq}} \leq N_{\text{seq}}$).

% ===============================================================================
\begin{figure*}[t]
    \centering
    \captionsetup{justification=justified}
    \captionsetup[subfigure]{justification=centering}
    \begin{subfigure}{0.99\columnwidth}
        \includegraphics[trim={0 0 0 0},clip, width=\columnwidth]{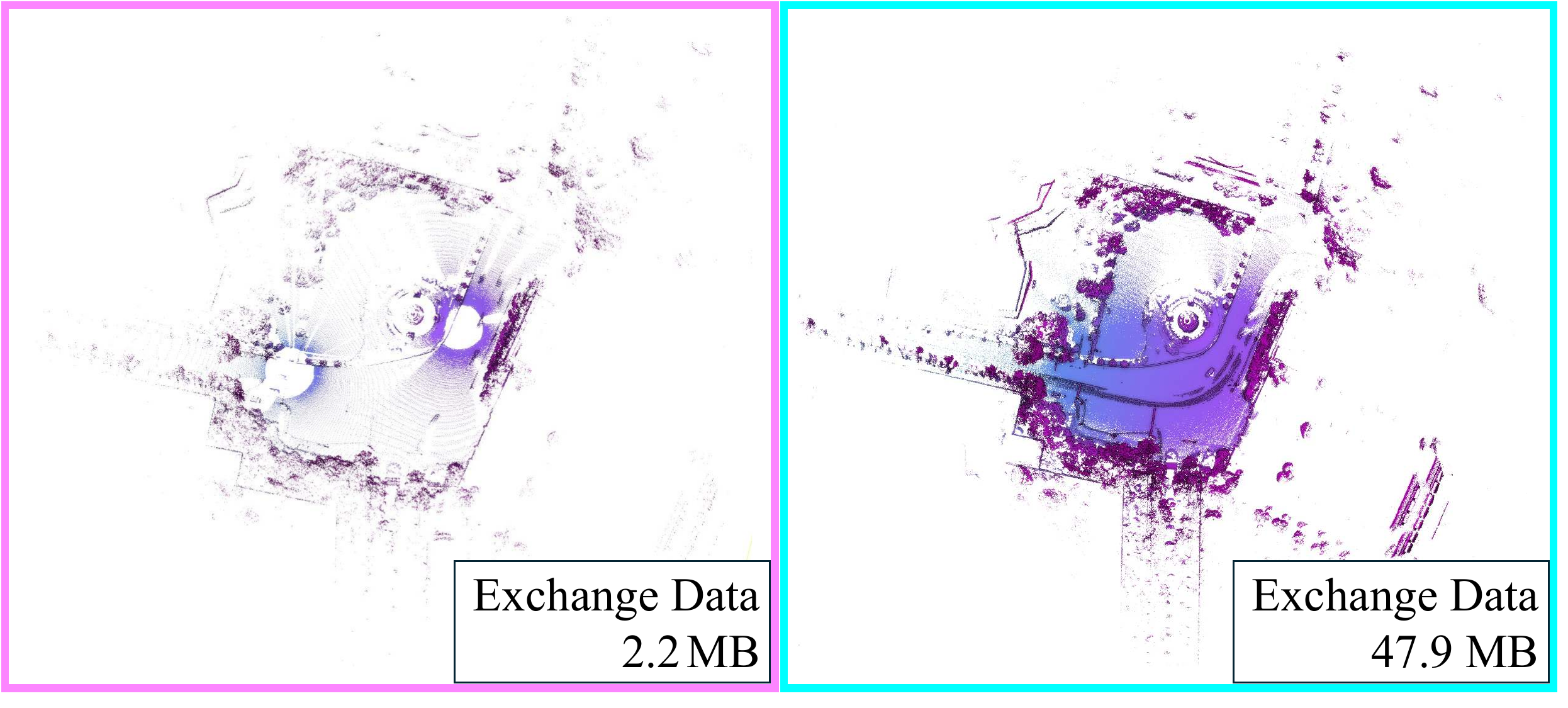}
        \vspace{-1.5em}
        \caption{Minimum Vertex Cover (MVC)}
    \end{subfigure}
    \begin{subfigure}{0.99\columnwidth}
        \includegraphics[trim={0 0 0 0},clip, width=\columnwidth]{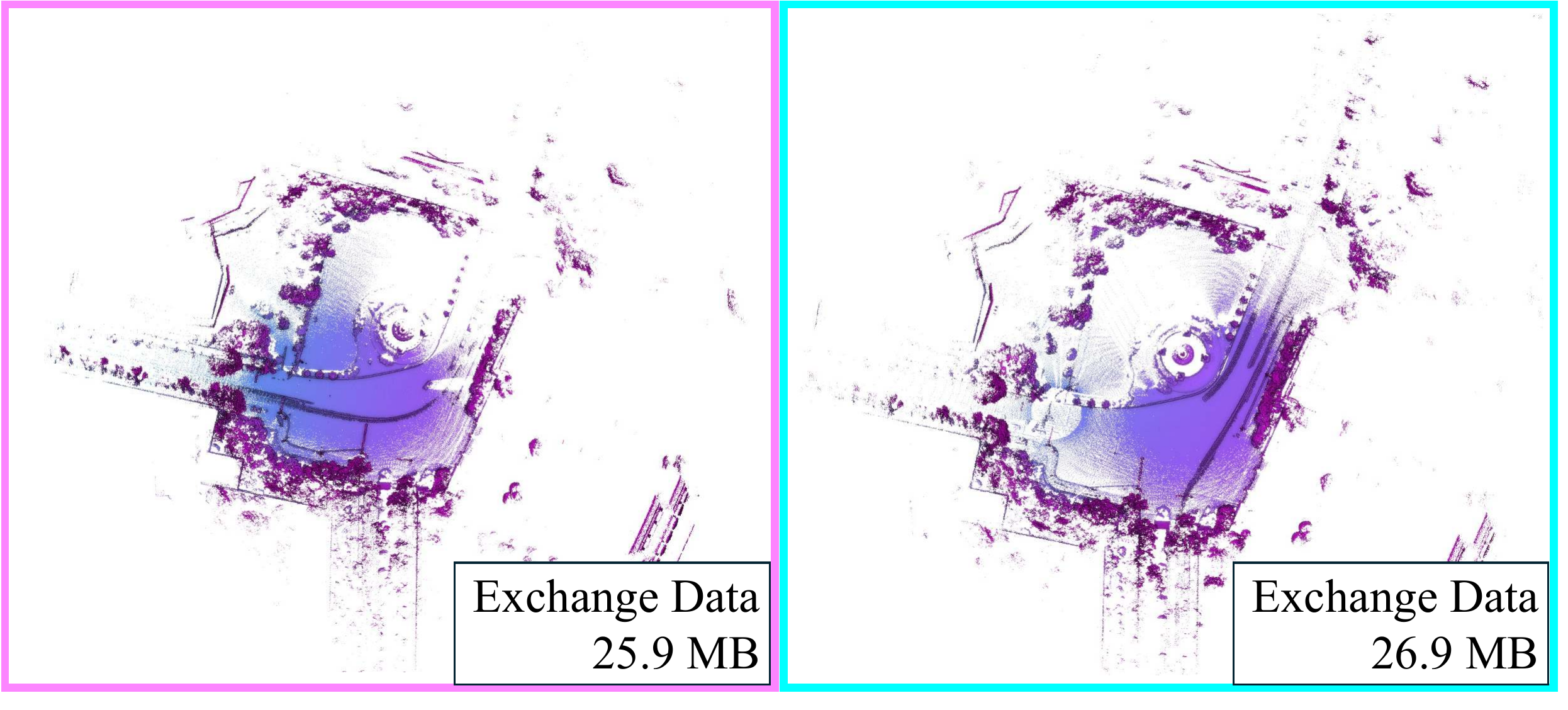}
        \vspace{-1.5em}
        \caption{Balanced Vertex Cover (BVC; Proposed)}
    \end{subfigure}
    \includegraphics[trim={0 0 0 0},clip, width=0.7\columnwidth]{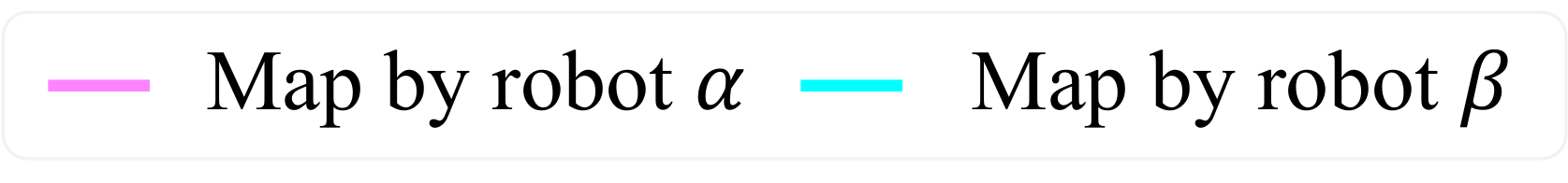}
  \caption{Comparison of submaps generated under (a)~standard minimum vertex cover (MVC) and (b)~balanced vertex cover~(BVC). 
  The magenta and cyan frames indicate scans from robot $\alpha$ and robot $\beta$, respectively, with the transmitted data size shown in MB. 
  While standard MVC achieves lower total transmission volume (2.2\,+\,47.9\,=\,50.1\,MB vs 25.9\,+\,26.9\,=\,52.8\,MB), it produces submaps with reduced overlap despite being taken from the same location, increasing the likelihood of failure in the submap merging stage. 
  Moreover, standard MVC creates monologue transmissions where one robot sends disproportionately large data volumes (2.2\,MB vs 47.9\,MB), concentrating transmission on a single robot, which can be undesirable under practical resource constraints. 
  In contrast, BVC ensures more even data distribution (25.9\,MB vs 26.9\,MB) and maintains sufficient overlap to achieve better registration, demonstrating the importance of geometric diversity over pure transmission minimization.}
    \label{fig:bmvc}
    \vspace{-3mm}
\end{figure*}
% ===============================================================================

We consider the resulting unweighted bipartite exchange graph.
By König's theorem~\citep{schrijver2003combinatorial}, which states that in unweighted bipartite graphs the size of a minimum vertex cover equals the size of a maximum matching, the minimum vertex cover size is given as follows:
\begin{equation}
N_{\text{vertex}} = N_{\text{MM}}.
\end{equation}
This result provides a tight lower bound on the number of scans that must be transmitted to preserve full coverage of all candidate inter-robot loop closures.

While this lower bound is theoretically optimal, standard MVC optimization can lead to a monologue problem, where the solution selects vertices predominantly from one partition while ignoring the other. 
For example, if robot $\alpha$ has keyframes with higher connectivity, the MVC might select most observations from $\mathcal{V}^{\alpha}_{\text{seq}}$ and few from $\,\mathcal{U}^{\beta}_{\text{seq}}$ (\figref{fig:bmvc}(a)). 
Consequently, it leads to unbalanced communication loads and reduces the geometric diversity of the selected scans, eventually degrading subsequent registration quality.

To avoid these issues while leveraging the tight bound $N_{\text{MM}}$, we define a BVC that prevents the monologue solution by constraining each robot to contribute a minimum share of the selected scans.
To this end, we employ Shannon entropy~\citep{shannon1948mathematical} as a principled measure of participation balance across robots.

Shannon entropy for a discrete distribution $p = \{p_r\}_{r=1}^{N_{\text{robot}}}$ is defined as follows:
\begin{equation}
H(p) = -\sum_{r} p_r \log_2 p_r,
\end{equation}
where $p_r$ represents an abstract proportion of contribution assigned to robot $r$ for balancing purposes.
The effective number of participating robots is given by $2^{H(p)}$.

Using this entropy-derived quantity, we impose the balance constraints as follows:
\begin{equation}
    \sum_{v \in \mathcal{V}^{\alpha}_{\text{seq}}} \pi(v) \ge \frac{N_{\text{MM}}}{2^{H(p)}}, \qquad
    \sum_{u \in \, \mathcal{U}^{\beta}_{\text{seq}}} \pi(u) \ge \frac{N_{\text{MM}}}{2^{H(p)}}.
\label{eq:balanced}
\end{equation}
Here, $p_r$ represents an abstract participation ratio used solely to prevent degenerate (monologue) solutions in scan selection.
Since our objective is scan cardinality rather than bandwidth, we do not interpret $p_r$ as an online estimate of link capacity.
For the two-robot setting in our experiments, we set $p_\alpha = p_\beta = 1/2$, yielding an equal-share constraint that promotes geometric diversity from both robots.

This choice encourages balanced scan selection from both robots and avoids degenerate solutions dominated by one robot.
We incorporate these balance constraints into~\eqref{eq:mvc}, yielding the BVC formulation.
Although our experiments focus on two robots (where $p_\alpha=p_\beta=1/2$ and the constraint reduces to an equal-share form), the entropy-based formulation provides a principled way to generalize the minimum-share constraint to larger teams and asymmetric participation settings \citep{cover1999elements,kelly1998rate}.

With these constraints, the BVC selects a minimum cost set of scans that still covers every candidate loop edge (full coverage of $\mathcal{E}^{\alpha,\beta}_{\text{seq}}$), while preventing monologue solutions where one robot dominates the transmission as shown in \figref{fig:bmvc}(b).
This balance ensures that subsequent clique selection in Section~\ref{sec:edge_clique} has access to geometrically diverse scans from both robots, improving registration robustness.
In addition, this approach distributes communication loads evenly across robots, preventing single-robot dominance and providing more stable, implementation-friendly scan exchange behavior.

In summary, given the sequential matching result $\mathcal{L}_{\text{seq}}$ from \eqref{eq:final_seq}, we extract the vertices incident to each edge defined in \eqref{eq:edge}.
We select these vertices by solving the BVC formulation: the optimization $J_{\text{V}}(\cdot)$ in \eqref{eq:vertex} minimizes the objective in \eqref{eq:mvc} while satisfying the balance constraints in \eqref{eq:balanced}, yielding the optimal solution $\pi^\star(\cdot)$. 
From $\pi^\star(\cdot)$, we construct the vertex cover sets as follows:
\begin{align}
\mathcal{V}^{\alpha}_{\text{V}} = \Big\{ v^{\alpha}_c \in \mathcal{V}^{\alpha}_{\text{seq}}  \mid \pi^\star(v^{\alpha}_c) = 1 \Big\}, \\
\mathcal{U}^{\beta}_{\text{V}}  = \Big\{ u^{\beta}_q  \in \,\mathcal{U}^{\beta}_{\text{seq}} \mid \pi^\star(u^{\beta}_q)  = 1 \Big\}.
\end{align}
The resulting loop set $\mathcal{L}_{\text{V}}$ contains all edges whose at least one endpoint is selected as follows:
\begin{equation}
\mathcal{L}_{\text{V}} = \Big\{ (c,q) \in \mathcal{L}_{\text{seq}} \mid v^{\alpha}_c \in \mathcal{V}^{\alpha}_{\text{V}} \text{ or } u^{\beta}_q \in \mathcal{U}^{\beta}_{\text{V}} \Big\}.
\label{eq:balanced_loop}
\end{equation}
This completes $\stepB$ of our three-stage framework.

To avoid notation clutter, we use tildes for indices of selected nodes at the BVC stage: ${\mathcal{I}}^\alpha_{\text{V}} := \{\, \tilde{c}\ \mid\ v^\alpha_{\tilde{c}}\in\mathcal{V}^\alpha_{\text{V}} \,\}$ and $ {\mathcal{J}}^\beta_{\text{V}}  := \{\, \tilde{q}\ \mid\ u^\beta_{\tilde{q}}\in\mathcal{U}^\beta_{\text{V}} \,\}$.
Accordingly, pairs in $\mathcal{L}_{\text{V}}$ will be written as $(\tilde{c},\tilde{q})$ with $\tilde{c}\in{\mathcal{I}}^\alpha_{\text{V}}$ and $\tilde{q}\in{\mathcal{J}}^\beta_{\text{V}}$.
% ==================================================================================================================================================================================

% ==================================================================================================================================================================================
\subsection{Maximum Edge-Weighted Clique Selection in Exchange Graph}
\label{sec:edge_clique}
\noindent
Building on the vertex cover results from $\stepB$, which reduces the search space to a tractable size, we proceed to $\stepC$ of our cascaded framework to optimize geometric consistency and further reduce transmission requirements.
Section~\ref{sec:vertex_cover} reduces communication overhead by selecting a subset of scans that covers all candidate inter-robot loops. 
However, $\stepB$ optimizes coverage from a graph-theoretic perspective in terms of vertex cardinality, not actual geometric overlap between two robots. 
While the graph nodes may be fully covered at the graph level (\ie vertex coverage), the corresponding geometric spaces may still have little spatial overlap, with most 3D points observing different parts of the physical scene.
Moreover, even with balanced selection, the transmission volume can remain substantial for large-scale missions with high-resolution sensors or long durations, potentially exceeding practical bandwidth constraints in field deployments.

To minimize both the geometric mismatch and communication overhead, we need a more sophisticated selection criterion beyond simple graph coverage. 
Inspired by the consistency checking approach in \citet{mangelson2018pairwise}, we recognize that positional consistency can measure whether two inter-robot loops maintain geometric overlap under transformation. 
% However, positional validation alone cannot ensure information-rich correspondences because geometrically consistent loops may observe different scene content or exhibit poor perceptual quality.
However, positional validation alone cannot ensure information-rich correspondences because geometrically consistent loops may observe different scene content or exhibit poor perceptual quality (\eg in a factory environment where identical shelf arrangements are repeated across multiple zones).

To address these limitations, we develop a novel pairwise consistency (\ie compatibility score) framework that combines both positional and perceptual consistency through the received pose databases $\mathcal{X}^{\alpha*}$ and $\mathcal{X}^{\beta*}$ described in Section~\ref{sec:intra_slam} and SOLiD descriptors, respectively.
Building on this compatibility score, we construct a compatibility graph using the poses and SOLiD descriptors corresponding to the loop set $\mathcal{L}_{\text{V}}$ as follows:
\begin{align}
\mathcal{G}_{\text{comp}} &= (\mathcal{V}_{\text{comp}},\mathcal{E}_{\text{comp}}, \mathcal{W}_{\text{comp}}), \\
\mathcal{V}_{\text{comp}} &= \mathcal{L}_{\text{V}} =\{\,\tilde\edge_{C}=(\tilde{c}_{C},\tilde{q}_{C})\,\}_{C=1}^{N_{\text{comp}}},
\end{align}
where $N_{\text{comp}} = |\mathcal{L}_{\text{V}}|$ is the number of loops selected from $\stepB$, and $\tilde{\edge}_C$ represents the $C$-th loop connecting keyframe $\tilde{c}_C$ from robot $\alpha$ and keyframe $\tilde{q}_C$ from robot $\beta$. 
We define the index set $\mathcal{I}_{\text{comp}} := \{1,2,\cdots,\vartheta_1, \vartheta_2, \dots, N_{\text{comp}}\}$ to enumerate all candidate loops.
Each vertex in $\mathcal{V}_{\text{comp}}$ represents a loop, each edge in $\mathcal{E}_{\text{comp}}$ connects two loop pairs (\eg $\tilde{\edge}_{\vartheta_1}$ and $\tilde{\edge}_{\vartheta_2}$), and each weight in $\mathcal{W}_{\text{comp}}$ measures the compatibility score between two loop pairs.
By solving a graph-theoretic problem on this graph, we select the final scan indices that must be transmitted, maximizing communication efficiency while preserving essential correspondences.

Without loss of generality, consider loops indexed by $\vartheta_1$ and $\vartheta_2$. 
We then denote the compatibility score, $\xi_{\vartheta_1,\vartheta_2} \in \mathcal{W}_{\text{comp}}$, between loops $\tilde{\edge}_{\vartheta_1}$ and $\tilde{\edge}_{\vartheta_2}$, defined as follows:
\begin{equation}
\xi_{\vartheta_1,\vartheta_2} = \exp(-\varkappa_{\vartheta_1} \cdot \kappa_{\vartheta_1,\vartheta_2} \cdot \varkappa_{\vartheta_2}),
\label{eq:comp_score}
\end{equation}
where $\kappa_{\vartheta_1,\vartheta_2}$ measures the positional consistency between the two loops and $\varkappa_{\vartheta_1}$ and $\varkappa_{\vartheta_2}$ represent the individual perceptual consistency for loops $\tilde{\edge}_{\vartheta_1}$ and $\tilde{\edge}_{\vartheta_2}$, respectively.

For the positional consistency (\ie $\kappa_{\vartheta_1,\vartheta_2}$), full SE(3) transformation checks can be employed as in \citet{mangelson2018pairwise}.
However, our SOLiD framework is designed to be rotationally invariant, meaning that robots may visit the same physical location from different orientations, and SOLiD would correctly identify these as loop closure candidates regardless of viewpoint differences.
A full SE(3)-based consistency check would penalize such geometrically valid loops due to orientation discrepancies, potentially rejecting correct correspondences that are rotationally displaced but spatially overlapping.
Therefore, inspired by the compatibility test in \citet{shi2021robin}, we focus on translational error, allowing rotational freedom while ensuring spatial proximity between loop candidates.

Let $\mathbf{T}^\alpha_{\tilde{c}} \in \mathcal{X}^{\alpha*}$ and $\mathbf{T}^\beta_{\tilde{q}} \in \mathcal{X}^{\beta*}$ denote the received poses from the transmitted databases, and let $\mathbf{t}^\alpha_{\tilde{c}}, \mathbf{t}^\beta_{\tilde{q}} \in \mathbb{R}^3$ denote their respective translation components.
We then define the positional consistency as follows:
\vspace{-0.1cm}
\begin{align}
\kappa_{\vartheta_1,\vartheta_2}
&=\Big|\,\|\mathbf{t}^\alpha_{\tilde{c}_{\vartheta_1}}-\mathbf{t}^\alpha_{\tilde{c}_{\vartheta_2}}\|_2
      - \|\mathbf{t}^\beta_{\tilde{q}_{\vartheta_1}}-\mathbf{t}^\beta_{\tilde{q}_{\vartheta_2}}\|_2 \,\Big|.
\label{eq:pos_consist}
\end{align}
\vspace{-0.3cm}

For perceptual consistency (\ie $\varkappa_{\vartheta_1}$ and $\varkappa_{\vartheta_2}$), since our framework employs SOLiD descriptors throughout the pipeline, we utilize SOLiD-based similarity in \eqref{eq:f_sim} as follows:
\begin{equation}
\varkappa_{\vartheta_1} \;:=\; w^{\alpha,\beta}_{\tilde c_{\vartheta_1},\,\tilde q_{\vartheta_1}} \in \mathbb{R}_{\ge 0}, \qquad \varkappa_{\vartheta_2} \;:=\; w^{\alpha,\beta}_{\tilde c_{\vartheta_2},\,\tilde q_{\vartheta_2}} \in \mathbb{R}_{\ge 0}.
\label{eq:per_consist}
\end{equation}
Finally, using \eqref{eq:pos_consist} and \eqref{eq:per_consist}, we define the complete compatibility score $\xi_{\vartheta_1,\vartheta_2}$ in \eqref{eq:comp_score} that combines both positional and perceptual consistency.

Since our goal is to identify mutually compatible loop sets, we seek subsets in which every pair of loops satisfies a compatibility condition.
To this end, we construct a compatibility graph by explicitly separating feasibility from compatibility strength.

We first define a binary feasibility mask $\mathbf{M}\in\{0,1\}^{N_{\text{comp}}\times N_{\text{comp}}}$ as follows:
\begin{equation}
\mathbf{M}_{\vartheta_1,\vartheta_2} =
\begin{cases}
1, & \text{if } \vartheta_1 < \vartheta_2 \;\wedge\;
\kappa_{\vartheta_1,\vartheta_2} < \tau_{\text{con}},\\
0, & \text{otherwise}.
\end{cases}
\label{eq:feasibility_mask}
\end{equation}
This mask determines whether a pair of loops is geometrically admissible and defines the edge set of the compatibility graph.

For admissible loop pairs, we assign a continuous compatibility score $\xi_{\vartheta_1,\vartheta_2}$ defined in \eqref{eq:comp_score}.
The final compatibility matrix $\mathbf{C}^{\triangle}$ is then defined as follows:
\begin{equation}
\mathbf{C}^{\triangle}
\;=\;
\mathbf{M} \odot \boldsymbol{\Xi},
\label{eq:compatibility_matrix}
\end{equation}
where $\boldsymbol{\Xi}_{\vartheta_1,\vartheta_2} := \xi_{\vartheta_1,\vartheta_2}$ and $\odot$ denotes the Hadamard product.
Equivalently, this can be written entrywise as follows:
\begin{equation}
\mathbf{C}^{\triangle}_{\vartheta_1,\vartheta_2}
=
\begin{cases}
\xi_{\vartheta_1,\vartheta_2},
& \text{if } \vartheta_1<\vartheta_2 \ \wedge\ \kappa_{\vartheta_1,\vartheta_2} < \tau_{\text{con}},\\
0,
& \text{otherwise}.
\end{cases}
\end{equation}

\begin{figure}[t]
    \centering
    \captionsetup{justification=justified}
    \captionsetup[subfigure]{justification=centering}
    \begin{subfigure}{0.46\columnwidth}
        \includegraphics[trim={0 0 0 0},clip, width=\columnwidth]{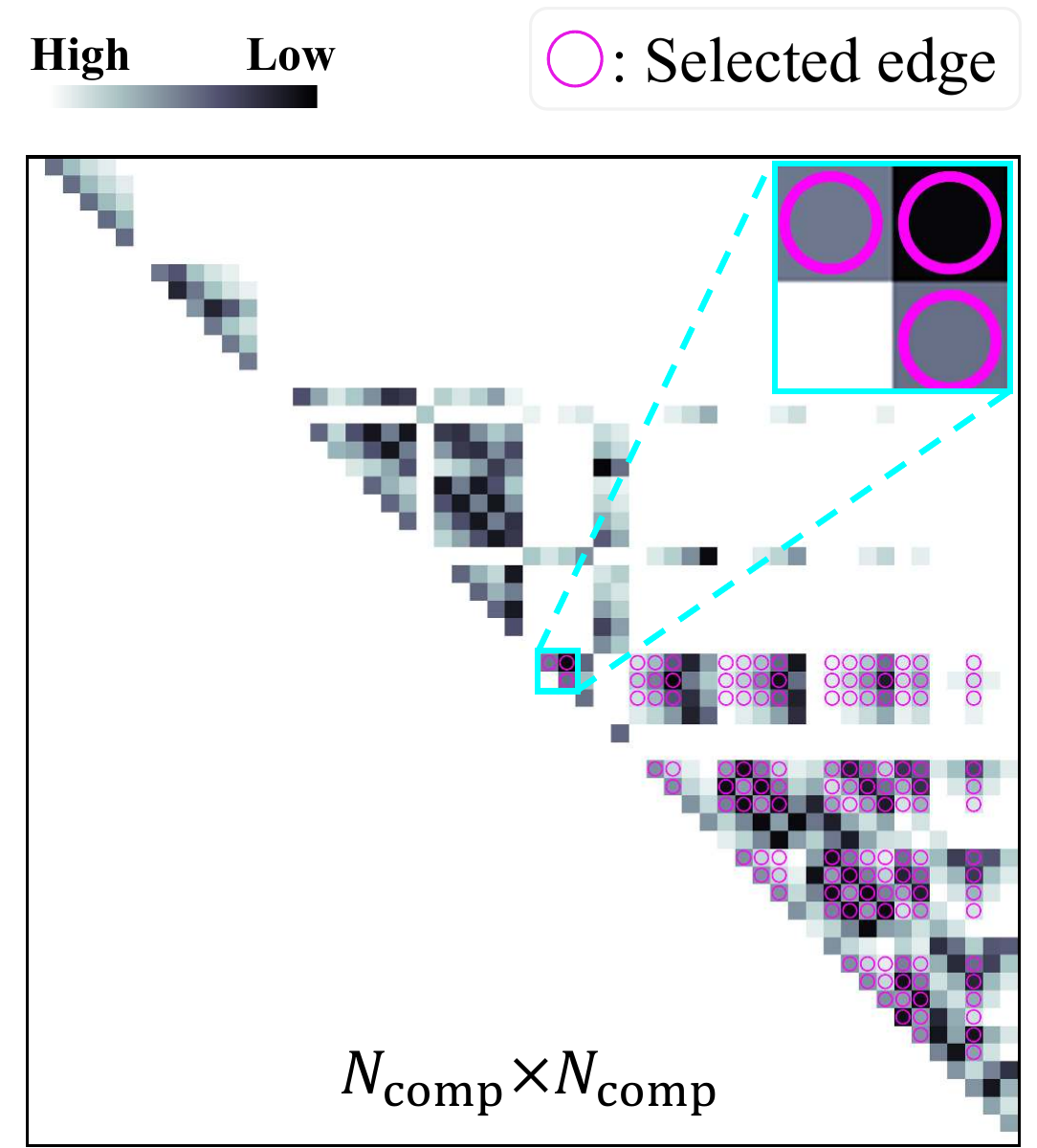}
        \vspace{-1.5em}
        \caption{}
    \end{subfigure}
    \begin{subfigure}{0.52\columnwidth}
        \includegraphics[trim={0 0 0 0},clip, width=\columnwidth]{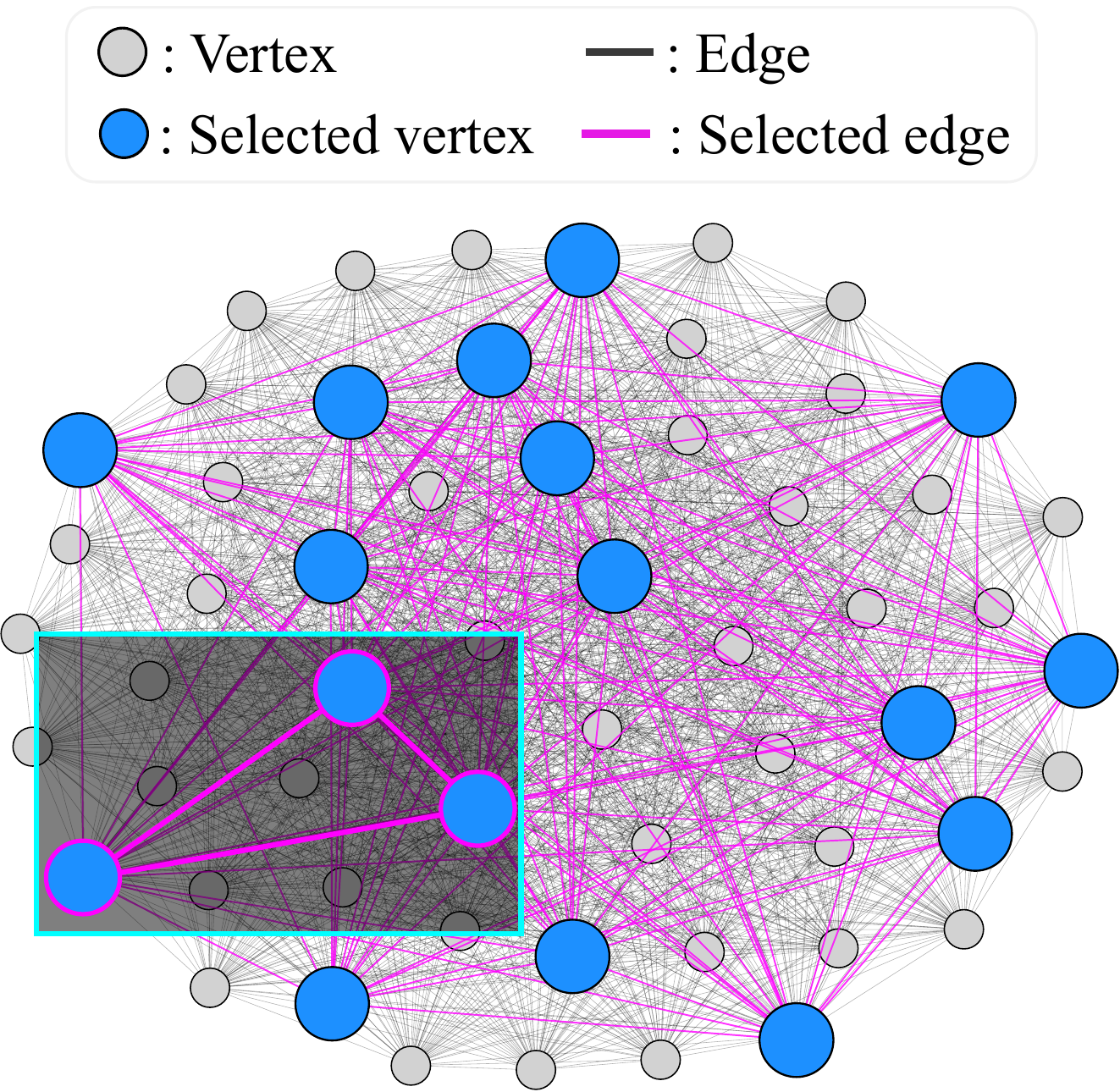}
        \vspace{-1.5em}
        \caption{}
    \end{subfigure}
  \caption{Visualization of the compatibility graph and its matrix representation for maximum edge-weighted clique selection. 
          (a)~Compatibility matrix where each cell represents a potential edge between two loops. 
          Selected edges (\ie magenta circles) indicate high compatibility pairs that form the clique.
          (b)~Compatibility graph representation where vertices (\ie grey nodes) represent inter-robot loops between robots $\alpha$ and $\beta$, and edges (\ie grey lines) encode compatibility scores.
          The maximum edge-weighted clique selection identifies optimal vertices (\ie blue nodes) and edges (\ie magenta lines) with the highest mutual consistency.
          For example, the magenta circles within the cyan box in (a) correspond directly to the connections between selected vertices highlighted in the cyan region of (b). 
          Only scans incident to the selected loops (\ie blue nodes) are transmitted to the server.}
  \label{fig:max_clique}
  \vspace{-3mm}
\end{figure}

Accordingly, we define the edge set as $\mathcal{E}_{\text{comp}} := \{(\vartheta_1,\vartheta_2)\mid \mathbf{M}_{\vartheta_1,\vartheta_2}=1\}$ and assign edge weights by $\mathcal{W}_{\text{comp}}(\vartheta_1,\vartheta_2) := \mathbf{C}^{\triangle}_{\vartheta_1,\vartheta_2}$.
In summary, as illustrated in \figref{fig:max_clique}(a), $\mathbf{C}^{\triangle}$ assigns each matrix cell a unified pairwise compatibility weight by combining positional consistency $\kappa_{\vartheta_1,\vartheta_2}$ and perceptual consistency $\varkappa_{\vartheta}$, while infeasible pairs are masked out by $\mathbf{M}$ (set to zero).

For loop selection, we could apply a standard maximum clique approach to identify the largest set of mutually compatible loops. 
However, similar to the limitation discussed in Section~\ref{sec:vertex_cover}, this approach optimizes graph-theoretic coverage rather than geometric overlap quality.
It prioritizes the cardinality of compatible loops without considering the actual spatial intersection between robot observations or the informativeness of individual correspondences.
To address this limitation, we adopt an edge-weighted clique formulation that incorporates both aspects. 
Unlike the standard maximum clique selection, which treats all compatible loops equally, our approach weights each loop by its compatibility. 
This ensures that selected loops are perceptually and geometrically consistent.

We now solve a maximum edge-weighted clique problem on the compatibility graph. 
Let $\mathcal{K} \subseteq \mathcal{I}_{\text{comp}}$ represent a subset of loop indices. 
The optimization is formulated as follows:
\begin{align}
f_K(\mathcal{K}) &:= \sum_{\vartheta_1 < \vartheta_2 \in \mathcal{K}} \mathbf{C}^{\triangle}_{\vartheta_1,\vartheta_2},
\\
\mathcal{K}^\star &\;=\;\operatorname*{arg\,max}_{\mathbf{M}_{\vartheta_1,\vartheta_2}=1\ \ \forall\,\vartheta_1<\vartheta_2\in\mathcal{K}}
f_{\text{K}}(\mathcal{K}),
\label{eq:mewc}
\end{align}

where the clique constraint $\mathbf{M}_{\vartheta_1,\vartheta_2}=1$ ensures that every pair of loops in $\mathcal{K}$ forms a clique, meaning all selected loops are mutually compatible.
The objective $f_{\text{K}}(\cdot)$ maximizes loop quality (via $\kappa_{\vartheta_1,\vartheta_2},\varkappa_{\vartheta_1}$, $\varkappa_{\vartheta_2}$), selecting informationally rich and mutually compatible loops as shown in \figref{fig:max_clique}(b).
By doing so, we can map the selected indices back to loops as:
\begin{align}
\mathcal{L}_{\text{K}}
\;:=\;
\{\, \tilde\edge_{\vartheta_C}\in\mathcal{L}_{\text{V}} \mid C\in\mathcal{K}^\star \,\}.
\label{eq:optim_subset}
\end{align}

% =======================================================
% =======================================================

% =======================================================
\begin{figure*}[t]
    \centering
    \captionsetup{justification=justified}
    \captionsetup[subfigure]{justification=centering}
    \begin{subfigure}{0.85\columnwidth}
        \includegraphics[trim={0 0 0 0},clip, width=\columnwidth]{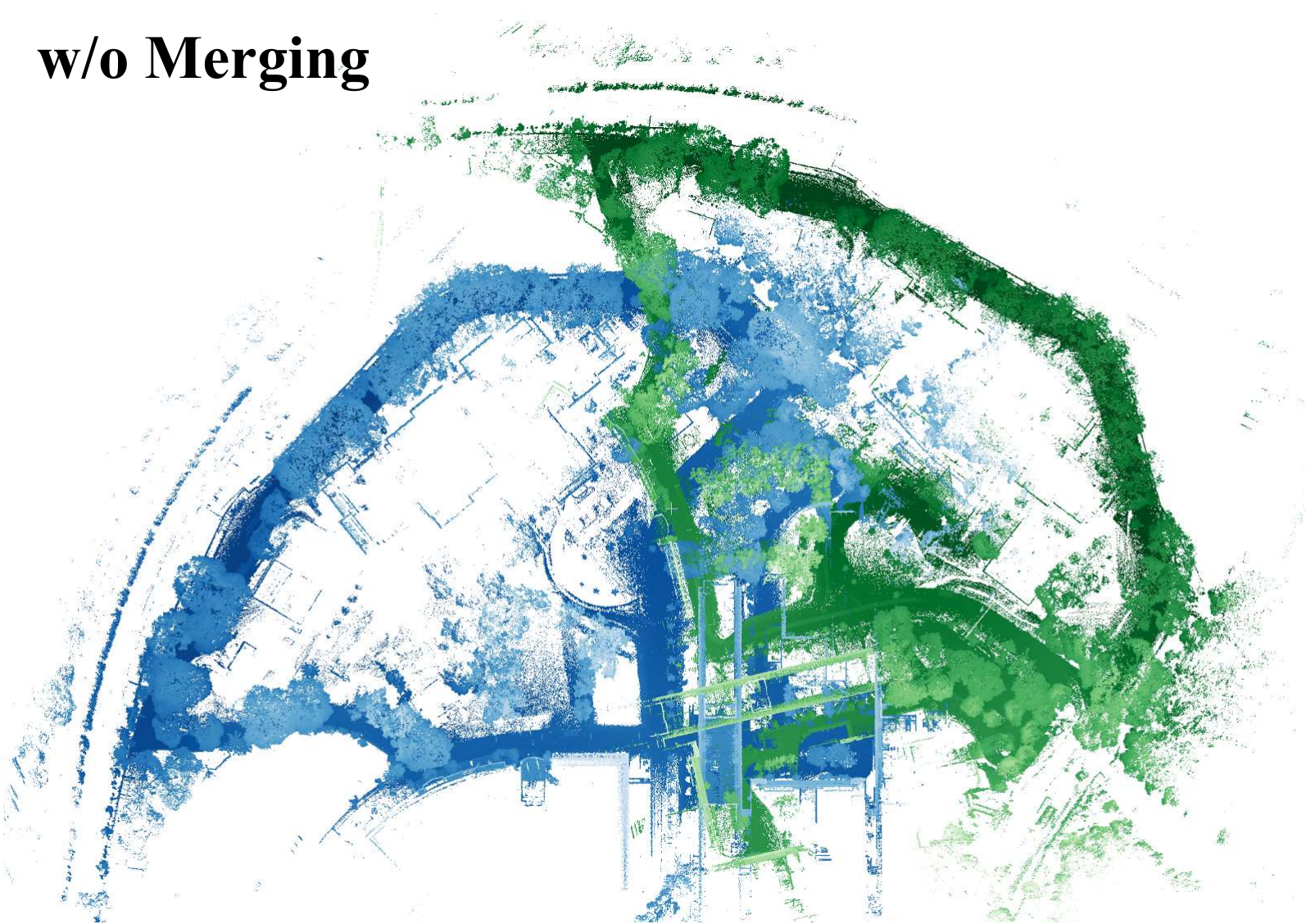}
        \vspace{-1.5em}
        \caption{}
    \end{subfigure}
    \begin{subfigure}{1.13\columnwidth}
        \includegraphics[trim={0 0 0 0},clip, width=\columnwidth]{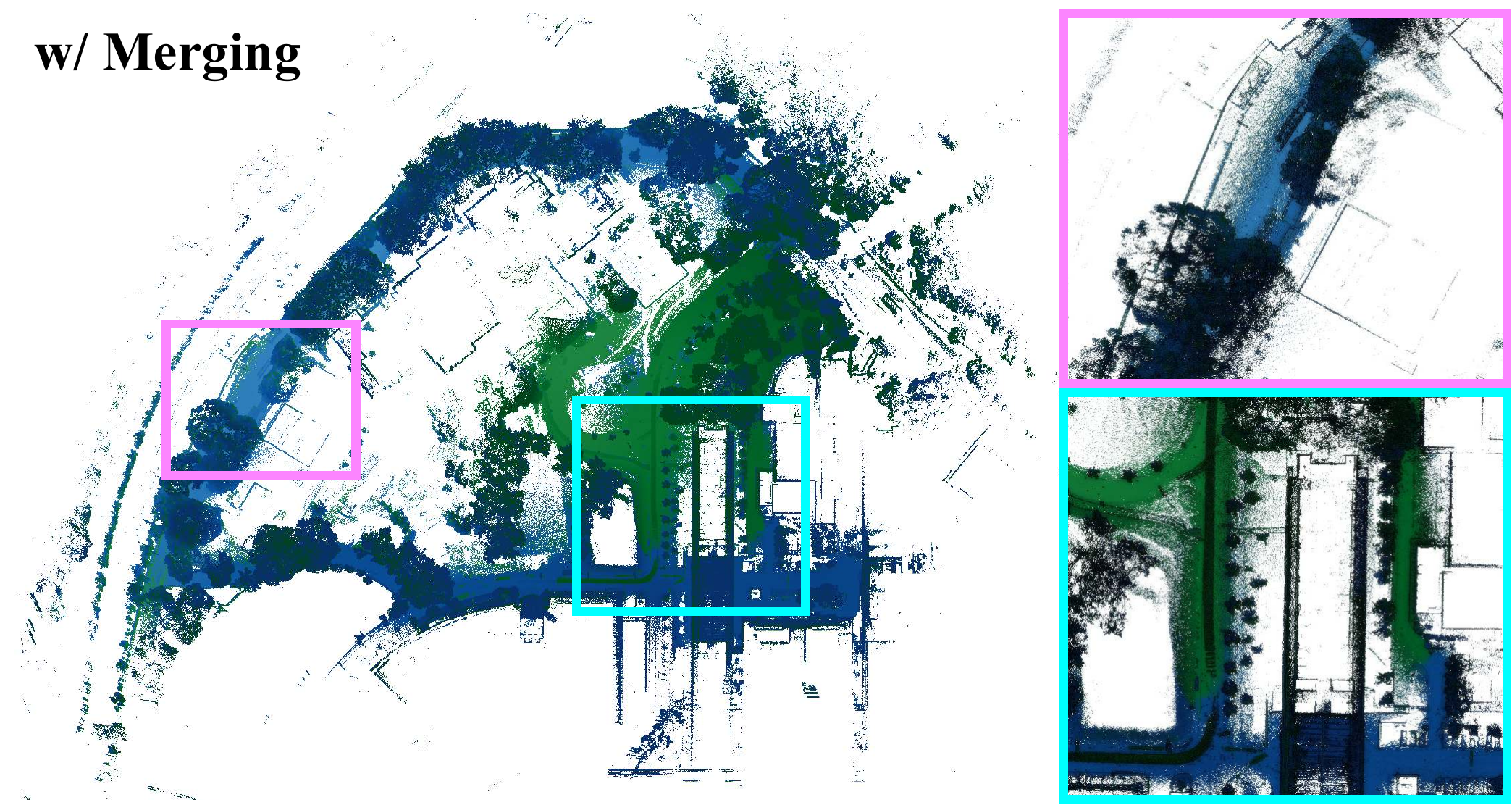}
        \vspace{-1.5em}
        \caption{}
    \end{subfigure}
    \includegraphics[trim={0 0 0 0},clip, width=0.7\columnwidth]{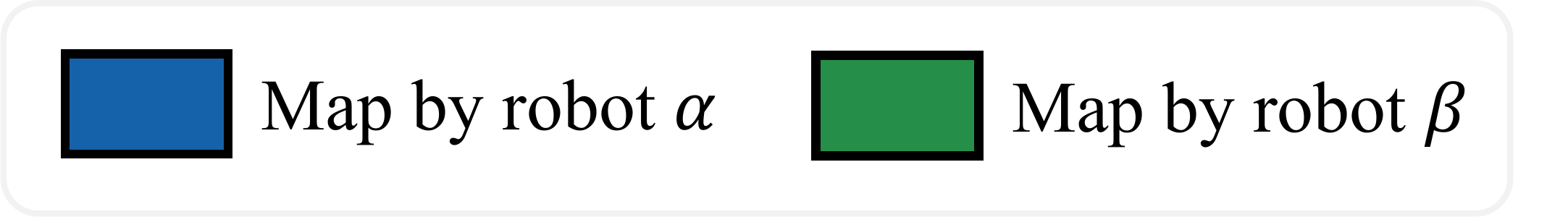}
  \vspace{-2mm}
    \caption{Results before and after applying \textit{Commerge} for multi-robot map merging in the Multi-Campus Dataset (MCD) acquired with a Livox Mid-70 LiDAR sensor.
            The magenta box highlights a building structure, and the cyan box highlights trees, both of which appear well-aligned after merging without distortion. 
            }
  \label{fig:mcd}
  \vspace{-4mm}
\end{figure*}
% =======================================================

In summary, using the loop set $\mathcal{L}_{\text{V}}$ from \eqref{eq:balanced} and the received pose databases $\mathcal{X}^{\alpha*}$ and $\mathcal{X}^{\beta*}$ from each robot, we construct the compatibility matrix as defined in \eqref{eq:compatibility_matrix}. 
We then identify the maximum edge-weighted clique from this compatibility graph through the optimization in \eqref{eq:mewc}.
This process is denoted by $J_{\text{K}}(\cdot)$, completing $\stepC$ as defined in \eqref{eq:clique}.
Accordingly, we write pairs in $\mathcal{L}_{\text{K}}$ as $(c^\star,q^\star)$ with
$c^\star\in\mathcal{I}^\alpha_{\text{K}} := \big\{\, c^\star \mid (c^\star,q^\star)\in\mathcal{L}_{\text{K}} \,\big\}$ and
$q^\star\in\mathcal{J}^\beta_{\text{K}} := \big\{\, q^\star \mid (c^\star,q^\star)\in\mathcal{L}_{\text{K}} \,\big\}$.

\subsection{Exchange Policy Generation}
\label{sec:exchange_policy}
\noindent
Once the optimal subset is identified through solving the three-stage cascaded problem, the server generates an exchange policy that instructs robots to transmit only the scans incident to the selected loops. 
Upon receiving this policy from the server, each robot transmits the minimal scan subsets back to the server, which are necessary to maintain inter-robot alignment while achieving substantial bandwidth reduction.

Specifically, given $\mathcal{L}_{\text{K}}$ in \eqref{eq:optim_subset}, the server requests only the unique endpoint scans of all selected loops as follows:
\begin{align}
\mathcal{I}^{\alpha}_{\text{K}}
&:= \big\{\, c^\star \ \big|\ (c^\star,q^\star)\in\mathcal{L}_{\text{K}} \,\big\}, \label{eq:exchange_policy_alpha} \\ 
\mathcal{J}^{\beta}_{\text{K}}
&:= \big\{\, q^\star \ \big|\ (c^\star,q^\star)\in\mathcal{L}_{\text{K}} \,\big\}.
\label{eq:exchange_policy_beta}
\end{align}
Robots then transmit only the scans corresponding to these selected indices back to the server, reducing communication load while preserving essential geometric information for robust alignment.

% ==================================================================================================================================================================================
\subsection{Submap-Based Map Merging}
\label{sec:map_merging}
\noindent
Upon receiving the selected scans specified by the exchange policy \eqref{eq:exchange_policy_alpha} and \eqref{eq:exchange_policy_beta}, the server constructs submaps by aggregating the corresponding point clouds transformed to the global frame as follows:
\begin{align}
\mathcal{M}_\alpha &\coloneqq \bigcup_{c^\star\in\mathcal{I}^\alpha_{\text{K}}} \{\mathbf{R}^{\alpha}_{c^\star} \mathbf{p}_{\text{s}} + \mathbf{t}^{\alpha}_{c^\star} \mid \mathbf{p}_{\text{s}} \in \mathbf{P}^{\alpha}_{c^\star}\},\\
\mathcal{M}_\beta &\coloneqq \bigcup_{q^\star\in\mathcal{J}^ \beta_{\text{K}}} \{\mathbf{R}^{\beta}_{q^\star} \mathbf{q}_{\text{s}} + \mathbf{t}^{\beta}_{q^\star} \mid \mathbf{q}_{\text{s}} \in \mathbf{P}^{ \beta}_{q^\star}\},
\end{align}
where $\mathbf{P}^{\alpha}_{c^\star}$ and $\mathbf{P}^{\beta}_{q^\star}$ denote the LiDAR point clouds at keyframes $c^\star$ and $q^\star$ in their respective local sensor frames, and $\mathbf{R}^{\alpha}_{c^\star}, \mathbf{t}^{\alpha}_{c^\star}$ and $\mathbf{R}^{\beta}_{q^\star}, \mathbf{t}^{\beta}_{q^\star}$ are the corresponding rotation matrices and translation vectors extracted from the pose databases $\mathcal{X}^{\alpha*}$ and $\mathcal{X}^{\beta*}$ used in Section~\ref{sec:edge_clique}.

With the compact submaps $\mathcal{M}_\alpha$ and $\mathcal{M}_\beta$ constructed from the minimal scan subsets in Section~\ref{sec:exchange_policy}, the server performs global alignment using a coarse-to-fine registration approach, completing the efficient map merging process. 
The coarse registration stage employs KISS-Matcher for robust initial alignment, followed by fine registration using small-GICP for precise pose refinement.

We begin with coarse registration using KISS-Matcher to establish initial correspondences between the two submaps. 
Let $\mathcal{A}_{\text{init}}$ be the set of initial point correspondences between the submaps $\mathcal{M}_\alpha$ and $\mathcal{M}_\beta$ considered by KISS-Matcher, and let $\mathcal{F}\subseteq\mathcal{A}_{\text{init}}$ denote the pairs flagged as outliers.
We then estimate the initial rigid transformation $\hat{\mathbf{T}}^{\alpha,\beta}_{\text{init}} = \begin{bmatrix} \hat{\mathbf{R}}^{\alpha,\beta}_{\text{init}} & \hat{\mathbf{t}}^{\alpha,\beta}_{\text{init}} \\ \mathbf{0}^\top & 1 \end{bmatrix} \in \text{SE}(3)$ as follows:
\begin{equation}
\hat{\mathbf{R}}^{\alpha,\beta}_{\text{init}},\;\hat{\mathbf{t}}^{\alpha,\beta}_{\text{init}}
\;=\;
\operatorname*{arg\,min}_{\substack{\mathbf{R}\in\mathrm{SO}(3)\\ \mathbf{t}\in\mathbb{R}^3}}
\sum_{(\mathbf{p},\,\mathbf{q})\in \mathcal{A}_{\text{init}}\setminus\mathcal{F}}
\rho\Big(\,\big\|\mathbf{R}\mathbf{p}+\mathbf{t}-\mathbf{q}\big\|_2^2\,\Big),
\label{eq:kiss_robust}
\end{equation}
where $\mathbf{p}\in\mathbb{R}^3$ and $\mathbf{q}\in\mathbb{R}^3$ are corresponding 3D points sampled from $\mathcal{M}_\alpha$ and $\mathcal{M}_\beta$, respectively, and $\rho(\cdot)$ is a robust loss used to suppress the influence of residual large errors. 
For detailed algorithmic descriptions of KISS-Matcher, please refer to our previous work~\citep{kiss_matcher}.

Next, we apply a lightweight local registration, \ie small-GICP~\citep{small_gicp}, to achieve fine alignment initialized at $\hat{\mathbf{T}}^{\alpha,\beta}_{\text{init}}$.
This coarse-to-fine approach combines the robustness of global registration for handling large initial misalignment with the precision of local optimization for accurate final alignment.
Let $\mathcal{A}_{\text{fine}}$ be the set of local correspondences established around $\hat{\mathbf{T}}^{\alpha,\beta}_{\text{init}}$, where each correspondence $(\mathbf{p}', \mathbf{q}') \in \mathcal{A}_{\text{fine}}$ consists of points $\mathbf{p}' = \hat{\mathbf{R}}^{\alpha,\beta}_{\text{init}} \mathbf{p} + \hat{\mathbf{t}}^{\alpha,\beta}_{\text{init}}
$ (transformed from $\mathcal{M}_\alpha$) and $\mathbf{q}' \in \mathcal{M}_\beta$. The refined solution $\hat{\mathbf{T}}_{\text{fine}}^{\alpha,\beta} := \begin{bmatrix} \hat{\mathbf{R}}_{\text{fine}}^{\alpha,\beta} & \hat{\mathbf{t}}_{\text{fine}}^{\alpha,\beta} \\ \mathbf{0}^\top & 1 \end{bmatrix} \in \text{SE}(3)$ is defined as follows:
\begin{equation}
\hat{\mathbf{R}}_{\text{fine}}^{\alpha,\beta},\;
\hat{\mathbf{t}}_{\text{fine}}^{\alpha,\beta}
\;=\;
\operatorname*{arg\,min}_{\substack{\mathbf{R}\in\mathrm{SO}(3)\\ \mathbf{t}\in\mathbb{R}^3}}
\sum_{(\mathbf{p}',\,\mathbf{q}')\in\mathcal{A}_{\text{fine}}}
\big\|\,\mathbf{R}\,\mathbf{p}'+\mathbf{t}-\mathbf{q}'\,\big\|_2^2.
\label{eq:fine_regi}
\end{equation}

Finally, we obtain the global transformation by composing the coarse and fine transformations using \eqref{eq:kiss_robust} and \eqref{eq:fine_regi} as:
\begin{equation}
\mathbf{T}_{\text{final}}^{\alpha,\beta} = \begin{bmatrix} \mathbf{R}_{\text{final}}^{\alpha,\beta} & \mathbf{t}_{\text{final}}^{\alpha,\beta} \\ \mathbf{0}^\top & 1 \end{bmatrix} = \hat{\mathbf{T}}_{\text{fine}}^{\alpha,\beta} \hat{\mathbf{T}}_{\text{init}}^{\alpha,\beta}.
\label{eq:final_transformation}
\end{equation}
As shown in \figref{fig:mcd}, we then transform $\mathcal{M}_\alpha$~(\figref{fig:mcd}(a)) using the final transformation in \eqref{eq:final_transformation} and fuse it with $\mathcal{M}_\beta$~(\figref{fig:mcd}(b)) as follows:
\begin{align}
\mathcal{M}_{\alpha,\text{aligned}} 
&\coloneqq \big\{\, \mathbf{R}_{\text{final}}^{\alpha,\beta}\,\mathbf{p} + \mathbf{t}_{\text{final}}^{\alpha,\beta} \ \big|\ \mathbf{p}\in\mathcal{M}_\alpha \,\big\}, \\
\mathcal{M}_{\text{merged}} 
&\coloneqq \mathcal{M}_{\alpha,\text{aligned}} \ \cup\ \mathcal{M}_\beta.
\end{align}
This pipeline leverages overlap-rich submaps to stabilize one-shot global registration, while achieving dramatic communication efficiency: transmitting only the scans corresponding to $\mathcal{K}^\star$ typically reduces data exchange from GB-scale to MB-scale.
\section{Implementation Details}
\subsection{Intra-Robot SLAM Configuration}
\label{sec:robot}
\noindent
For intra-robot SLAM, we adopt a standard pose graph optimization framework implemented in GTSAM\footnote{\url{https://gtsam.org/}} \citep{dellaert2012gtsam}. 
Each robot incrementally builds its local submap by registering LiDAR scans and optimizing the trajectory using odometry and loop closure constraints. 
The GTSAM backend allows efficient non-linear least squares optimization with factor graphs, providing accurate and consistent intra-robot maps that serve as inputs to our multi-robot pipeline.

\subsection{Graph-Theoretic Problem Solvers}
\label{sec:server}
\noindent
To minimize the number of scans that robots must transmit to the server, we formulate and solve graph-theoretic optimization problems using the SOLiD descriptors and poses received from multiple robots.
Specifically, we construct the exchange graph where vertices represent robot keyframes and edges denote potential inter-robot loops, then solve the three-stage cascade optimization using Gurobi\footnote{\url{https://www.gurobi.com/}}.

In particular, the BVC problem is NP-hard because vertex cover is a canonical NP-hard problem whose search space grows combinatorially with the number of vertices.
To address this, we formulate the BVC as a mixed-integer linear program (MILP) with binary decision variables $\pi(x)\in\{0,1\}$ indicating scan selection.
The MILP is solved using Gurobi's branch-and-bound procedure, which returns the optimal solution whenever it completes within the solver time limit, and a near-optimal solution otherwise.

Similarly, the maximum edge-weighted clique selection problem is also NP-hard, as finding the clique with the maximum sum of edge weights requires exploring an exponential number of vertex combinations. 
However, since the exchange graph is substantially pruned after the MVC stage, the problem size becomes more manageable. 
We also formulate maximum edge-weighted clique selection as a mixed-integer quadratic program (MIQP), where binary variables indicate clique membership and quadratic terms capture edge weights between selected vertices.
Gurobi handles the quadratic constraints via its branch-and-cut procedure, returning the optimal solution whenever it completes within the solver time limit, and a near-optimal solution otherwise.

\subsection{Parameters of Commerge}
\label{sec:parameters}
\noindent
Table~\ref{tab:parameters} summarizes the parameters used throughout our pipeline, organized by processing robot-side in Section~\ref{sec:robot} and server-side in Section~\ref{sec:server}.
Further implementation details are available in our open-source release.

\definecolor{myemerald}{rgb}{0.753, 0.898, 0.804}
\definecolor{mylightgreen}{rgb}{0.894, 0.933, 0.745}
\definecolor{myyellow}{rgb}{0.996, 0.972, 0.780}
\newcommand{\firstc}{\cellcolor{myemerald!100}}
\newcommand{\secondc}{\cellcolor{mylightgreen!100}}
\newcommand{\thirdc}{\cellcolor{myyellow!100}}
\newcommand{\oursdc}{\cellcolor{white!100}}

\newcommand{\rmsemetric}{\boldsymbol{e}_{\text{rmse}}}
\newcommand{\commmetric}{\boldsymbol{t}_{\text{comm}}}
\newcommand{\exchmetric}{\boldsymbol{d}_{\text{exch}}}

\newcommand{\hardA}{\texttt{Intel-i7}}
\newcommand{\hardB}{\texttt{IPC-6000}}
\newcommand{\hardC}{\texttt{Jet-Nano}}

\section{Experiments}
\subsection{Experiments Setup}
\label{sec:experiments_setup}
\noindent
We first specify the hardware platform for the server that performs multi-robot map merging operations.
To verify practical field deployability, we selected three representative hardware configurations with varying computational capabilities: 
a desktop workstation with Intel Core i7-12700KF (12 cores, 3.6\,GHz, and 32\,GB RAM; $\hardA$), an industrial mini PC (IPC-6000 series with Intel Core i5-8265U and 8\,GB RAM; $\hardB$), and an NVIDIA Jetson Nano (quad-core ARM Cortex-A57 with 4\,GB RAM; $\hardC$).
Note that we assume each platform serves as the central server that coordinates data selection and performs map merging across robots.

We evaluate algorithmic aspects~(Sections~\ref{sec:perform_comparison}, \ref{sec:robustness}) such as overall map merging performance on $\hardA$ to ensure consistent evaluation conditions across all methods, while system-level aspects~(Sections~\ref{sec:scalability}, \ref{sec:resource_efficiency}) including scalability, resource efficiency, and runtime are evaluated across all three platforms ($\hardA$, $\hardB$, and $\hardC$) to compare deployment feasibility under varying hardware constraints.

\subsection{Evaluation Metrics}
\label{sec:eval_metrics}
\noindent
Next, we employ three evaluation metrics for map merging performance, used in Section~\ref{sec:results}, and six evaluation metrics for the ablation study on place recognition and registration, used in Section~\ref{sec:ablation}.
In all metrics, {\setlength{\fboxsep}{1pt}\colorbox{myemerald}{\textbf{1st}}}, {\setlength{\fboxsep}{1pt}\colorbox{mylightgreen}{2nd}}, and {\setlength{\fboxsep}{1pt}\colorbox{myyellow}{3rd}} indicate ranking.

The first metric for map merging is defined as follows:
\begin{equation}
\rmsemetric = f_\text{RMSE}\left(\mathbf{T}^\text{GT}_{\alpha} \mathbf{T}^\alpha_\beta \mathbf{T}_j^\beta, \mathbf{T}_j^{\beta,\text{GT}}\right),
\label{eq:rmsemetric}
\end{equation}
which evaluates inter-robot alignment quality through the following procedure~(Section~\ref{sec:perform_comparison}).
Specifically, given robot $\alpha$'s estimated trajectory and its ground truth trajectory, we compute the optimal alignment transformation~$\mathbf{T}^\text{GT}_{\alpha}$ using Umeyama alignment~\citep{umeyama1991least}.
Each method estimates the inter-robot transformation $\mathbf{T}^\alpha_\beta$ that aligns robot $\alpha$'s frame to robot $\beta$'s frame. 
To evaluate this estimated transformation, we apply both $\mathbf{T}^\text{GT}_{\alpha}$ and $\mathbf{T}^\alpha_\beta$ to robot $\beta$'s estimated trajectory, $\mathbf{T}_j^\beta$, and compute the root mean square error (RMSE) against robot $\beta$'s ground truth trajectory, $\mathbf{T}_j^{\beta,\text{GT}}$, using the EVO toolkit~\citep{grupp2017evo}.

Next, the second metric is defined as follows:
\begin{equation}
    \commmetric = \sum_{r=1}^{N_{\text{robot}}} (t_{N_{\text{tx}}}^{(r)} - t_1^{(r)}),
\label{eq:commmetric}
\end{equation}
which measures the total communication time elapsed from the first transmission to the last transmission across all robots, where $t_{N_{\text{tx}}}^{(r)}$ and $t_1^{(r)}$ are the final and initial transmission timestamps for robot $r$~(Sections~\ref{sec:scalability} and \ref{sec:resource_efficiency}).

In addition, the third metric is defined as follows:
\begin{equation}
    \exchmetric = \sum_{r=1}^{N_{\text{robot}}} \sum_{n=1}^{N_{\text{tx}}^{(r)}} b_n^{(r)},
\label{eq:exchmetric}
\end{equation}
which measures the cumulative payload size of all transmitted messages, where $b_n^{(r)}$ represents the payload size~(in bytes) of the $n$-th message (\ie global descriptor or scan) from robot $r$~(Sections~\ref{sec:scalability} and \ref{sec:resource_efficiency}).

In particular, \eqref{eq:rmsemetric} is evaluated on $\hardA$, while \eqref{eq:commmetric} and \eqref{eq:exchmetric} are evaluated on an actual wireless communication infrastructure with $\hardB$, whose detailed configuration is described in Section~\ref{sec:comm_setup}.

% ====================================================================
\begin{table}[t]
\captionsetup{width=.49\textwidth, justification=justified}
\caption{Parameter configuration for the Commerge framework. For loop detection, $\tau_{\text{SOLiD}}^{\text{intra}}$ is stricter than $\tau_{\text{SOLiD}}^{\text{inter}}$ to ensure high-quality local maps, while inter-robot parameters can be more permissive due to sequential matching's filtering capability. Note that $\tau_{\text{ICP}}$ varies based on LiDAR scanning patterns (lower for repetitive patterns, higher for non-repetitive scan patterns).}
\renewcommand{\arraystretch}{1.2}
\centering\resizebox{0.49\textwidth}{!}{\tiny
\begin{tabular}{l|l|ccc}
\toprule
\midrule
                  &                                                          & Params.                       & Description                        & Value          \\ \midrule
\multirow{9}{*}{\rotatebox[origin=c]{90}{Commerge}}                    
                  & \multirow{5}{*}{\rotatebox[origin=c]{90}{$\robot$ (robot)}}   & $\Delta\nu$                           & Voxel size for point cloud downsampling                
                                                                                  & 1.0\,m \\                    
                  &                                                               & $\Delta d$                            & Keyframe spacing distance              
                                                                                  & 1.0\,m  \\ 
                  &                                                               & $\tau_{\text{ICP}}$                   & ICP fitness threshold for loop verification              
                                                                                  & 0.3\,m\,/\,2.0\,m  \\ 
                  &                                                               & $\tau_{\text{window}}$                & Temporal window for loop exclusion              
                                                                                  & 100 \\ 
                  &                                                               & $\tau_{\text{SOLiD}}^{\text{intra}}$  & Intra-robot loop detection threshold       
                                                                                  & 0.05\,/\,0.01  \\ \cmidrule{2-5}
                  & \multirow{3}{*}{\rotatebox[origin=c]{90}{$\server$ (server)}} & $\tau_{\text{SOLiD}}^{\text{inter}}$  & Inter-robot loop detection threshold               
                                                                                  & 0.10\,/\,0.05 \\
                  &                                                               & $\tau_{\text{area}}$                  & Minimum cluster area for sequential matching               
                                                                                  & 10{,}000 \\
                  &                                                               & $\tau_{\text{con}}$               & Positional consistency threshold              
                                                                                  & 5\,m \\ \midrule
\bottomrule
\end{tabular}}
\label{tab:parameters}
\vspace{-0.5cm}
\end{table}
% ====================================================================
% ===============================================================
\begin{table*}[t]
\captionsetup{width=\textwidth, justification=justified}\caption{Summary of datasets used in our experiments. We evaluate Commerge across diverse environments with varying scale~(\ie map size or duration time), team sizes (denoted by $\robot$), sensor types ($\spin$: spinning LiDAR, $\livox$: non-repetitive scan pattern LiDAR), trajectory overlap degrees ($\complete$: complete overlap where all robots traverse the almost same area, $\moderate$: moderate overlap with substantial shared regions, $\smalloverlap$: partial overlap with minimal shared trajectories), and environments.
}

\renewcommand{\arraystretch}{1.2}
\centering\resizebox{\textwidth}{!}{\tiny
\begin{tabular}{l|cccccccccccccc}
\toprule \midrule
Dataset          & \multicolumn{4}{c}{\makecell{HeLiPR     \\ \citep{jung2024helipr}}}    & \multicolumn{4}{c}{\makecell{Kimera-Multi \\ \citep{tian23arxiv_kimeramultiexperiments}}}    
                 & \multicolumn{2}{c}{\makecell{WildPlaces \\ \citep{knights2023wild}}}   & \multicolumn{2}{c}{\makecell{MCD \\ \citep{nguyen2024mcd}}}                    
                 & \multicolumn{2}{c}{\makecell{Botanic Garden \\ \citep{liu2024botanicgarden}}}         
              \\ \cmidrule(lr){2-5} \cmidrule(lr){6-9} \cmidrule(lr){10-11} \cmidrule(lr){12-13}  \cmidrule(lr){14-15}
Sequence         & \multicolumn{2}{c}{\texttt{Roundabout}}   & \multicolumn{2}{c}{\texttt{Town}}          & \multicolumn{2}{c}{\texttt{Outdoor}} & \multicolumn{2}{c}{\texttt{Tunnel}}
                 & \multicolumn{2}{c}{\texttt{Venman}}       & \multicolumn{2}{c}{\texttt{NTU}}           & \multicolumn{2}{c}{\texttt{1006-01}} \\ \hline

GT Alignment     & \multicolumn{2}{c}{\checkmark}            & \multicolumn{2}{c}{\checkmark}             & \multicolumn{2}{c}{\checkmark} 
                 & \multicolumn{2}{c}{\xmark}                & \multicolumn{2}{c}{\xmark}                 & \multicolumn{2}{c}{\checkmark}
                 & \multicolumn{2}{c}{\xmark}  \\ 

Map size         & \multicolumn{2}{c}{7.6\,GB}   & \multicolumn{2}{c}{7.3\,GB}    & \multicolumn{2}{c}{217.0\,MB} & \multicolumn{2}{c}{337.3\,MB}
                 & \multicolumn{2}{c}{26.4\,GB}  & \multicolumn{2}{c}{211.9\,MB}  & \multicolumn{2}{c}{130.6\,MB} \\ 

Duration Avg.    & \multicolumn{2}{c}{2,729.761\,sec}  & \multicolumn{2}{c}{2,414.119\,sec}  & \multicolumn{2}{c}{1,159.260\,sec} & \multicolumn{2}{c}{916.697\,sec}
                 & \multicolumn{2}{c}{2,293.620\,sec}  & \multicolumn{2}{c}{597.900\,sec}    & \multicolumn{2}{c}{600.300\,sec} \\ 

Robot Num.       & \multicolumn{2}{c}{$\robot$ $\robot$ $\robot$}                                & \multicolumn{2}{c}{$\robot$ $\robot$ $\robot$}  
                 & \multicolumn{2}{c}{$\robot$ $\robot$ $\robot$ $\robot$ $\robot$}              & \multicolumn{2}{c}{$\robot$ $\robot$ $\robot$ $\robot$ $\robot$}
                 & \multicolumn{2}{c}{$\robot$ $\robot$ $\robot$}                                & \multicolumn{2}{c}{$\robot$ $\robot$ $\robot$ $\robot$ $\robot$} 
                 & \multicolumn{2}{c}{$\robot$ $\robot$ $\robot$} \\ 

Overlap Deg.     & \multicolumn{2}{c}{Complete (\,$\complete$\,)}     & \multicolumn{2}{c}{Complete (\,$\complete$\,)} 
                 & \multicolumn{2}{c}{Partial (\,$\smalloverlap$\,)}  & \multicolumn{2}{c}{Partial (\,$\smalloverlap$\,)}        
                 & \multicolumn{2}{c}{Complete (\,$\complete$\,)}     & \multicolumn{2}{c}{Moderate (\,$\moderate$\,)}
                 & \multicolumn{2}{c}{Partial (\,$\smalloverlap$\,)}   \\ 

LiDAR sensor     & \multicolumn{2}{c}{Ouster OS2-128 (\,$\spin$\,)}  & \multicolumn{2}{c}{Ouster OS2-128 (\,$\spin$\,)} & \multicolumn{2}{c}{Velodyne VLP-16 (\,$\spin$\,)} & \multicolumn{2}{c}{Velodyne VLP-16 (\,$\spin$\,)}
                 & \multicolumn{2}{c}{Submap (\,$\spin$\,)}          & \multicolumn{2}{c}{Livox Mid-70 (\,$\livox$\,)}   & \multicolumn{2}{c}{Livox Avia (\,$\livox$\,)} \\ 

Environment      & \multicolumn{2}{c}{Urban}                          & \multicolumn{2}{c}{Urban} 
                 & \multicolumn{2}{c}{Campus}                         & \multicolumn{2}{c}{Indoor}        
                 & \multicolumn{2}{c}{Jungle}                         & \multicolumn{2}{c}{Campus}
                 & \multicolumn{2}{c}{Park}   \\ 

Central Map      & \multicolumn{2}{c}{\raisebox{-0.5\height}{\includegraphics[width=2.0cm]{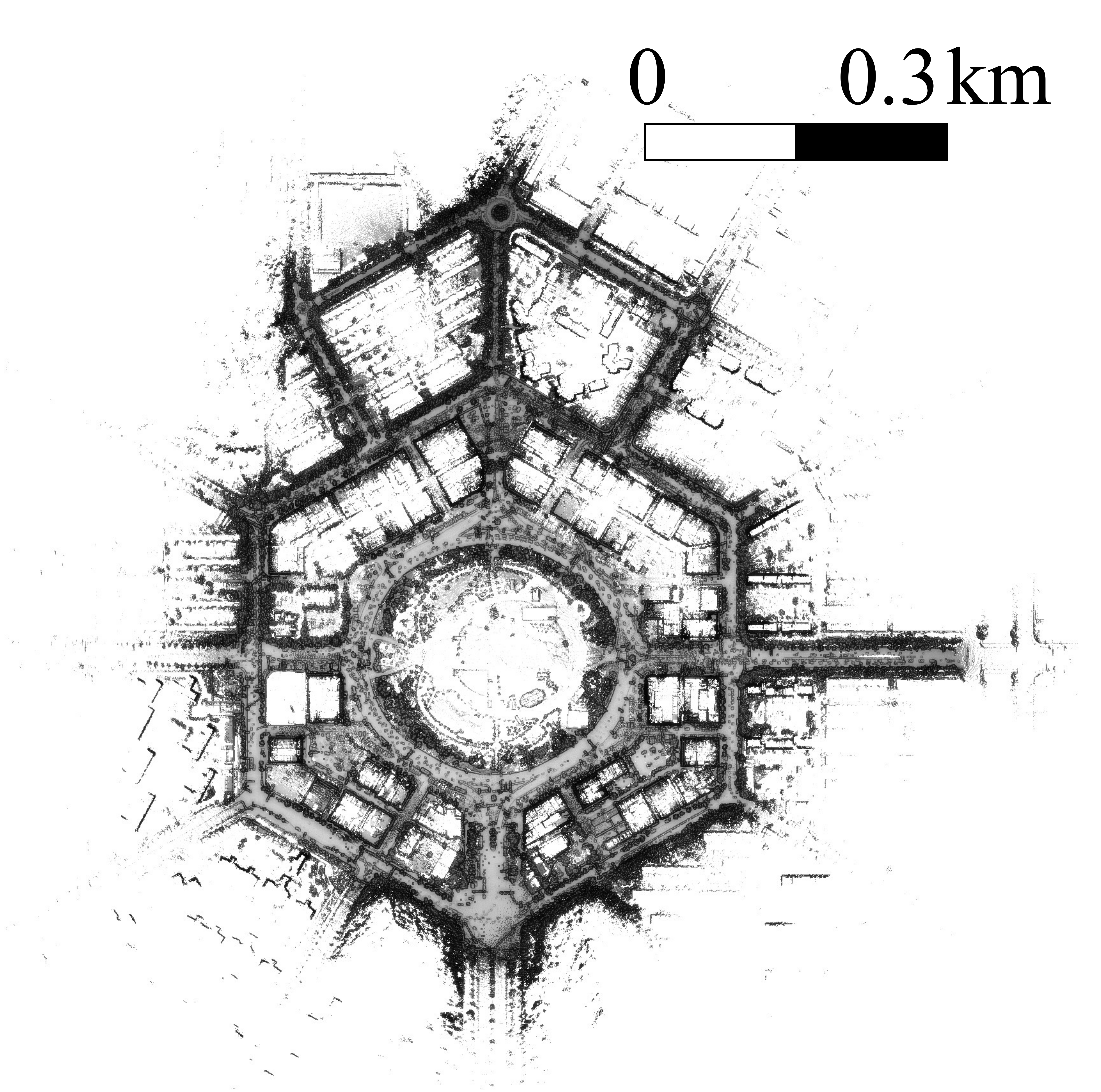}}}
                 & \multicolumn{2}{c}{\raisebox{-0.5\height}{\includegraphics[width=2.0cm]{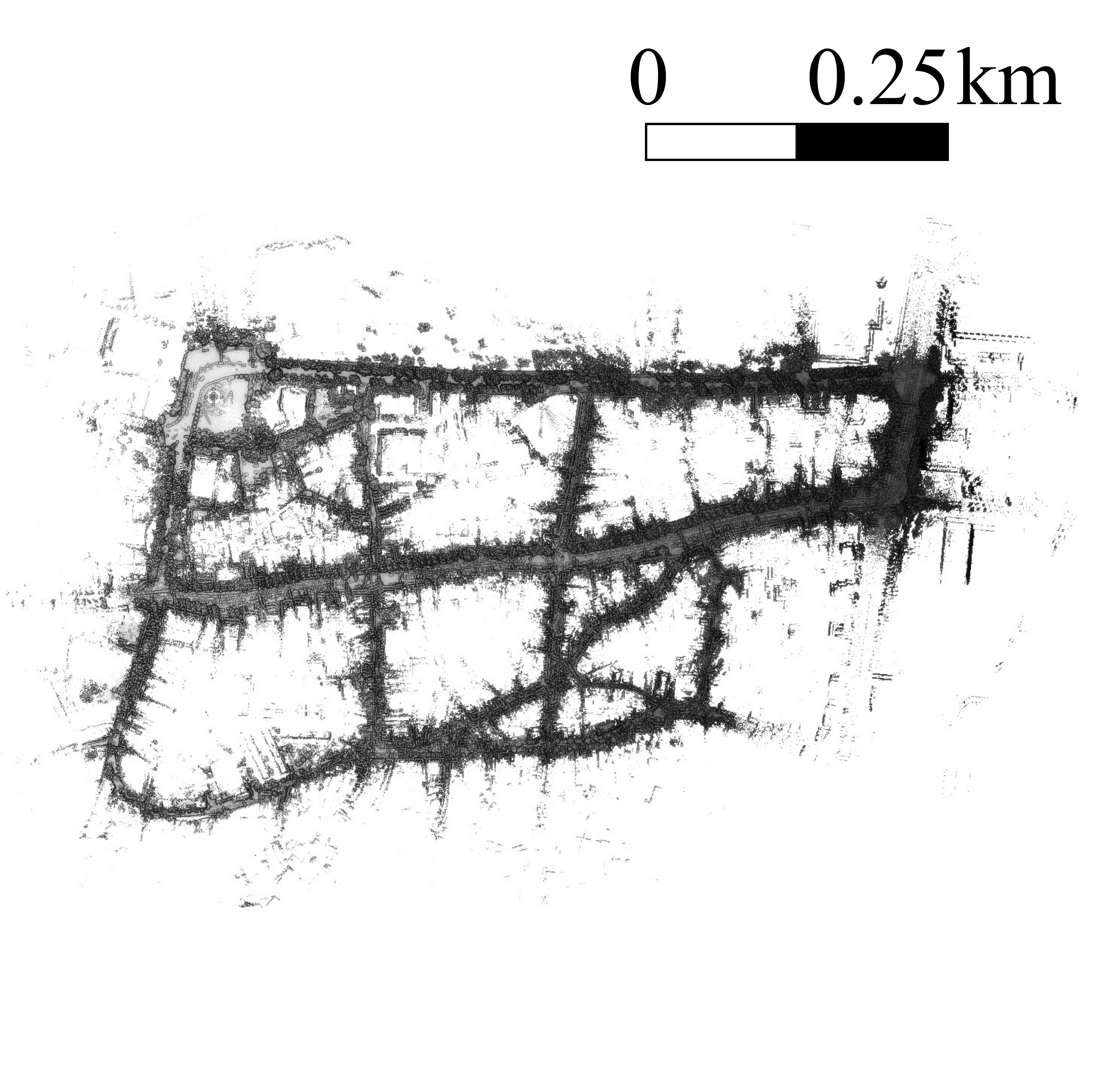}}}
                 & \multicolumn{2}{c}{\raisebox{-0.5\height}{\includegraphics[width=2.0cm]{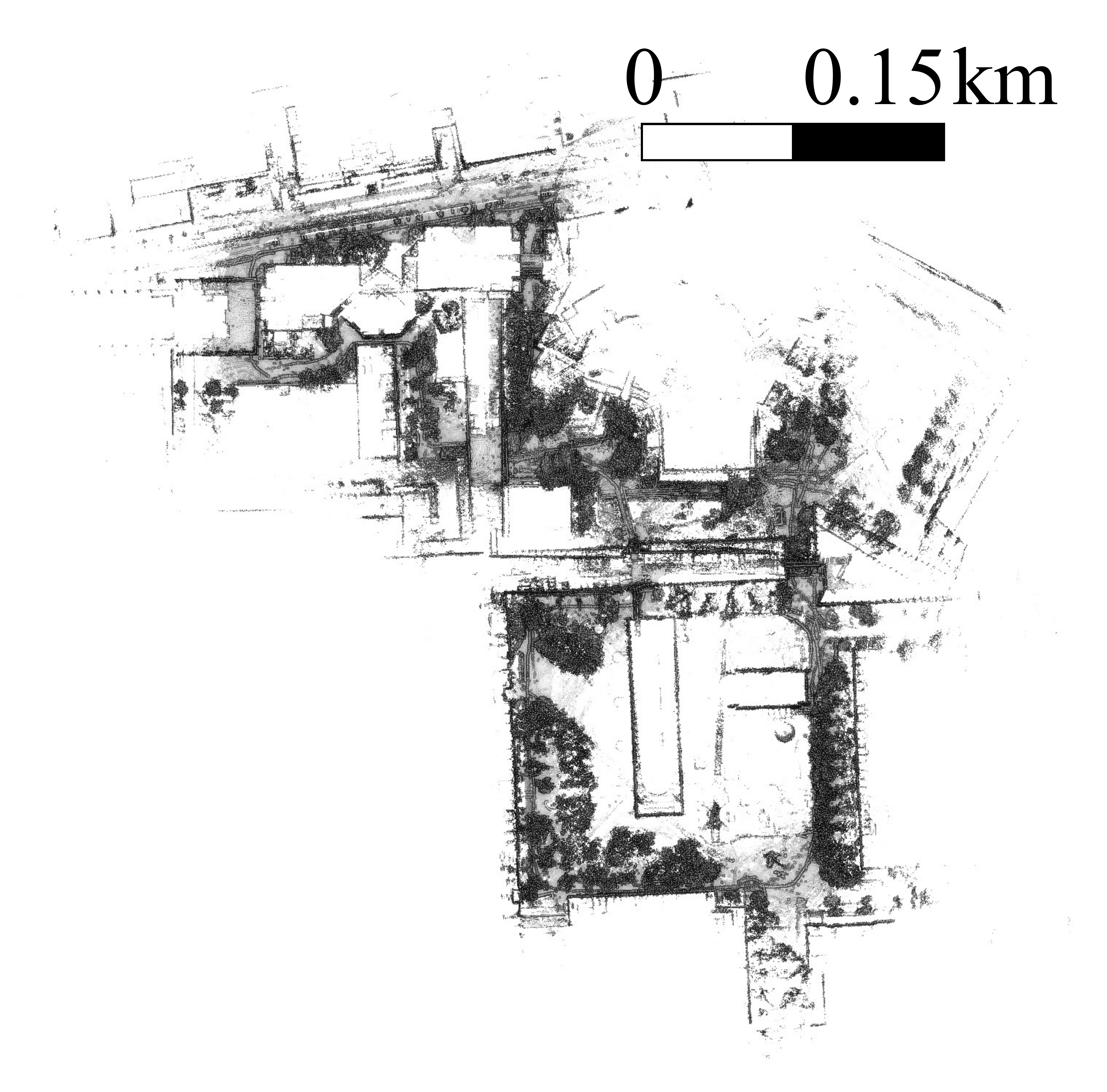}}}
                 & \multicolumn{2}{c}{\raisebox{-0.5\height}{\includegraphics[width=2.0cm]{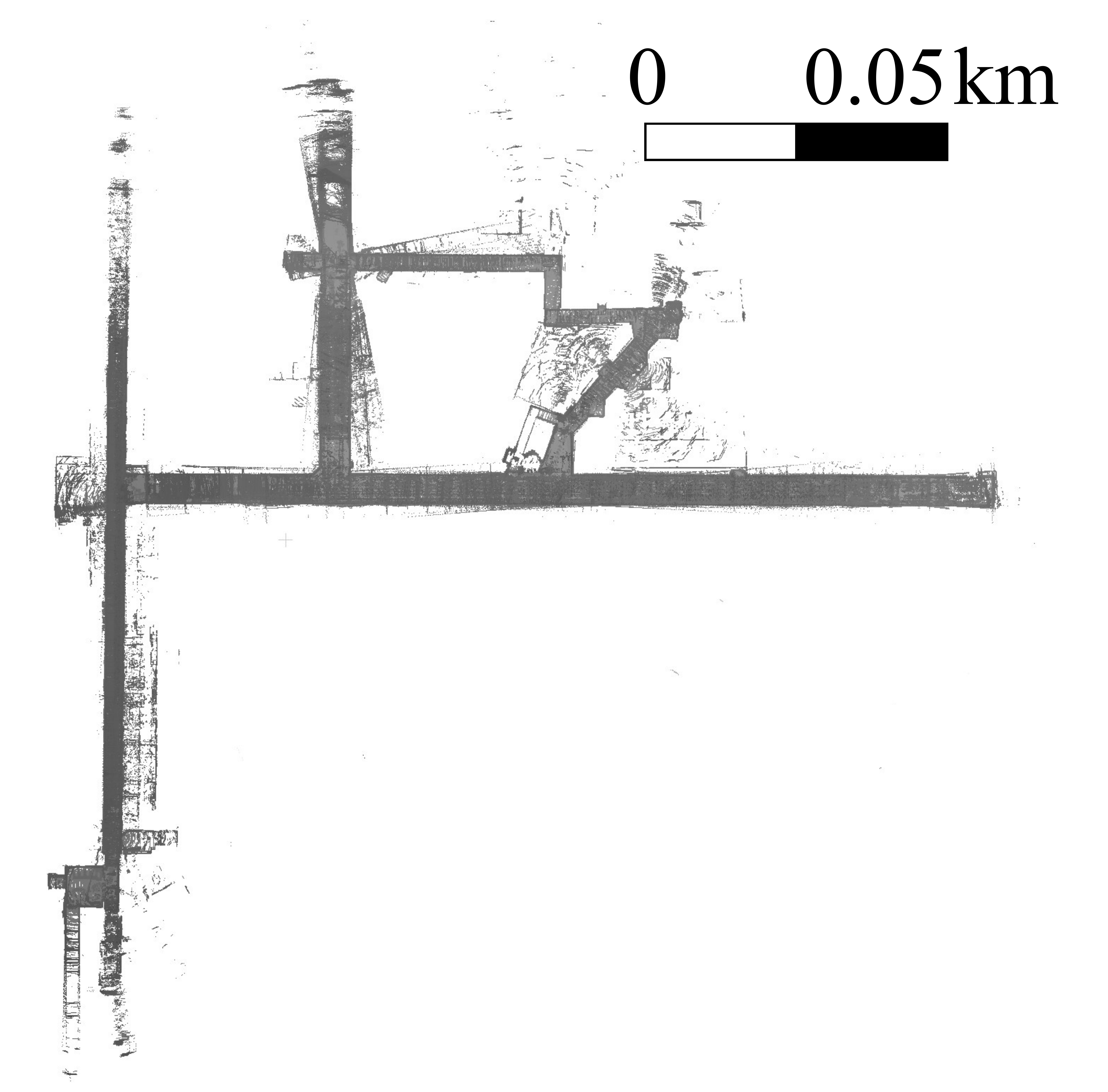}}}
                 & \multicolumn{2}{c}{\raisebox{-0.5\height}{\includegraphics[width=2.0cm]{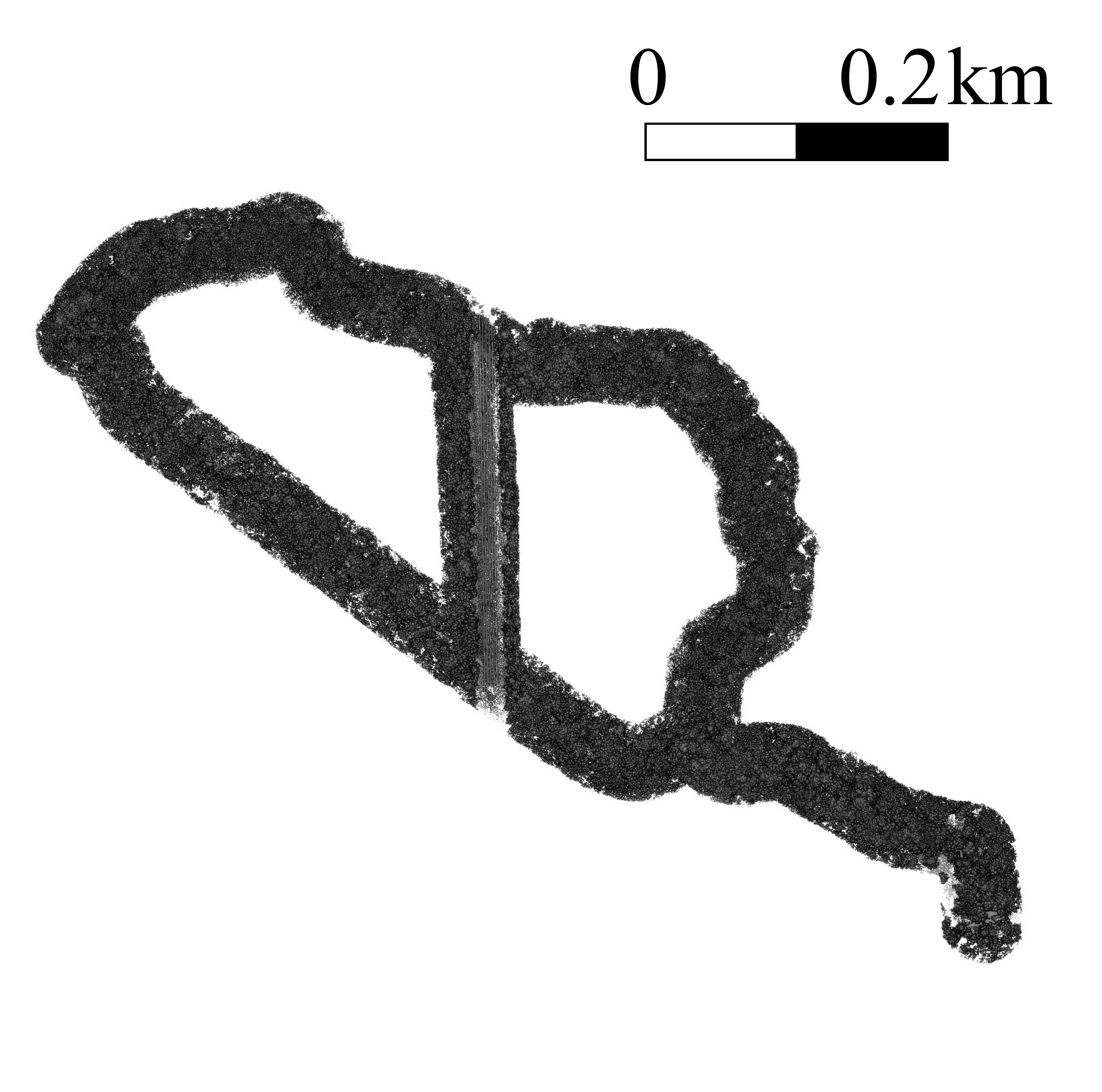}}}
                 & \multicolumn{2}{c}{\raisebox{-0.5\height}{\includegraphics[width=2.0cm]{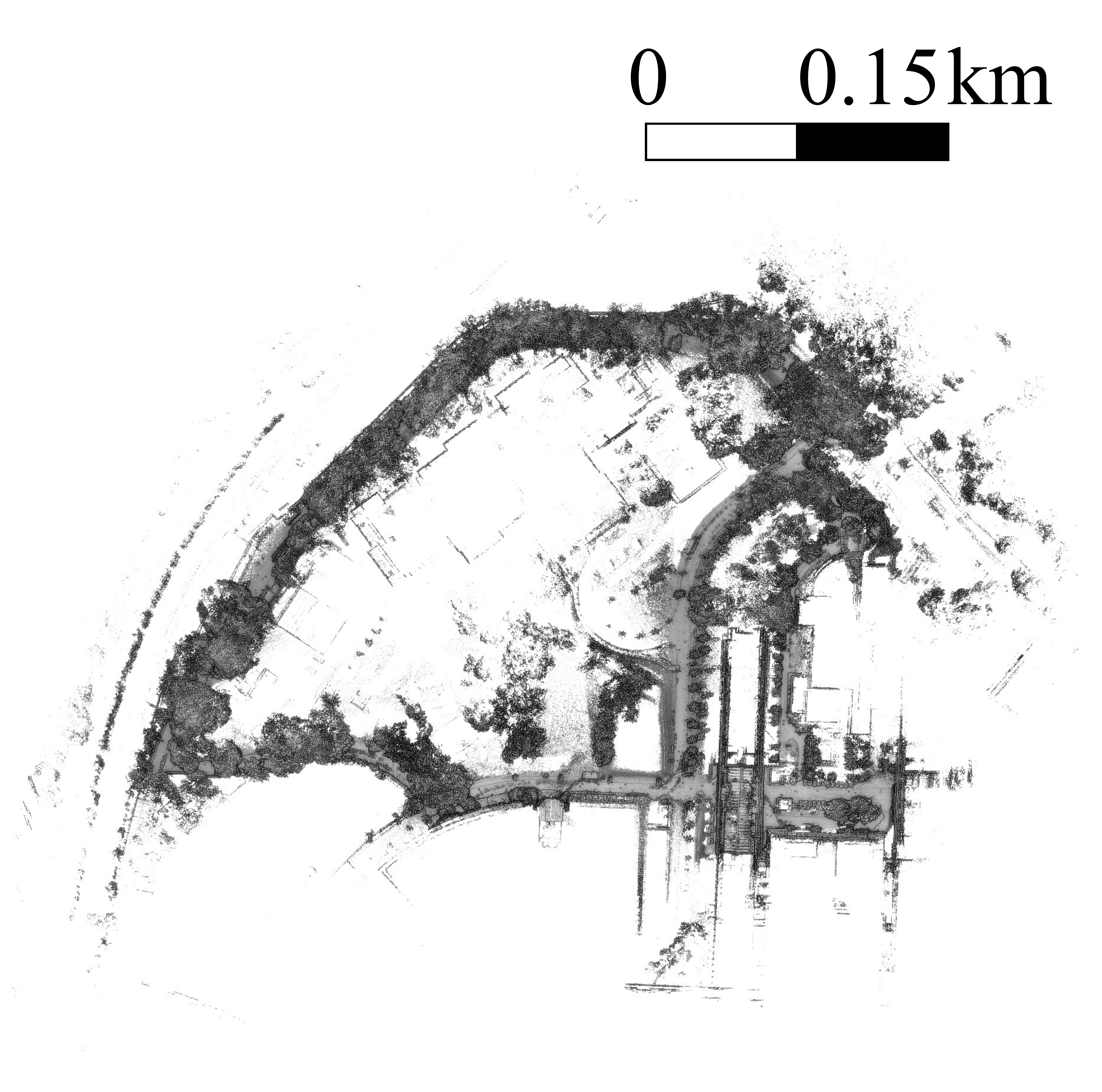}}}
                 & \multicolumn{2}{c}{\raisebox{-0.5\height}{\includegraphics[width=2.0cm]{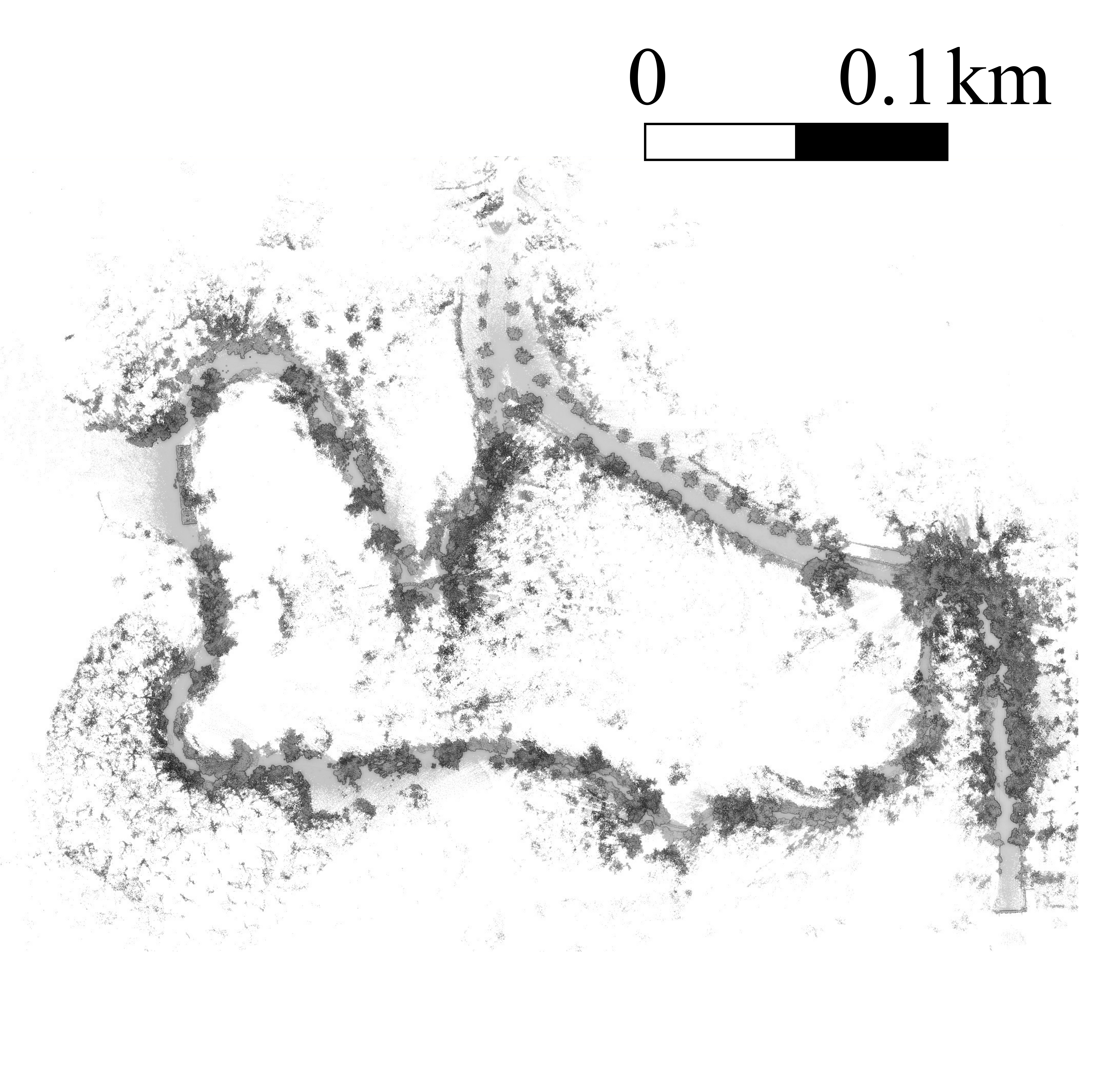}}}  \\

\midrule \bottomrule
\end{tabular}}
\label{tab:datasets}
\vspace{-2mm}
\end{table*}
% ===============================================================
% % =============================================================
\begin{figure*}[t]
    \centering
    \captionsetup{justification=justified}
    \captionsetup[subfigure]{justification=centering}
    \begin{subfigure}{0.495\columnwidth}
        \includegraphics[trim={0 0 0 0},clip, width=\columnwidth]{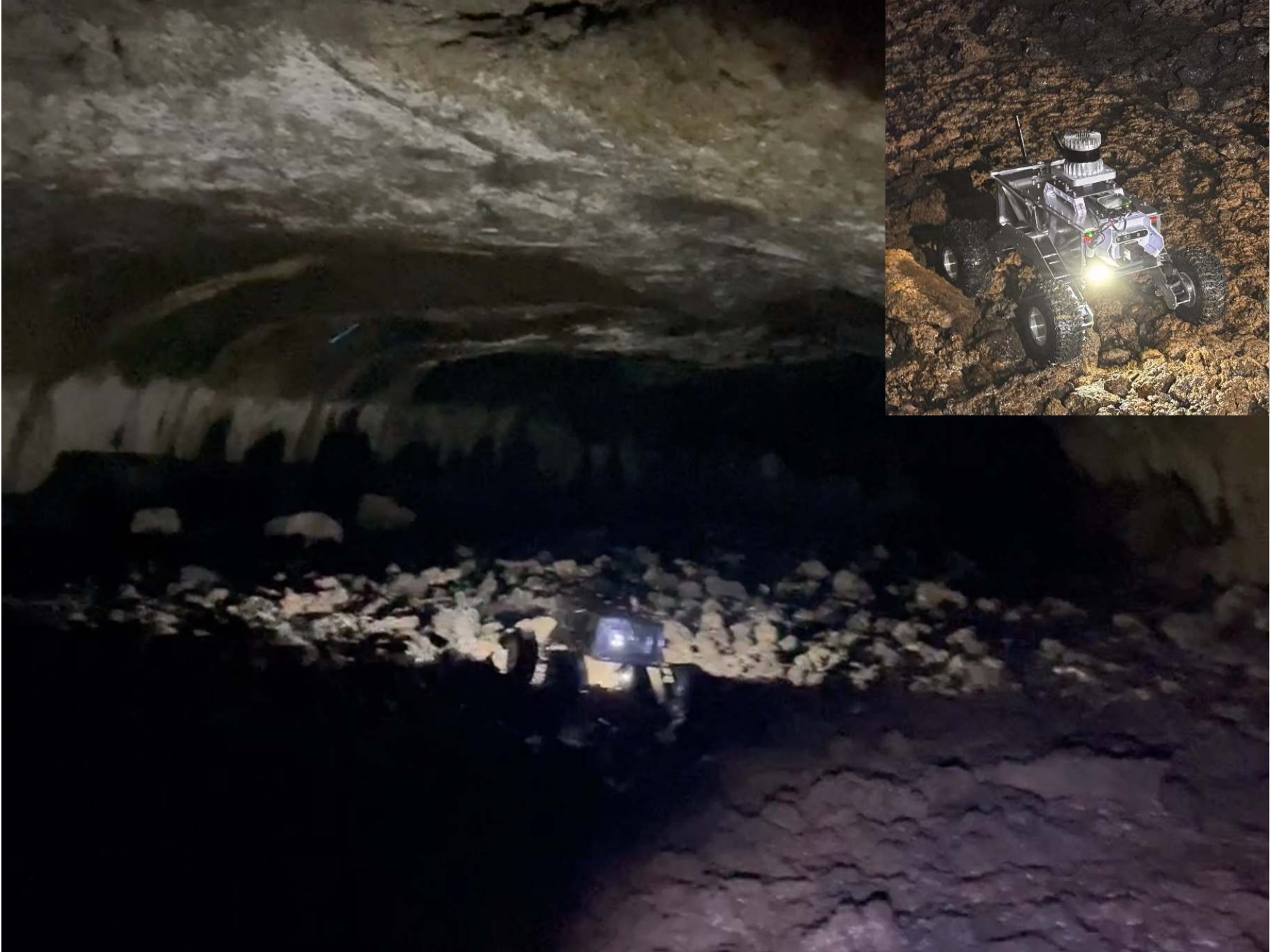}
        \vspace{-1.5em}
        \caption{INHA-\texttt{Cave}}
    \end{subfigure}
    \begin{subfigure}{0.495\columnwidth}
        \includegraphics[trim={0 0 0 0},clip, width=\columnwidth]{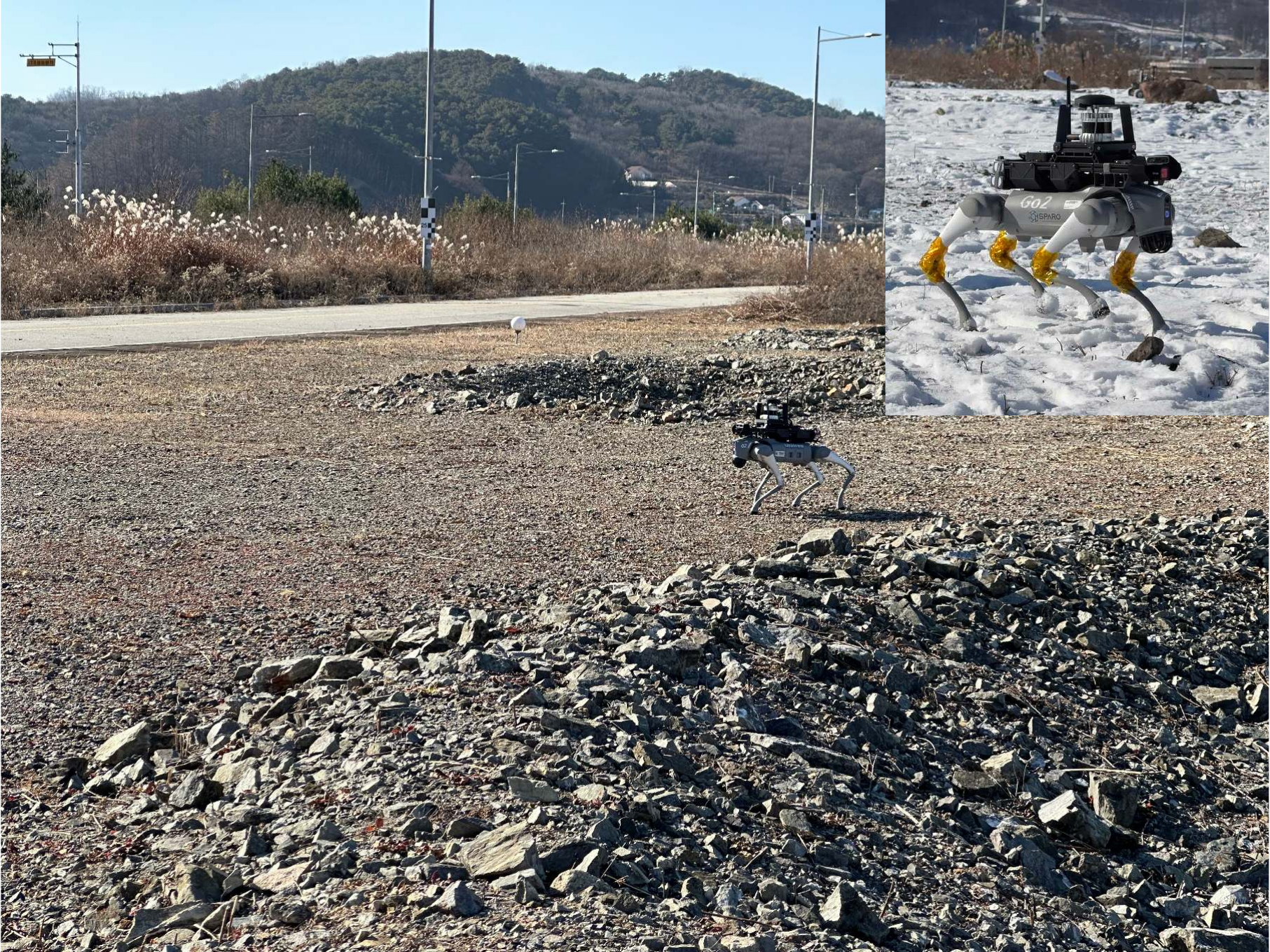}
        \vspace{-1.5em}
        \caption{INHA-\texttt{Planetary}}
    \end{subfigure}
    \begin{subfigure}{0.495\columnwidth}
        \includegraphics[trim={0 0 0 0},clip, width=\columnwidth]{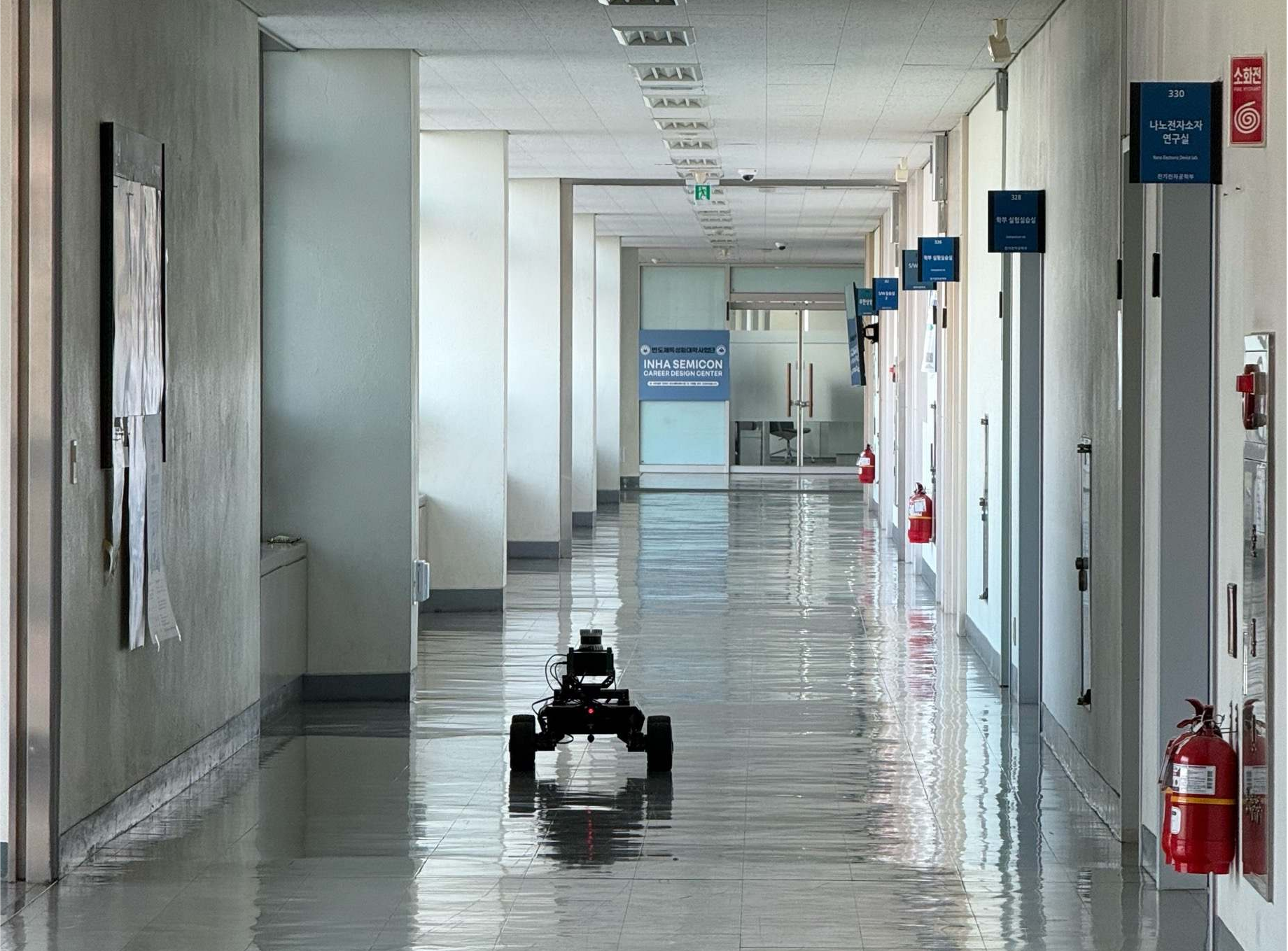}
        \vspace{-1.5em}
        \caption{INHA-\texttt{Indoor}}
    \end{subfigure}
    \begin{subfigure}{0.495\columnwidth}
        \includegraphics[trim={0 0 0 0},clip, width=\columnwidth]{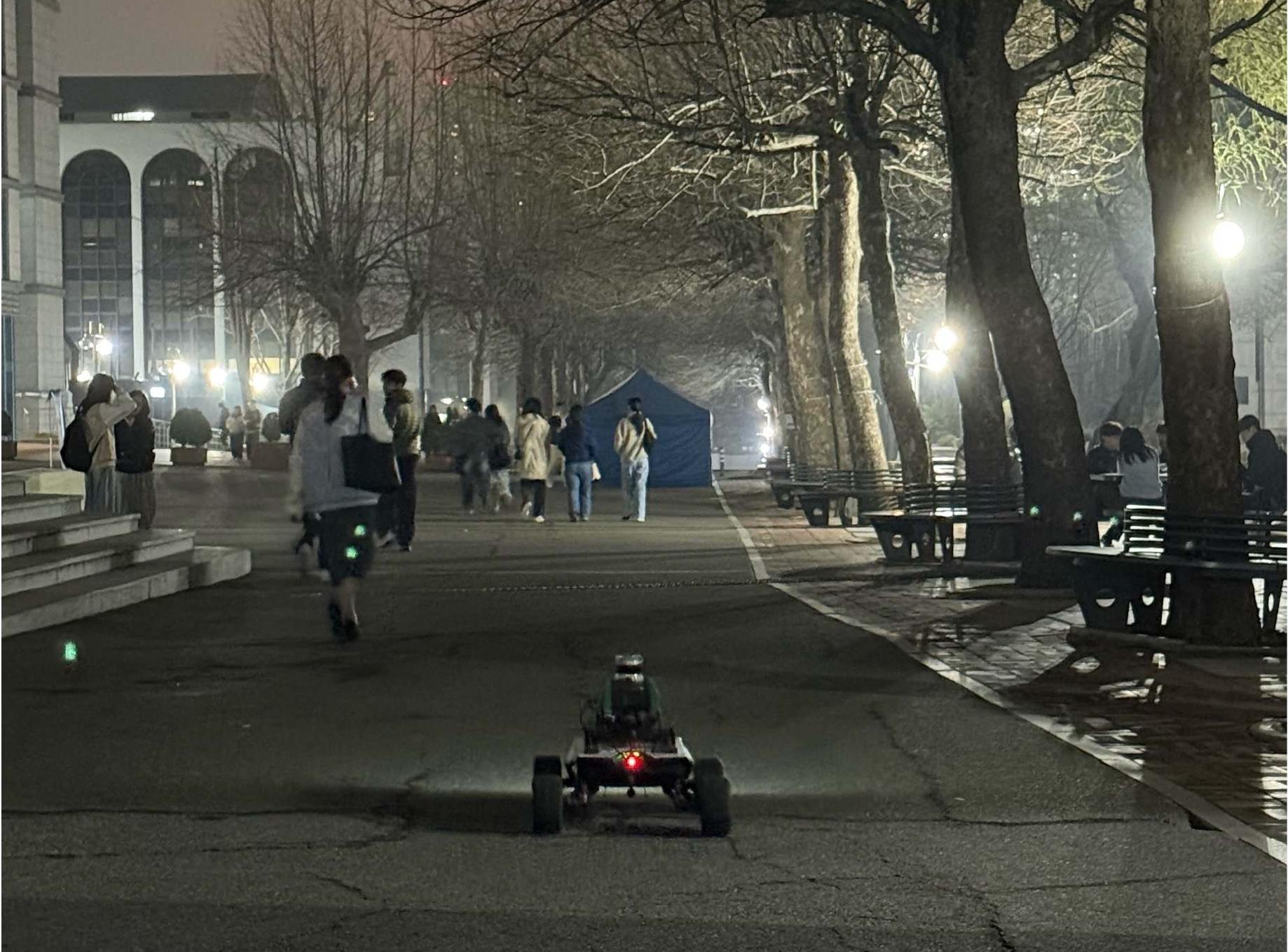}
        \vspace{-1.5em}
        \caption{INHA-\texttt{Outdoor}}
    \end{subfigure}
    \vspace{-0.15cm}
    \caption{Real-world robot deployments for in-house multi-robot dataset acquisition.
    (a)~Subterranean lava cave with a custom rover.
    (b)~Planetary-analog terrain with a legged robot.
    (c)~Indoor corridor with a wheeled robot.
    (d)~Nighttime campus with a wheeled robot.}
    \label{fig:robot_deploy}
    \vspace{-4mm}
\end{figure*}
% % =============================================================
% % =============================================================
\begin{figure*}[t]
    \centering
    \captionsetup{justification=justified}
    \captionsetup[subfigure]{justification=centering}
    \begin{subfigure}{0.38\columnwidth}
        \includegraphics[trim={0 0 0 0},clip, width=\columnwidth]{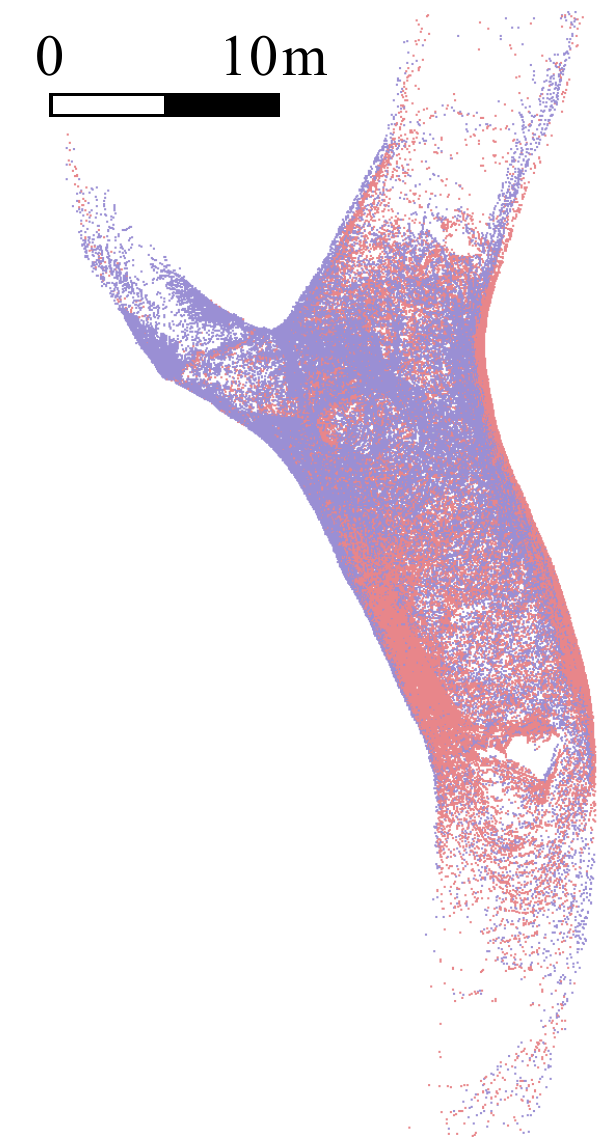}
        \vspace{-1.5em}
        \caption{INHA-\texttt{Cave}}
    \end{subfigure}
    \begin{subfigure}{0.41\columnwidth}
        \includegraphics[trim={0 0 0 0},clip, width=\columnwidth]{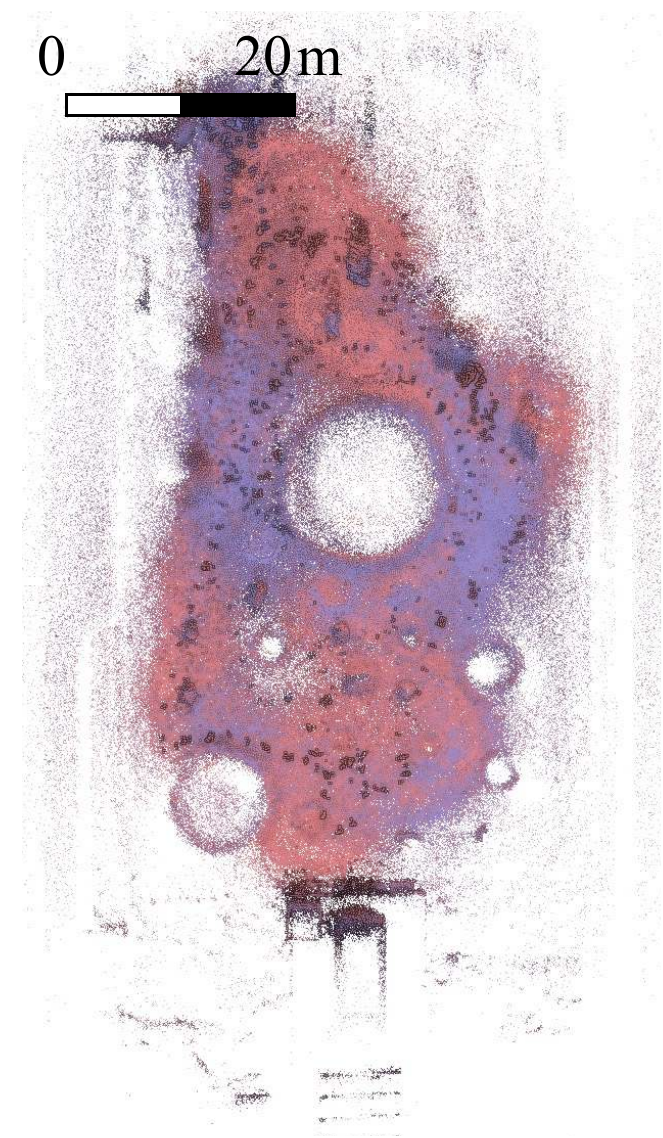}
        \vspace{-1.5em}
        \caption{INHA-\texttt{Planetary}}
    \end{subfigure}
    \begin{subfigure}{0.525\columnwidth}
        \includegraphics[trim={0 0 0 0},clip, width=\columnwidth]{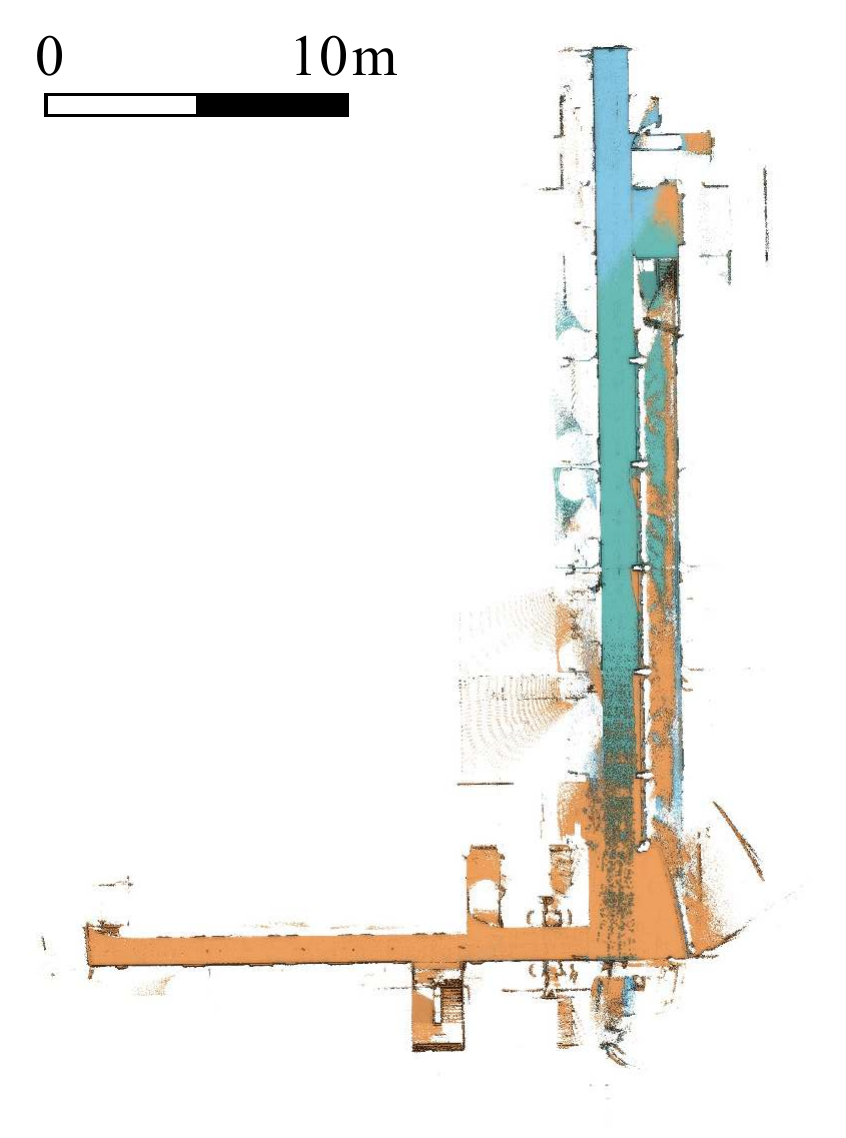}
        \vspace{-1.5em}
        \caption{INHA-\texttt{Indoor}}
    \end{subfigure}
    \begin{subfigure}{0.645\columnwidth}
        \includegraphics[trim={0 0 0 0},clip, width=\columnwidth]{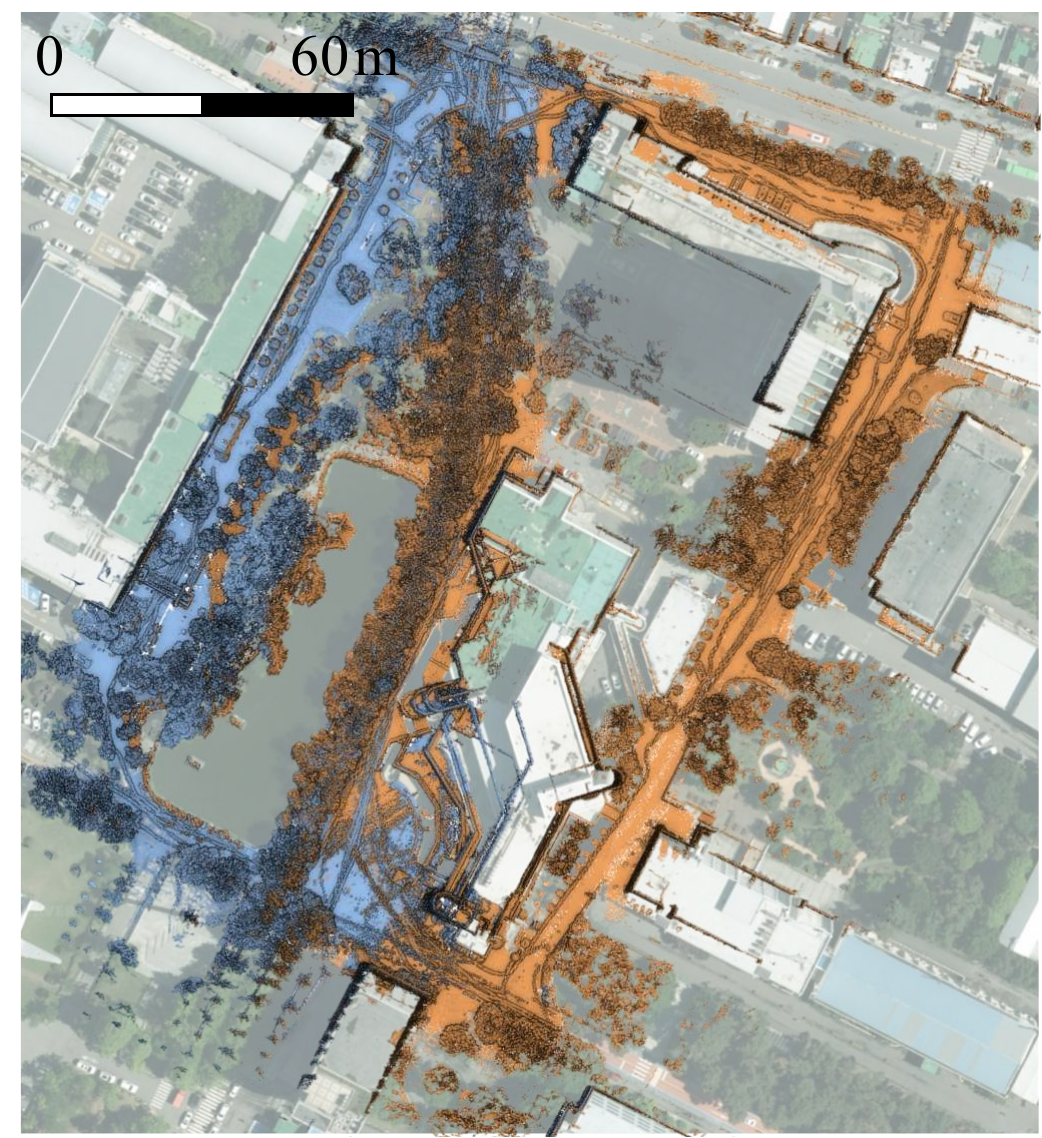}
        \vspace{-1.5em}
        \caption{INHA-\texttt{Outdoor}}
    \end{subfigure}
    \vspace{-0.15cm}
    \caption{Multi-robot map merging results produced by Commerge on each in-house dataset, where different colors indicate maps acquired by different robots.
    (a)~Cave. (b)~Planetary-analog terrain. (c)~Indoor corridor. (d)~Outdoor campus, overlaid on a satellite map for reference.}
    \label{fig:map_deploy}
    \vspace{-3mm}   
\end{figure*}
% % =============================================================
% % =============================================================
\begin{figure*}[h]
    \centering
    \captionsetup{justification=justified}
    \captionsetup[subfigure]{justification=centering}
    \includegraphics[trim={0 80 0 80},clip, width=1.99\columnwidth]{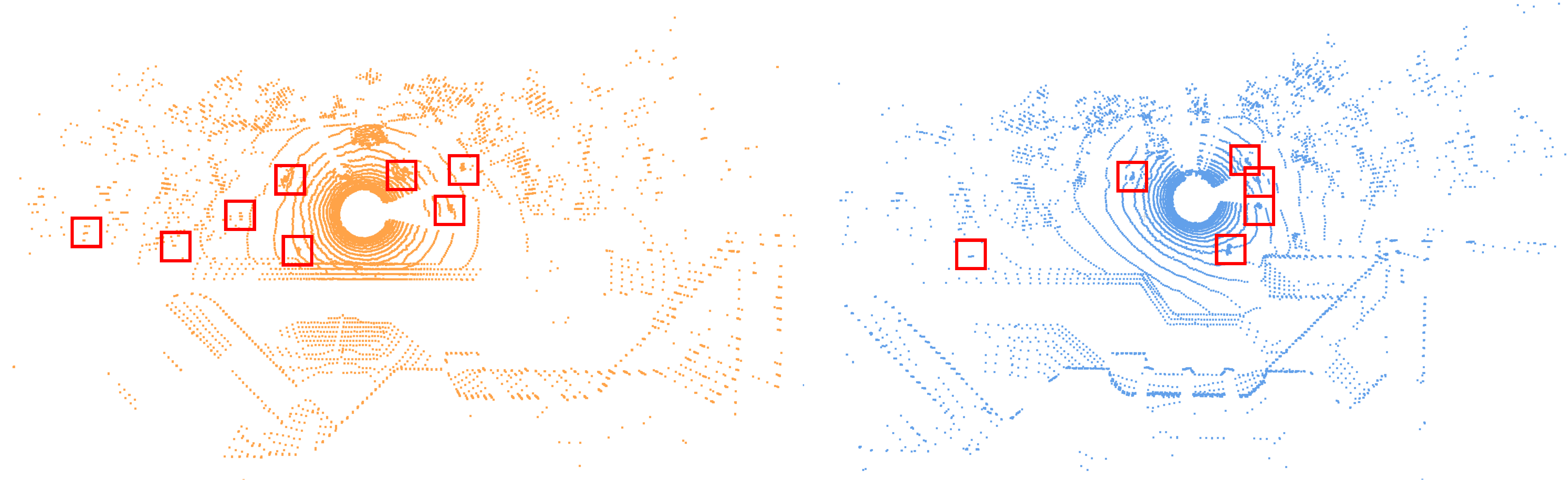}
    \vspace{-0.15cm}
    \caption{Point clouds from robot~$\alpha$ (left, orange) and robot~$\beta$ (right, blue) in the INHA-\texttt{Outdoor} sequence. 
                 Red boxes indicate dynamic pedestrians.}
    \label{fig:dynamic}
    \vspace{-3mm}
\end{figure*}
% % =============================================================
Furthermore, the three metrics for the place recognition ablation study are defined as follows:
\begin{itemize}
    \item Recall@1 = $\frac{\text{TP}_1}{N_{\text{GT}}}$,
    \item AUC Score = $\int_0^1 \text{Precision}(r)$,
    \item F1 Score = $2 \times \frac{\text{Precision} \times \text{Recall}}{(\text{Precision} + \text{Recall})}$,
\end{itemize}
where $N_{\text{GT}}$ is the number of ground truth loop closures in the dataset, $\text{TP}_1$ is the number of queries where the top-ranked match is a true positive, $\text{Precision}(r)$ denotes 
the precision at recall level $r$, $\text{Precision} = \frac{\text{TP}}{\text{TP} + \text{FP}}$, $\text{Recall} = \frac{\text{TP}}{\text{TP} + \text{FN}}$, and TP, FP, and FN denote true positives, false positives, and false negatives from descriptor matching against ground truth correspondences.

Finally, the three metrics for the registration ablation, namely relative translation error (RTE), relative rotation error (RRE), and success rate, are defined as follows:
\begin{itemize}
    \item RTE = $\frac{1}{N_{\text{success}}} \sum_{n=1}^{N_{\text{success}}} \left\| \mathbf{t}_{n,\text{GT}} - \hat{\mathbf{t}}_n \right\|_2^2$,
    \item RRE = $\frac{180}{\pi N_{\text{success}}} \sum_{n=1}^{N_{\text{success}}} \left| \cos^{-1}\left(\frac{\text{Tr}(\hat{\mathbf{R}}_n^\top \mathbf{R}_{n,\text{GT}}) - 1}{2}\right) \right|$,
    \item Success Rate = $\frac{N_{\text{success}}}{N_{\text{total}}} \times 100\,\%$,
\end{itemize}
where $N_{\text{success}}$ is the number of registration attempts that converge within acceptable error thresholds (\ie RTE $< 2$\,m and RRE $< 5^\circ$), $N_{\text{total}}$ is the total number of registration attempts, $\mathbf{t}_{n,\text{GT}}$ and $\hat{\mathbf{t}}_n$ are the ground truth and estimated translations, and $\mathbf{R}_{n,\text{GT}}$ and $\hat{\mathbf{R}}_n$ are the ground truth and estimated rotations.

\subsection{Public Datasets}
\label{sec:datasets}
\noindent
In addition, we conducted experiments across various datasets to evaluate the performance of \textit{Commerge}.
Table~\ref{tab:datasets} summarizes the datasets used in our experiments, including diverse characteristics that affect multi-robot mapping performance~(\eg map size, mission duration, robot number, overlap degree, LiDAR sensor type, and environment).

Since public datasets with truly simultaneous multi-robot data acquisition are scarce, we follow the standard evaluation protocol used in prior multi-robot mapping literature: single-robot sequences are converted into multi-robot scenarios through temporal alignment \citep{huang2021disco, zhong2023dcl, kim2025skid, wei2024large}.
Specifically, all robot sequences are time-shifted to align with a designated central sequence, creating synchronized multi-robot trajectories.
This processing enables controlled evaluation of map merging algorithms under various perspectives while maintaining reproducibility and fair comparison across methods.

Note that quantitative evaluation is limited for certain datasets: the Botanic Garden dataset lacks ground truth alignment between robot trajectories, while WildPlaces and Kimera-Multi's \texttt{Tunnel} sequences present geometrically degenerate environments where LiDAR odometry exhibits substantial drift, preventing reliable assessment.
In these cases, we employ qualitative assessment through visual consistency and quantitative evaluation using ground truth trajectories with arbitrary rotations, respectively.

% =============================================================
\begin{table*}[t]
\captionsetup{width=\textwidth, justification=justified}\caption{Map merging performance comparison (\ie $\rmsemetric$) against state-of-the-art multi-robot map merging methods on the $\hardA$ platform across four representative datasets with varying scales and complexity (from small-scale \texttt{NTU} to large-scale \texttt{Town}, left to right).
Out-of-Memory~(OOM) indicates methods that exhausted 32\,GB RAM and failed to execute.
}
\renewcommand{\arraystretch}{1.2}
\centering\resizebox{\textwidth}{!}{\tiny
\begin{tabular}{l|cccccccccccccc}
\toprule
\midrule
                  Sequence        & \multicolumn{3}{c}{\texttt{NTU}}
                                    & \multicolumn{5}{c}{\texttt{Outdoor}} 
                                    & \multicolumn{3}{c}{\texttt{Roundabout}}
                                    & \multicolumn{3}{c}{\texttt{Town}}   \\ \cmidrule(lr){2-4} \cmidrule(lr){5-9} \cmidrule(lr){10-12} \cmidrule(lr){13-15}
                  Robot Info.     & \texttt{01-02}                & \texttt{01-10}           & Avg.
                                    & \texttt{A01-A02}              & \texttt{A01-H01}         & \texttt{A01-S01}         & \texttt{A01-S02}         & Avg.
                                    & \texttt{01-02}                & \texttt{01-03}           & Avg.
                                    & \texttt{01-02}                & \texttt{01-03}           & Avg.                                                               \\  \midrule

% \multirow{6}{*}{\rotatebox[origin=c]{90}{$\rmsemetric$}}
                  LT-mapper       & \thirdc 3.679\,m           & \thirdc 3.340\,m         & \thirdc 3.510\,m                                                       
                                    & 121.195\,m                 & \secondc 4.328\,m        & 270.756\,m               & 86.612\,m                & 121.973\,m       
                                    & \secondc 7.787\,m          & \secondc 11.691\,m       & \secondc 9.739\,m
                                    & \thirdc 9.653\,m           & \thirdc 9.352\,m         & \thirdc 9.503\,m                                                        \\
                  ELite           & 149.194\,m                 & 245.532\,m               & 197.363\,m
                                    & 168.183\,m                 & 333.118\,m               & \thirdc 251.057\,m       & 49.770\,m                & 200.532\,m        
                                    & OOM                        & OOM                      & OOM
                                    & OOM                        & OOM                      & OOM                                                              \\
                  LAMM            & 172.979\,m                 & 227.212\,m               & 200.096\,m
                                    & \firstc \textbf{4.154}\,m           & 380.226\,m               & 254.106\,m               & \secondc 10.512\,m       & \thirdc 79.925\,m
                                    & OOM                        & OOM                      & OOM
                                    & OOM                        & OOM                      & OOM                                                              \\
                  Uni-Mapper      & 63.354\,m                  & 226.885\,m               & 145.120\,m
                                    & 131.386\,m                 & \firstc \textbf{2.884}\,m         & 254.398\,m               & \firstc \textbf{10.509}\,m        & 99.794\,m         
                                    & OOM                        & OOM                      & OOM
                                    & OOM                        & OOM                      & OOM                                                              \\
                  KISS-Matcher    & \firstc \textbf{0.679}\,m  & \firstc \textbf{1.817}\,m & \firstc \textbf{1.248}\,m
                                    & \thirdc 20.427\,m          & 11.870\,m                & \firstc \textbf{22.020}\,m        & 37.532\,m                & \secondc 22.926\,m
                                    & \thirdc 15.726\,m          & \thirdc 19.753\,m        & \thirdc 17.740\,m
                                    & \secondc 3.940\,m          & \secondc 8.883\,m        & \secondc 6.412\,m                                                \\
                  Ours            & \secondc 1.783\,m          & \secondc 2.594\,m         & \secondc 2.189\,m
                                    & \secondc 9.624\,m          & \thirdc 7.818\,m          & \secondc 30.265\,m       & \thirdc 14.853\,m               & \firstc \textbf{15.640}\,m  
                                    & \firstc \textbf{1.843}\,m           & \firstc \textbf{3.807}\,m          & \firstc \textbf{2.825}\,m
                                    & \firstc \textbf{2.657}\,m           & \firstc \textbf{8.580}\,m          & \firstc \textbf{5.619}\,m                     \\ \midrule
\bottomrule
\end{tabular}}
\label{tab:overall}
\vspace{-3mm}
\end{table*}
% =============================================================

\subsection{In-House Datasets}
\label{sec:inhouse}
\noindent
We further validate the practicality of Commerge on our own in-house multi-robot datasets~(\ie INHA-\texttt{Cave}, INHA-\texttt{Planetary}, INHA-\texttt{Indoor}, and INHA-\texttt{Outdoor}) as summarized in Fig.~\ref{fig:robot_deploy} and Fig.~\ref{fig:map_deploy}, respectively.
To this end, we conduct network-aware experiments in Section~\ref{sec:practicality} under two complementary settings.
Note that INHA-\texttt{Cave} and INHA-\texttt{Planetary} datasets shown in Figs.~\ref{fig:robot_deploy}(a) and~\ref{fig:robot_deploy}(b) do not permit real wireless infrastructure due to the harshness of the environments, so we emulate delay, bandwidth limitation, and packet loss using NetEm~\citep{hemminger2005network}.
In contrast, INHA-\texttt{Indoor} and INHA-\texttt{Outdoor} datasets shown in Figs.~\ref{fig:robot_deploy}(c) and \ref{fig:robot_deploy}(d) allow us to construct an actual wireless network infrastructure that reproduces non-line-of-sight~(NLOS) conditions and real packet dropout.
In particular, INHA-\texttt{Outdoor} dataset further contains a large number of dynamic pedestrians, as shown in \figref{fig:dynamic}, which makes map merging even more challenging.
The hardware setup for communication is detailed in the following section.

\subsection{Communication Setup}
\label{sec:comm_setup}
\begin{figure}[t]
  \centering
  \includegraphics[width=\linewidth, trim=0 0 0 0, clip]{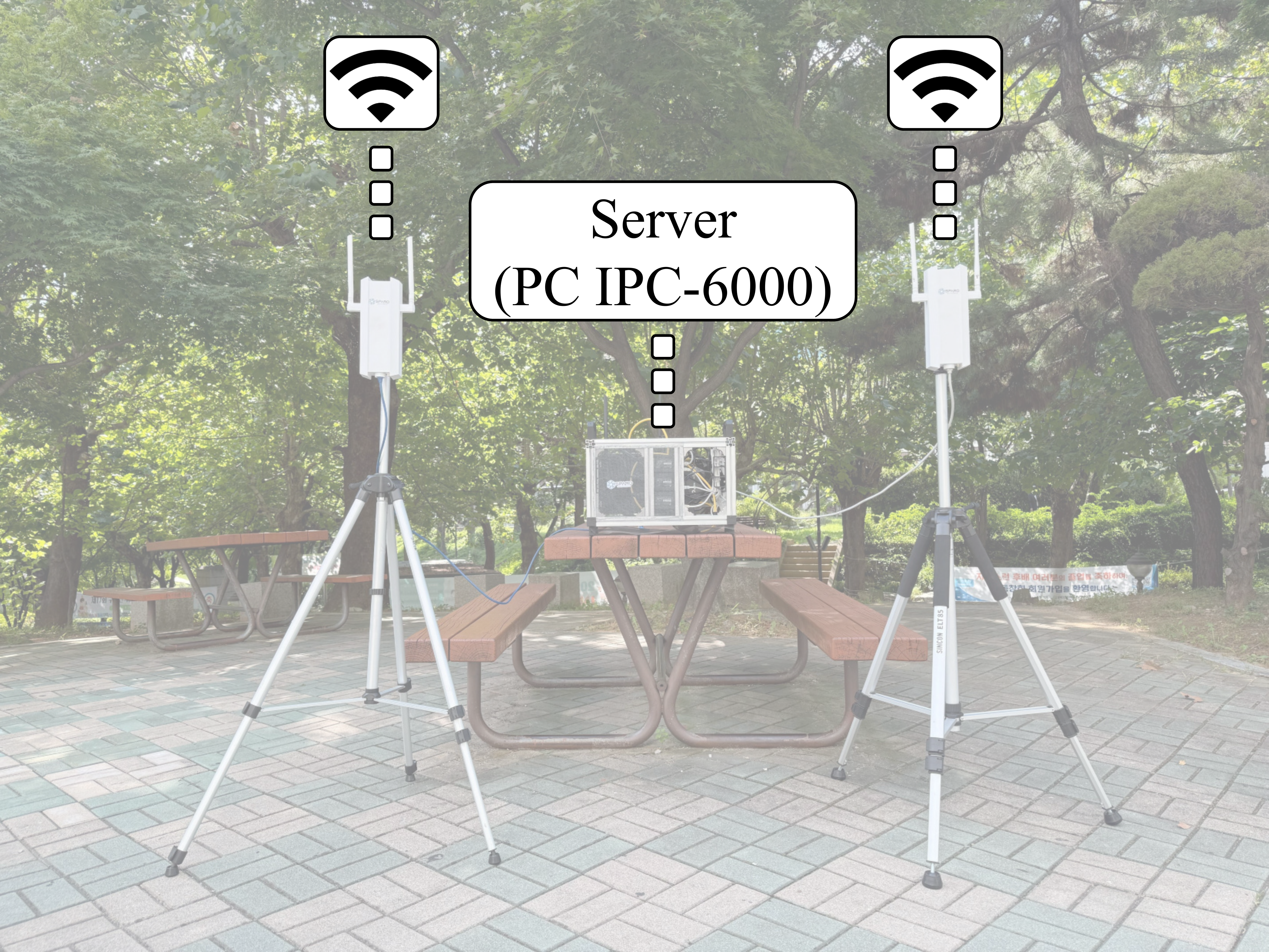}
  \captionsetup{justification=justified}
  \vspace{-0.6cm}
  \caption{Field communication setup used in our experiments. 
           A central server (\ie industrial mini PC $\hardB$) communicates with robots through two Omada WiFi Access Points~(left/right), forming a WLAN for outdoor trials.
           This infrastructure was used to exchange global descriptors and LiDAR scans via ROS/ZeroMQ and to measure communication time~($\commmetric$) and total exchange volume~($\exchmetric$).}
  \label{fig:comm_setup}
    \vspace{-4mm}
\end{figure}

\noindent
Finally, we configured a controlled wireless infrastructure using TP-Link Omada EAP245 WiFi access points, as shown in \figref{fig:comm_setup}, to evaluate $\commmetric$, $\exchmetric$, and practicality under realistic network conditions.
Since $\hardA$ lacks portability for field deployment due to its desktop form factor, we adopt $\hardB$ as the server platform for our communication experiments, balancing computational capability with practical deployability.
A detailed evaluation of system behavior under realistic network degradation scenarios is presented in Sections~\ref{sec:scalability}, \ref{sec:resource_efficiency}, and \ref{sec:practicality}.

\section{Experimental Results}
\label{sec:results}
\noindent
In our experiments, we compare against well-known centralized multi-robot mapping frameworks, as server-based coordination enables effective global management.
The baseline methods are as follows.
LT-Mapper~\citep{ltmapper} extends Scan Context to multi-robot scenarios using anchor-based pose graph optimization and dynamic object filtering.
ELite~\citep{gil2025ephemerality} employs point-level temporal-spatial context to filter ephemeral objects for lifelong mapping.
Uni-Mapper~\citep{kang2025uni} integrates dynamic object removal and multi-modal LiDAR fusion based on DynaSTD.
LAMM~\citep{wei2024large} combines BTC descriptors~\citep{yuan2024btc} with dynamic removal for efficient place recognition and dynamic filtering.
KISS-Matcher~\citep{kiss_matcher} performs fast map-level global registration for robot alignment.
% ====================================================================
% ====================================================================
% =============================================================
\begin{figure*}[t]
    \centering
    \captionsetup{justification=justified}
    \captionsetup[minipage]{justification=justified}
    % ---------- LEFT (a) ----------
    \begin{minipage}{0.9\columnwidth}
        \centering
        \captionsetup{justification=centering}
        \includegraphics[width=\linewidth]{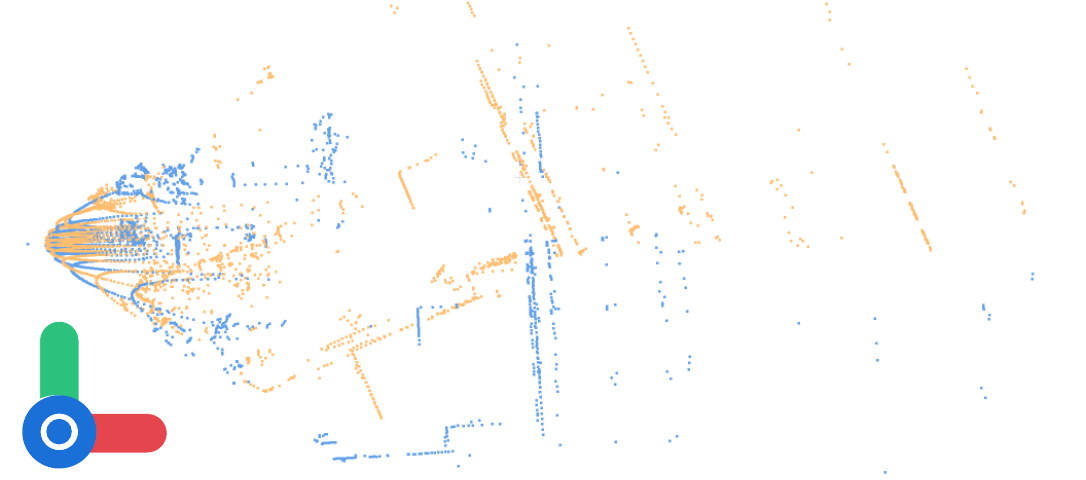}
        \vspace{-1.5em}
        \caption*{\footnotesize\textbf{(a)} Raw scan}
    \end{minipage}
    \hfill
    % ---------- RIGHT (b,c stacked) ----------
    \begin{minipage}{1.08\columnwidth}
        \centering
        \captionsetup{justification=centering}

        \includegraphics[width=\linewidth]{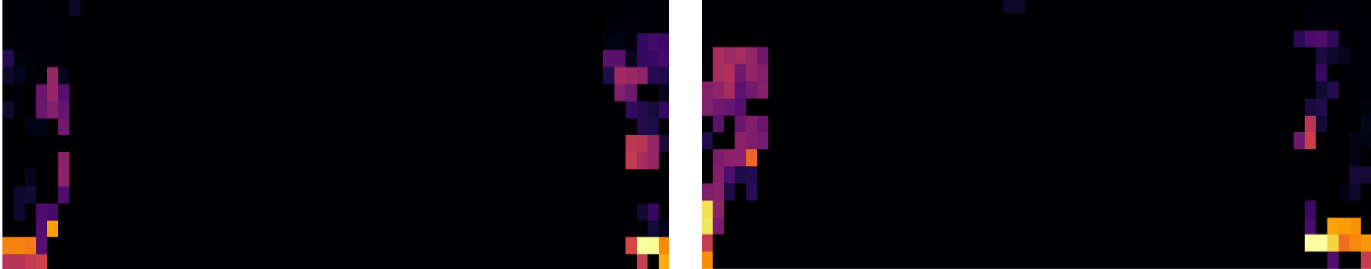}
        \vspace{-1.5em}
        \caption*{\footnotesize\textbf{(b)} Scan Context~\citep{kim2018scan}}

        \vspace{0.5em}

        \includegraphics[width=\linewidth]{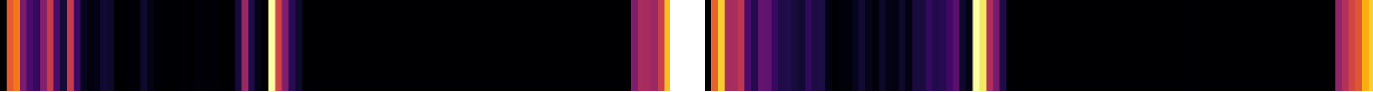}
        \vspace{-1.5em}
        \caption*{\footnotesize\textbf{(c)} SOLiD~\citep{kim2024narrowing}}

    \end{minipage}

    \caption{Qualitative comparison between Scan Context~\citep{kim2018scan} and SOLiD~\citep{kim2024narrowing} descriptors.
    (a)~Top view of a query scan~(blue) and its matched candidate scan~(orange) from \texttt{NTU 01-10}.
    (b)~Scan Context descriptors of the query~(left) and candidate retrieved from database~(right).
    (c)~SOLiD descriptors of the query~(left) and candidate retrieved from database~(right).
    }
    \label{fig:diff_scan_pattern}
    \vspace{-2mm}
\end{figure*}
% =============================================================
% =============================================================
\begin{figure*}[t]
    \centering    
    \captionsetup{justification=justified}
    \captionsetup[subfigure]{justification=centering}
    \begin{subfigure}{0.49\columnwidth}
        \includegraphics[trim={20 120 100 75},clip, width=\columnwidth]{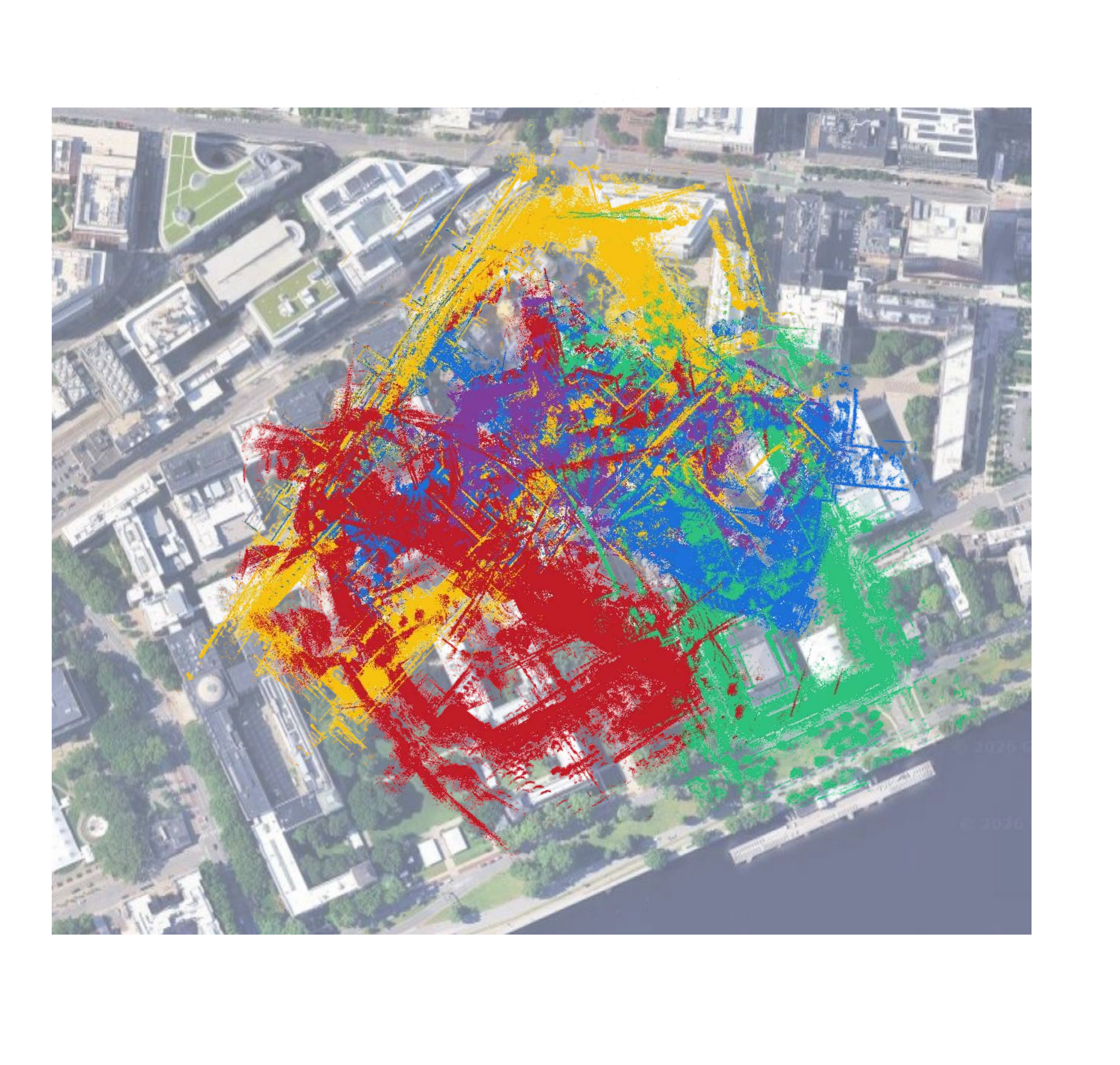}
        \vspace{-1.5em}
        \caption{LT-Mapper \\ \citep{ltmapper}}
    \end{subfigure}
    \begin{subfigure}{0.49\columnwidth}
        \includegraphics[trim={120 120 140 230},clip, width=\columnwidth]{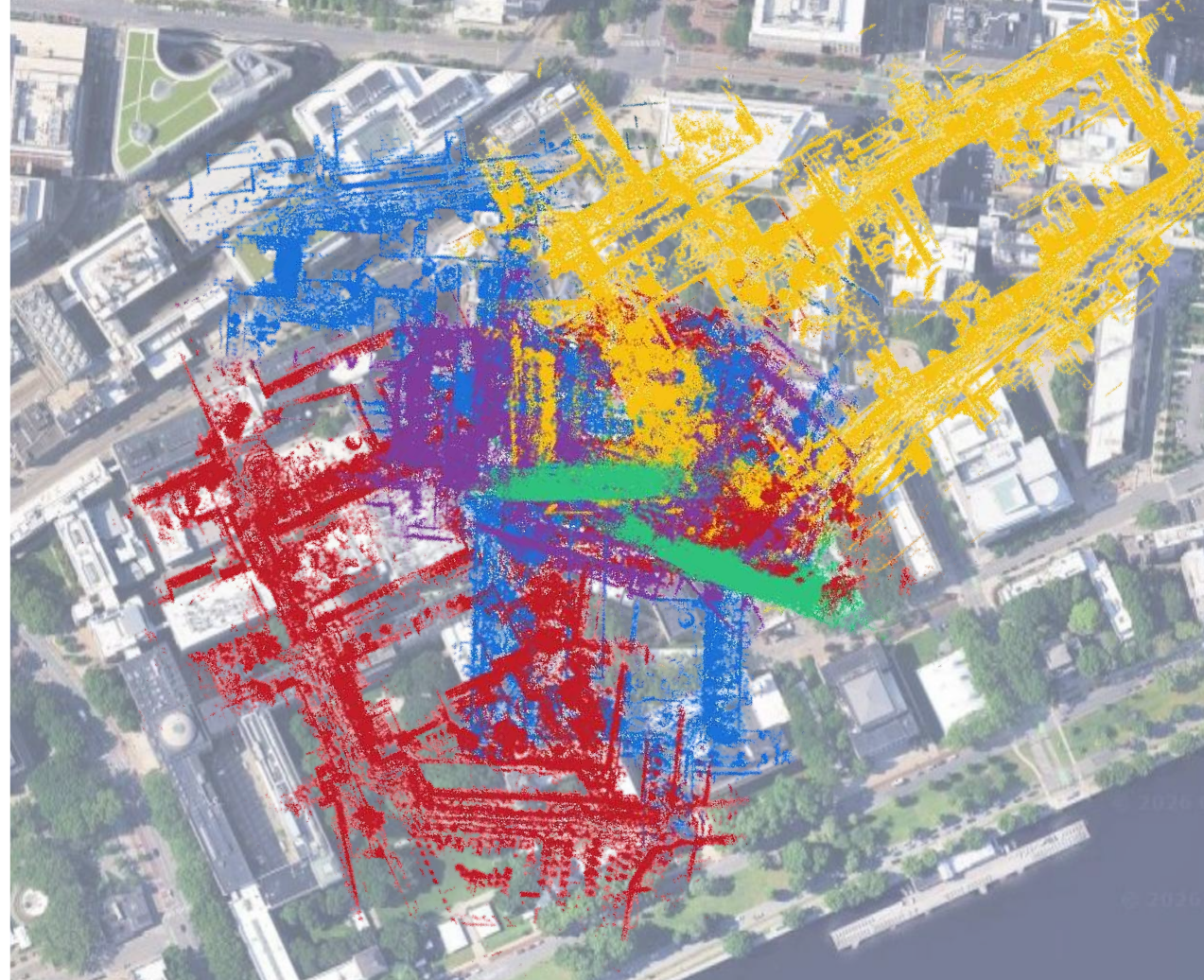}
        \vspace{-1.5em}
        \caption{ELite \\ \citep{gil2025ephemerality}}
    \end{subfigure}
    \begin{subfigure}{0.49\columnwidth}
        \includegraphics[trim={40 0 0 40},clip, width=\columnwidth]{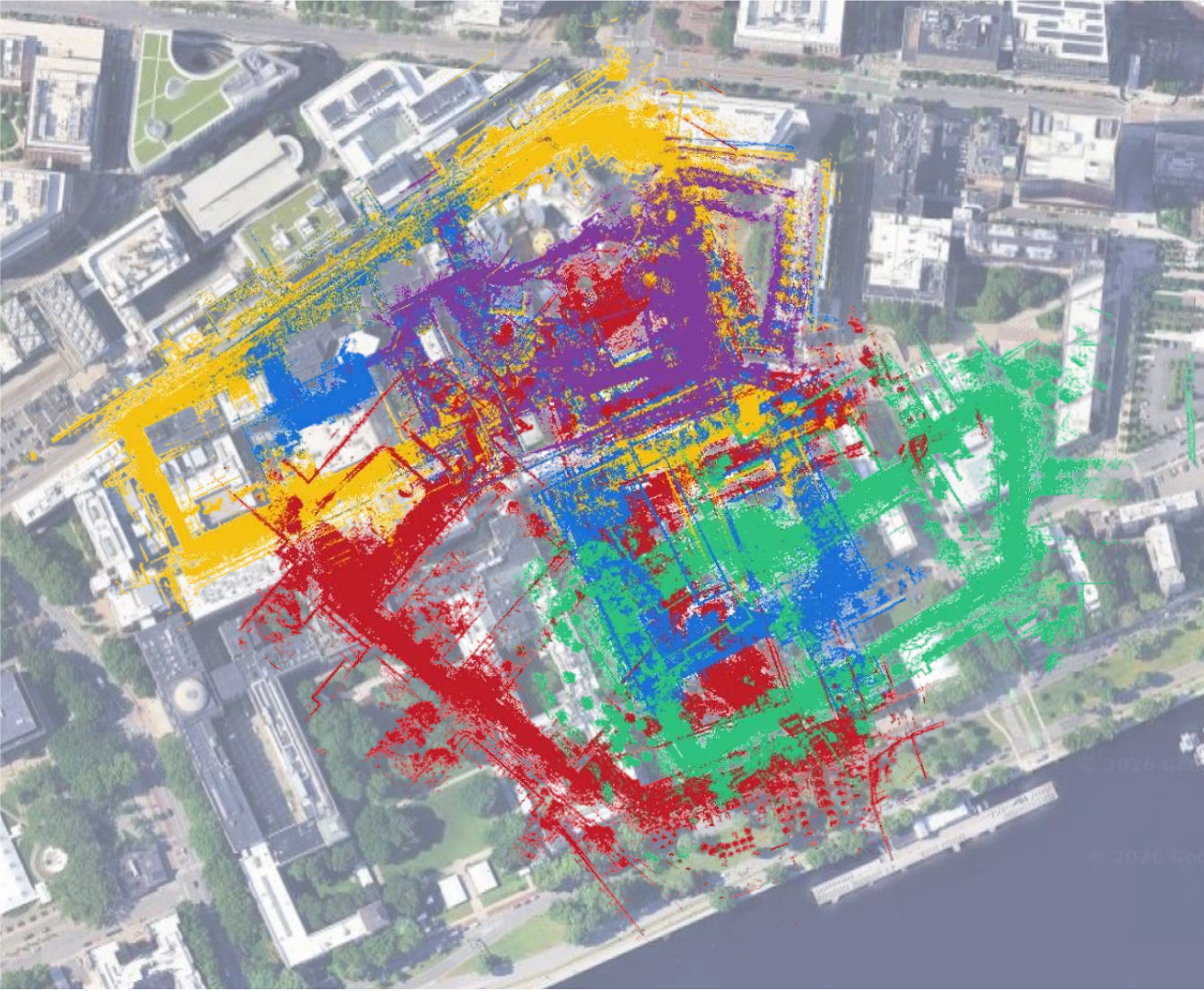}
        \vspace{-1.5em}
        \caption{Uni-Mapper \\ \citep{kang2025uni}}
    \end{subfigure}
    \begin{subfigure}{0.49\columnwidth}
        \includegraphics[trim={80 80 120 135},clip, width=\columnwidth]{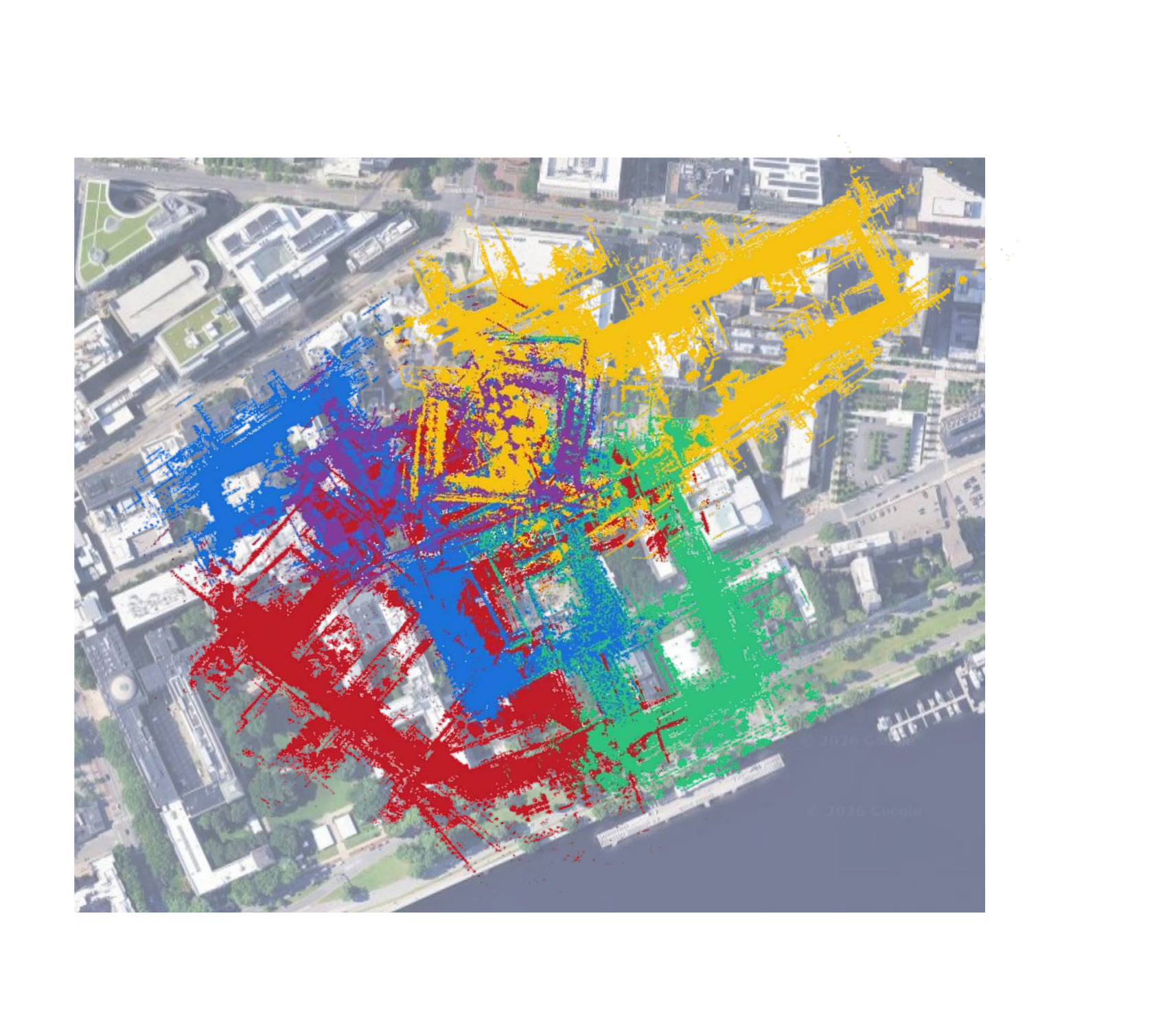}
        \vspace{-1.5em}
        \caption{LAMM \\ \citep{wei2024large}}
    \end{subfigure}
    
    \begin{subfigure}{0.98\columnwidth}
        \includegraphics[trim={240 240 0 240},clip, width=\columnwidth]{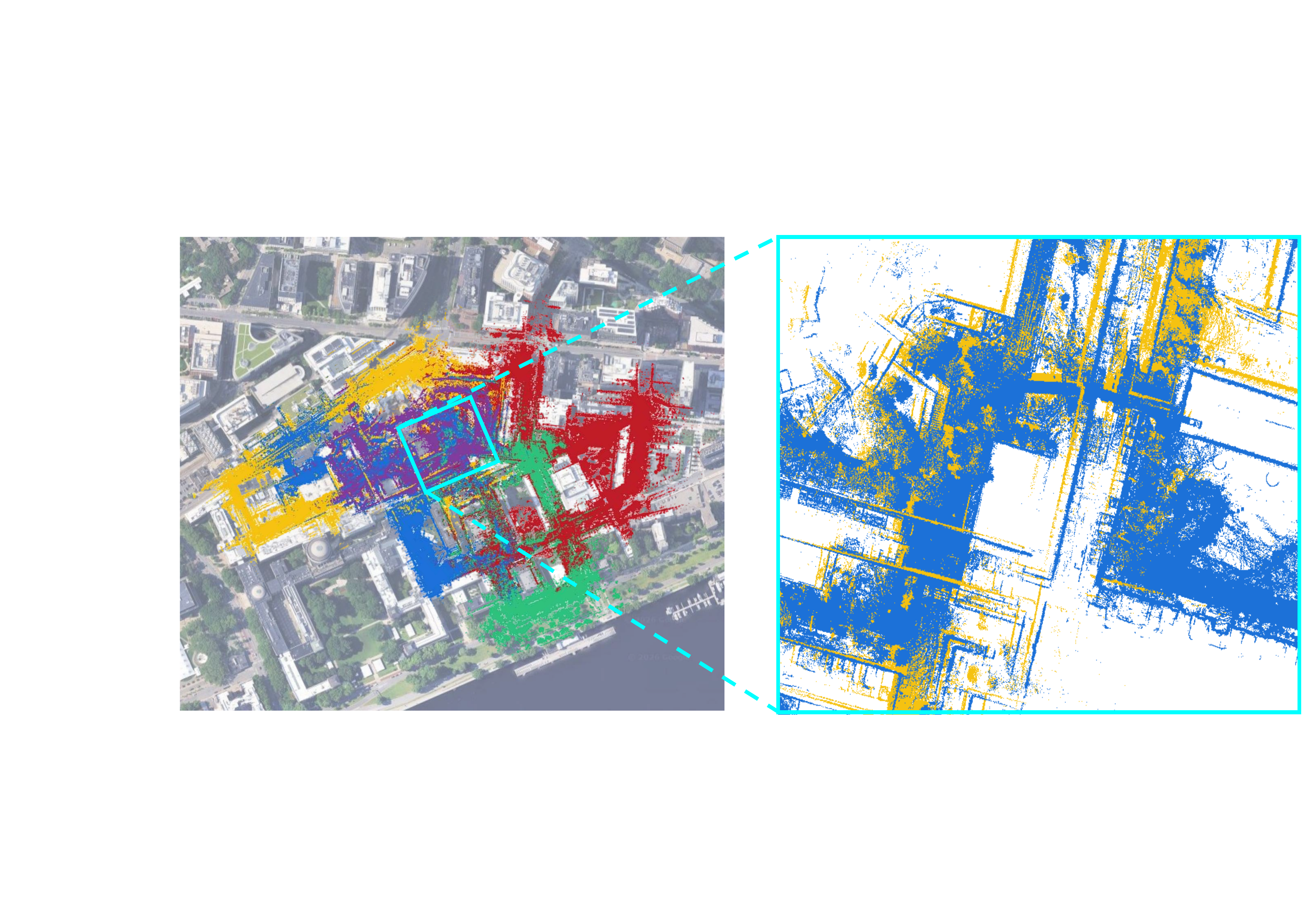}
        \vspace{-1.5em}
        \caption{KISS-Matcher~\citep{kiss_matcher}}
    \end{subfigure}
    \begin{subfigure}{1\columnwidth}
        \includegraphics[trim={240 240 0 240},clip, width=\columnwidth]{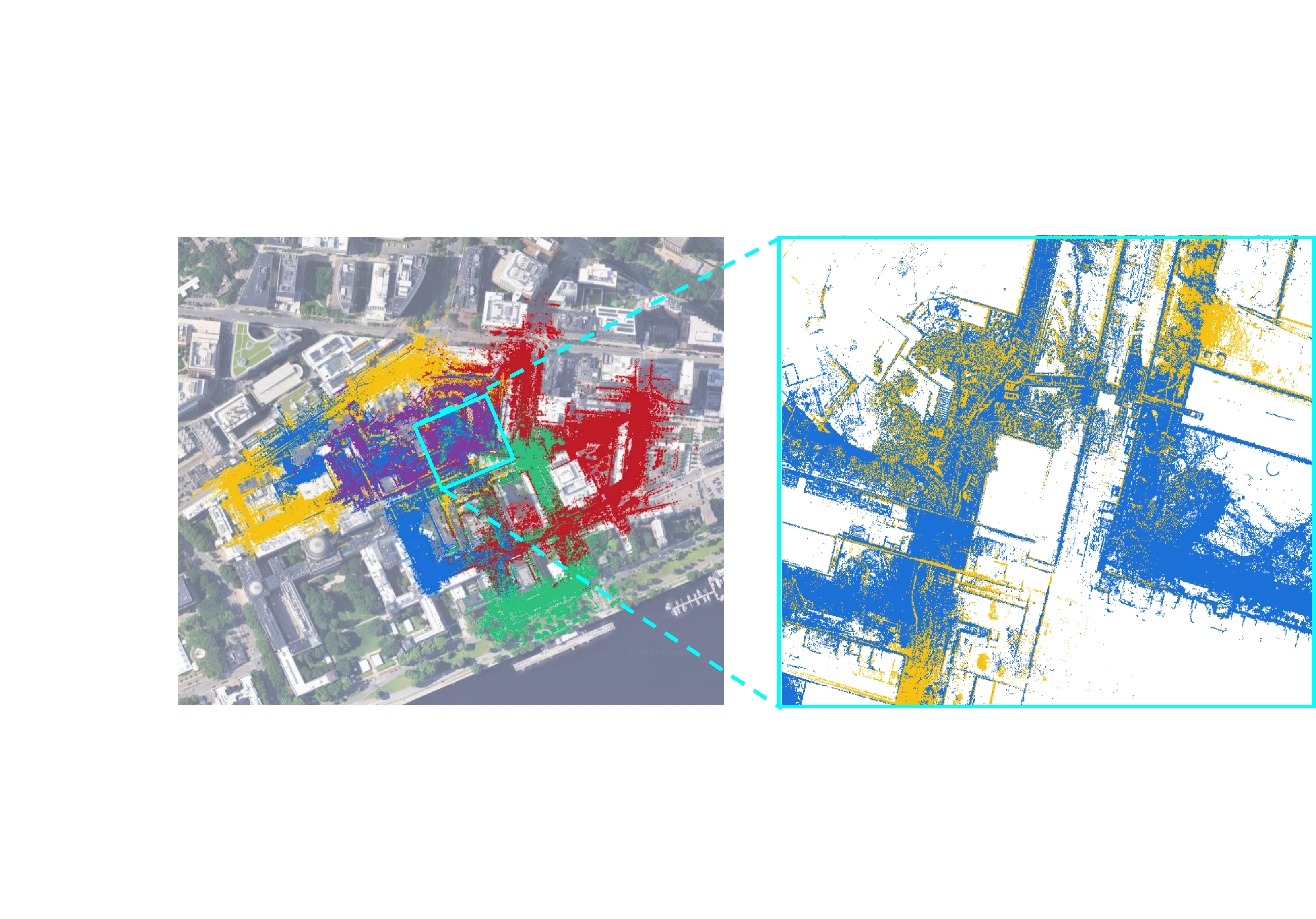}
        \vspace{-1.5em}
        \caption{Ours}
    \end{subfigure}

    \includegraphics[trim={0 0 0 0},clip, width=1.7\columnwidth]{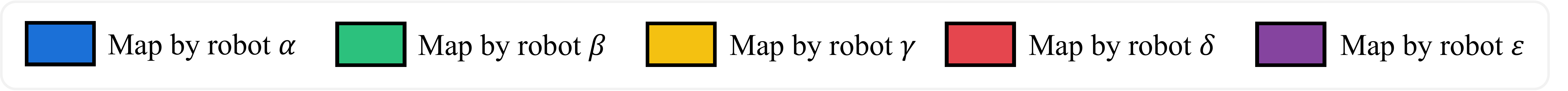}
  \caption{Multi-robot map merging results on the Kimera-Multi-\texttt{Outdoor} sequence acquired with an Ouster OS1-64 LiDAR sensor. 
  (a)--(d) Baseline methods exhibit substantial misalignment and structural inconsistencies.
  (e) KISS-Matcher and (f) our method yield tighter alignment in the overlap regions, with zoom-in views highlighting the resulting structural consistency.
  }
    \label{fig:kimera_outdoor}
    \vspace{-3mm}
\end{figure*}
% =============================================================
% =============================================================
\begin{figure*}[t]
    \centering
    \captionsetup[subfigure]{justification=centering}    
    \begin{subfigure}{0.66\columnwidth}
        \includegraphics[trim={0 0 0 0},clip, width=\columnwidth]{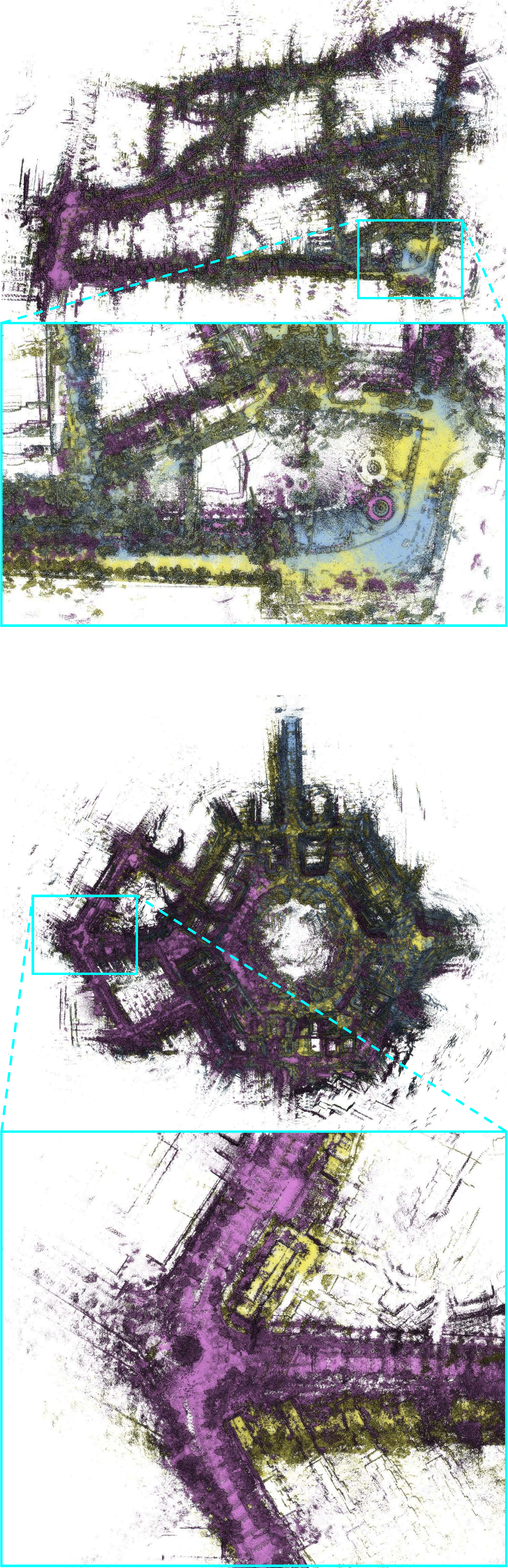}
        \vspace{-1.5em}
        \caption{LT-Mapper \\ \citep{ltmapper}}
    \end{subfigure}
    \begin{subfigure}{0.66\columnwidth}
        \includegraphics[trim={0 0 0 0},clip, width=\columnwidth]{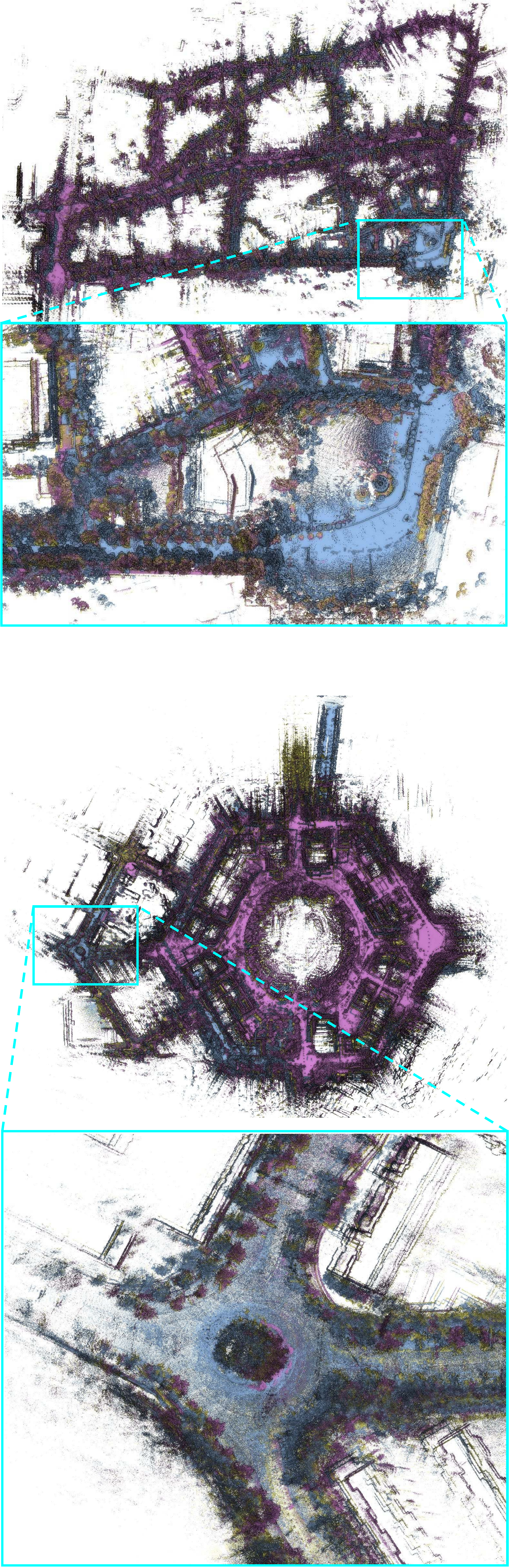}
        \vspace{-1.5em}
        \caption{KISS-Matcher \\ \citep{kiss_matcher}}
    \end{subfigure}
    \begin{subfigure}{0.66\columnwidth}
        \includegraphics[trim={0 0 0 0},clip, width=\columnwidth]{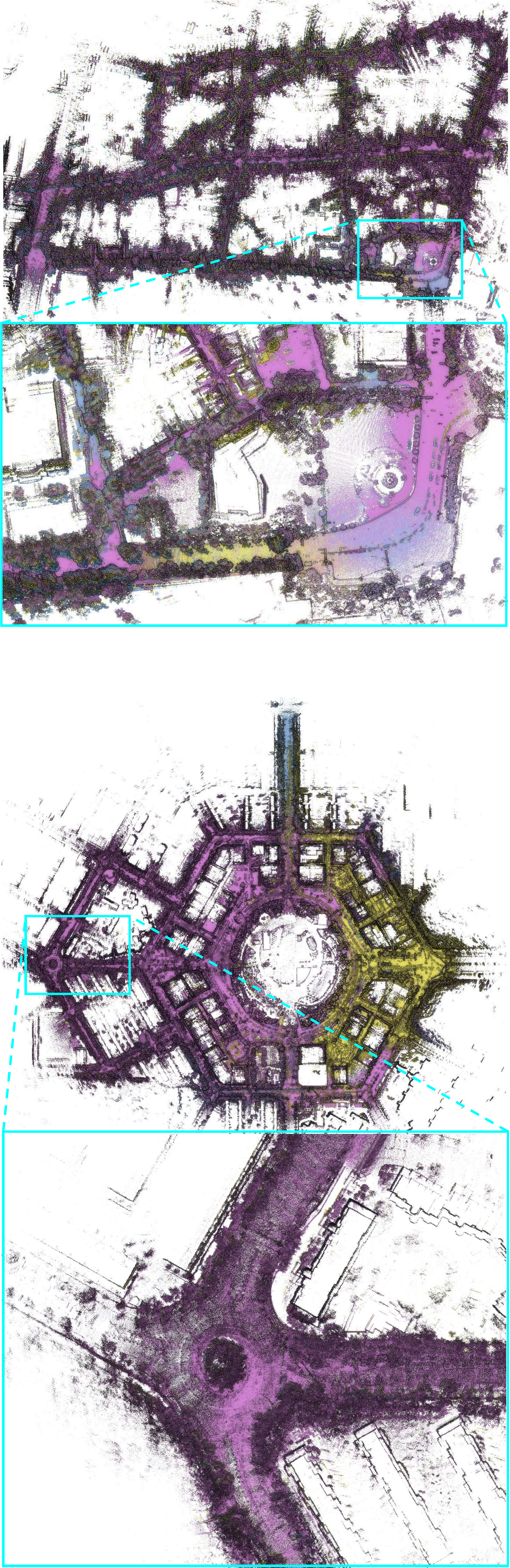}
        \vspace{-1.5em}
        \caption{Ours \\ ~}
    \end{subfigure}
    \includegraphics[trim={0 0 0 0},clip, width=1\columnwidth]{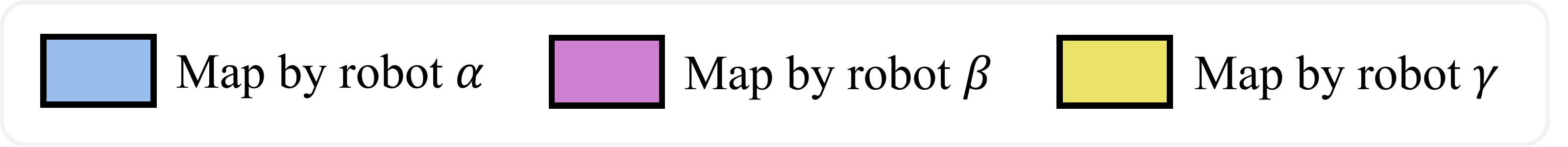}
    \vspace{-2mm}
      \caption{(a)--(c)~Multi-robot map merging performance comparison between baseline approaches and ours for three robots~($\alpha$, $\beta$, $\gamma$) on the \texttt{Town} (top) and \texttt{Roundabout} (bottom) sequences acquired with an Ouster OS2-128 LiDAR sensor. 
      As highlighted in zoom boxes, note that our method demonstrates high inter-robot alignment quality.}
    \label{fig:helipr}
    \vspace{-3mm}
\end{figure*}
% =============================================================

Here, the main differences between our proposed \textit{Commerge} and these methods are explained.
LT-Mapper, ELite, Uni-Mapper, and LAMM continuously transmit scan data at each communication step during the mission. 
KISS-Matcher transmits voxelized maps segment-by-segment when robots are in proximity, where each segment covers the trajectory between consecutive encounter points.
Note that KISS-Matcher assumes robots can detect proximity through signal strength indicators (\eg RSSI) or similar metrics.
In contrast, our method transmits only lightweight SOLiD descriptors at each communication step and sends carefully selected scans only after identifying consistent loop closures.

For fair comparison, all methods used FAST-LIO2~\citep{xu2022fast} as the odometry estimator with 1.0\,m keyframe spacing to obtain poses, and all scans corresponding to each pose were voxelized with a 1.0\,m voxel size.
Additionally, global descriptors were extracted from individual scans at each keyframe without point accumulation from adjacent frames.
Note that to improve data transmission efficiency, KISS-Matcher first accumulates the voxelized scans into a map based on estimated poses, then applies additional 1.0\,m voxelization to the accumulated map before transmission.

\subsection{Map Merging Performance Analysis with State-of-the-Art Methods}
\label{sec:perform_comparison}
\noindent
First, Table~\ref{tab:overall} reports $\rmsemetric$ across four datasets, where our method consistently ranks within the top three and demonstrates competitive or better performance than existing approaches.

Specifically, the leftmost columns show competitive performance on the \texttt{NTU} sequences acquired with non-repetitive scan pattern LiDAR~($\livox$), verifying robustness beyond conventional spinning LiDAR.
In contrast, given the raw scans shown in \figref{fig:diff_scan_pattern}(a), Scan Context~\citep{kim2018scan}, which serves as the core place recognition module of ELite and LT-Mapper, fails to produce consistent descriptors between the query and candidate under non-repetitive scan patterns, as shown in \figref{fig:diff_scan_pattern}(b).
In comparison, SOLiD~\citep{kim2024narrowing} adopted in our Commerge yields stable descriptors, as shown in \figref{fig:diff_scan_pattern}(c).
The middle columns, together with \figref{fig:kimera_outdoor}, present the Kimera-Multi \texttt{Outdoor} sequence, where our method attains the best average $\rmsemetric$ even under partial trajectory overlap~($\smalloverlap$), indicating that large errors in one robot pair are less likely to propagate to others.
The rightmost columns, together with \figref{fig:helipr}, highlight the \texttt{Roundabout} and \texttt{Town} sequences, where our method attains the lowest $\rmsemetric$ despite large-scale nature.

Therefore, we conclude that Commerge is a map merging framework robust to scan pattern, trajectory overlap, and operational scale.

\subsection{Robustness in Challenging Scenarios}
\label{sec:robustness}
\begin{figure*}[t]
    \centering
    \captionsetup{justification=justified}
    \captionsetup[subfigure]{justification=centering}
    \begin{subfigure}{0.79\columnwidth}
        \includegraphics[trim={0 0 0 0},clip, width=\columnwidth]{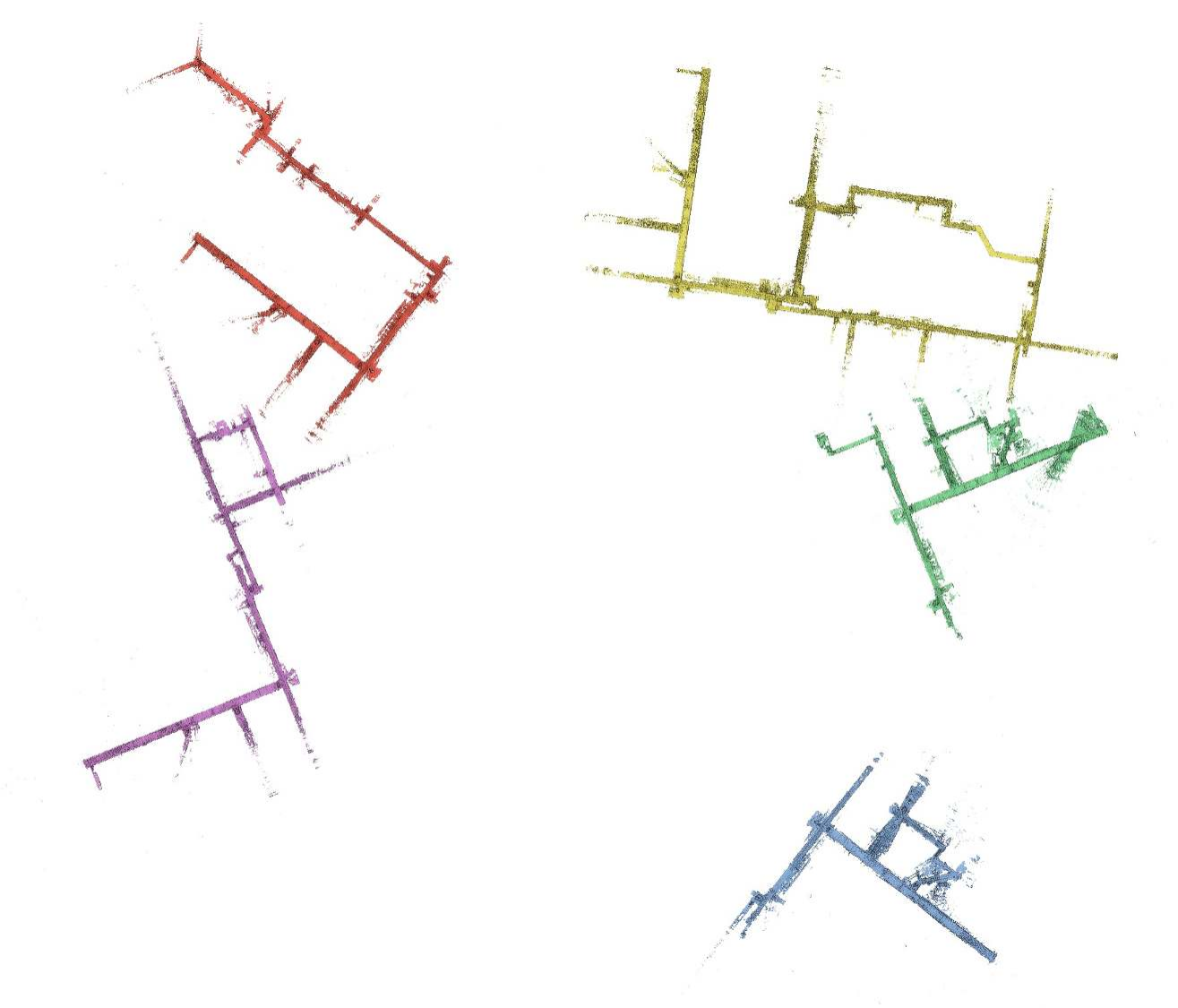}
        \vspace{-1.5em}
        \caption{Input}
    \end{subfigure}
    \begin{subfigure}{1.19\columnwidth}
        \includegraphics[trim={0 0 0 0},clip, width=\columnwidth]{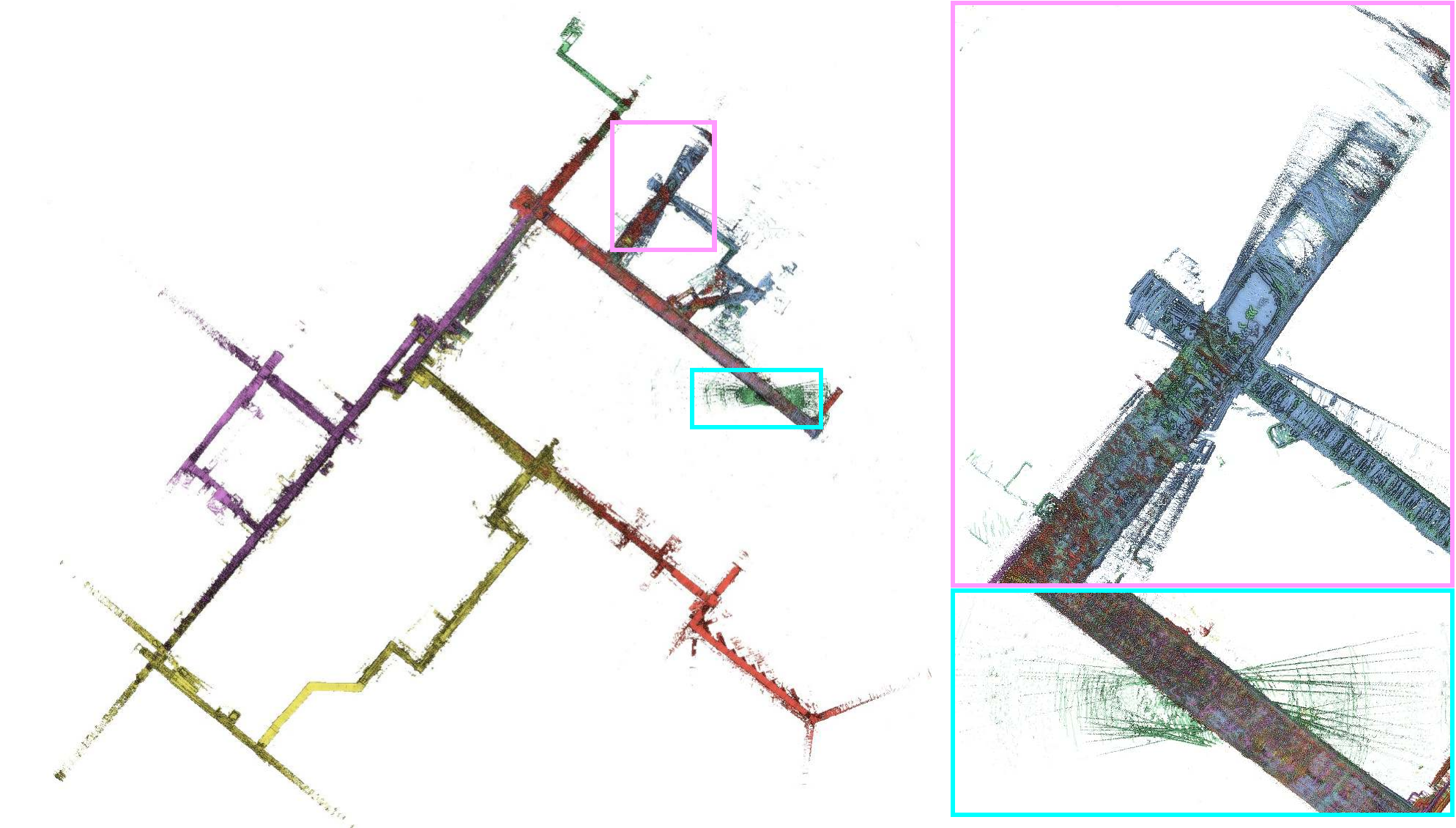}
        \vspace{-1.5em}
        \caption{Proposed}
    \end{subfigure}
    \includegraphics[trim={0 0 0 0},clip, width=1.7\columnwidth]{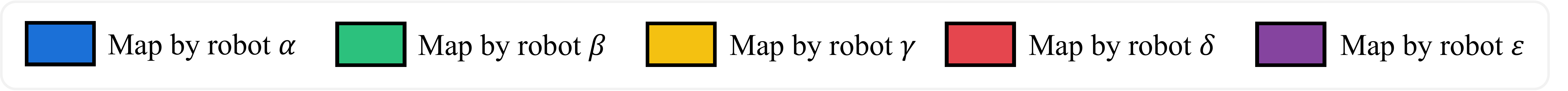}
    \caption{Multi-robot map merging in geometrically degenerate indoor environments using the Kimera-Multi-\texttt{Tunnel} dataset. 
               (a) Input: Two locally consistent robot maps built from ground truth trajectories with deliberately injected yaw bias to simulate misalignment. 
               (b) Output: Successfully merged map showing tight color co-registration between robots. 
               The cyan and magenta boxes highlight regions where even ground truth-based maps exhibit inherent degeneracy and misalignment due to indoor geometric ambiguity, yet our method successfully recovers valid inter-robot transformation and produces a coherent global map.}
     \label{fig:indoor}
     \vspace{-3mm}
\end{figure*}

\begin{figure*}[t]
    \centering
    \captionsetup{justification=justified}
    \captionsetup[subfigure]{justification=centering}
    \begin{subfigure}{0.5\columnwidth}
        \includegraphics[trim={0 0 0 0},clip, width=\columnwidth]{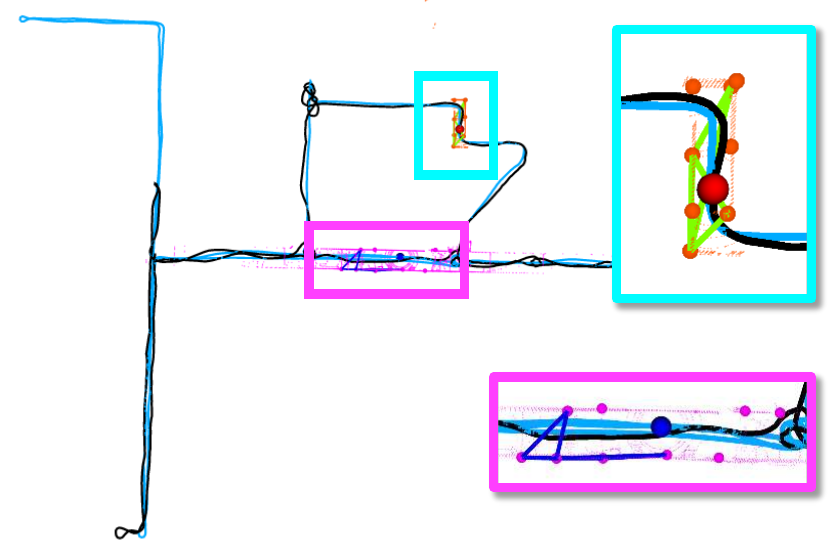}
        \vspace{-1.5em}
        \caption{}
    \end{subfigure}
    \begin{subfigure}{0.56\columnwidth}
        \includegraphics[trim={0 0 0 0},clip, width=\columnwidth]{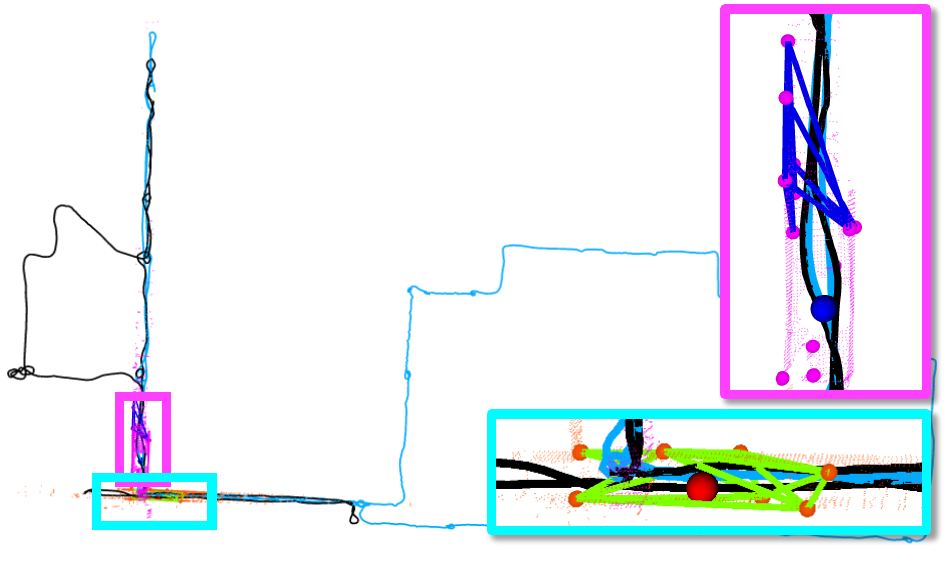}
        \vspace{-1.5em}
        \caption{}
    \end{subfigure}
    \begin{subfigure}{0.43\columnwidth}
        \includegraphics[trim={0 0 0 0},clip, width=\columnwidth]{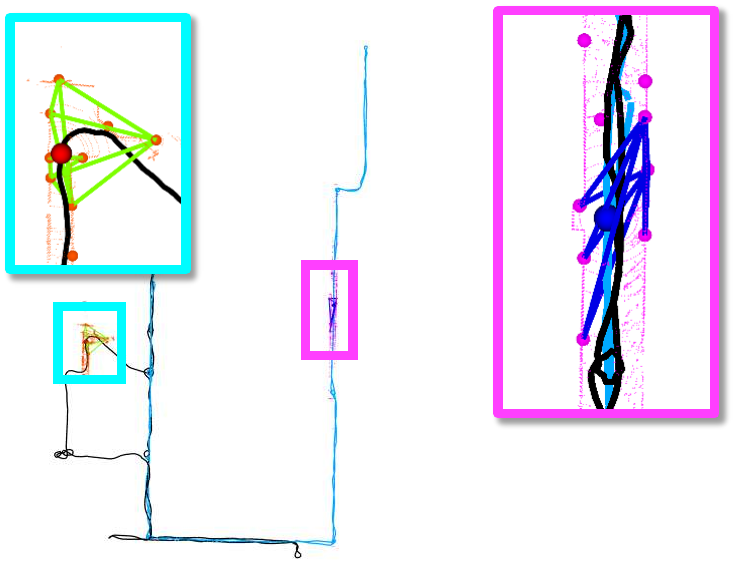}
        \vspace{-1.5em}
        \caption{}
    \end{subfigure}
    \begin{subfigure}{0.43\columnwidth}
        \includegraphics[trim={0 0 0 0},clip, width=\columnwidth]{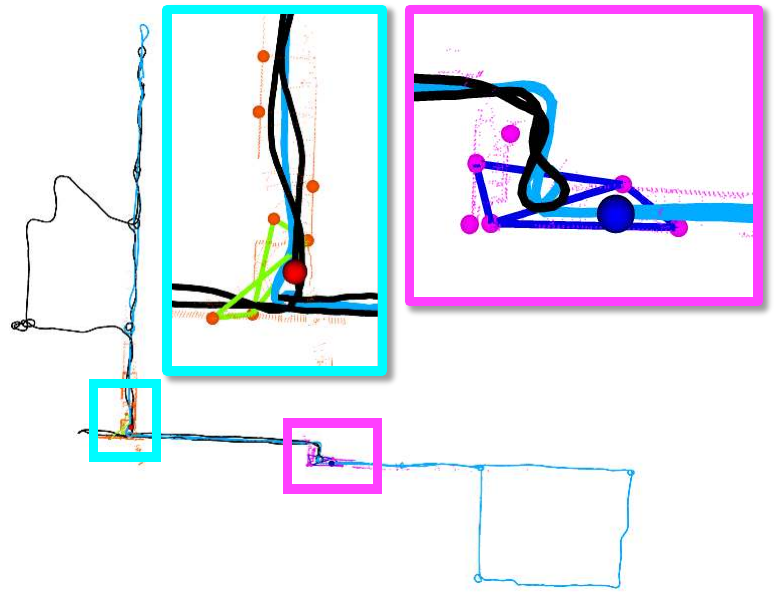}
        \vspace{-1.5em}
        \caption{}
    \end{subfigure}
    \caption{STD descriptor's false positive matches in repetitive indoor environments across four Kimera-Multi's \texttt{Tunnel} sequences: (a) \texttt{A01-A02}, (b) \texttt{A01-AP01}, (c) \texttt{A01-H01}, (d) \texttt{A01-S02}. 
             Black and blue trajectories represent the two robots' paths, respectively. 
             Cyan and magenta boxes highlight locations that STD incorrectly identifies as the same place despite being at different positions along the trajectories. 
             The zoomed views reveal structurally similar structures, which cause STD to generate nearly identical geometric descriptions for distinct locations. 
             These false matches explain STD's failure in repetitive indoor environments.}
    \label{fig:std_indoor}
    \vspace{-1mm}
\end{figure*}

\noindent
One notable aspect of our Commerge is sufficient robustness in challenging scenarios.
As shown in \figref{fig:indoor}, our method successfully merges robot maps in repetitive indoor structures that cause perceptual aliasing, even under partial trajectory overlap~($\smalloverlap$).
In contrast, STD~\citep{yuan2023std}, which serves as the core place recognition module of Uni-Mapper and LAMM, fails to produce reliable loop closures in such indoor environments, as shown in \figref{fig:std_indoor}.

% =============================================================
% =============================================================
% =============================================================
% =============================================================
% =============================================================
\begin{table*}[t]
\captionsetup{width=\textwidth, justification=justified}\caption{Scalability and resource efficiency comparison (\ie $\exchmetric$ and $\commmetric$) against state-of-the-art multi-robot map merging methods on the $\hardB$ platform across four representative datasets with varying scales and complexity (from small-scale \texttt{NTU} to large-scale \texttt{Town}, left to right).
Time out~(T/O) indicates communication time exceeding 1 hour, rendering the method impractical for real-time operation.
}
\renewcommand{\arraystretch}{1.2}
\centering\resizebox{\textwidth}{!}{\tiny
\begin{tabular}{l|l|cccccccccccccc}
\toprule
\midrule
                  & Sequence        & \multicolumn{3}{c}{\texttt{NTU}}
                                    & \multicolumn{5}{c}{\texttt{Outdoor}} 
                                    & \multicolumn{3}{c}{\texttt{Roundabout}}
                                    & \multicolumn{3}{c}{\texttt{Town}}   \\ \cmidrule(lr){3-5} \cmidrule(lr){6-10} \cmidrule(lr){11-13} \cmidrule(lr){14-16}
                  & Robot Info.     & \texttt{01-02}                & \texttt{01-10}           & Avg.
                                    & \texttt{A01-A02}              & \texttt{A01-H01}         & \texttt{A01-S01}         & \texttt{A01-S02}         & Avg.
                                    & \texttt{01-02}                & \texttt{01-03}           & Avg.
                                    & \texttt{01-02}                & \texttt{01-03}           & Avg.                                                               \\  \midrule
\multirow{3}{*}{\rotatebox[origin=c]{90}{$\exchmetric$}}
                  & Baselines        & \thirdc 252.4\,MB            & \thirdc 332.0\,MB            & \thirdc 292.2\,MB
                                     & \thirdc 328.4\,MB            & \thirdc 389.6\,MB            & \thirdc 424.0\,MB        & \thirdc 308.3\,MB        & \thirdc 362.7\,MB
                                     & \thirdc 13.9\,GB             & \thirdc 15.4\,GB             & \thirdc 14.7\,GB
                                     & \thirdc 14.3\,GB             & \thirdc 15.2\,GB             & \thirdc 14.8\,GB                     \\
                  & KISS-Matcher     & \secondc 5.5\,MB             & \secondc 7.5\,MB             & \secondc 6.5\,MB 
                                     & \secondc 4.5\,MB             & \secondc 4.8\,MB             & \secondc 5.0\,MB         & \secondc 5.6\,MB         & \secondc 5.0\,MB              
                                     & \secondc 67.1\,MB            & \secondc 77.6\,MB            & \secondc 72.4\,MB
                                     & \secondc 43.8\,MB            & \secondc 42.7\,MB            & \secondc 43.3\,MB                     \\
                  & Ours             & \firstc \textbf{2.9}\,MB 
                                     & \firstc \textbf{3.9}\,MB  
                                     & \firstc \textbf{3.4}\,MB  

                                     & \firstc \textbf{2.2}\,MB  
                                     & \firstc \textbf{2.4}\,MB  
                                     & \firstc \textbf{2.9}\,MB  
                                     & \firstc \textbf{2.8}\,MB  
                                     & \firstc \textbf{2.6}\,MB  

                                     & \firstc \textbf{19.0}\,MB 
                                     & \firstc \textbf{14.3}\,MB 
                                     & \firstc \textbf{16.8}\,MB 

                                     & \firstc \textbf{12.9}\,MB 
                                     & \firstc \textbf{14.4}\,MB 
                                     & \firstc \textbf{13.7}\,MB  \\ \midrule
\multirow{3}{*}{\rotatebox[origin=c]{90}{$\commmetric$}}
                  & Baselines        & \thirdc 5.284\,min         & \thirdc 7.891\,min       & \thirdc 6.588\,min 
                                     & \thirdc 24.172\,min        & \thirdc 27.112\,min      & \thirdc 31.908\,min      & \thirdc 22.441\,min      & \thirdc 26.408\,min 
                                     & T/O                        & T/O                      & T/O         
                                     & T/O                        & T/O                      & T/O                                     \\
                  & KISS-Matcher     & \secondc 7.738\,sec        & \secondc 11.490\,sec     & \secondc 9.682\,sec
                                     & \secondc 17.812\,sec       & \secondc 15.149\,sec     & \secondc 11.621\,sec     & \secondc 21.353\,sec     & \secondc 16.100\,sec                 
                                     & \secondc 79.390\,sec       & \secondc 82.114\,sec     & \secondc 81.875\,sec
                                     & \secondc 45.212\,sec       & \secondc 42.113\,sec     & \secondc 43.673\,sec                    \\
                  & Ours             & \firstc \textbf{3.449}\,sec   & \firstc \textbf{4.637}\,sec       & \firstc \textbf{4.042}\,sec
                                     & \firstc \textbf{2.616}\,sec   & \firstc \textbf{2.854}\,sec       & \firstc \textbf{3.449}\,sec
                                     & \firstc \textbf{3.330}\,sec   & \firstc \textbf{3.092}\,sec                 
                                     & \firstc \textbf{22.592}\,sec  & \firstc \textbf{17.003}\,sec      & \firstc \textbf{19.976}\,sec
                                     & \firstc \textbf{15.343}\,sec  & \firstc \textbf{17.122}\,sec      & \firstc \textbf{16.290}\,sec  \\ \midrule
\bottomrule
\end{tabular}}
\label{tab:efficiency}
\vspace{-4mm}
\end{table*}
% =============================================================
% =============================================================
\begin{figure*}[!t]
    \centering
    \captionsetup{justification=justified}
    \captionsetup[subfigure]{justification=centering}
    \begin{subfigure}{1.904\columnwidth}
        \includegraphics[trim={0 0 0 0},clip, width=\columnwidth]{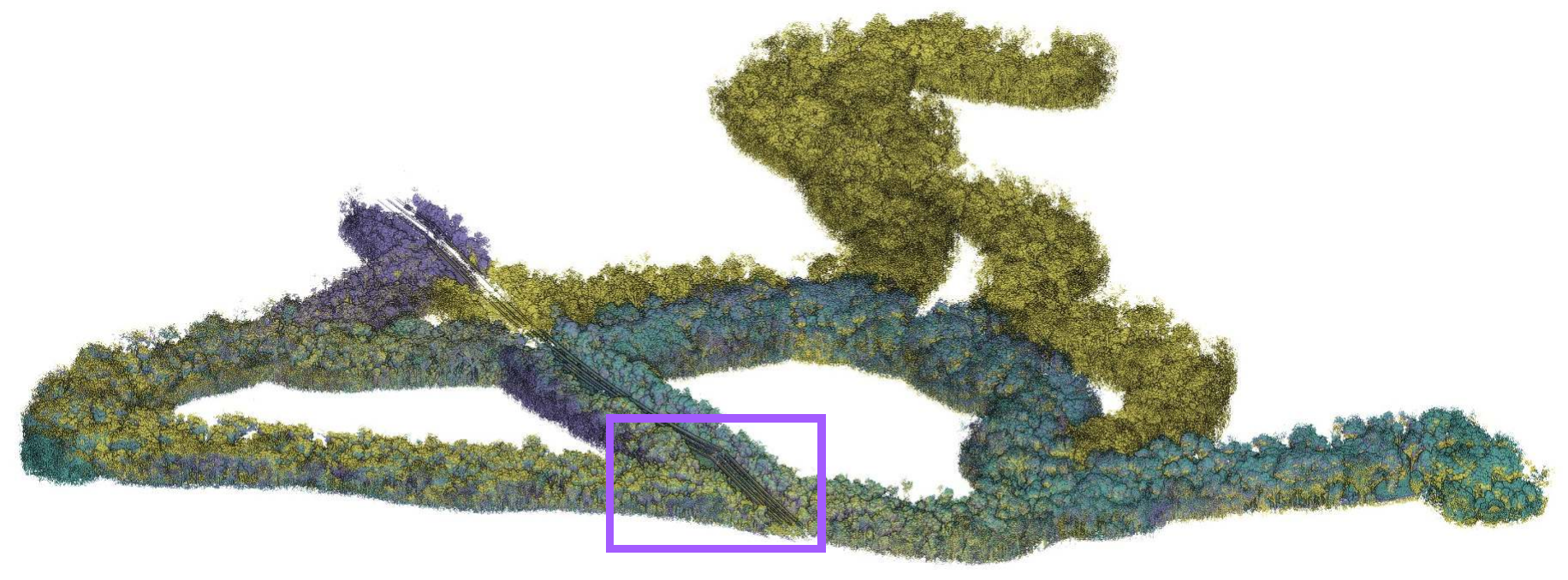}
        \vspace{-1.5em}
        \caption{}
        \vspace{-0.9cm}
    \end{subfigure}
    
    \begin{subfigure}{0.47\columnwidth}
        \includegraphics[trim={0 0 0 0},clip, width=\columnwidth]{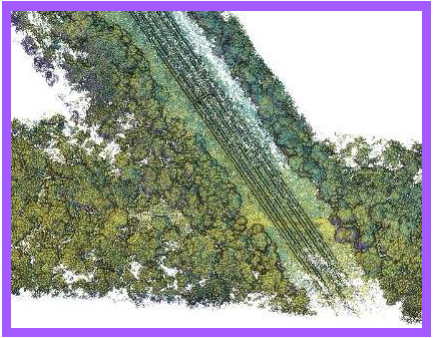}
        \vspace{-1.5em}
        \caption{}
    \end{subfigure}
    \begin{subfigure}{0.47\columnwidth}
        \includegraphics[trim={0 0 0 0},clip, width=\columnwidth]{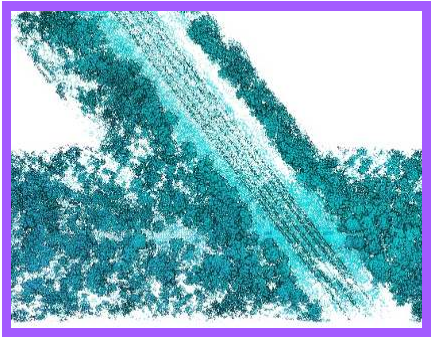}
        \vspace{-1.5em}
        \caption{}
    \end{subfigure}
    \begin{subfigure}{0.47\columnwidth}
        \includegraphics[trim={0 0 0 0},clip, width=\columnwidth]{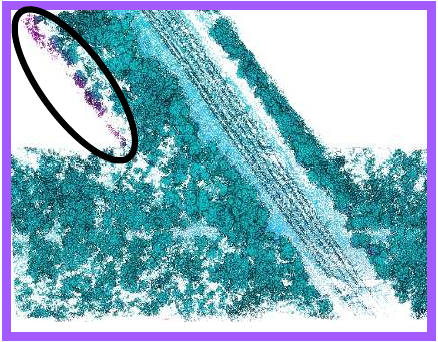}
        \vspace{-1.5em}
        \caption{}
    \end{subfigure}
    \begin{subfigure}{0.494\columnwidth}
        \includegraphics[trim={0 0 0 0},clip, width=\columnwidth]{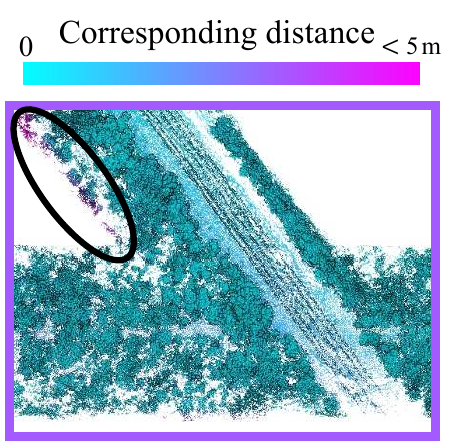}
        \vspace{-1.5em}
        \caption{}
    \end{subfigure}
    \includegraphics[trim={0 0 0 0},clip, width=1.05\columnwidth]{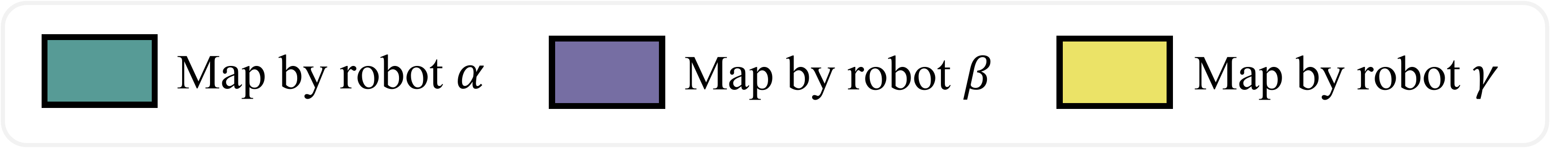}
    \vspace{-2mm}
    \caption{Visualization of map merging results on the WildPlaces dataset~\citep{knights2023wild}.
    (a)~Full merged map with improved color scheme.
    (b)~Zoomed-in view of the merged result.
    (c)--(e)~Point-to-point corresponding distances between robot pairs: \texttt{V02-V03}~(6 month gap), \texttt{V03-V04}~(8 month gap), and \texttt{V02-V04}~(14 month gap). 
    Elliptical regions in (d) and (e) highlight areas where \texttt{V04} exhibits larger corresponding distances due to environmental changes over longer temporal gaps.
    }
    \label{fig:wildplaces}
    \vspace{-2mm}
\end{figure*}
% =============================================================
% =============================================================
\begin{figure*}[!htp]
    \centering
    \captionsetup{justification=justified}
    \captionsetup[subfigure]{justification=centering}

    \begin{subfigure}{0.65\columnwidth}
        \includegraphics[trim={0 0 0 0},clip, width=\columnwidth]{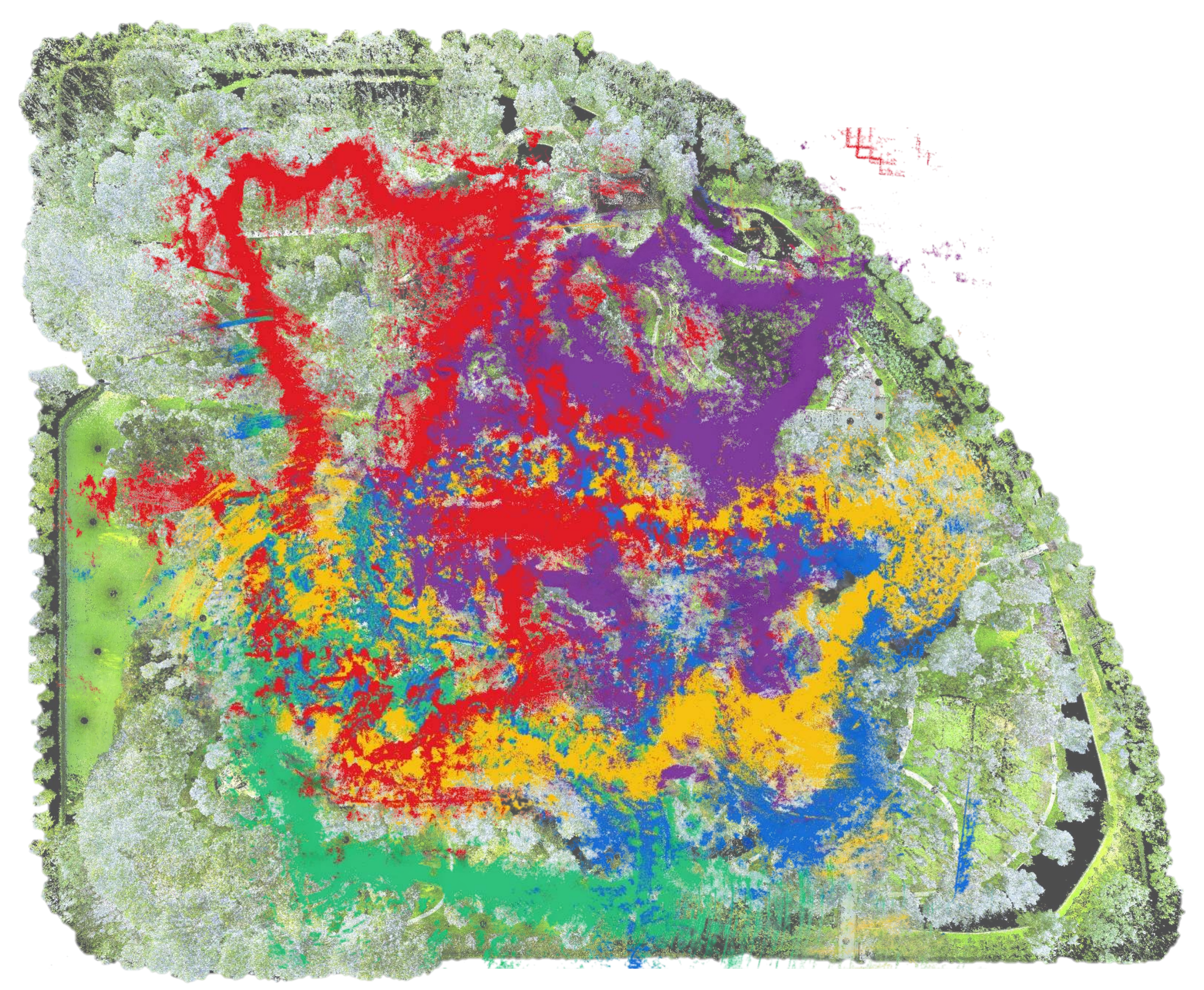}
        \vspace{-1.5em}
        \caption{LT-Mapper~\citep{ltmapper}}
    \end{subfigure}
    \begin{subfigure}{0.65\columnwidth}
        \includegraphics[trim={0 0 0 0},clip, width=\columnwidth]{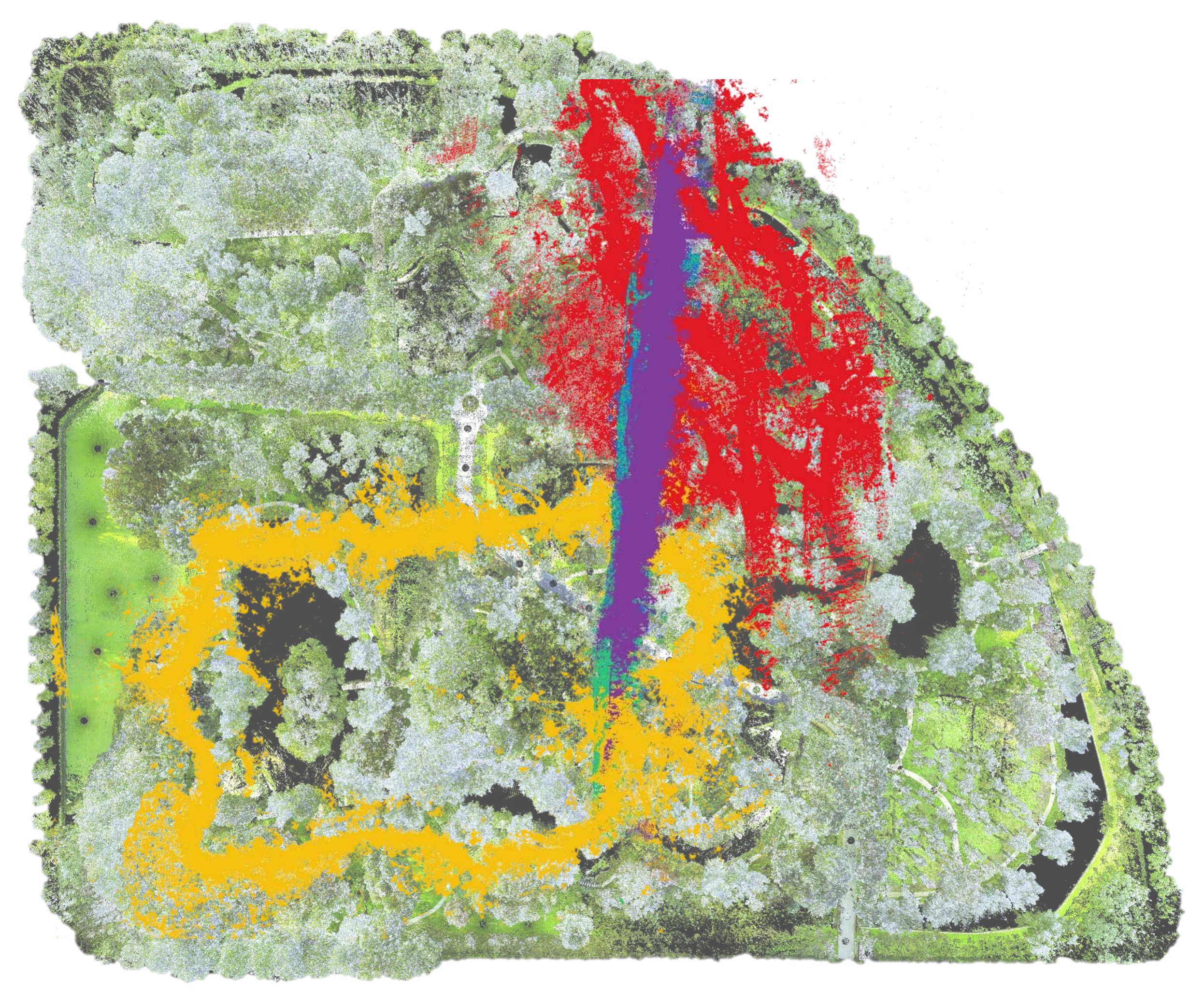}
        \vspace{-1.5em}
        \caption{ELite~\citep{gil2025ephemerality}}
    \end{subfigure}
    \begin{subfigure}{0.65\columnwidth}
        \includegraphics[trim={0 0 0 0},clip, width=\columnwidth]{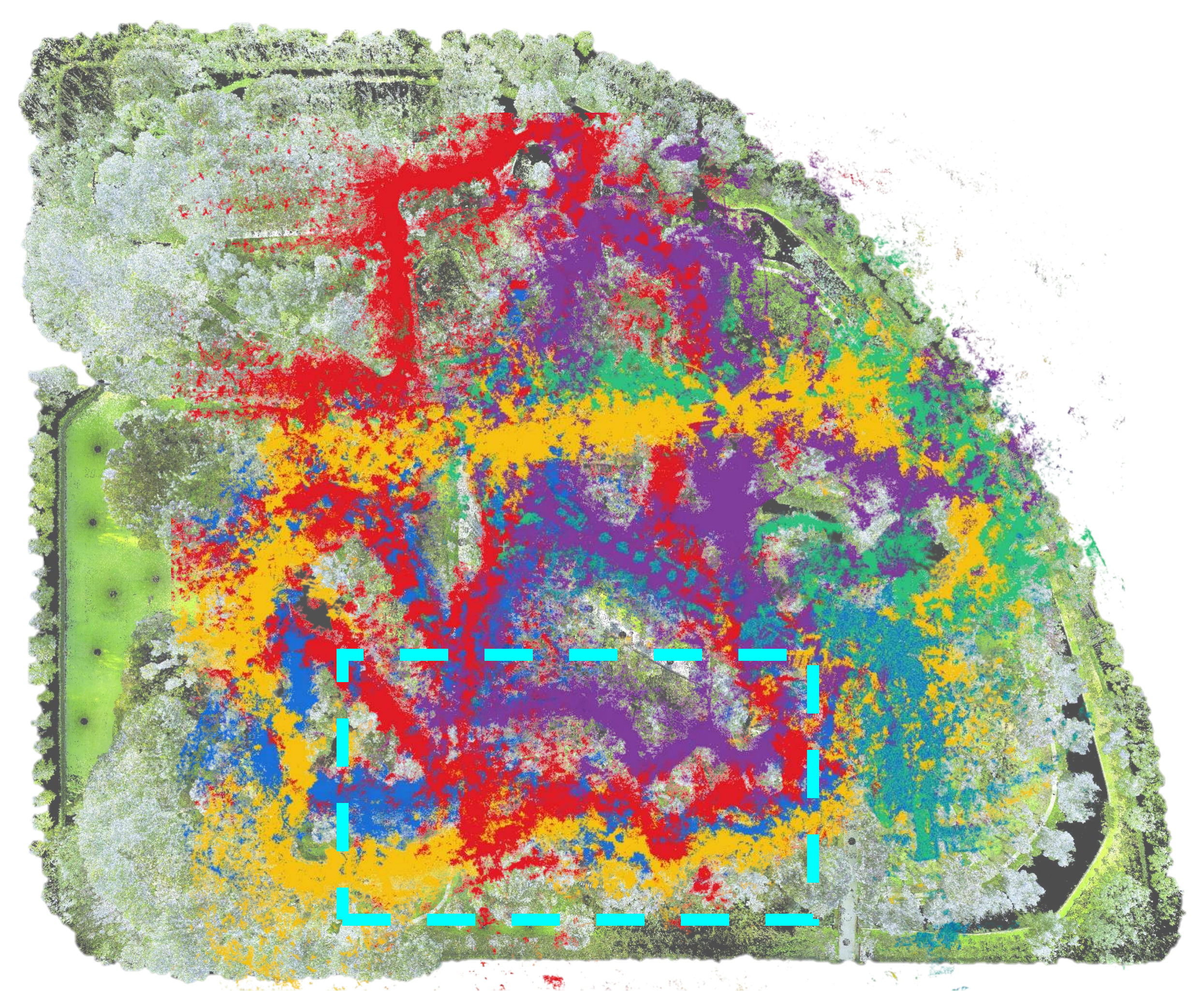}
        \vspace{-1.5em}
        \caption{Uni-Mapper~\citep{kang2025uni}}
    \end{subfigure}
    
    \begin{subfigure}{0.65\columnwidth}
        \includegraphics[trim={0 0 0 0},clip, width=\columnwidth]{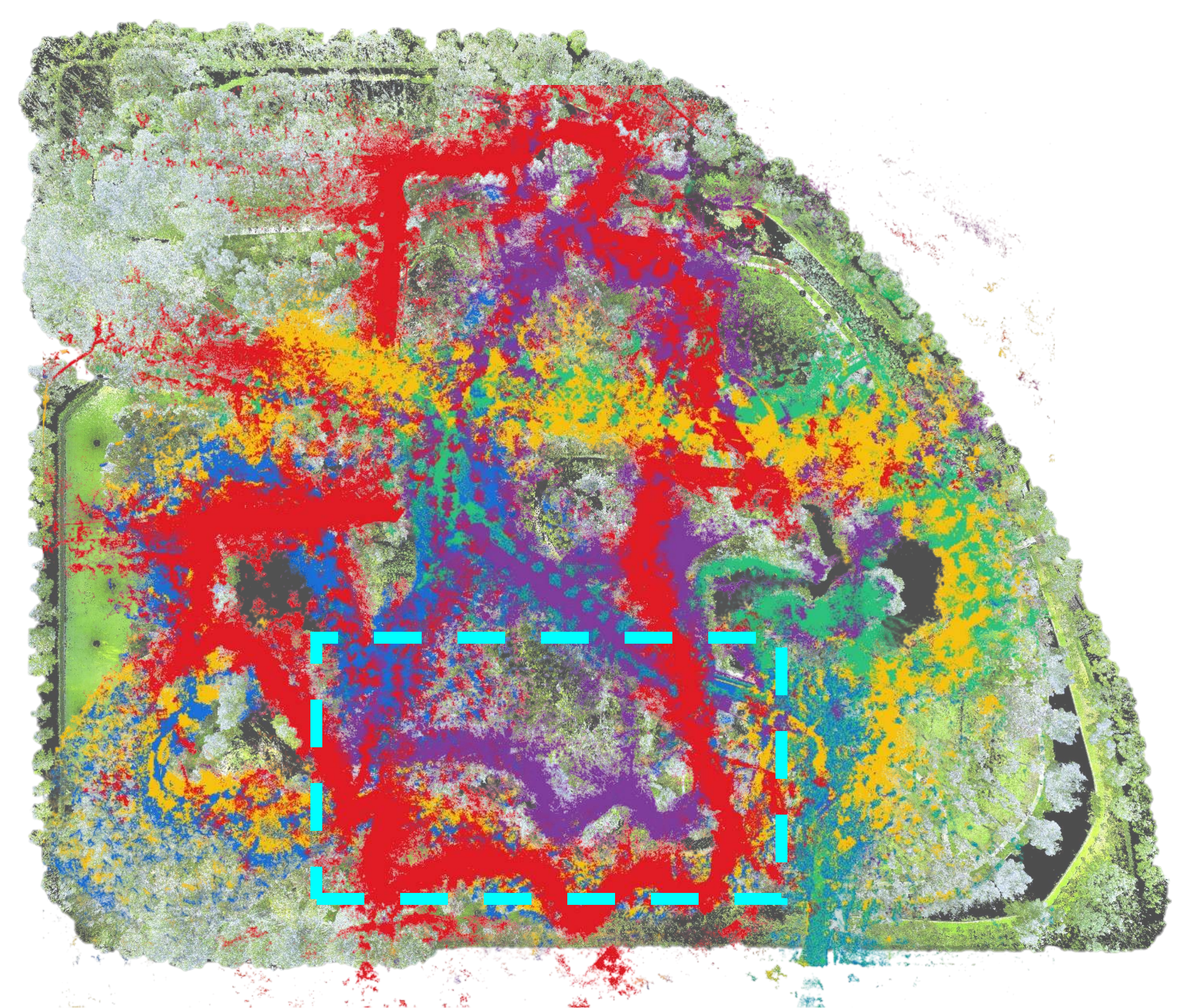}
        \vspace{-1.5em}
        \caption{LAMM~\citep{wei2024large}}
    \end{subfigure}
    \begin{subfigure}{0.65\columnwidth}
        \includegraphics[trim={0 0 0 0},clip, width=\columnwidth]{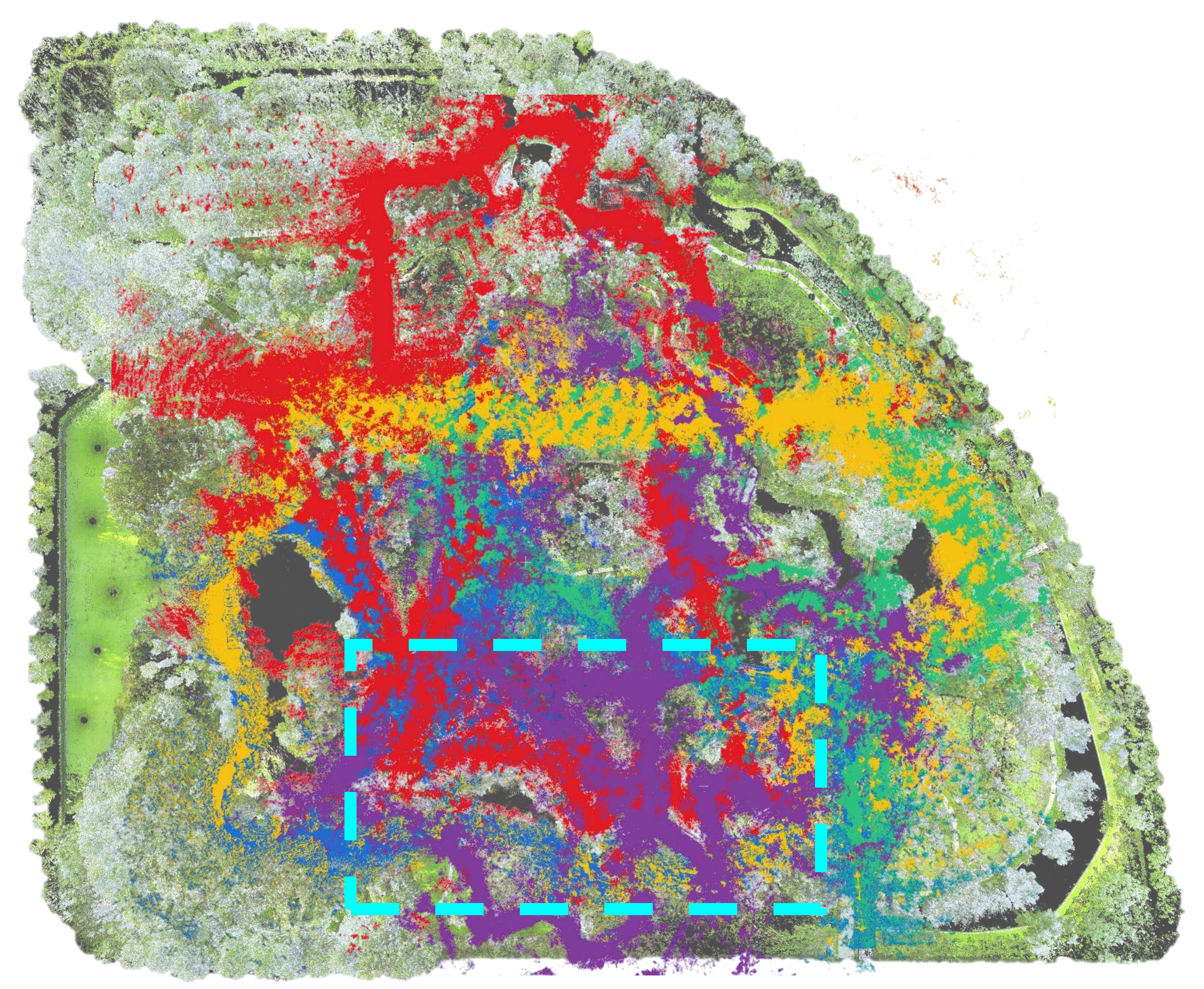}
        \vspace{-1.5em}
        \caption{KISS-Matcher~\citep{kiss_matcher}}
    \end{subfigure}
    \begin{subfigure}{0.65\columnwidth}
        \vspace{-2mm}
        \includegraphics[trim={0 0 0 0},clip, width=\columnwidth]{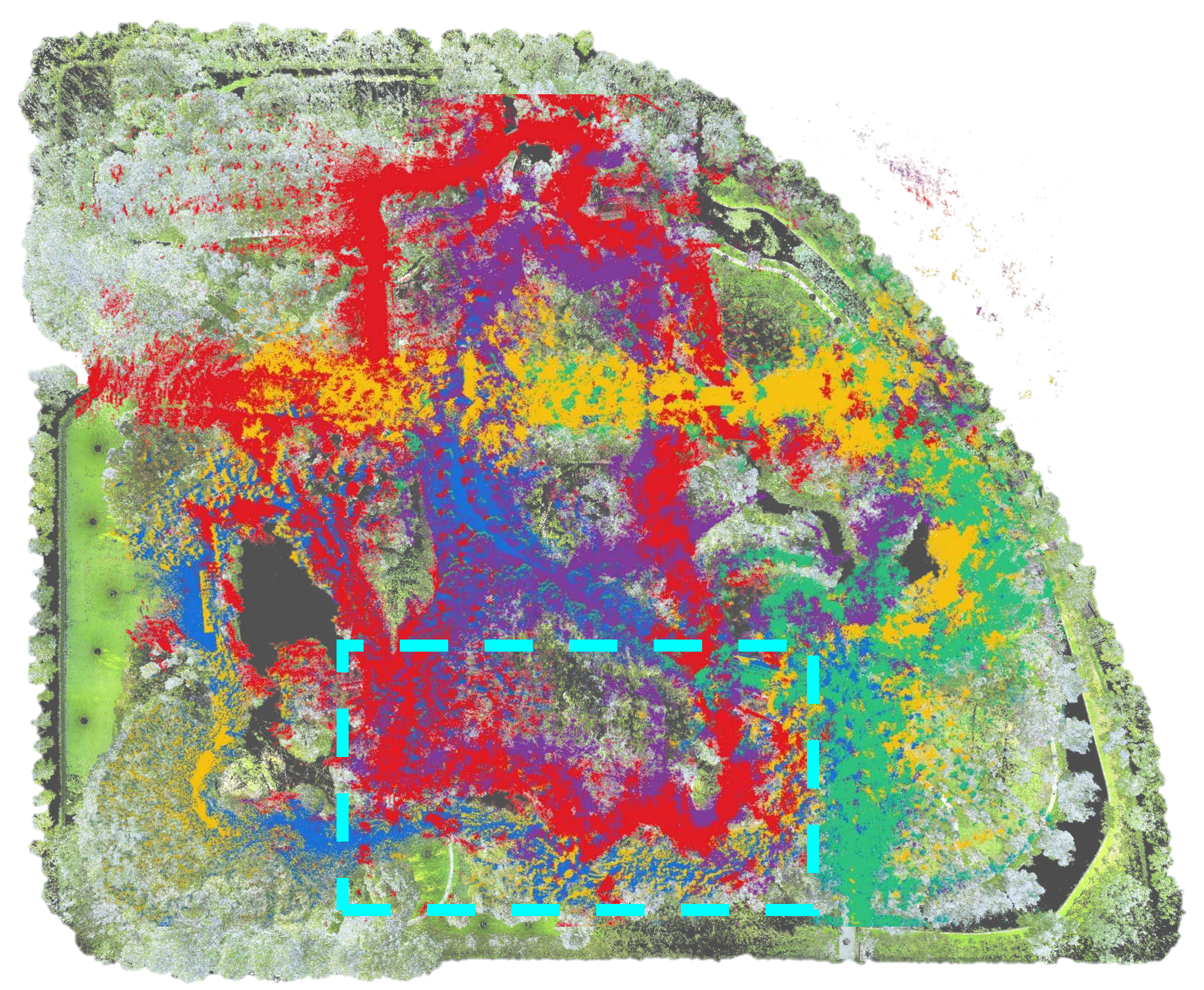}
        \vspace{-1.5em}
        \caption{Ours}
    \end{subfigure}

    \includegraphics[trim={0 0 0 0},clip, width=1.7\columnwidth]{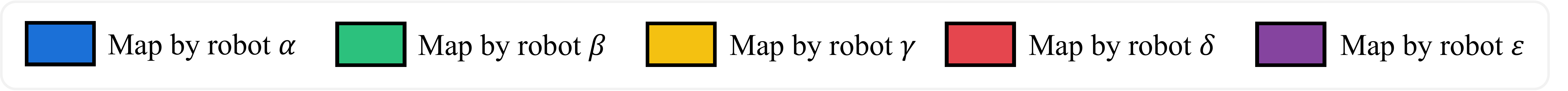}
  \caption{Multi-robot map merging results on the Botanic Garden dataset acquired with a Livox Avia LiDAR sensor.
           (a--f) Mapping results from different methods, where colored point clouds represent different robots.
           The red and purple boxes (cyan boxes) highlight critical overlap regions that should align if inter-robot alignment is successful.}
  \label{fig:botanic}
\end{figure*}
% =============================================================

Furthermore, as shown in Figs.~\ref{fig:wildplaces} and \ref{fig:botanic}, our method successfully performs map merging in unstructured vegetated environments such as forests, where dense foliage and repetitive natural textures degrade geometric distinctiveness.
Specifically, \figref{fig:wildplaces} demonstrates robustness on the large-scale WildPlaces dataset, where LiDAR and LiDAR-inertial odometry methods typically fail due to the lack of geometric structure, while \figref{fig:botanic} validates robustness under partial trajectory overlap~($\smalloverlap$) with non-repetitive scan pattern LiDAR~($\livox$).

Notably, these results confirm that Commerge reliably operates across diverse conditions where LiDAR-based perception typically struggles. These conditions include perceptually aliased indoor structures, unstructured outdoor environments, large operational scale, partial trajectory overlap, and non-repetitive scan patterns.

\subsection{Scalability and Transmission Efficiency Analysis}
\label{sec:scalability}
\begin{figure*}[t]
    \centering
    \captionsetup{justification=justified}
    \captionsetup[subfigure]{justification=centering}
    \begin{subfigure}{0.5\columnwidth}
        \includegraphics[trim={0 0 0 0},clip, width=\columnwidth]{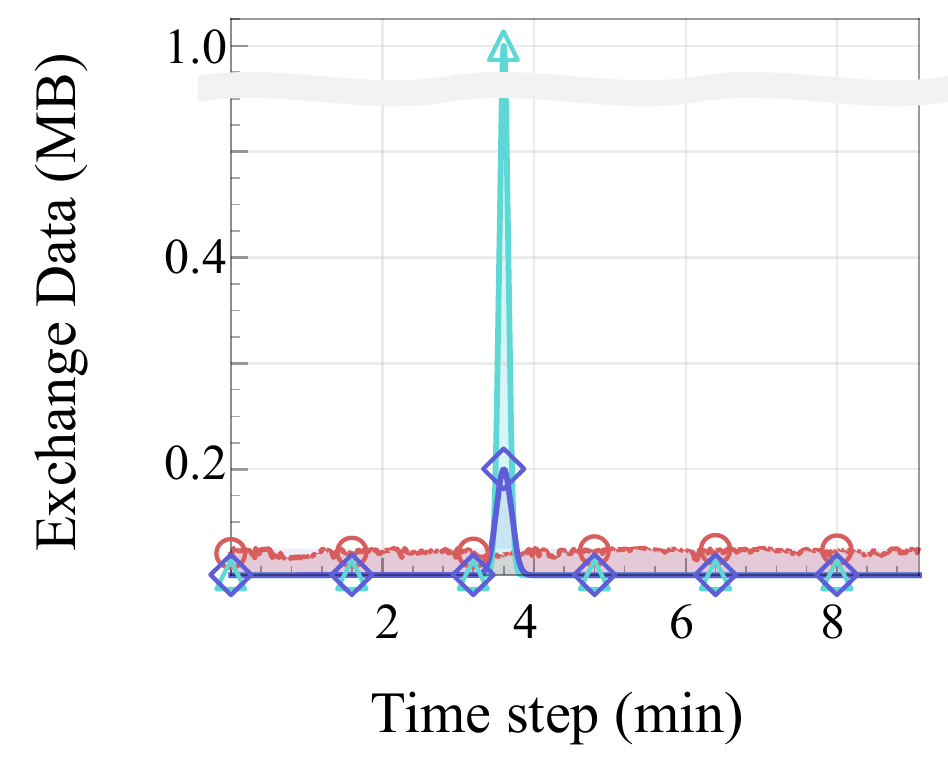}
        \vspace{-1.5em}
        \caption{}
    \end{subfigure}
    \begin{subfigure}{0.5\columnwidth}
        \includegraphics[trim={0 0 0 0},clip, width=\columnwidth]{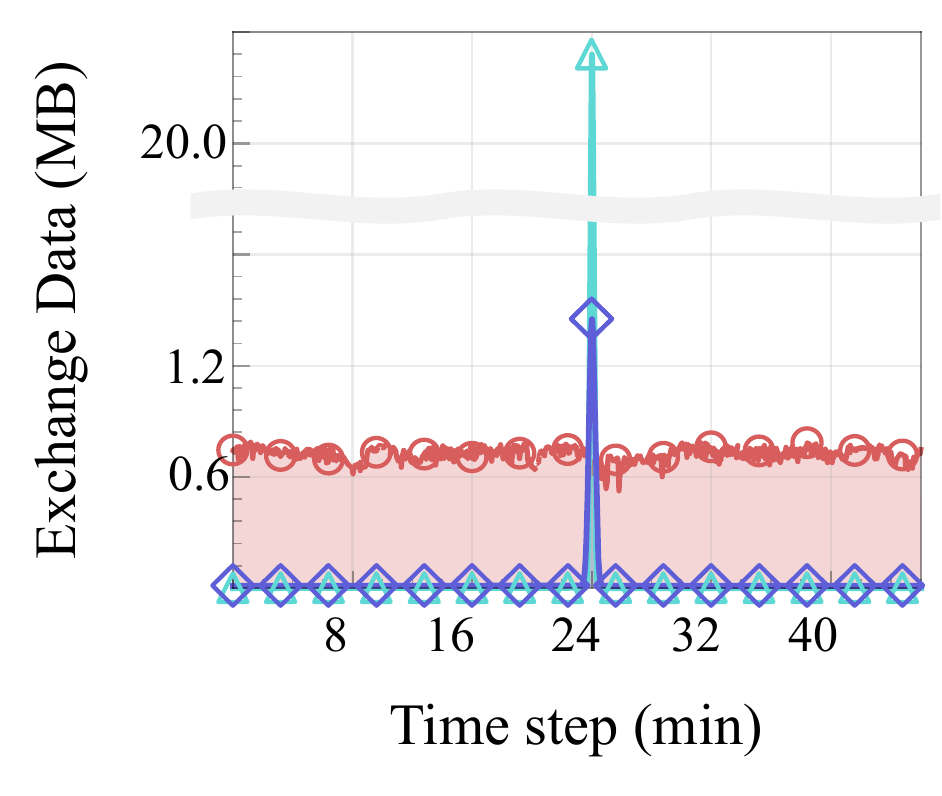}
        \vspace{-1.5em}
        \caption{}
    \end{subfigure}
    \begin{subfigure}{0.97\columnwidth}
        \includegraphics[trim={0 0 0 0},clip, width=\columnwidth]{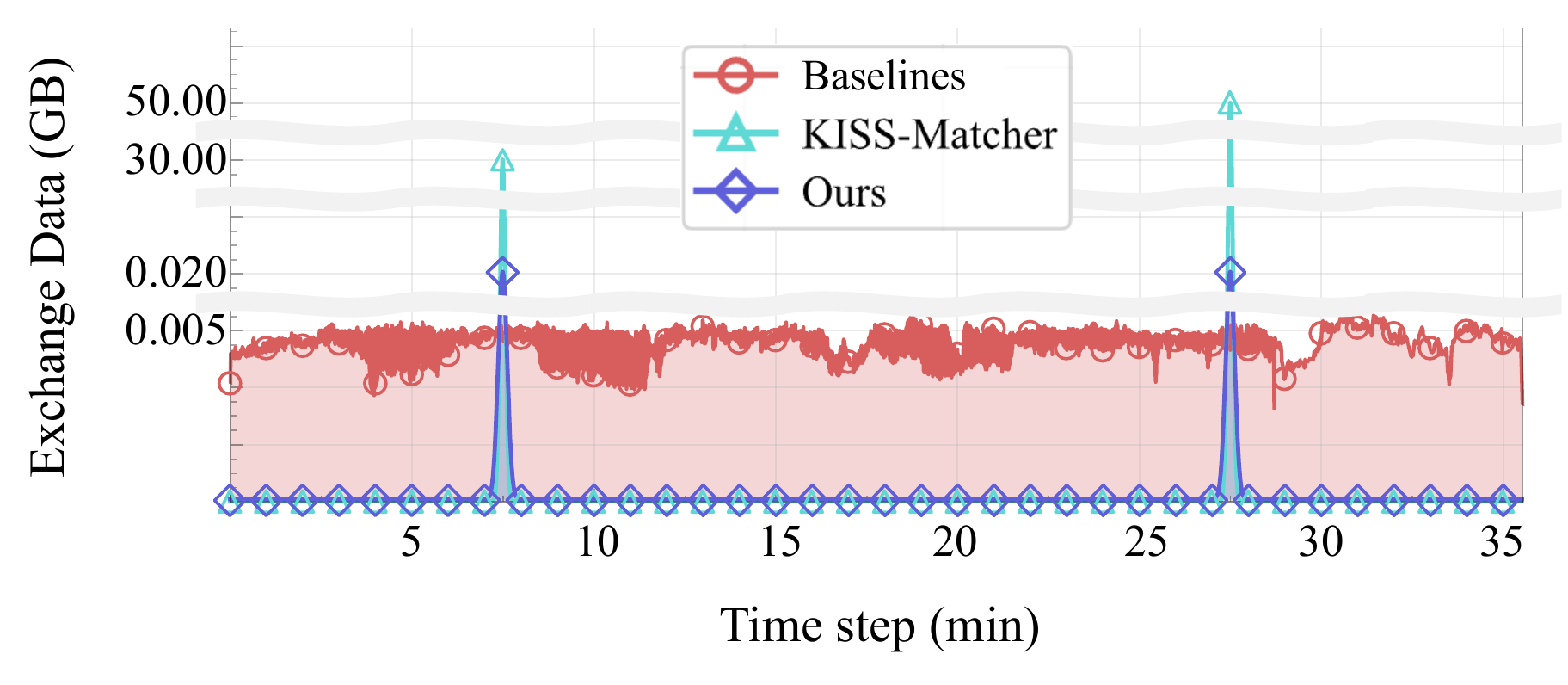}
        \vspace{-1.5em}
        \caption{}
    \end{subfigure}

    \caption{Exchange data over time for each merging round on the (a)~\texttt{NTU 01-10}~\citep{nguyen2024mcd}, (b)~\texttt{Town 01-03}~\citep{jung2024helipr}, and (c)~\texttt{V02-V03-V04}~\citep{knights2023wild} sequences. 
            }
    \label{fig:mem_time}
    \vspace{-3mm}
\end{figure*}

\noindent
Next, the top section of Table~\ref{tab:efficiency} and \figref{fig:mem_time} show that our Commerge transmits substantially less data overall compared to the baselines, and also reduces the per-round burst volume compared to KISS-Matcher, which transmits all matched scans at once upon loop detection.
Notably, this advantage becomes more pronounced as operational scale increases from \figref{fig:mem_time}(a) to \figref{fig:mem_time}(c), demonstrating that our selective data exchange strategy scales favorably with growing map size and team size.
% =============================================================
% =============================================================
% % =============================================================
\begin{figure}[t]
    \centering
    \captionsetup{justification=justified}
    \captionsetup[subfigure]{justification=centering}
    \begin{subfigure}{0.52\columnwidth}
        \includegraphics[trim={0 0 0 0},clip, width=\columnwidth]{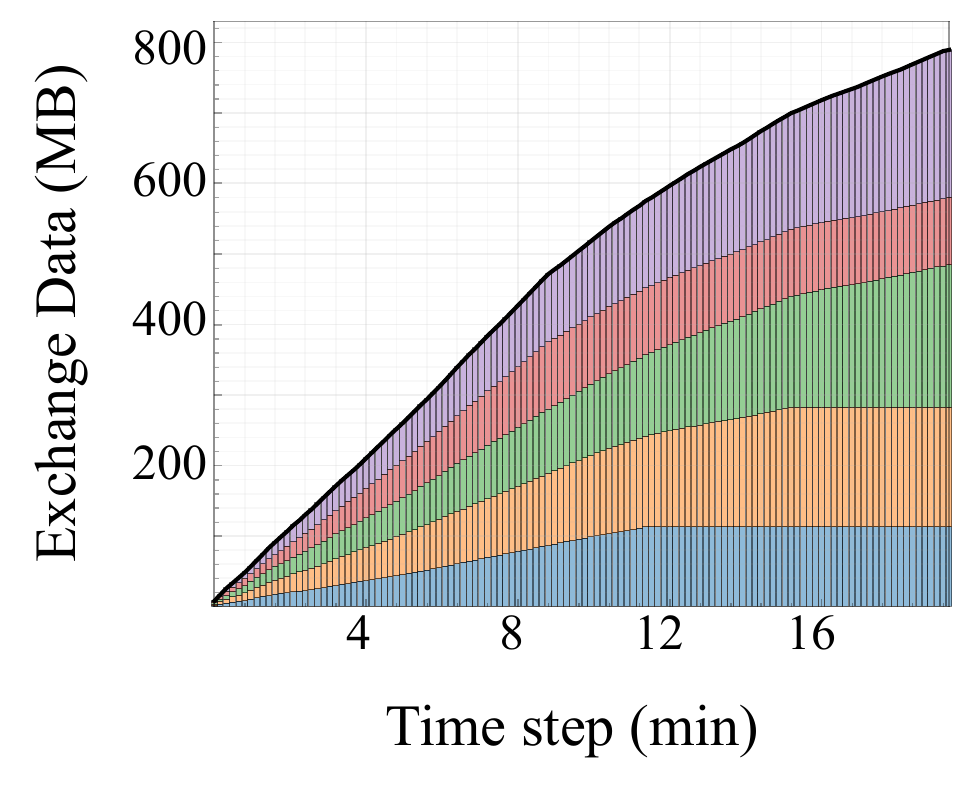}
        \vspace{-1.5em}
        \caption{}
    \end{subfigure}
    \begin{subfigure}{0.46\columnwidth}
        \includegraphics[trim={0 0 0 0},clip, width=\columnwidth]{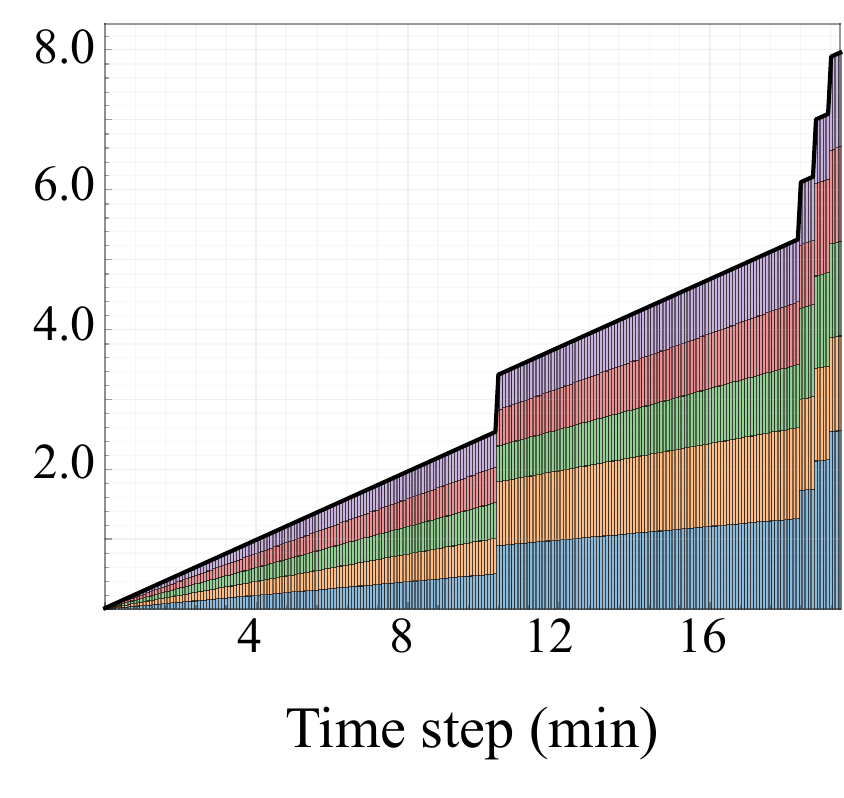}
        \vspace{-1.5em}
        \caption{}
    \end{subfigure}
    \includegraphics[trim={0 0 0 0},clip, width=0.98\columnwidth]{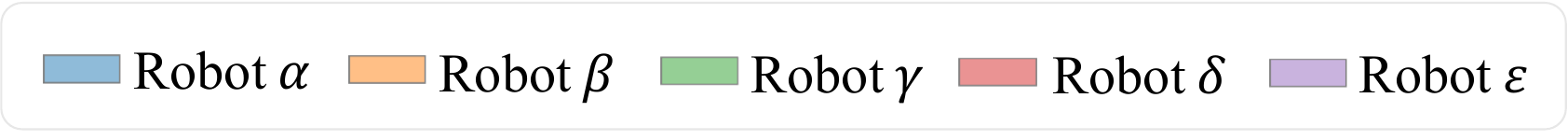}
      \caption{Cumulative data exchange from five robots to the server over time on the Kimera-Multi-\texttt{Outdoor} sequence. 
               The colored regions represent individual robot contributions~(\ie from robot $\alpha$ to robot $\epsilon$).
               (a)~Baseline methods, where each robot continuously transmits all scans, result in stacked growth reaching approximately 800\,MB. 
               (b)~Our method's communication pattern exhibits a continuous exchange of lightweight SOLiD descriptors (448 bytes per robot per keyframe), resulting in gradual baseline growth, punctuated by four discrete jumps where optimally selected scans are transmitted. 
               }
    \label{fig:num_robot}
\end{figure}
% % =============================================================
% =============================================================
\begin{table}[t]
\captionsetup{width=0.49\textwidth, justification=justified}
\caption{Onboard feasibility across three hardware tiers, $\hardC$, $\hardB$, and $\hardA$, on the \texttt{Town 01-02} sequence.
A checkmark (\gl{\cmark}) indicates the method executed without Out-of-Memory (OOM); an x mark (\rl{\xmark}) indicates failure to run due to the OOM.
}
\centering\resizebox{0.49\textwidth}{!}{\tiny
\begin{tabular}{l|l|c|c|c}
\toprule \midrule
                  & Hardware        & \hardC & \hardB & \hardA \\ \midrule
                    \multirow{6}{*}{\rotatebox[origin=c]{90}{Method}}
                  & LT-Mapper                 & \rl{\xmark} & \rl{\xmark} & \firstc \gl{\cmark} \\
                  & ELite                     & \rl{\xmark} & \rl{\xmark} & \rl{\xmark} \\
                  & Uni-Mapper                & \rl{\xmark} & \rl{\xmark} & \rl{\xmark} \\
                  & LAMM                      & \rl{\xmark} & \rl{\xmark} & \rl{\xmark} \\
                  & KISS-Matcher              & \rl{\xmark} & \firstc\gl{\cmark} & \firstc\gl{\cmark} \\ 
                  & \oursdc Ours              & \firstc\gl{\cmark} & \firstc\gl{\cmark} & \firstc\gl{\cmark} \\
\midrule \bottomrule
\end{tabular}}
\label{tab:feasibility}
\vspace{-0.2cm}
\end{table}
% ===========================================================

% % =============================================================
\begin{figure}[t!]
    \centering
    \captionsetup{justification=justified}
    \captionsetup[subfigure]{justification=centering}
    \begin{subfigure}{0.48\columnwidth}
        \includegraphics[trim={0 0 0 0},clip, width=\columnwidth]{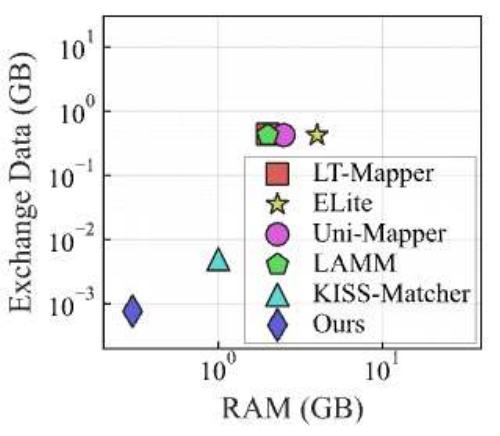}
        \vspace{-1.5em}
        \caption{}
    \end{subfigure}
    \begin{subfigure}{0.48\columnwidth}
        \includegraphics[trim={0 0 0 0},clip, width=\columnwidth]{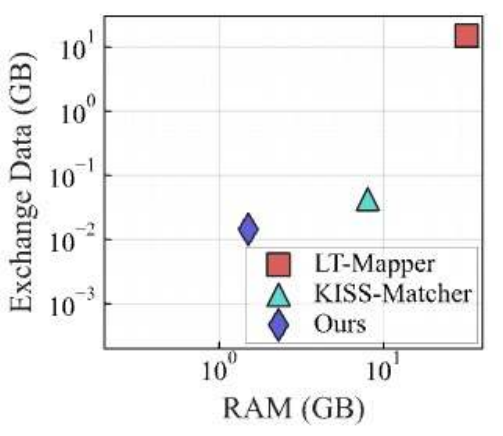}
        \vspace{-1.5em}
        \caption{}
    \end{subfigure}
    \caption{Comparison of peak RAM usage (x-axis) and total exchange data sent to the server (y-axis, both in log scale) for different map merging methods on $\hardA$ across two datasets: (a) Kimera-Multi-\texttt{Outdoor} (VLP-16) and (b)~HeLiPR~(OS2-128). 
                In (b), several baseline methods are not shown as they encountered OOM failures due to the GB-level data scale exceeding available memory capacity.}
    \label{fig:ram_exchange_tradeoff}
    \vspace{-3mm}
\end{figure}
% % =============================================================
% % =============================================================
\begin{figure}[t!]
    \centering
    \captionsetup{justification=justified}
    \captionsetup[subfigure]{justification=centering}
    \begin{subfigure}{0.58\columnwidth}
        \includegraphics[trim={0 0 0 0},clip, width=\columnwidth]{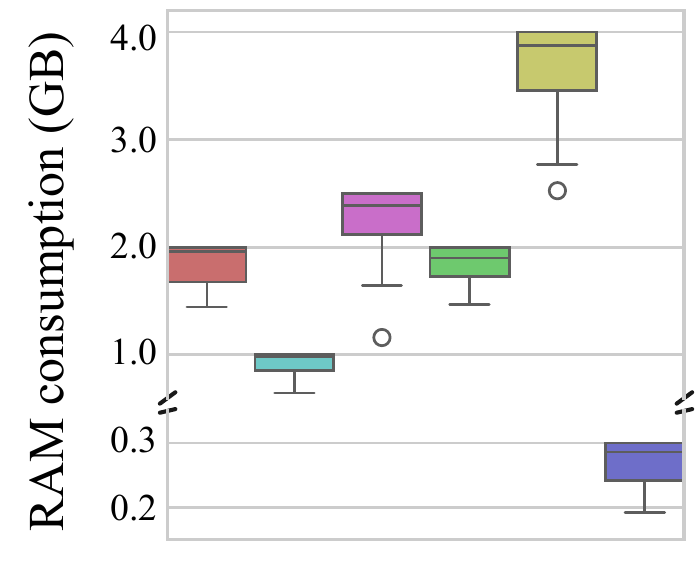}
        \vspace{-1.5em}
        \caption{}
    \end{subfigure}
    \begin{subfigure}{0.385\columnwidth}
        \includegraphics[trim={0 0 0 0},clip, width=\columnwidth]{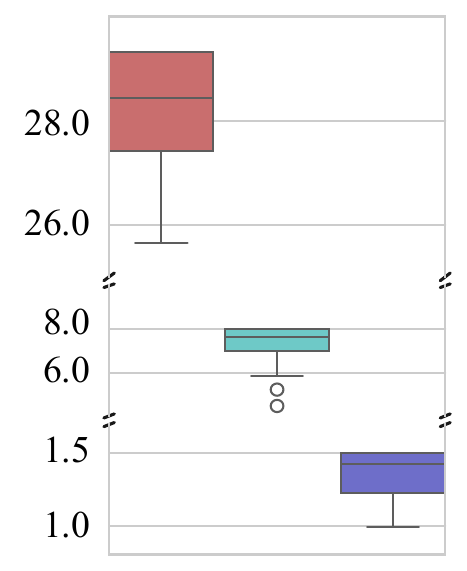}
        \vspace{-1.5em}
        \caption{}
    \end{subfigure}

    \includegraphics[trim={0 0 0 0},clip, width=0.7\columnwidth]{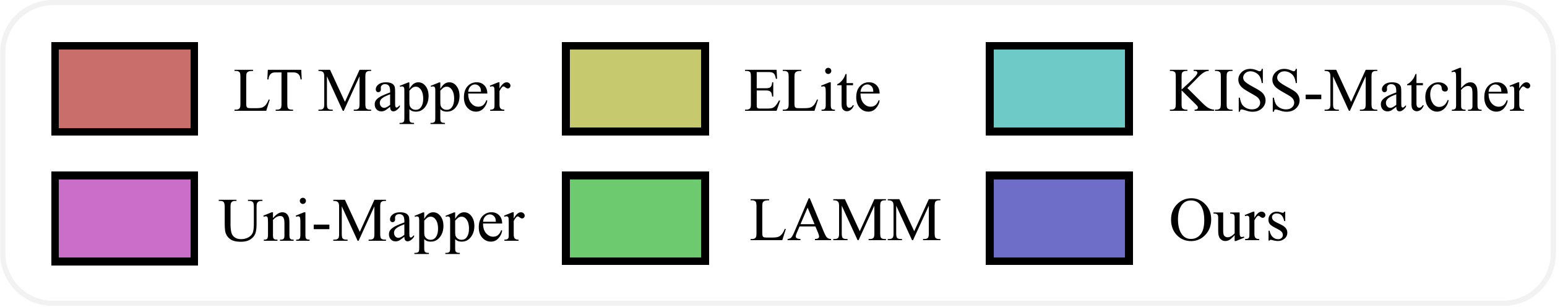}
    \caption{RAM consumption of each method across merging rounds on the (a)~Kimera-Multi-\texttt{Outdoor}~\citep{tian23arxiv_kimeramultiexperiments} and (b)~\texttt{Town}~\citep{jung2024helipr} sequences.}
    \label{fig:ram_boxplot}
    \vspace{-3mm}
\end{figure}
% % =============================================================

Furthermore, as shown in \figref{fig:num_robot}, our method scales favorably with growing team size.
The baselines exhibit steeply stacked growth as each robot continuously transmits all scans~(\figref{fig:num_robot}(a)), whereas our method maintains a much lower growth rate with occasional discrete jumps corresponding to the transmission of optimally selected scans~(\figref{fig:num_robot}(b)).

Overall, Commerge achieves both scalability, along the axes of operational scale and team size, and transmission efficiency by avoiding bursty transmission patterns, establishing its suitability for large-scale multi-robot deployment.

\subsection{Communication and Resource Efficiency Analysis}
\label{sec:resource_efficiency}
\noindent
Finally, the bottom section of Table~\ref{tab:efficiency} and Table~\ref{tab:feasibility} demonstrate the resource efficiency of our Commerge.
Owing to the substantially reduced data volume, our method achieves the shortest communication time $\commmetric$ across all sequences without time out~(T/O), while the minimal memory footprint enables deployment even on embedded platforms such as $\hardC$, where all baseline methods fail due to Out-of-Memory~(OOM).

Specifically, as shown in Figs.~\ref{fig:ram_exchange_tradeoff} and \ref{fig:ram_boxplot}, the minimal exchange data volume of our Commerge leads to the smallest peak RAM consumption, avoiding the OOM failures encountered by the baselines.

Taken together, these findings confirm that Commerge is communication- and resource-efficient, enabling reliable deployment even on resource-constrained embedded platforms.
\section{Practicality}
\label{sec:practicality}
\noindent
Building on the analysis in Section~\ref{sec:results}, we further validate the practicality of our Commerge on in-house datasets under two complementary settings.
First, we evaluate robustness under emulated network degradation in harsh environments where deploying real wireless infrastructure is infeasible.
Since centralized systems often fail to provide fair comparisons in real-world deployments due to system-level failures~\citep{tian23arxiv_kimeramultiexperiments}, we compare against the recent distributed system Multi-Proxy~\citep{wang2026communication} under identical NetEm-based degradation~\citep{hemminger2005network}.
Second, we verify the end-to-end operability of Commerge on an actual wireless communication infrastructure constructed with real access points~(Section~\ref{sec:comm_setup}).

% =============================================================
\begin{figure*}[!htbp]
    \centering
    \captionsetup{justification=justified}
    \captionsetup[subfigure]{justification=centering}
    \begin{subfigure}{0.94\columnwidth}
        \includegraphics[trim={0 0 0 0},clip, width=\columnwidth]{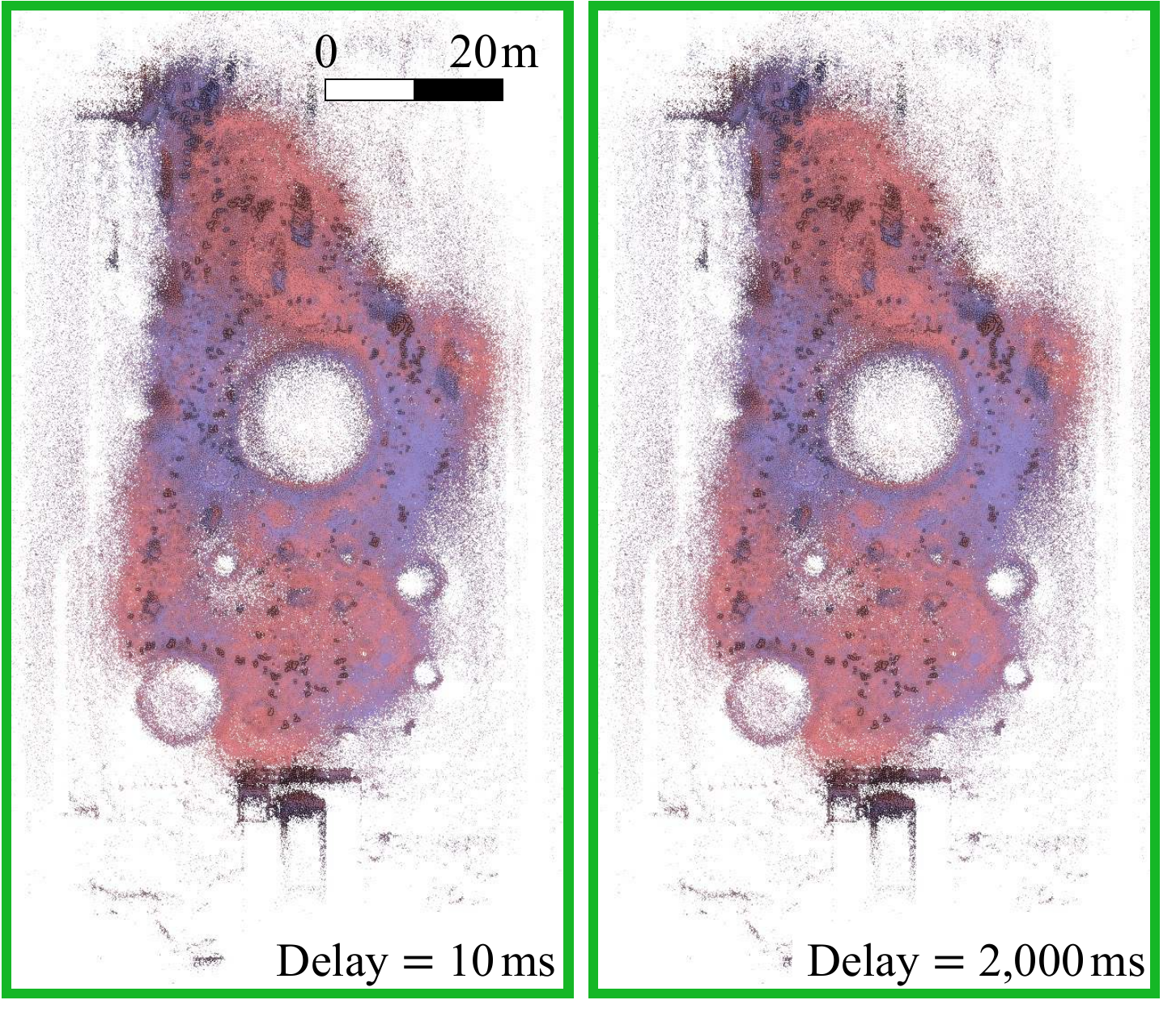}
        \vspace{-1.5em}
        \caption{Multi-Proxy \citep{wang2026communication}}
    \end{subfigure}
    \begin{subfigure}{0.94\columnwidth}
        \includegraphics[trim={0 0 0 0},clip, width=\columnwidth]{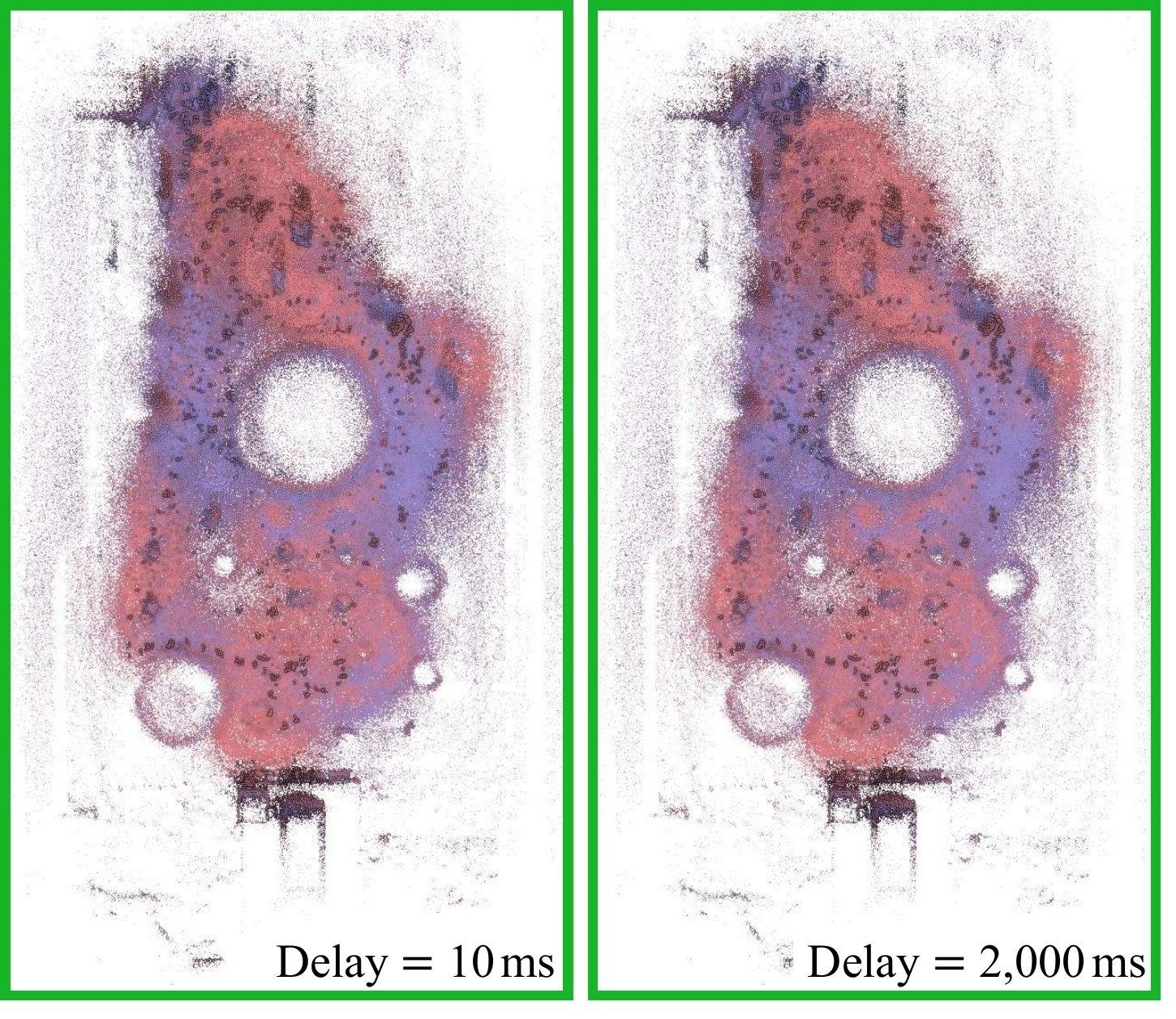}
        \vspace{-1.5em}
        \caption{Proposed}
    \end{subfigure}

    \begin{subfigure}{0.94\columnwidth}
        \includegraphics[trim={0 0 0 0},clip, width=\columnwidth]{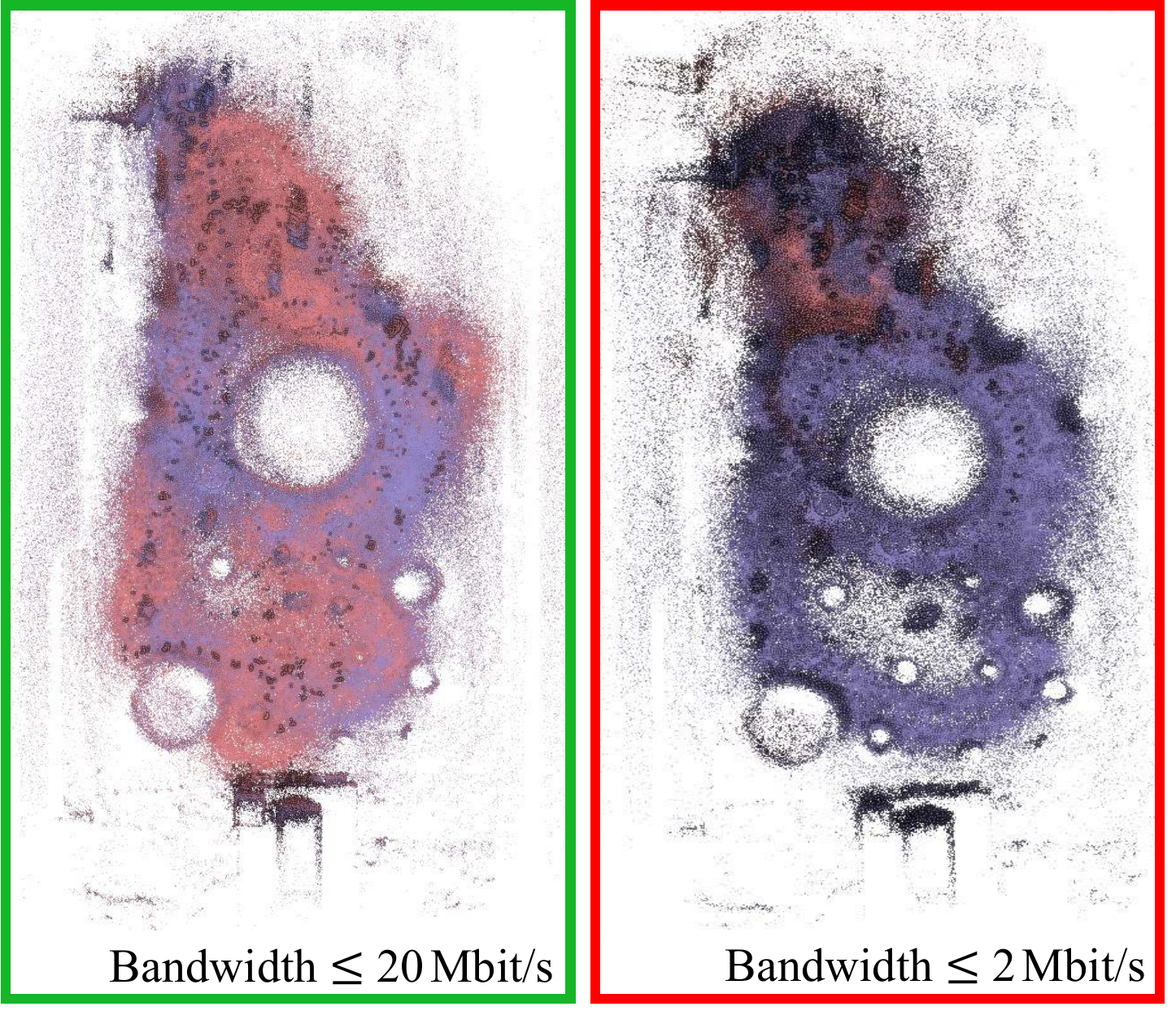}
        \vspace{-1.5em}
        \caption{Multi-Proxy \citep{wang2026communication}}
    \end{subfigure}
    \begin{subfigure}{0.94\columnwidth}
        \includegraphics[trim={0 0 0 0},clip, width=\columnwidth]{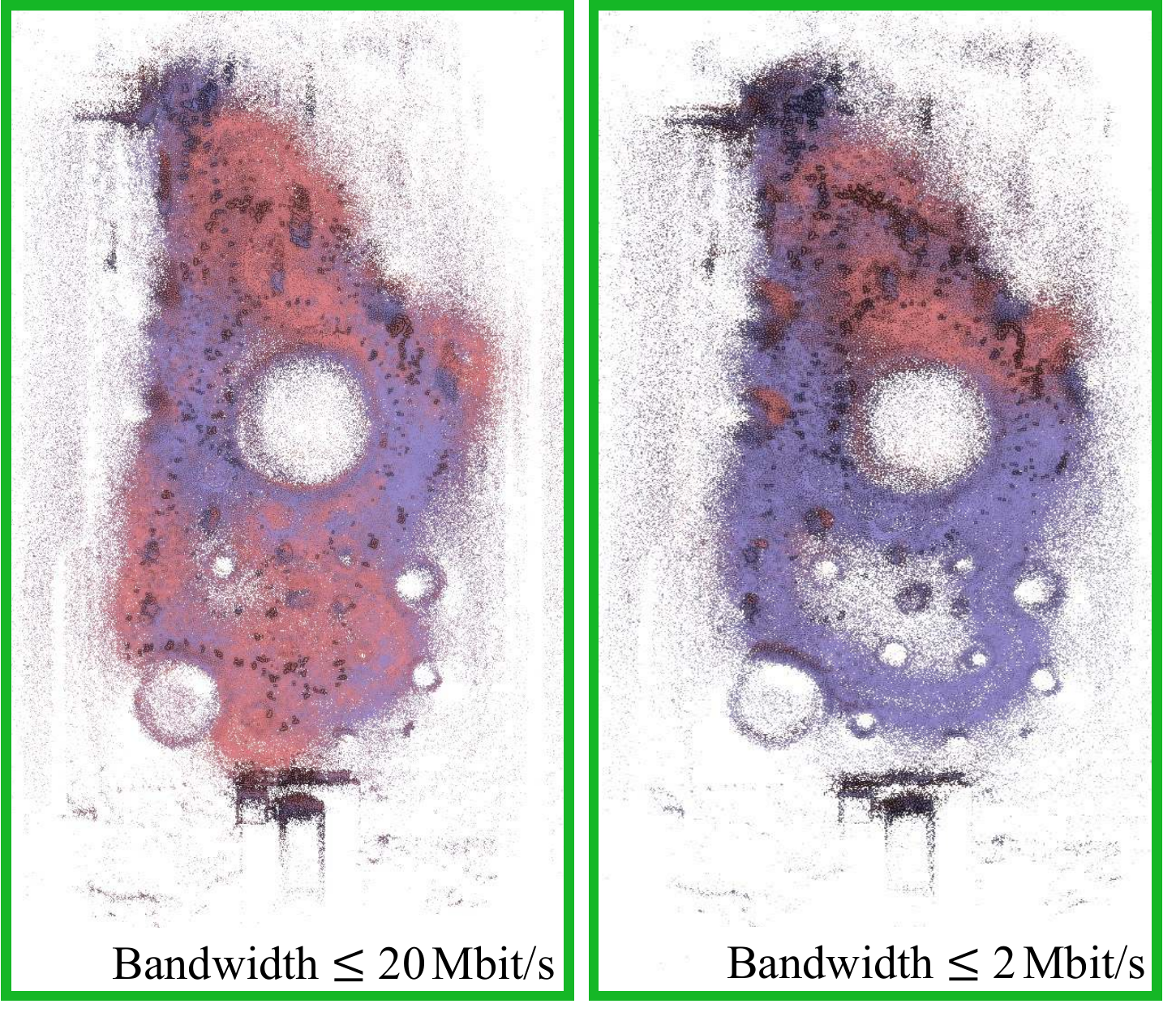}
        \vspace{-1.5em}
        \caption{Proposed}
    \end{subfigure}

    \begin{subfigure}{0.94\columnwidth}
        \includegraphics[trim={0 0 0 0},clip, width=\columnwidth]{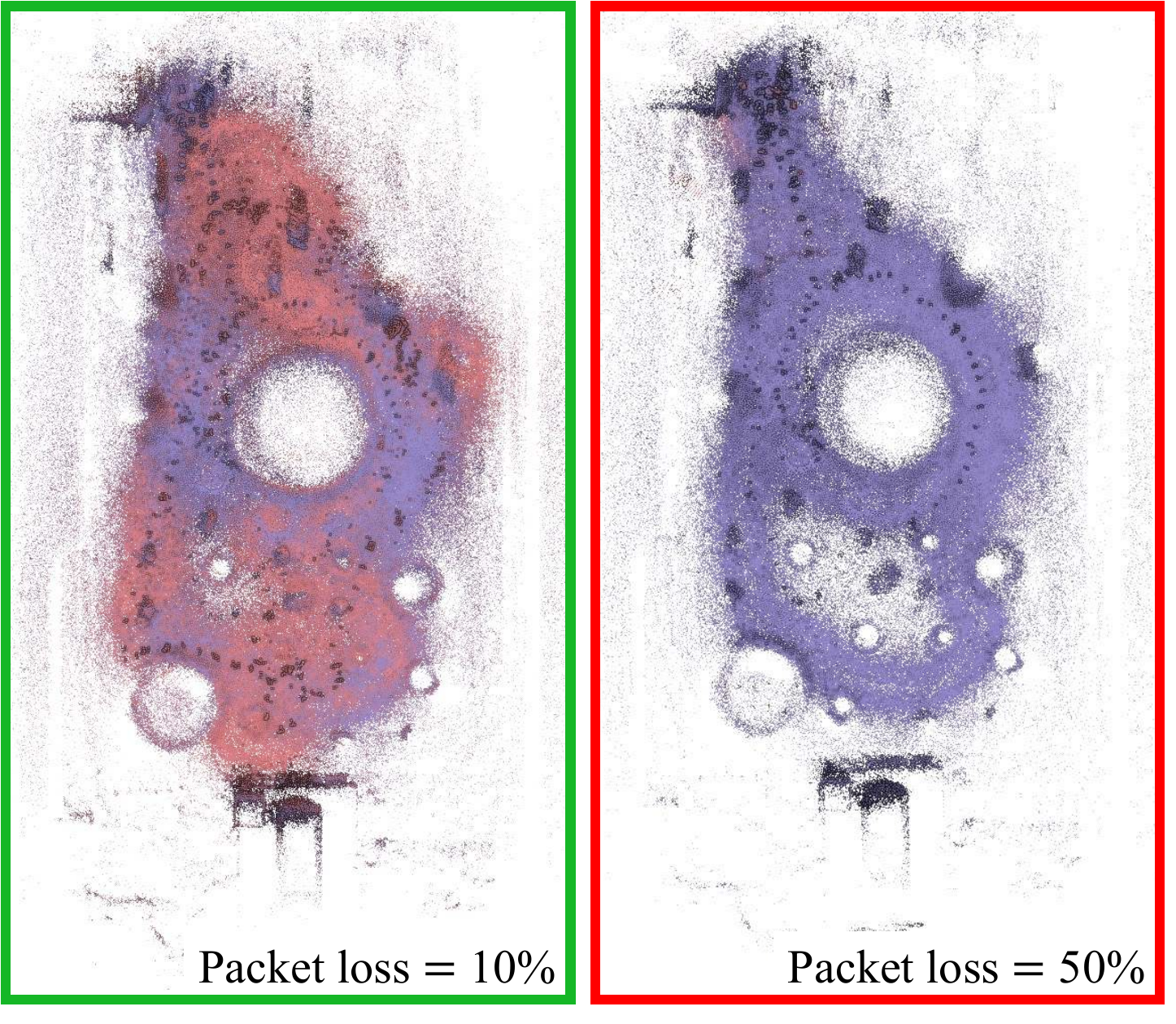}
        \vspace{-1.5em}
        \caption{Multi-Proxy \citep{wang2026communication}}
    \end{subfigure}
    \begin{subfigure}{0.94\columnwidth}
        \includegraphics[trim={0 0 0 0},clip, width=\columnwidth]{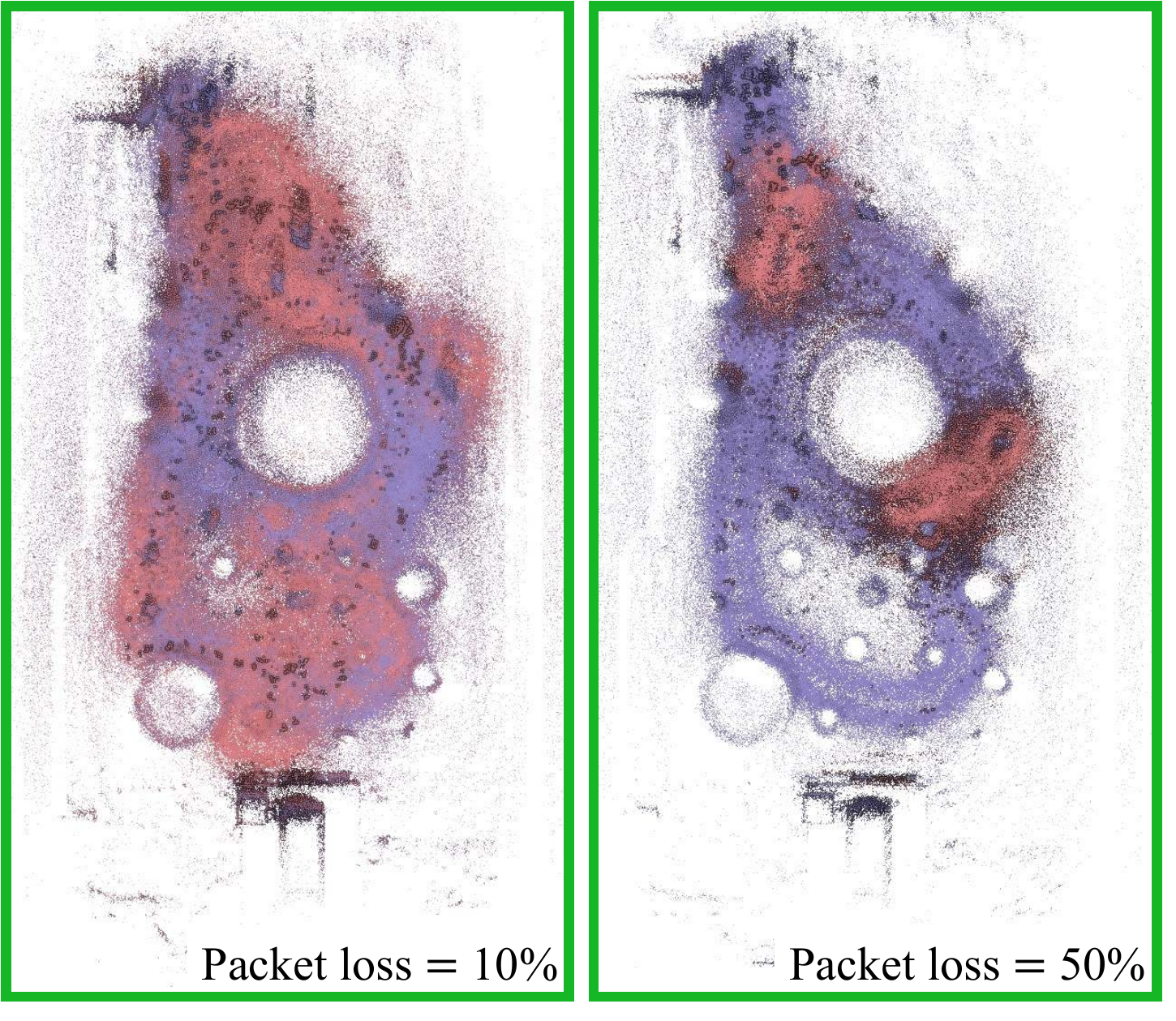}
        \vspace{-1.5em}
        \caption{Proposed}
    \end{subfigure}
    \includegraphics[trim={0 0 0 0},clip, width=0.7\columnwidth]{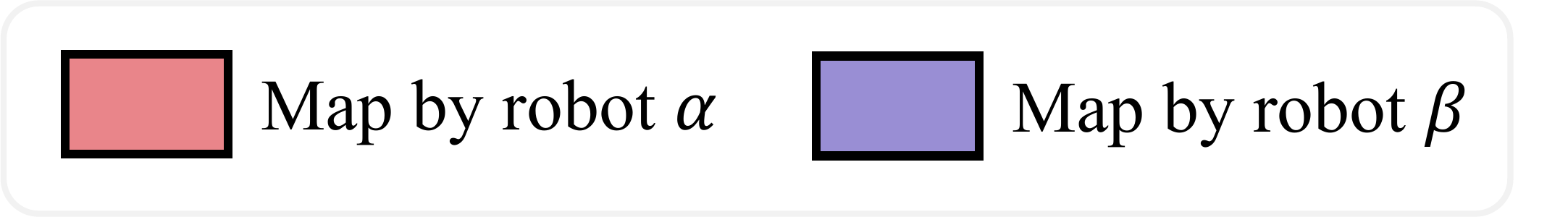}
    \caption{Qualitative comparison with Multi-Proxy~\citep{wang2026communication} under network degradation simulated via NetEm~\citep{hemminger2005network} on the planetary terrain dataset.
    (a)--(b)~Network delay (10\,ms vs.\ 2{,}000\,ms), 
    (c)--(d)~bandwidth limitation ($\leq$20\,Mbit/s vs.\ $\leq$2\,Mbit/s), 
    (e)--(f)~packet loss (10\% vs.\ 50\%).
    Green and red boxes indicate successful and failed merging.
    }
    \label{fig:multi_proxy}
\end{figure*}
% =============================================================

% % =============================================================
\begin{figure*}[t]
    \centering
    \captionsetup{justification=justified}
    \captionsetup[subfigure]{justification=centering}
    \begin{subfigure}{0.38\columnwidth}
        \includegraphics[trim={0 0 0 0},clip, width=\columnwidth]{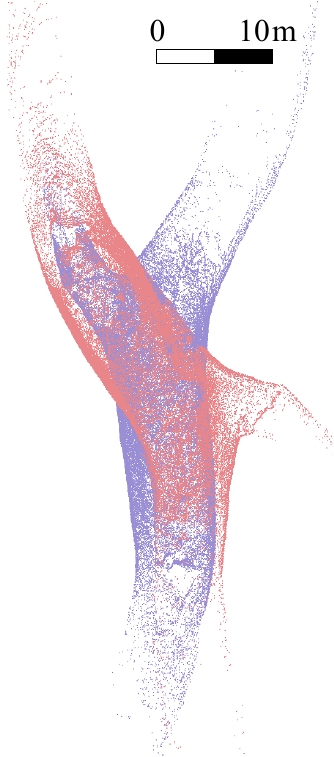}
        % \vspace{-1em}
        \caption{}
    \end{subfigure}
    \begin{subfigure}{0.38\columnwidth}
        \includegraphics[trim={0 0 0 0},clip, width=\columnwidth]{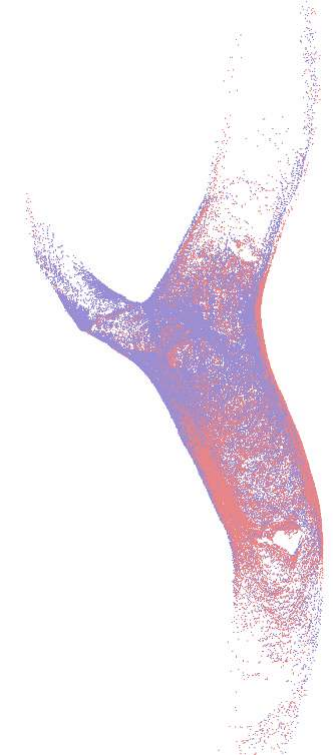}
        % \vspace{-1em}
        \caption{}
    \end{subfigure}
    \begin{subfigure}{0.38\columnwidth}
        \includegraphics[trim={0 0 0 0},clip, width=\columnwidth]{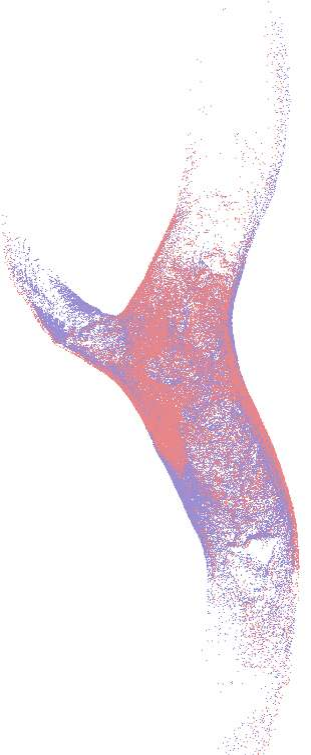}
        % \vspace{-1em}
        \caption{}
    \end{subfigure}
    \begin{subfigure}{0.38\columnwidth}
        \includegraphics[trim={0 0 0 0},clip, width=\columnwidth]{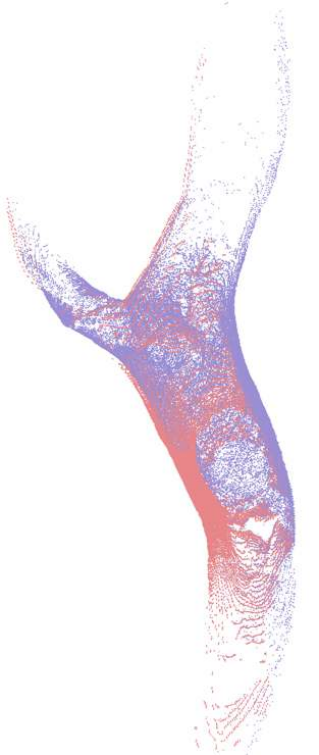}
        % \vspace{-1em}
        \caption{}
    \end{subfigure}
    \begin{subfigure}{0.38\columnwidth}
        \includegraphics[trim={0 0 0 0},clip, width=\columnwidth]{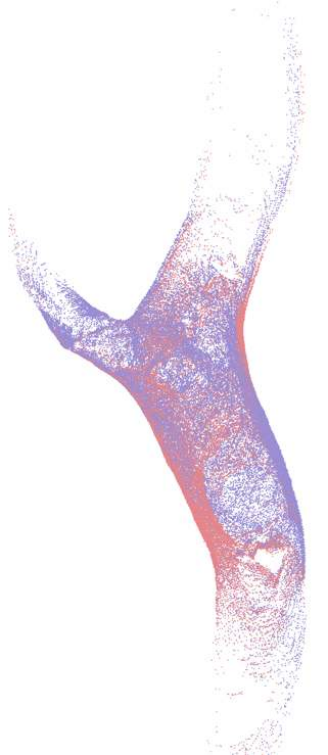}
        % \vspace{-1em}
        \caption{}
    \end{subfigure}
    \includegraphics[trim={0 0 0 0},clip, width=0.7\columnwidth]{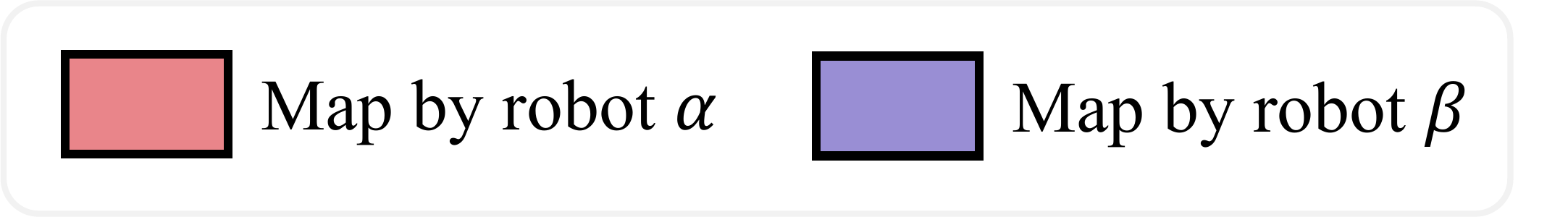}
    \caption{Multi-robot map merging results on a subterranean cave dataset under various network conditions.
    (a)~Before map merging.
    (b)~Map merging without network degradation.
    (c)~Map merging under 2000\,ms delay.
    (d)~Map merging under 2\,Mbit/s bandwidth limit.
    (e)~Map merging under 50\% packet loss.}
    \label{fig:cave}
    \vspace{-3mm}
\end{figure*}
% % =============================================================
% % =============================================================
\begin{figure*}[t]
    \centering
    \captionsetup{justification=justified}
    \captionsetup[subfigure]{justification=centering}
    \begin{subfigure}{0.80\columnwidth}
        \includegraphics[trim={0 0 0 0},clip, width=\columnwidth]{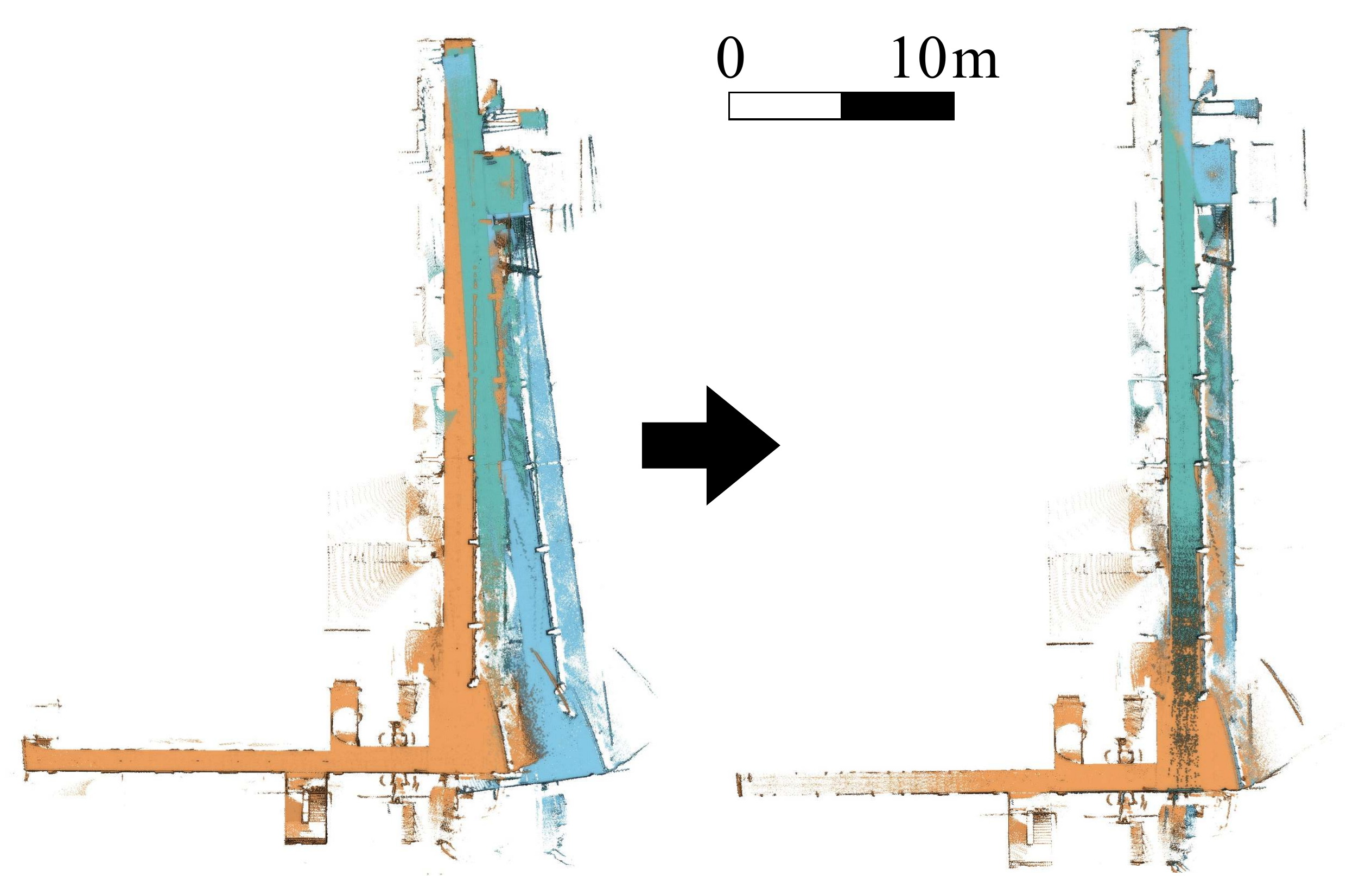}
        % \vspace{-1em}
        \caption{}
    \end{subfigure}
    \begin{subfigure}{1.1\columnwidth}
        \includegraphics[trim={0 0 0 0},clip, width=\columnwidth]{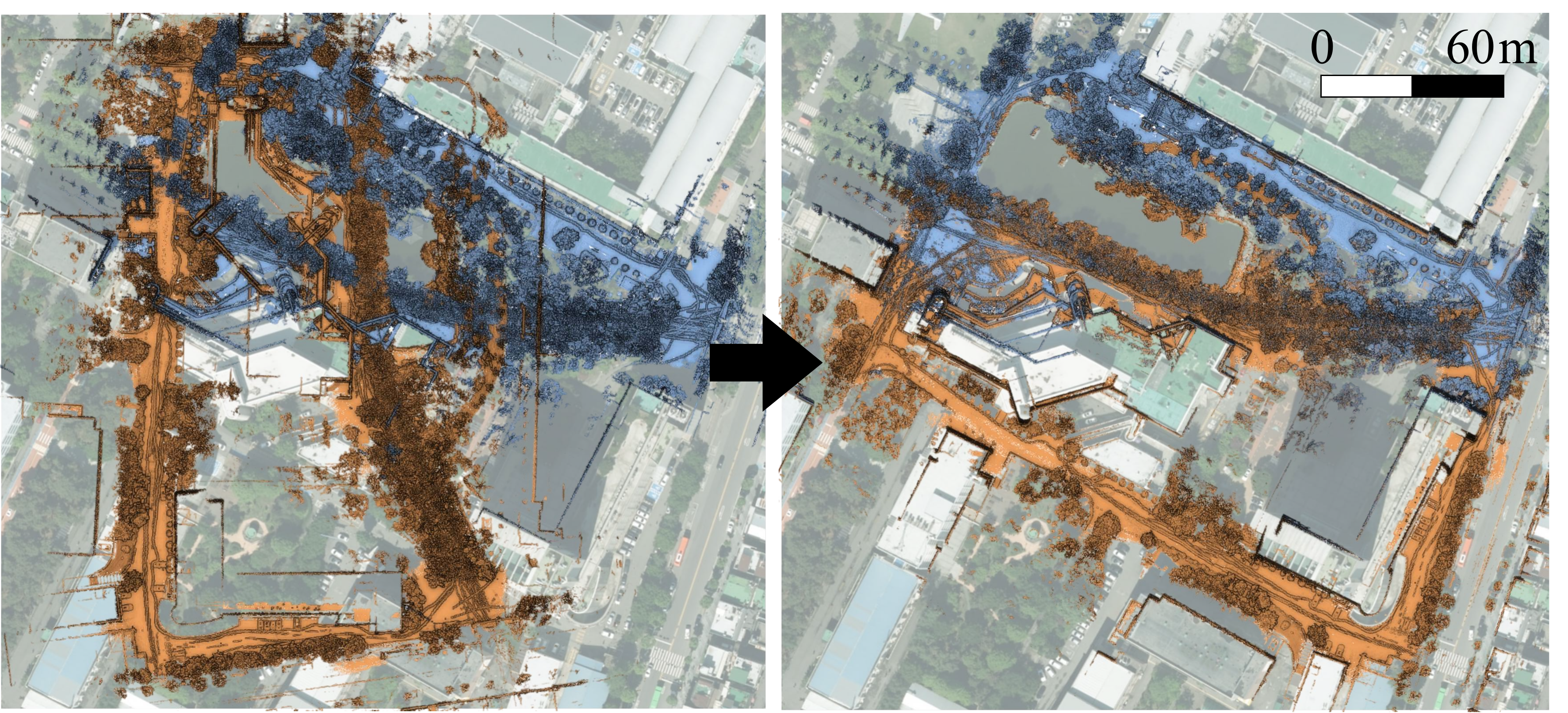}
        % \vspace{-1em}
        \caption{}
    \end{subfigure}
    \includegraphics[trim={0 0 0 0},clip, width=1\columnwidth]{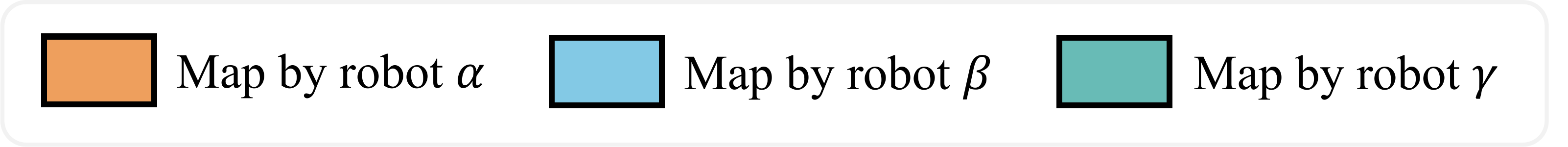}
    \caption{Multi-robot map merging results on the (a)~INHA-\texttt{Indoor} and (b)~INHA-\texttt{Outdoor}.
    Left: individual robot maps before merging. Right: merged map produced by Commerge.}
    \label{fig:inha_robot_map}
    \vspace{-3mm}
\end{figure*}
% % =============================================================
% % =============================================================
\begin{figure*}[t]
    \centering
    \captionsetup{justification=justified}
    \captionsetup[subfigure]{justification=centering}
    \begin{subfigure}{0.8\columnwidth}
        \includegraphics[trim={0 0 0 0},clip, width=\columnwidth]{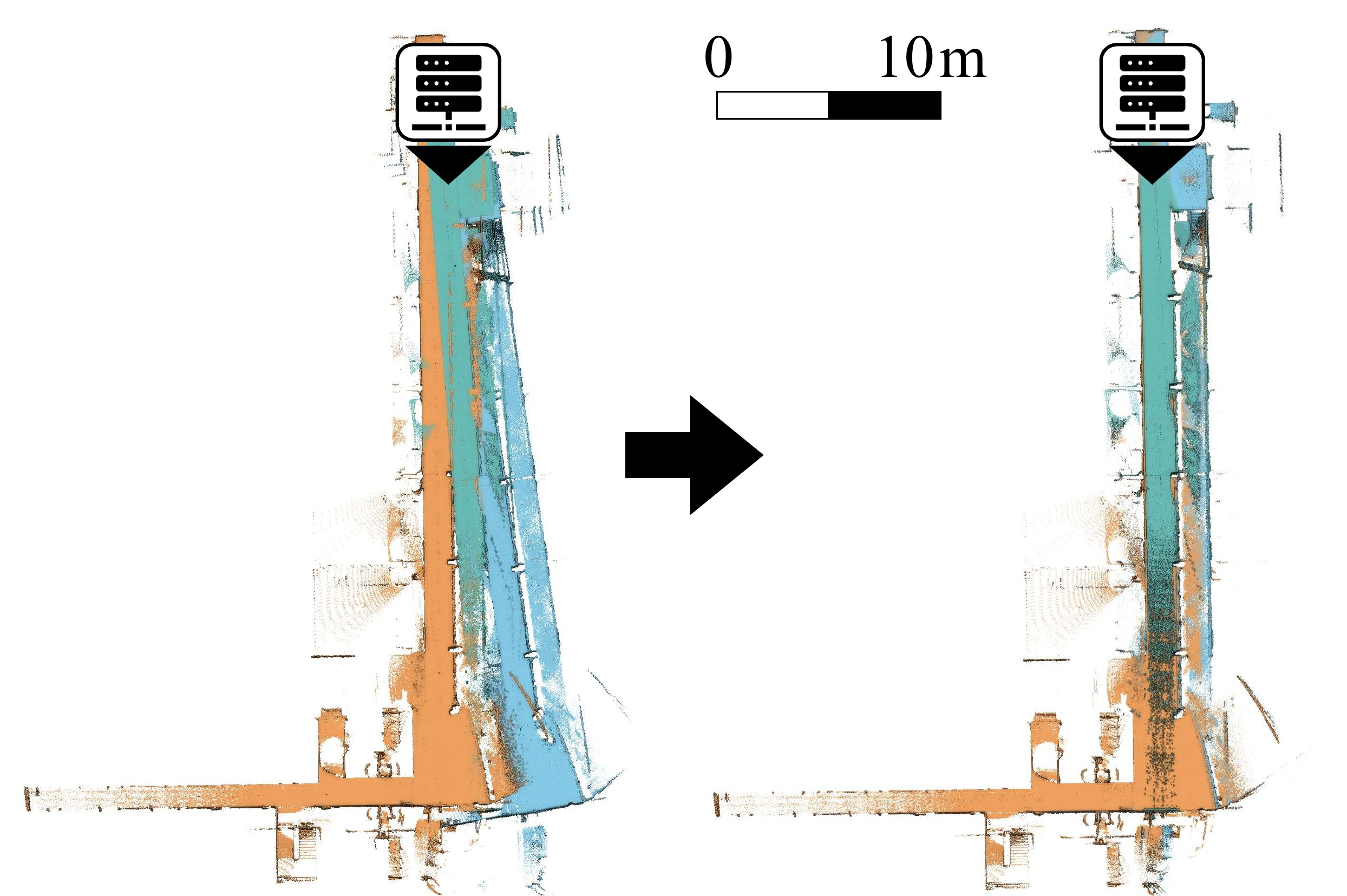}
        % \vspace{-1em}
        \caption{}
    \end{subfigure}
    \begin{subfigure}{1.1\columnwidth}
        \includegraphics[trim={0 0 0 0},clip, width=\columnwidth]{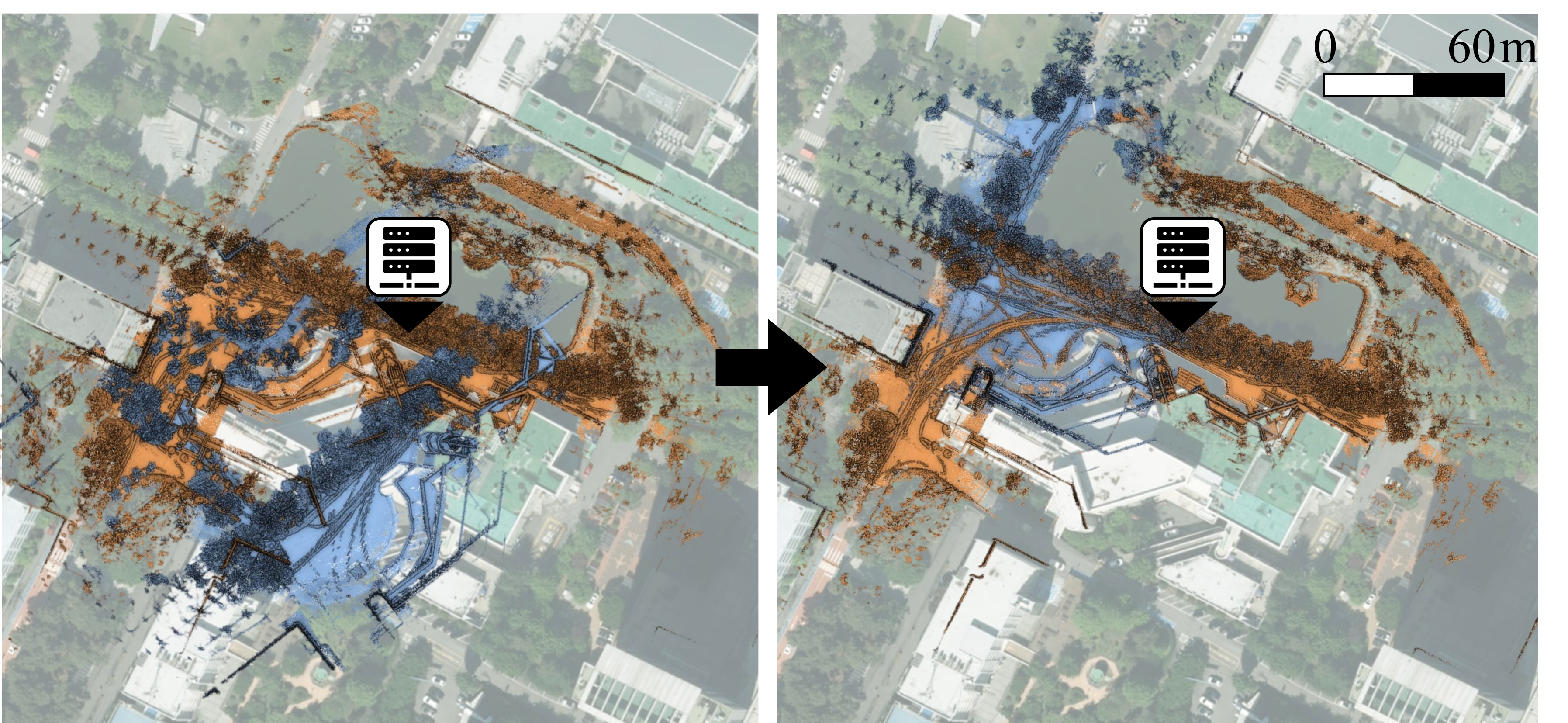}
        % \vspace{-1em}
        \caption{}
    \end{subfigure}
    \includegraphics[trim={0 0 0 0},clip, width=1\columnwidth]{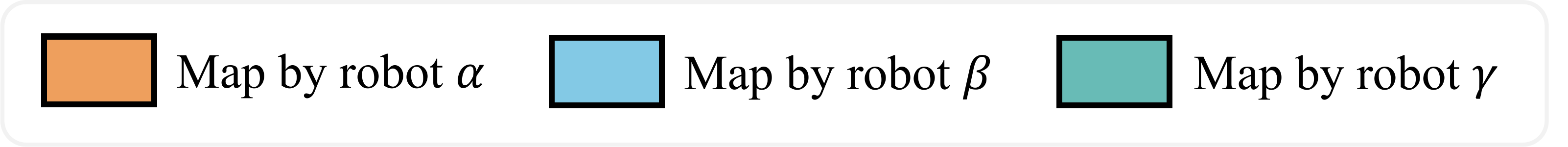}
    \caption{Multi-robot map merging results on the (a)~INHA-\texttt{Indoor} and (b)~INHA-\texttt{Outdoor} datasets, where each robot transmits its data to the server over a real wireless network and the server performs map merging using the received data.
    Left: individual robot maps before merging.
    Right: merged map produced by Commerge.
    The $\server$ icon indicates the actual location of the server on each map, where signals are blocked by walls or buildings, causing non-line-of-sight~(NLOS)-induced disconnection and dropout.}
    \label{fig:inha_server_map}
    \vspace{-3mm}
\end{figure*}
% % =============================================================

\subsection{Evaluation under Emulated Network Degradation}
\label{sec:emulated_network}
\noindent
First, as shown in Figs.~\ref{fig:multi_proxy} and \ref{fig:cave}, our Commerge successfully performs map merging under various communication degradation scenarios, including network delay, bandwidth limitation, and packet loss.
This robustness stems from the compactness of our descriptor: as shown in Table~\ref{tab:drop_ratio}, SOLiD occupies only 448\,B per descriptor, whereas Multi-Proxy transmits 1{,}585\,B. 
Since a smaller descriptor fits within fewer network packets, it is inherently less susceptible to bandwidth saturation and packet loss.
Consequently, under mild conditions (\ie 2\,ms delay, $\leq$20\,Mbit/s bandwidth, and 10\% packet loss), both methods experience no descriptor drop since their packets are delivered within the available network budget.
However, as the degradation intensifies (\ie 2{,}000\,ms delay, $\leq$2\,Mbit/s bandwidth, and 50\% packet loss), the larger descriptors of Multi-Proxy are more likely to be truncated or lost during transmission, resulting in substantially higher drop rates than our method (\eg 92.52\% vs.\ 62.64\% drop under 2\,Mbit/s bandwidth, and 99.43\% vs.\ 75.91\% drop under 50\% packet loss), leaving our method with a sufficient pool of surviving descriptors to recover successful map merging.

Overall, Commerge remains practically operable under realistic network impairments that typically disrupt multi-robot mapping systems.

\subsection{Evaluation on Real Wireless Infrastructure}
\label{sec:real_wireless}
\noindent
In addition, as shown in Figs.~\ref{fig:inha_robot_map} and \ref{fig:inha_server_map}, our method also operates reliably on the real wireless infrastructure described in Section~\ref{sec:comm_setup}, where Figs.~\ref{fig:inha_robot_map} and \ref{fig:inha_server_map} present the per-robot and server-side merged maps, respectively.
Notably, despite the non-line-of-sight~(NLOS) conditions caused by indoor walls in \figref{fig:inha_server_map}(a) and by outdoor buildings in \figref{fig:inha_server_map}(b), which induce the disconnection and packet drops reported in Table~\ref{tab:drop}, our Commerge still achieves successful map merging in both environments.

Taken together, these findings confirm the end-to-end deployability of Commerge on practical wireless networks, validating its readiness for real-world multi-robot missions.

% % =============================================================
\begin{table}[t]
\captionsetup{width=0.49\textwidth, justification=justified}
\caption{Average descriptor size and drop ratio under various network degradation conditions on the INHA-\texttt{Planetary} dataset. 
Drop ratio is defined as the percentage of descriptors that fail to be successfully delivered. A lower value indicates better communication robustness.}
\centering\resizebox{0.49\textwidth}{!}{\tiny
\begin{tabular}{l|l|c|c}
\toprule \midrule
\multicolumn{2}{l|}{Method} & \makecell{Multi-Proxy \\ \citep{wang2026communication}} & Proposed \\ \midrule
\multicolumn{2}{l|}{Avg. descriptor size [B] $\downarrow$} & 1{,}585 & \textbf{448} \\ \midrule
\multirow{7}{*}{\rotatebox[origin=c]{90}{Drop ratio [\%] $\downarrow$}} 
    & Delay (2\,ms)                & 0     & 0     \\
    & Delay (2,000\,ms)            & 2.54  & \textbf{0.31}  \\ \cmidrule(lr){2-4}
    & Bandwidth ($\leq$20\,Mbit/s) & 0     & 0     \\
    & Bandwidth ($\leq$2\,Mbit/s)  & 92.52 & \textbf{62.64} \\ \cmidrule(lr){2-4}
    & Packet loss (10\%)           & 0     & 0     \\
    & Packet loss (50\%)           & 99.43 & \textbf{75.91} \\
\midrule \bottomrule
\end{tabular}}
\label{tab:drop_ratio}
\vspace{-2mm}
\end{table}
% % =============================================================
% =============================================================
\begin{table}[t]
\captionsetup{width=.49\textwidth, justification=justified}
\caption{Descriptor dropout statistics under real communication conditions on the INHA dataset. Total scans transmitted by each robot~(\robot) and successfully received by the server~(\server) are reported, along with the resulting dropout ratio.}
\centering\resizebox{0.49\textwidth}{!}{\tiny
\begin{tabular}{l|c|c|c|c|c}
\toprule \midrule
Sequence                  & \multicolumn{3}{c|}{\texttt{Indoor}}                          & \multicolumn{2}{c}{\texttt{Outdoor}}          \\ \cmidrule(lr){2-4} \cmidrule(lr){5-6}
Robot                     & robot $\alpha$     & robot $\beta$      & robot $\gamma$      & robot $\alpha$         & robot $\beta$        \\ \midrule
Total scans ($\robot$)    & 3,265              & 2,050              & 1,346               & 7,204                  & 5,773                \\
Total scans ($\server$)   & 2,326              & 1,876              & 1,234               & 2,029                  & 1,541                \\ \midrule
Dropout ratio [\%]        & 28.8               & 8.5                & 8.3                 & 71.8                   & 73.3                 \\
\midrule \bottomrule
\end{tabular}}
\label{tab:drop}
\vspace{-4mm}
\end{table}
% =============================================================

\newcommand{\mvc}{\raisebox{-0.5ex}{\includegraphics[height=2.5ex]{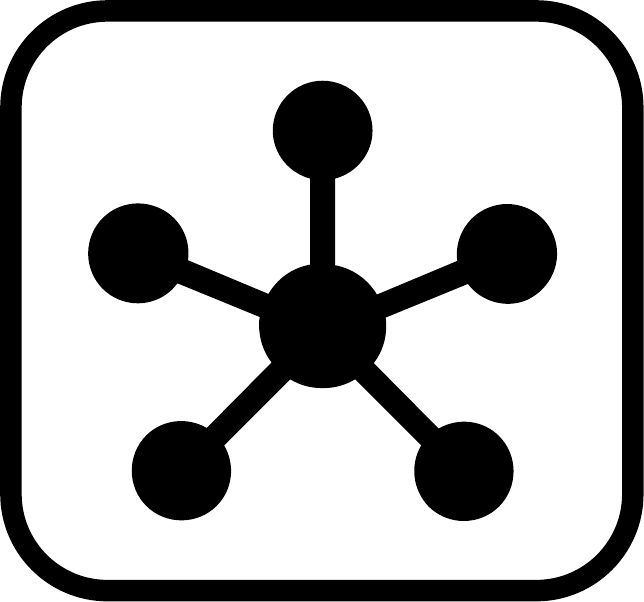}}}
\newcommand{\clique}{\raisebox{-0.5ex}{\includegraphics[height=2.5ex]{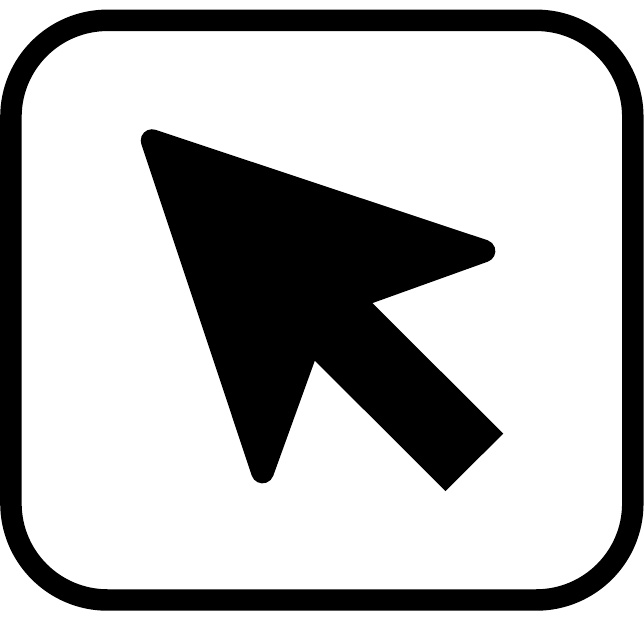}}}
\newcommand{\seq}{\raisebox{-0.5ex}{\includegraphics[height=2.5ex]{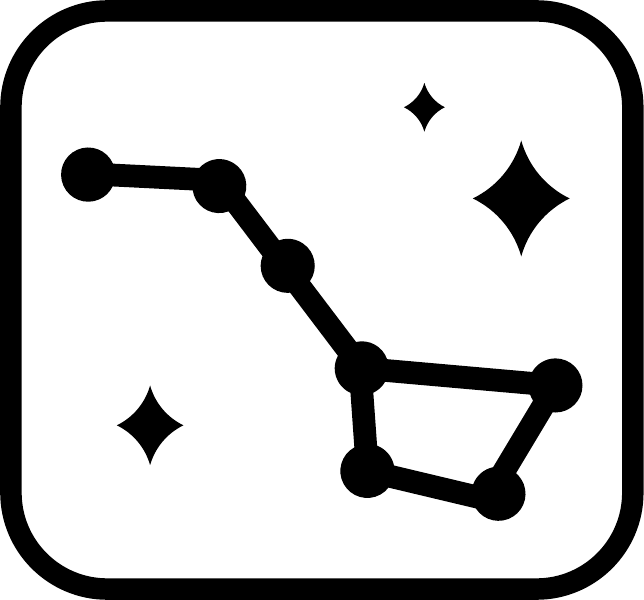}}}

\newcommand{\scenarioA}{\textsf{scenario~(A)}}
\newcommand{\scenarioB}{\textsf{scenario~(B)}}
\newcommand{\scenarioC}{\textsf{scenario~(C)}}
\newcommand{\scenarioD}{\textsf{scenario~(D)}}

\section{Ablation Studies}
\label{sec:ablation}
\noindent
In addition, we conduct ablation studies along three axes.
First, we analyze the contribution of each stage in our three-stage loop closure selection pipeline through (iii)~sequential matching, (iv)~balanced vertex cover, and (v)~maximum edge-weighted clique selection on datasets ranging from the small-scale \texttt{NTU} to the large-scale \texttt{Town} sequences~(Section~\ref{sec:ablation_pipeline}).
Second, we verify the robustness of our Commerge framework through (vi)~descriptor-agnostic applicability on the Kimera-Multi-\texttt{Outdoor} sequence, where multi-robot partial trajectory overlap~($\smalloverlap$) causes Scan Context-based methods such as LT-Mapper and ELite to fail, and (vii)~reliability across mission duration on the large-scale HeLiPR~\texttt{Town} sequence~(Section~\ref{sec:ablation_robustness}).
Finally, we validate the choice of constituent modules through (i)~place recognition performance on the MCD~\texttt{NTU} sequences acquired with non-repetitive scan pattern LiDAR~($\livox$) and (ii)~registration performance on the Kimera-Multi-\texttt{Tunnel} sequence, where repetitive geometric features make registration particularly challenging~(Section~\ref{sec:ablation_modules}).

% =============================================================
\begin{figure*}[t]
    \centering
    \captionsetup{justification=justified}
    \captionsetup[subfigure]{justification=centering}
    \begin{subfigure}{0.98\columnwidth}
        \includegraphics[trim={0 0 0 0},clip, width=\columnwidth]{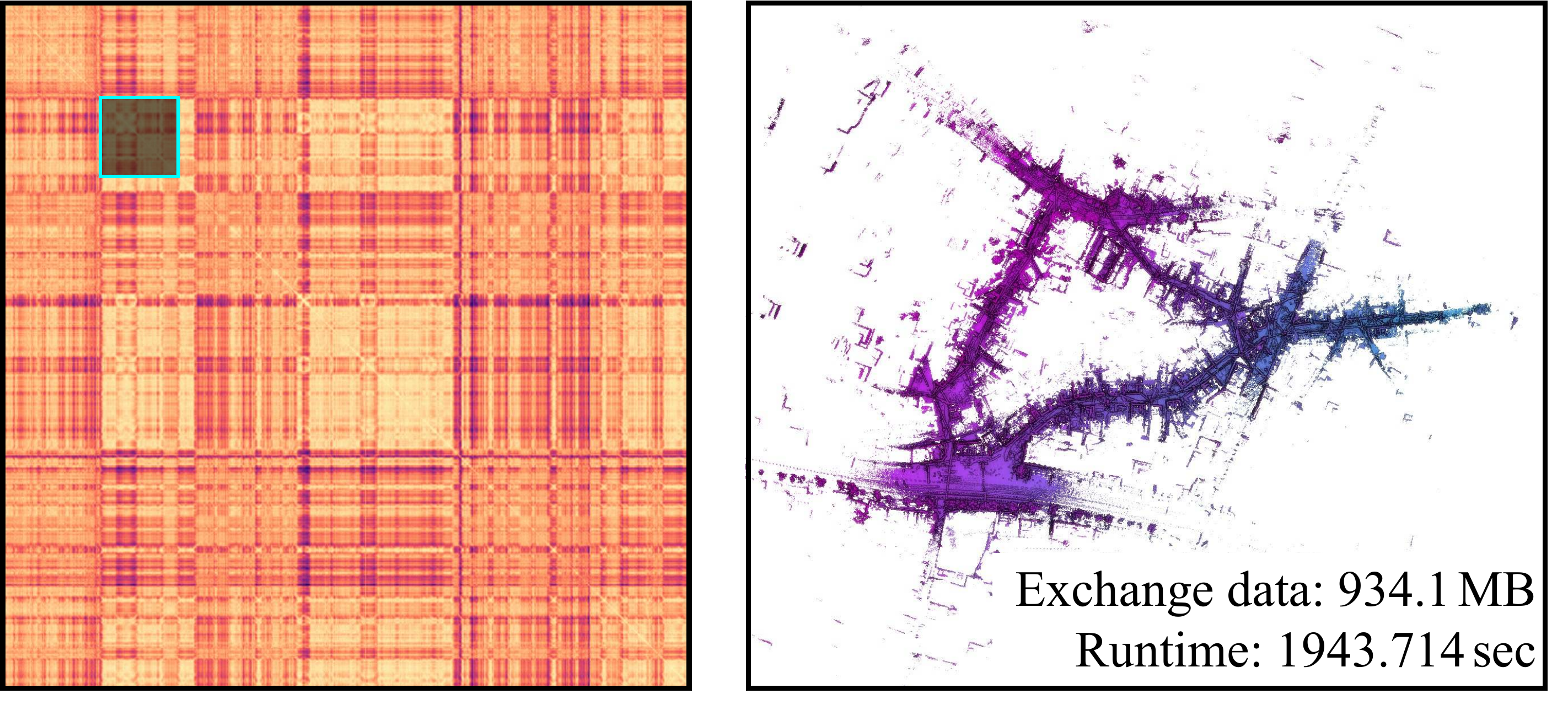}
        \vspace{-1.5em}
        \caption{AutoMerge \citep{yin2023automerge}}
    \end{subfigure}
    \begin{subfigure}{0.98\columnwidth}
        \includegraphics[trim={0 0 0 0},clip, width=\columnwidth]{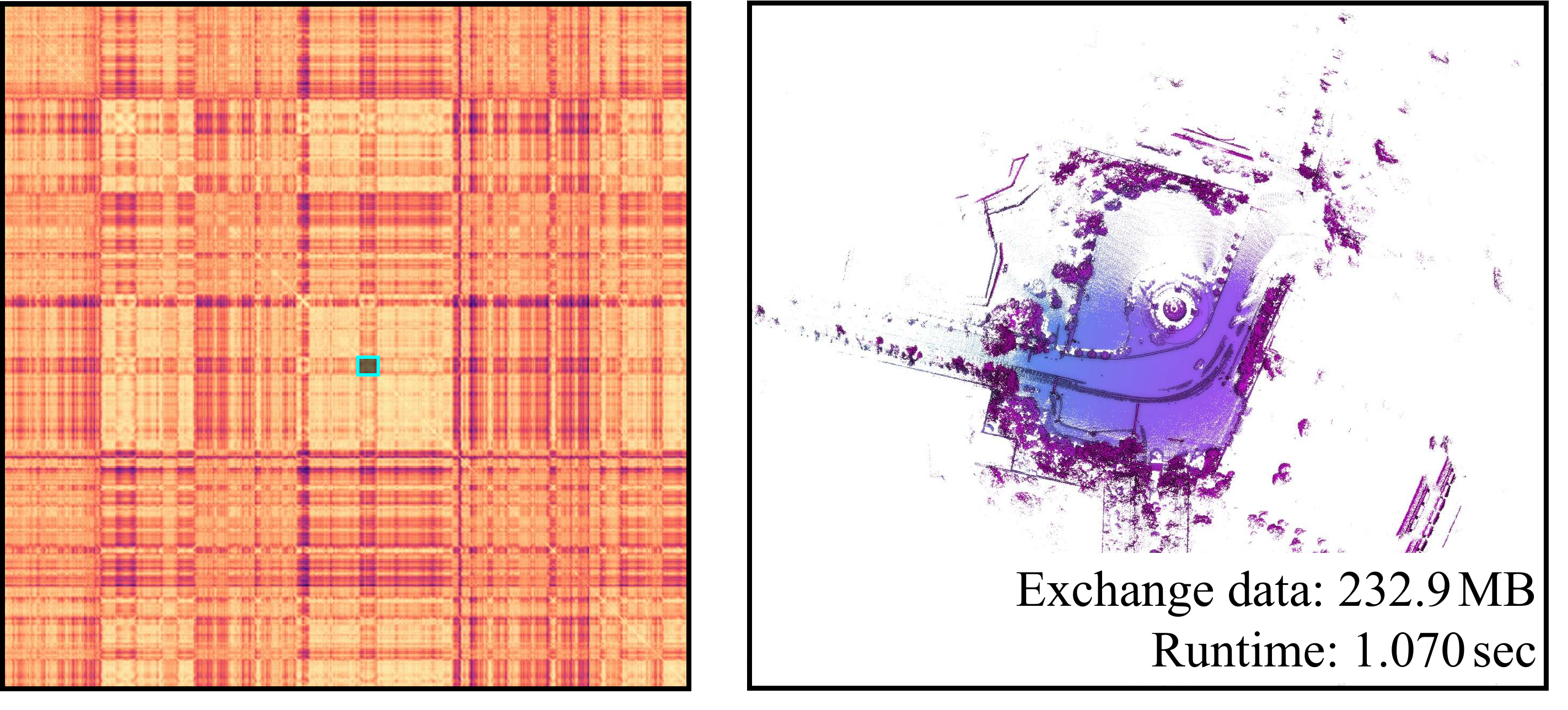}
        \vspace{-1.5em}
        \caption{Proposed}
    \end{subfigure}
    \caption{Sequential matching comparison on the HeLiPR dataset~\citep{jung2024helipr} between (a)~AutoMerge~\citep{yin2023automerge} and (b)~our proposed method.
    For each method, the left panel shows the affinity matrix where the highlighted box indicates the sequentially matched region, and the right panel shows the merged submap constructed from the submaps of robot~$\alpha$ and robot~$\beta$ corresponding to the selected region.} 
    \label{fig:seq_compare}
    % \vspace{-3mm}
\end{figure*}
% =============================================================
% =============================================================
\begin{table*}[t]
\centering
\captionsetup{justification=justified}
\captionsetup[subtable]{justification=centering}
\caption{Ablation study on the three-stage loop closure selection pipeline. 
Each row incrementally adds a stage to the baseline sequential matching ($\seq$): balanced vertex cover ($\mvc$), maximum edge-weighted clique ($\clique$), and their combination. 
We utilize map merging accuracy~($\rmsemetric$), computation time~($\commmetric$), and data exchange volume~($\exchmetric$) as evaluation metrics.}

\begin{subtable}[t]{0.48\textwidth}
  \caption{\texttt{NTU 01-10}~\citep{nguyen2024mcd}}
  \label{tab:ablation2_mcd}
  \centering
  \resizebox{\textwidth}{!}{\tiny
  \begin{tabular}{l|ccc}
    \toprule
    Evaluation & $\rmsemetric$ & $\commmetric$ & $\exchmetric$ \\
    \midrule
      $\seq$                   &  \hspace{0.15cm}  \secondc 1.826\,m    \hspace{0.15cm}  
                               &  \hspace{0.25cm}  \thirdc 24.137\,sec  \hspace{0.25cm}  
                               &   \hspace{0.15cm}  20.3\,MB            \hspace{0.15cm}     \\
      +\,$\mvc$                & 2.148\,m                      & \secondc 14.268\,sec              &  \thirdc  12.0\,MB         \\
      +\,$\clique$             & \firstc  \textbf{1.792}\,m    & 131.141\,sec                      &  \firstc  \textbf{3.9\,MB}                        \\
      +\,$\mvc$\,+\,$\clique$  & \firstc  \textbf{1.792}\,m    & \firstc \textbf{4.637}\,sec       &  \firstc  \textbf{3.9\,MB}     \\ \midrule
    \bottomrule
  \end{tabular}}
\end{subtable}
\hfill
\begin{subtable}[t]{0.48\textwidth}
  \caption{\texttt{Town 01-03}~\citep{jung2024helipr}}
  \label{tab:ablation2_helipr}
  \centering
  \resizebox{\textwidth}{!}{\tiny
  \begin{tabular}{l|ccc}
    \toprule
    Evaluation & $\rmsemetric$ & $\commmetric$ & $\exchmetric$ \\
    \midrule
      $\seq$                   & \hspace{0.15cm}  \secondc 2.988\,m           \hspace{0.15cm} 
                               & \hspace{0.15cm}  \thirdc 276.932\,sec        \hspace{0.15cm}              
                               & \hspace{0.15cm}   232.9\,MB                  \hspace{0.15cm}   \\
      +\,$\mvc$                & 3.464\,m                      & \secondc 140.190\,sec             &  \secondc 117.9\,MB      \\
      +\,$\clique$             & \firstc \textbf{2.887}\,m     & 283{,}637.921\,sec                   &  \thirdc 209.8\,MB                     \\
      +\,$\mvc$\,+\,$\clique$  & \thirdc 3.005\,m              &  \firstc \textbf{17.122}\,sec     &   \firstc \textbf{14.4\,MB}        \\ \midrule
    \bottomrule
  \end{tabular}}
\end{subtable}
\label{tab:ablation2}
\vspace{-0.3cm}
\end{table*}
% =============================================================

\subsection{Stage-wise Contribution of the Three-Stage Pipeline}
\label{sec:ablation_pipeline}
\noindent

\subsubsection{Sequential Matching}
\label{sec:ablation_seqmatch}
\noindent 
First, as shown in \figref{fig:seq_compare}, we compare our sequential matching stage with the SeqSLAM-based adaptive loop closure detection pipeline of AutoMerge~\citep{yin2023automerge}, which refines candidate windows via K-means clustering and RANSAC-based geometric verification.
While AutoMerge retains many low-value candidates and yields a dense affinity pattern~(\figref{fig:seq_compare}(a)), our method produces a substantially sparser and more structured affinity matrix~(\figref{fig:seq_compare}(b)), reducing exchange data from 934.1\,MB to 232.9\,MB and runtime from 1{,}943.714\,sec to 1.070\,sec, while also improving $\rmsemetric$ from 5.727\,m to 2.988\,m.

Therefore, our sequential matching stage selects fewer but more reliable loop closure candidates, jointly improving communication efficiency, computational cost, and downstream alignment accuracy.

\subsubsection{Balanced Vertex Cover and Maximum Edge-Weighted Clique Selection}
\label{sec:ablation_bvc_mec}
\begin{figure}[t]
    \centering
    \captionsetup{justification=justified}
    \captionsetup[subfigure]{justification=centering}
    \begin{subfigure}{0.525\columnwidth}
        \includegraphics[trim={0 0 0 0},clip, width=\columnwidth]{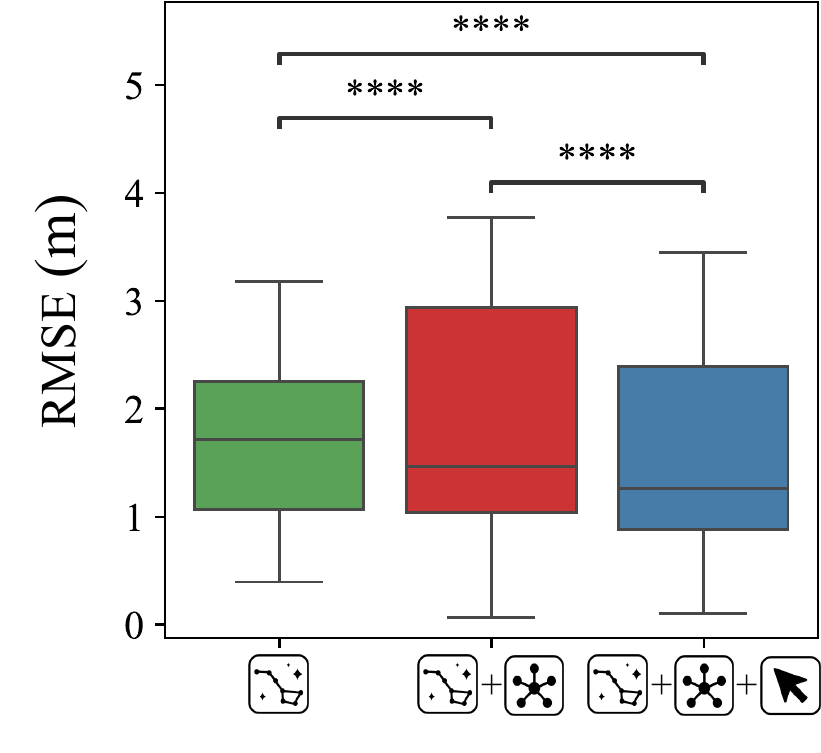}
        \vspace{-1.5em}
        \caption{}
    \end{subfigure}
    \begin{subfigure}{0.455\columnwidth}
        \includegraphics[trim={0 0 0 0},clip, width=\columnwidth]{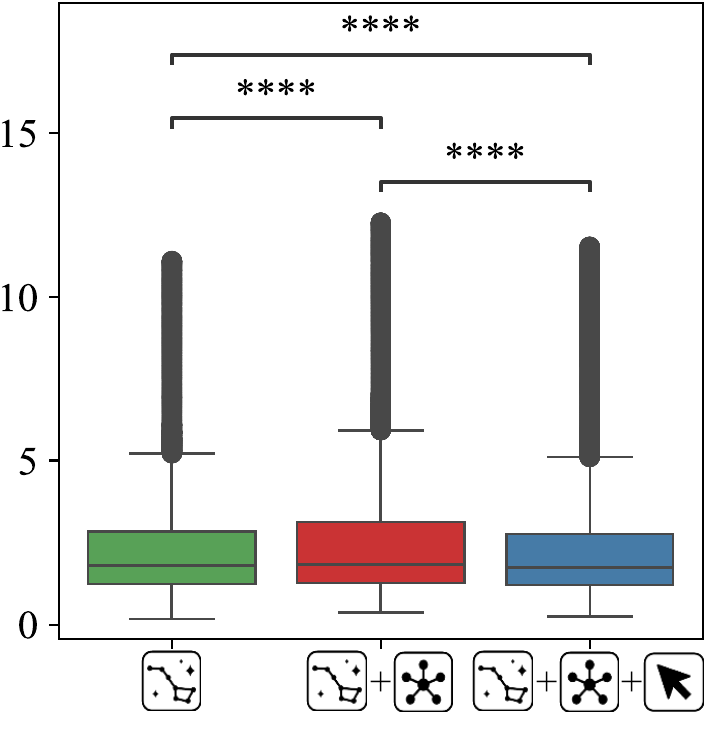}
        \vspace{-1.5em}
        \caption{}
    \end{subfigure}
    \includegraphics[trim={0 0 0 0},clip, width=0.98\columnwidth]{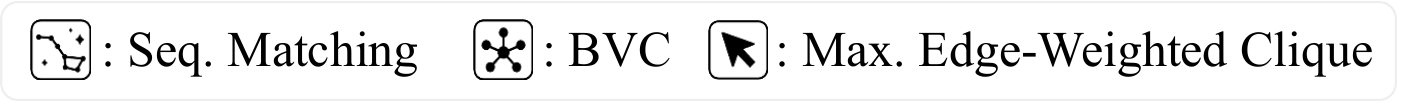}
    \caption{Box plot analysis of RMSE distribution across different pipeline configurations on (a) MCD \texttt{NTU 01-10} and (b) HeLiPR \texttt{Town 01-03} sequences. 
               $\seq$, $\mvc$, and $\clique$ represent the sequential matching, balanced minimum vertex cover, and maximum edge-weighted clique selection, respectively.
               The $*$$*$$*$$*$ annotations indicate measurements with a $p$-value $<$ 10$^{-4}$ after a paired $t$-test.}
    \label{fig:box_plot}
    \vspace{-3mm}
\end{figure}

\begin{figure*}[t]
    \centering
    \captionsetup{justification=justified}
    \captionsetup[subfigure]{justification=centering}
    \includegraphics[trim={0 0 0 0},clip, width=1.94\columnwidth]{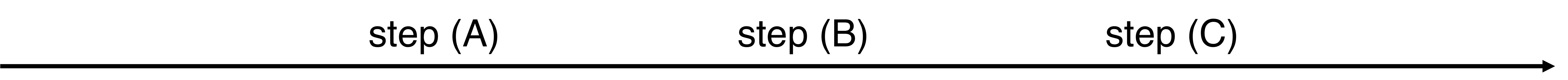}
    \begin{subfigure}{0.58\columnwidth}
        \includegraphics[trim={0 0 0 25},clip, width=\columnwidth]{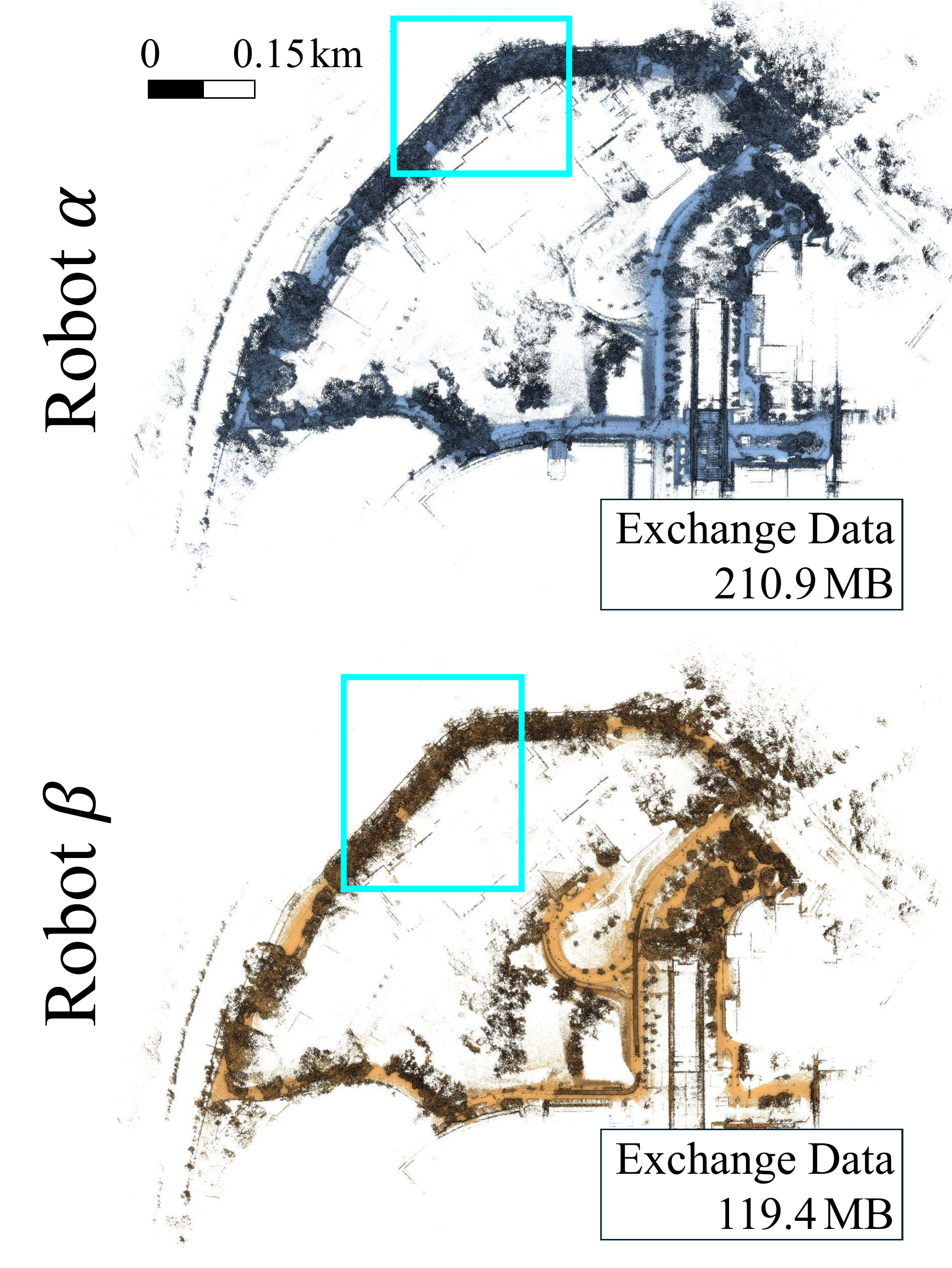}
        \vspace{-1.5em}
        \caption{}
    \end{subfigure}
    \begin{subfigure}{0.4375\columnwidth}
        \includegraphics[trim={0 0 0 0},clip, width=\columnwidth]{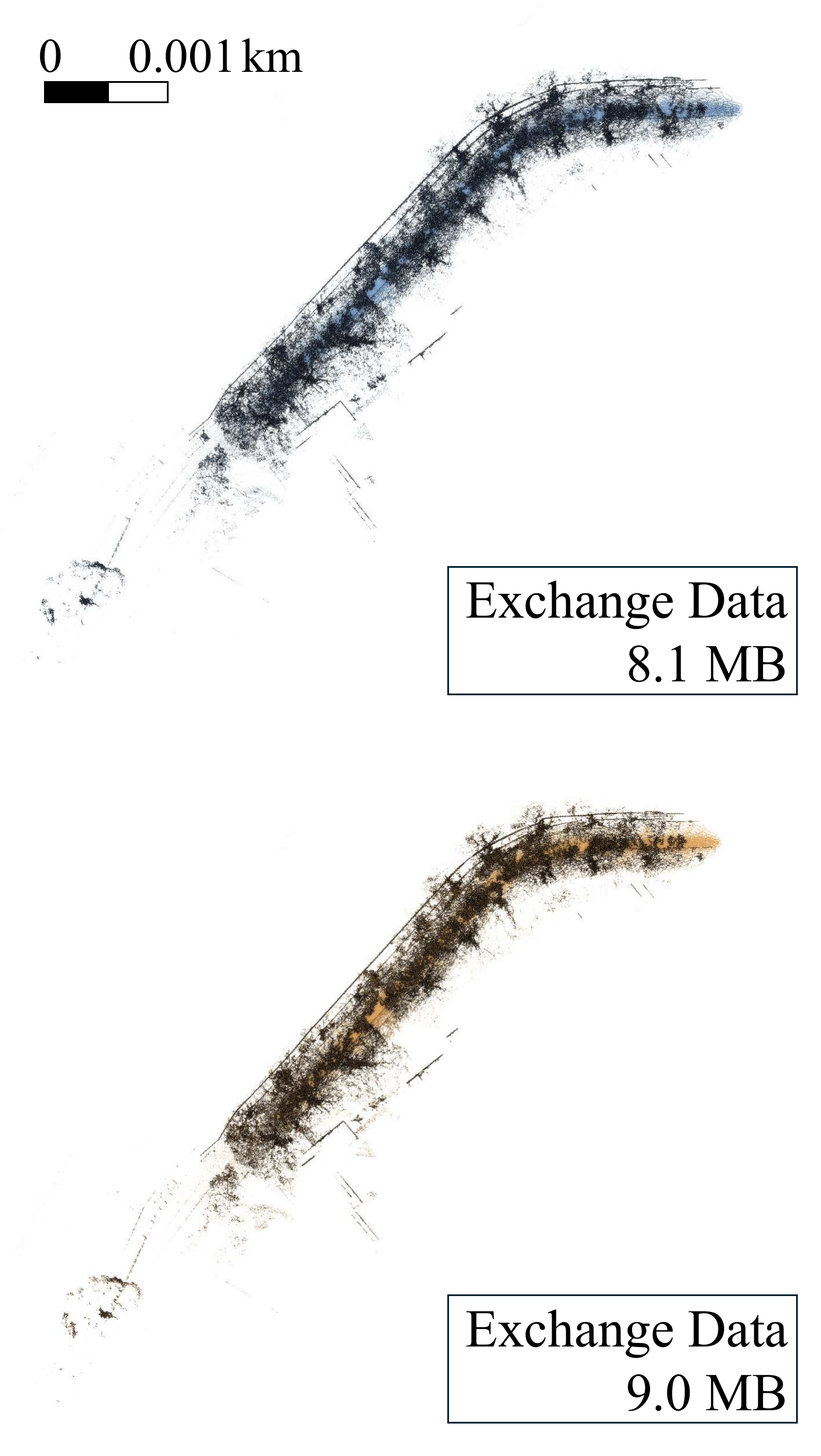}
        \vspace{-1.5em}
        \caption{}
    \end{subfigure}
    \begin{subfigure}{0.475\columnwidth}
        \includegraphics[trim={0 0 0 0},clip, width=\columnwidth]{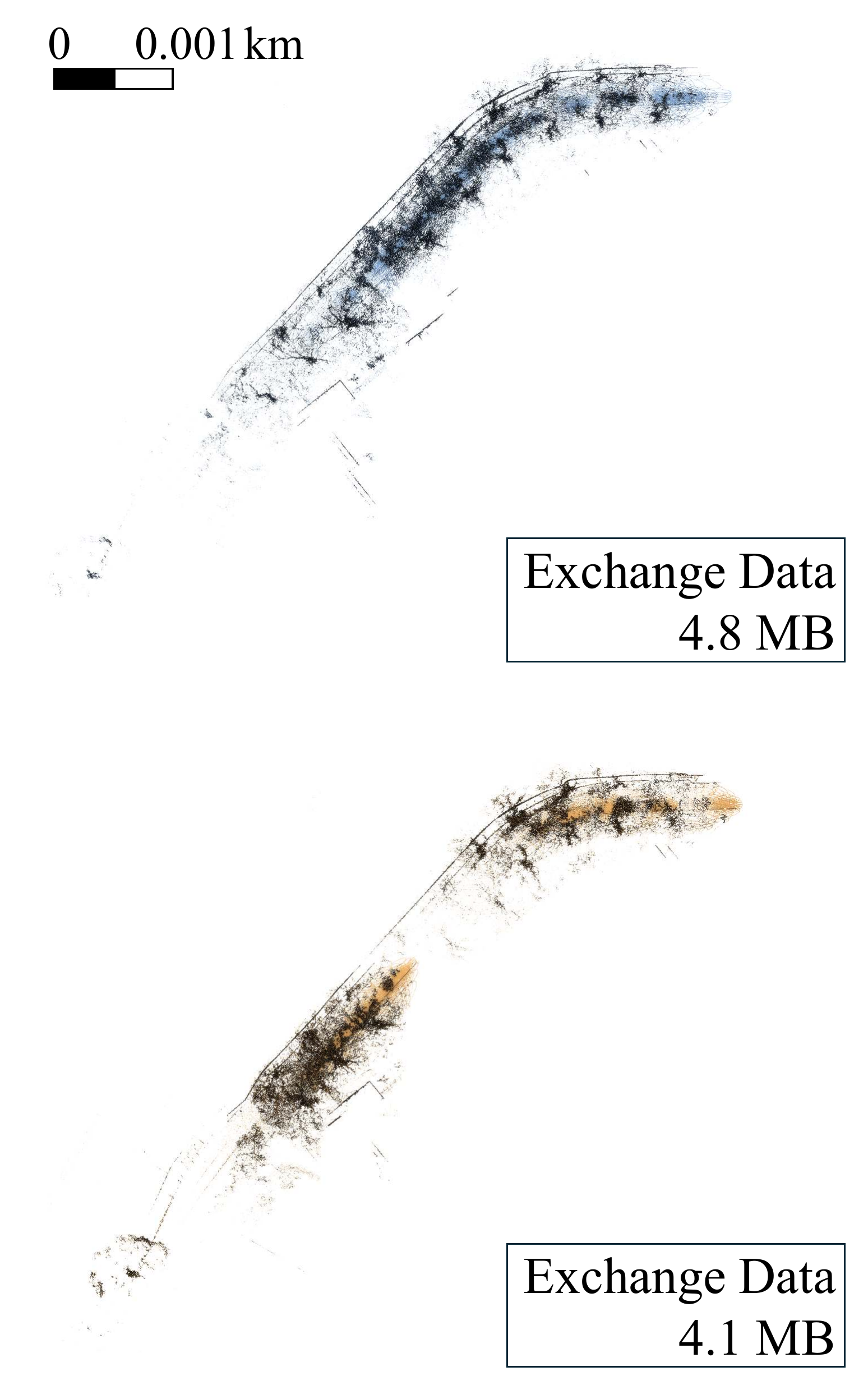}
        \vspace{-1.5em}
        \caption{}
    \end{subfigure}
    \begin{subfigure}{0.4425\columnwidth}
        \includegraphics[trim={0 0 0 0},clip, width=\columnwidth]{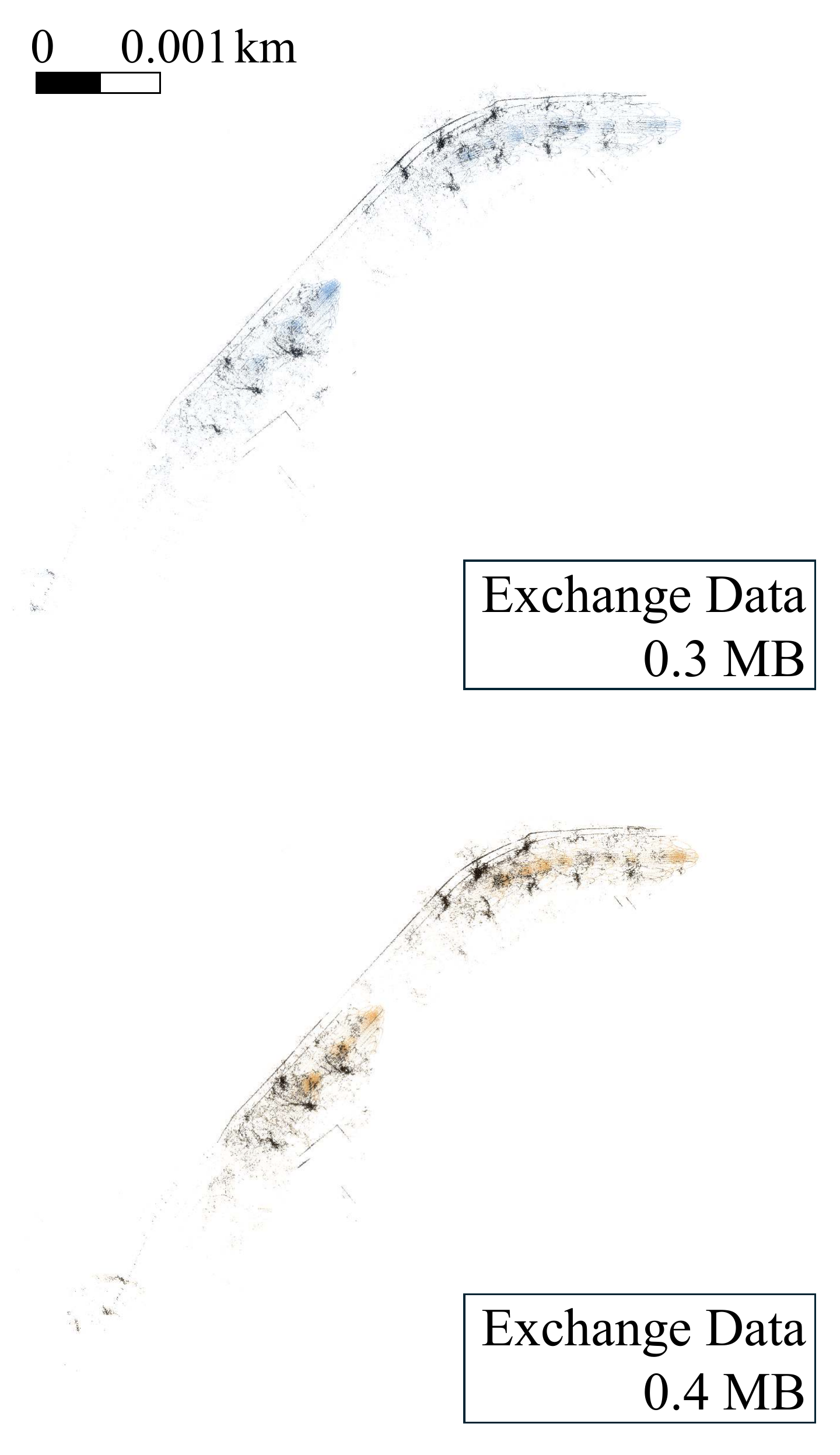}
        \vspace{-1.5em}
        \caption{}
    \end{subfigure}
    \caption{Submaps and associated exchange data sizes across modules in our framework. 
           (a) From intra-robot SLAM, each robot would require approximately 100\,MB--200\,MB of exchanged data. The cyan boxes highlight sequentially overlapping regions that are selected in $\stepA$.
           (b) After sequential matching, \ie $\stepA$, the exchange volume is reduced to 8.1\,MB for robot $\alpha$ and 9.0\,MB for robot $\beta$.
           (c) With balanced minimum vertex cover, \ie $\stepB$, the required data is further reduced to 4.8\,MB and 4.1\,MB, respectively.
           (d) Finally, the maximum edge-weighted clique selection, \ie $\stepC$ reduces the exchanged data to just 0.4\,MB per robot.
           This progressive reduction demonstrates the communication efficiency of our proposed map merging pipeline, achieving an overall 99.6\% reduction in transmission cost (from 100\,MB--200\,MB to 0.4\,MB), while still enabling successful submap merging.}
    \label{fig:mcd_ablation}
    \vspace{-0.5cm}
\end{figure*}

\noindent
Next, Table~\ref{tab:ablation2} and Figs~\ref{fig:box_plot} and \ref{fig:mcd_ablation} analyze the contribution of the remaining two stages, $\mvc$~(balanced vertex cover) and $\clique$~(maximum edge-weighted clique selection), on top of sequential matching~($\seq$).

As described in Section~\ref{sec:vertex_cover}, $\mvc$ regulates per-robot contribution via entropy-based balance constraints to avoid the monologue problem inherent in standard MVC.
As reported in Tables~\ref{tab:ablation2}(a) and~\ref{tab:ablation2}(b), the $\seq\,+\,\mvc$ configuration consistently reduces both $\commmetric$ and $\exchmetric$ compared to $\seq$ alone, confirming that $\mvc$ effectively regulates the per-robot transmission load.

Likewise, as described in Section~\ref{sec:edge_clique}, $\clique$ incorporates descriptor similarity as edge weights to select a more distinctive and reliable subset among geometrically valid candidates, instead of treating all compatible loops equally as in a standard maximum clique.
As reported in Tables~\ref{tab:ablation2}(a) and~\ref{tab:ablation2}(b), the $\seq\,+\,\clique$ configuration achieves the lowest $\rmsemetric$ and $\exchmetric$ but incurs prohibitive $\commmetric$~(\eg 283{,}637.921\,sec on \texttt{Town 01-03}) because the clique search over unfiltered candidates becomes combinatorially expensive.

Finally, the full pipeline $\seq\,+\,\mvc\,+\,\clique$ combines the complementary strengths of both stages: $\mvc$ regulates the candidate pool to keep the clique search tractable, while $\clique$ selects the most reliable subset from the regulated pool.
Consequently, the full pipeline attains the best $\commmetric$ and $\exchmetric$ across both datasets while maintaining $\rmsemetric$ comparable to the best individual configuration, as further illustrated by the progressive data reduction in \figref{fig:mcd_ablation} from 100--200\,MB~(\figref{fig:mcd_ablation}(a)) down to 0.4\,MB~(\figref{fig:mcd_ablation}(d)) per robot, corresponding to a 99.6\% reduction in transmission cost while preserving successful map merging.

In summary, $\mvc$ and $\clique$ play complementary roles in our pipeline, jointly enabling Commerge to achieve minimal communication cost, maximal transmission efficiency, and reliable map merging across both small- and large-scale environments.

% % =============================================================
% % =============================================================
% =============================================================
% =============================================================
% =============================================================
\begin{figure*}[t]
    \centering
    \captionsetup{justification=justified}
    \captionsetup[subfigure]{justification=centering}
    \begin{subfigure}{0.28\columnwidth}
        \includegraphics[trim={0 0 0 25},clip, width=\columnwidth]{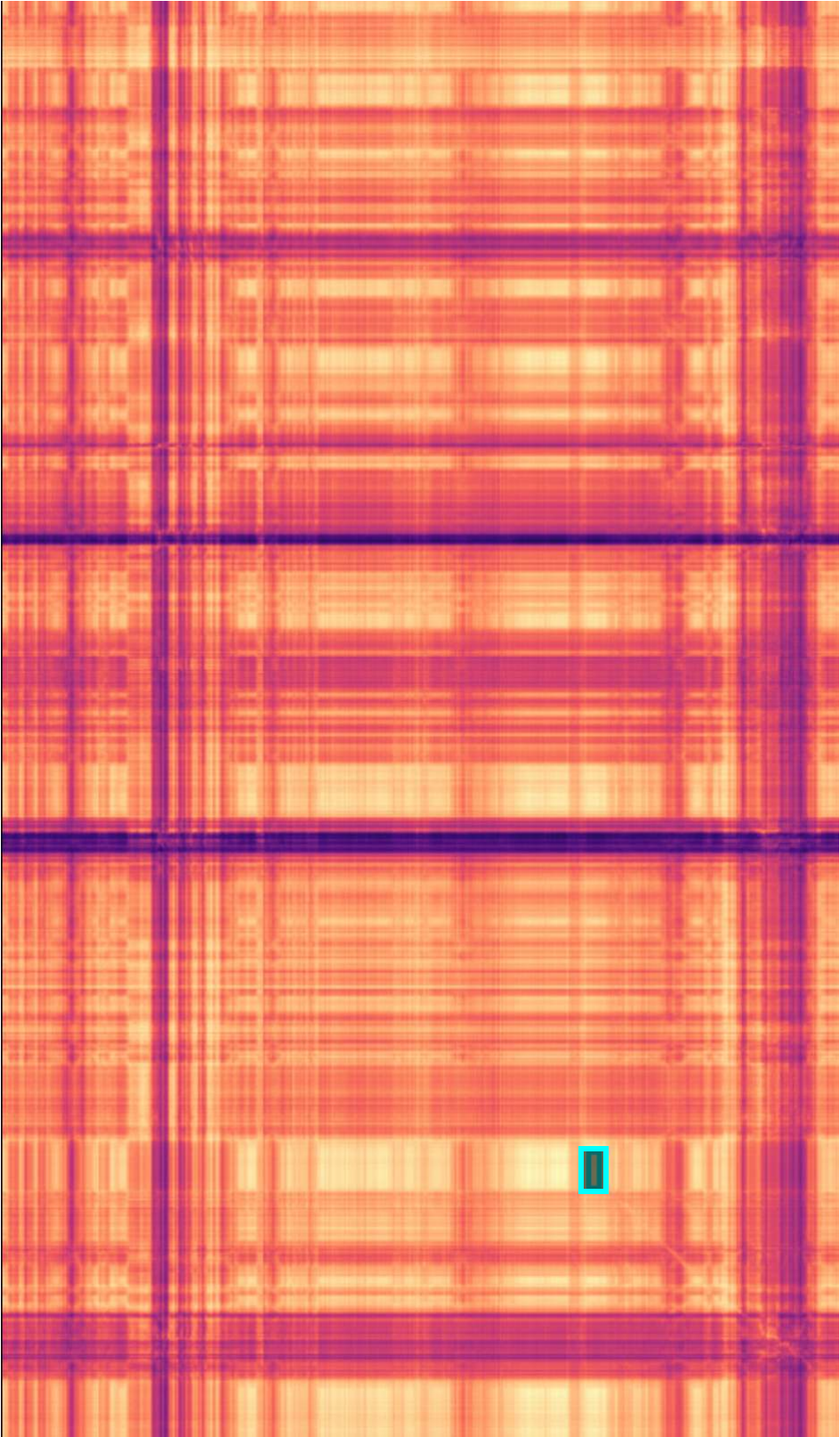}
        \vspace{-1.5em}
        \caption{}
    \end{subfigure}
    \begin{subfigure}{0.5375\columnwidth}
        \includegraphics[trim={0 0 0 0},clip, width=\columnwidth]{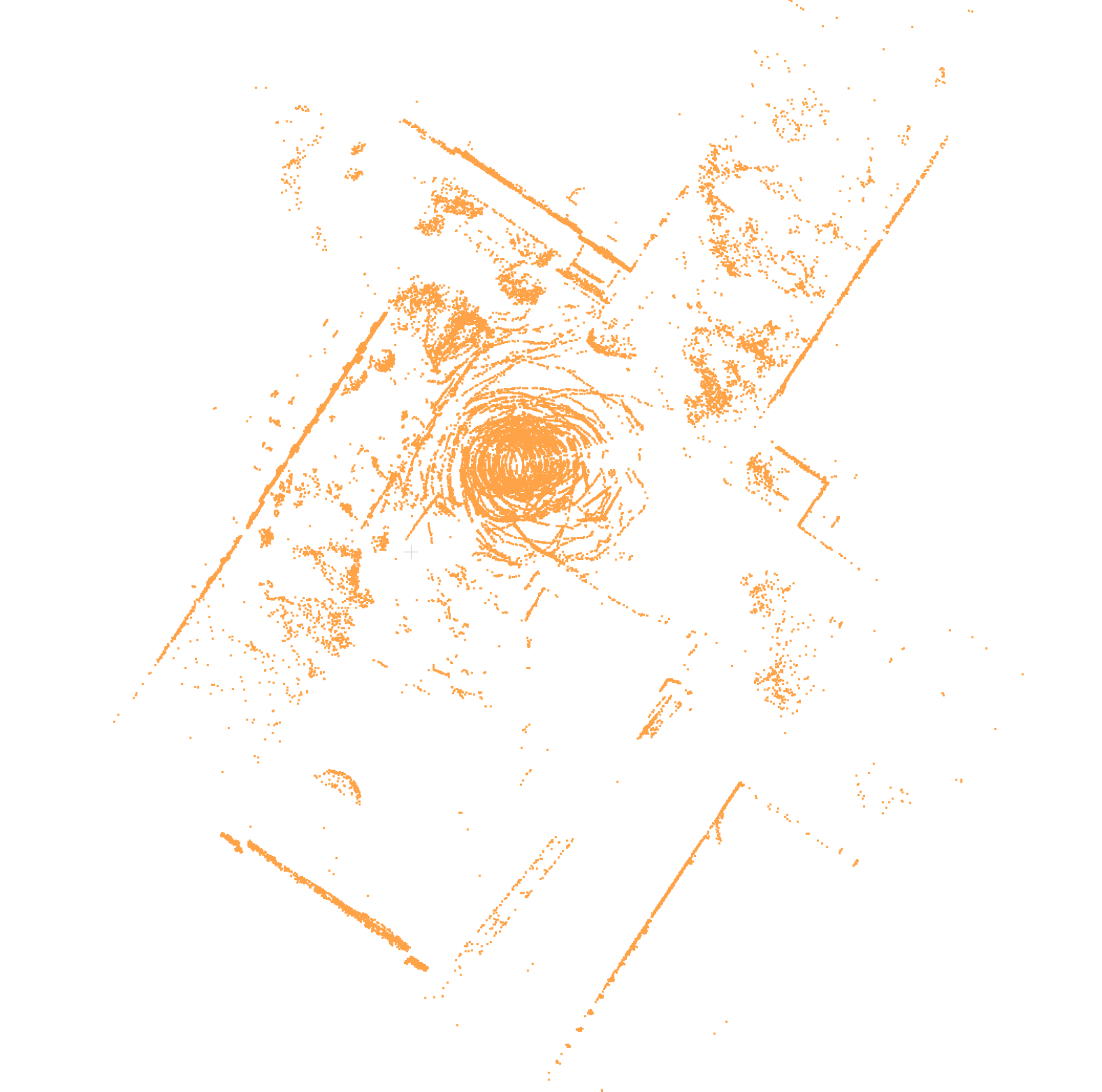}
        \vspace{-1.5em}
        \caption{}
    \end{subfigure}
    \begin{subfigure}{0.575\columnwidth}
        \includegraphics[trim={0 0 0 0},clip, width=\columnwidth]{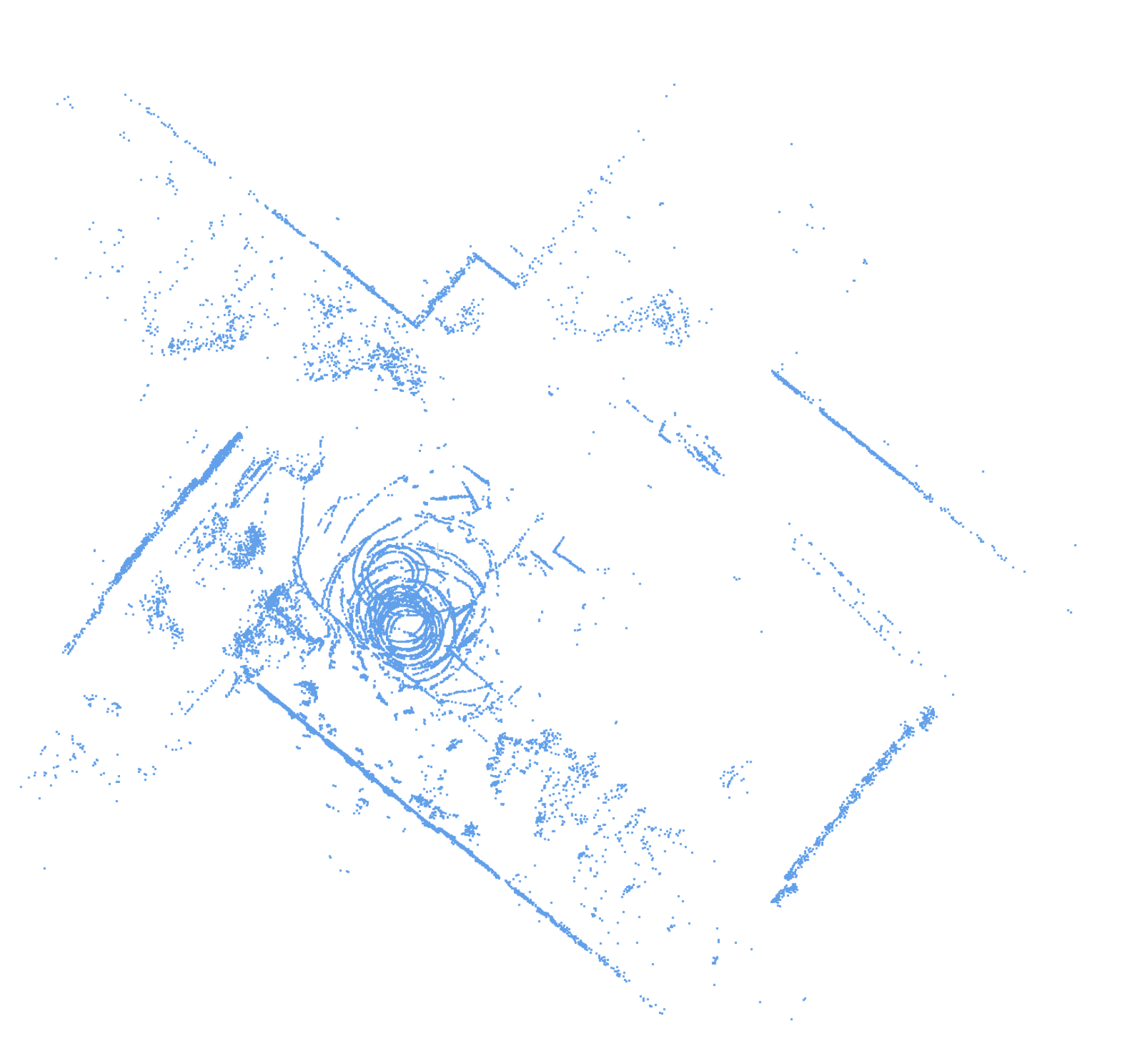}
        \vspace{-1.5em}
        \caption{}
    \end{subfigure}
    \begin{subfigure}{0.5425\columnwidth}
        \includegraphics[trim={0 0 0 0},clip, width=\columnwidth]{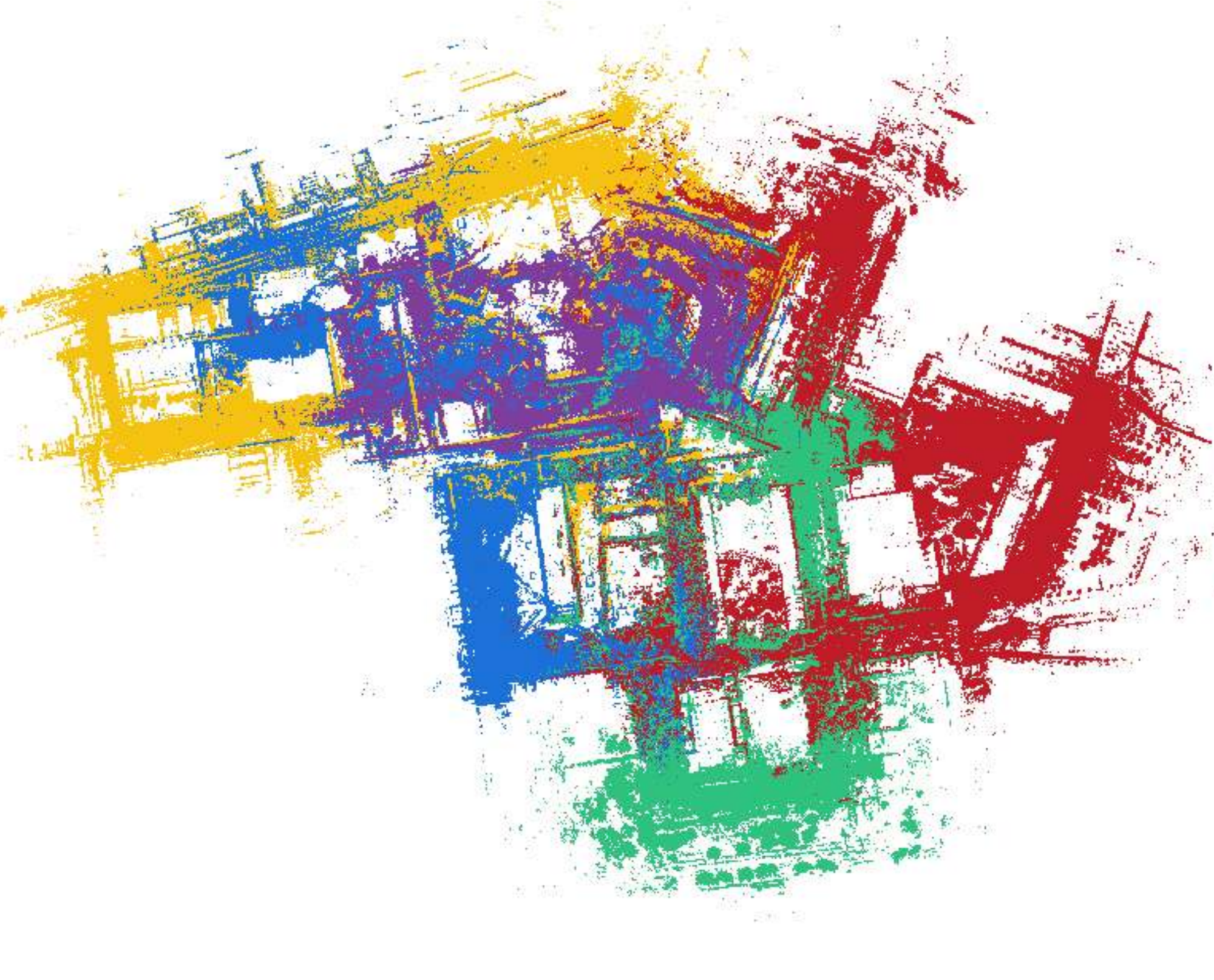}
        \vspace{-1.5em}
        \caption{}
    \end{subfigure}
    \caption{Validation of our three-stage optimization with Scan Context descriptors on Kimera-Multi-\texttt{Outdoor} partial overlap scenario to demonstrate the effectiveness of descriptor-agnosticism. 
           (a) Affinity matrix constructed from Scan Context similarities, where sequential matching identifies the optimal cluster (cyan box) representing sequentially coherent inter-robot correspondences despite the less discriminative descriptor. 
           (b)--(c) Carefully selected scans from robot $\alpha$ and robot $\beta$ extracted through our cascaded optimization, forming a compact submap covering critical overlap, respectively. 
           (d) Final merged map showing successful global alignment across all robots using the Scan Context descriptor.}
    \label{fig:sc_seq_match}
    \vspace{-0.3cm}
\end{figure*}
% =============================================================

\subsection{Robustness of the Commerge Framework}
\label{sec:ablation_robustness}
\noindent

\subsubsection{Descriptor-Agnostic Applicability}
\label{sec:ablation_descriptor}
\noindent
In addition, \figref{fig:sc_seq_match} demonstrates that our Commerge remains applicable even when SOLiD is replaced with a weaker descriptor such as Scan Context, which causes LT-Mapper and ELite to fail on the Kimera-Multi-\texttt{Outdoor} sequence due to partial trajectory overlap~($\smalloverlap$).
From the Scan Context affinity matrix with the sequentially matched cluster highlighted in \figref{fig:sc_seq_match}(a), our pipeline extracts scans from robot~$\alpha$~(\figref{fig:sc_seq_match}(b)) and robot~$\beta$~(\figref{fig:sc_seq_match}(c)), and merges the selected submaps into the globally consistent map shown in \figref{fig:sc_seq_match}(d).

In other words, the three-stage optimization of Commerge provides robustness independent of the specific descriptor choice, establishing the descriptor-agnostic applicability of our framework.

\subsubsection{Reliability across Mission Duration}
\label{sec:ablation_temporal}
\begin{figure*}[t]
    \centering
    \captionsetup{justification=justified}
    \captionsetup[subfigure]{justification=centering}
    \begin{subfigure}{0.975\columnwidth}
        \includegraphics[trim={0 0 0 0},clip, width=\columnwidth]{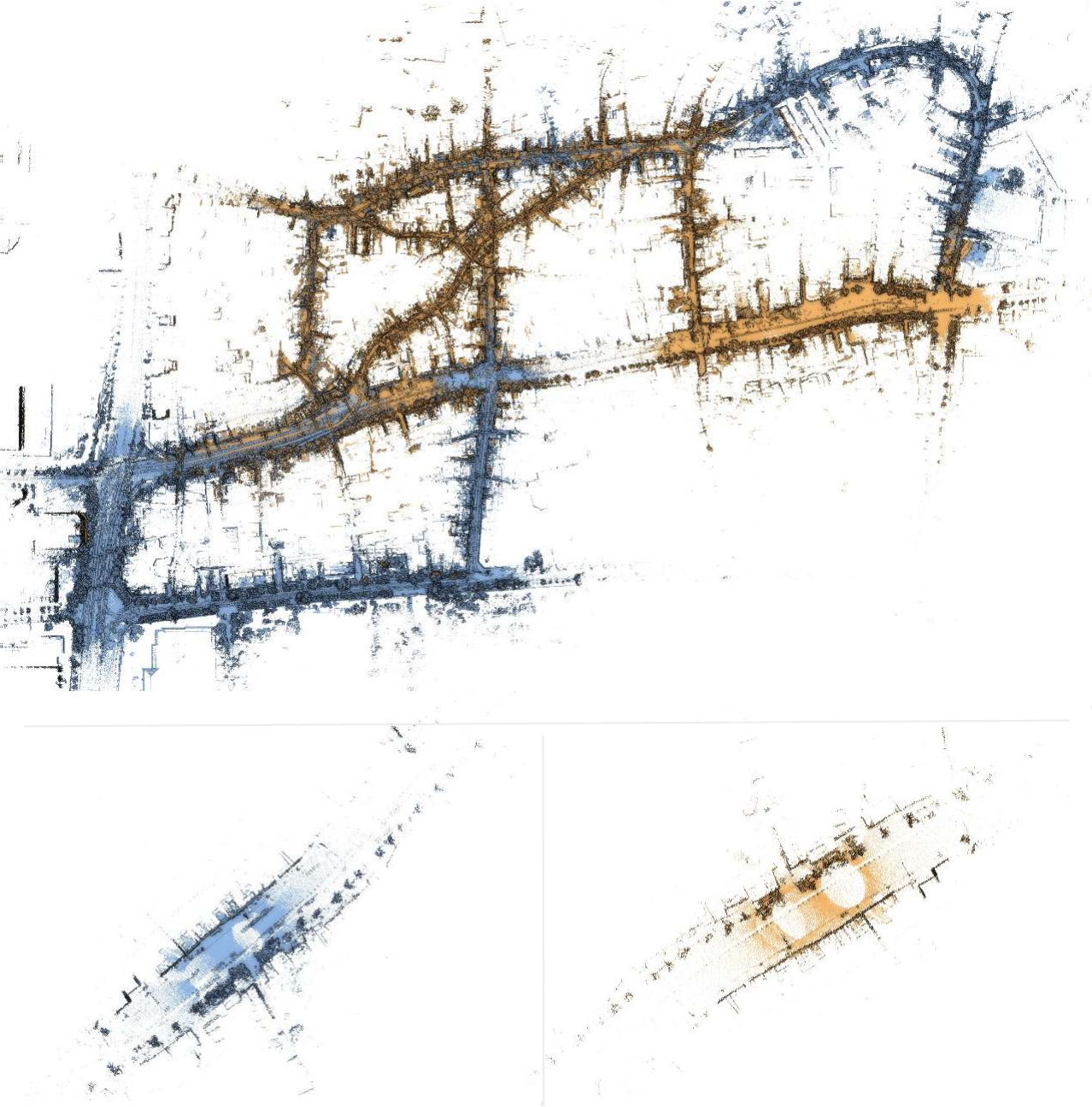}
        \vspace{-1.5em}
        \caption{}
    \end{subfigure}
    \begin{subfigure}{0.985\columnwidth}
        \includegraphics[trim={0 0 0 0},clip, width=\columnwidth]{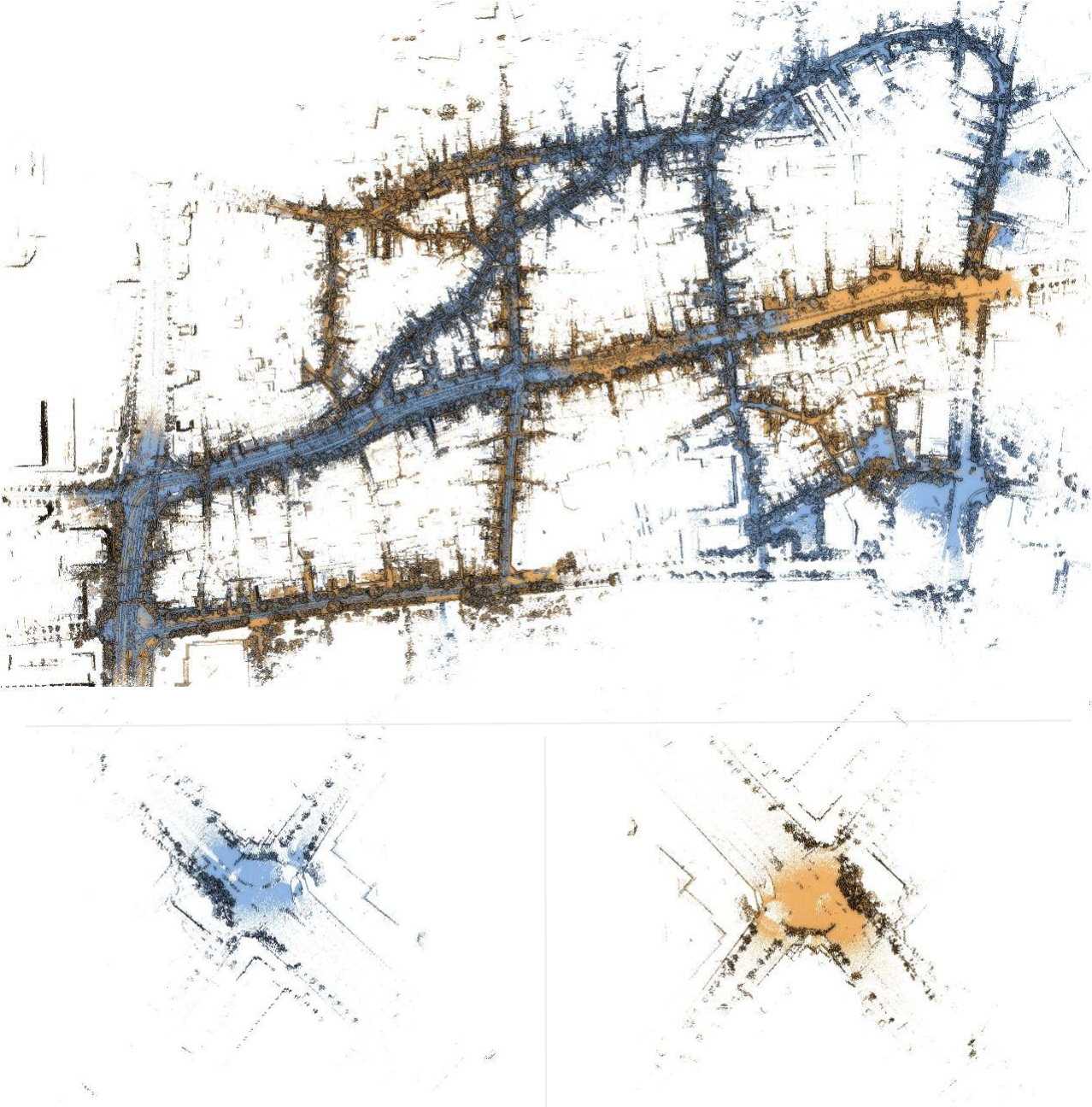}
        \vspace{-1.5em}
        \caption{}
    \end{subfigure}

    \begin{subfigure}{0.98\columnwidth}
        \includegraphics[trim={0 0 0 0},clip, width=\columnwidth]{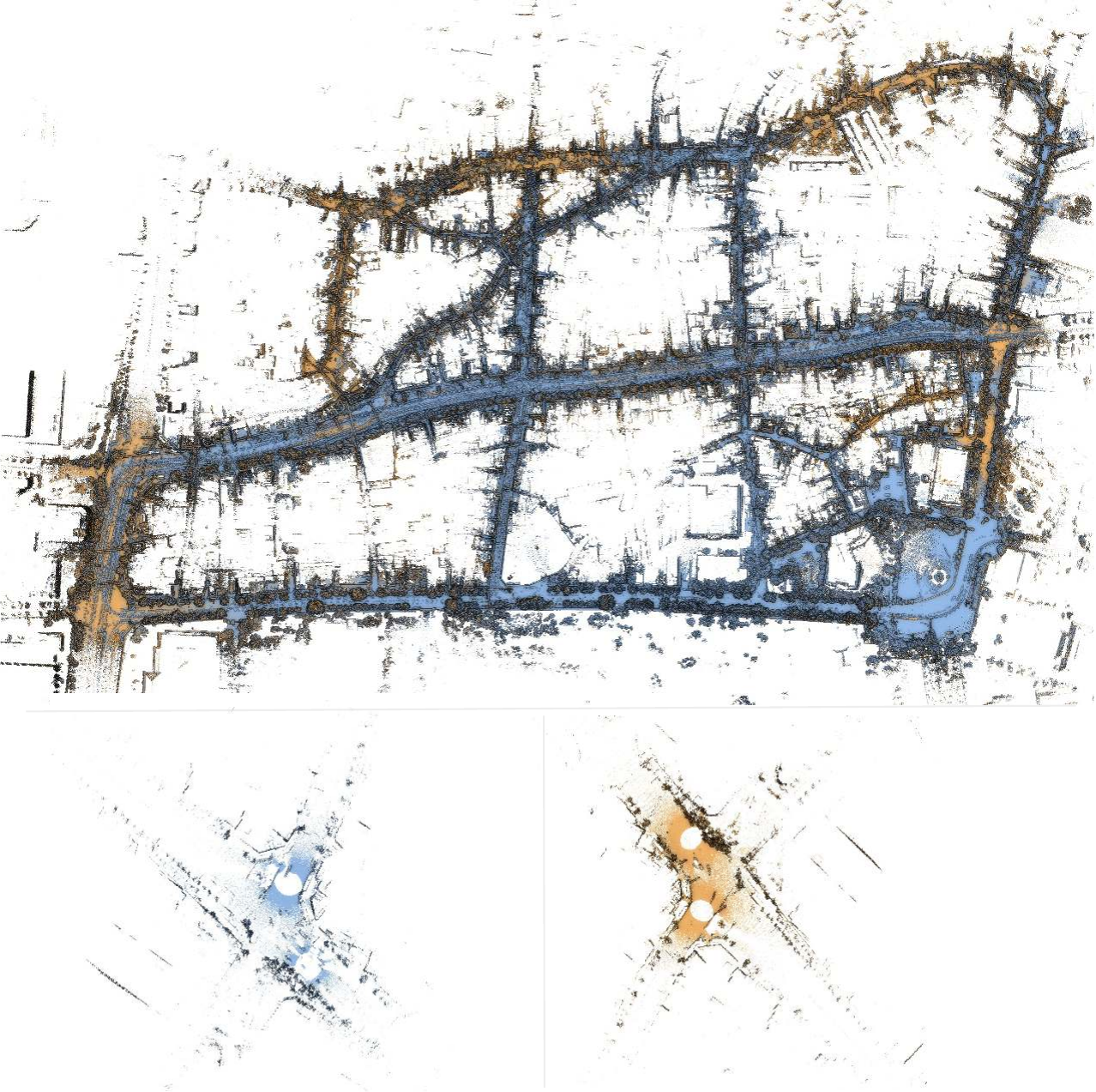}
        \vspace{-1.5em}
        \caption{}
    \end{subfigure}
    \begin{subfigure}{0.98\columnwidth}
        \includegraphics[trim={0 0 0 0},clip, width=\columnwidth]{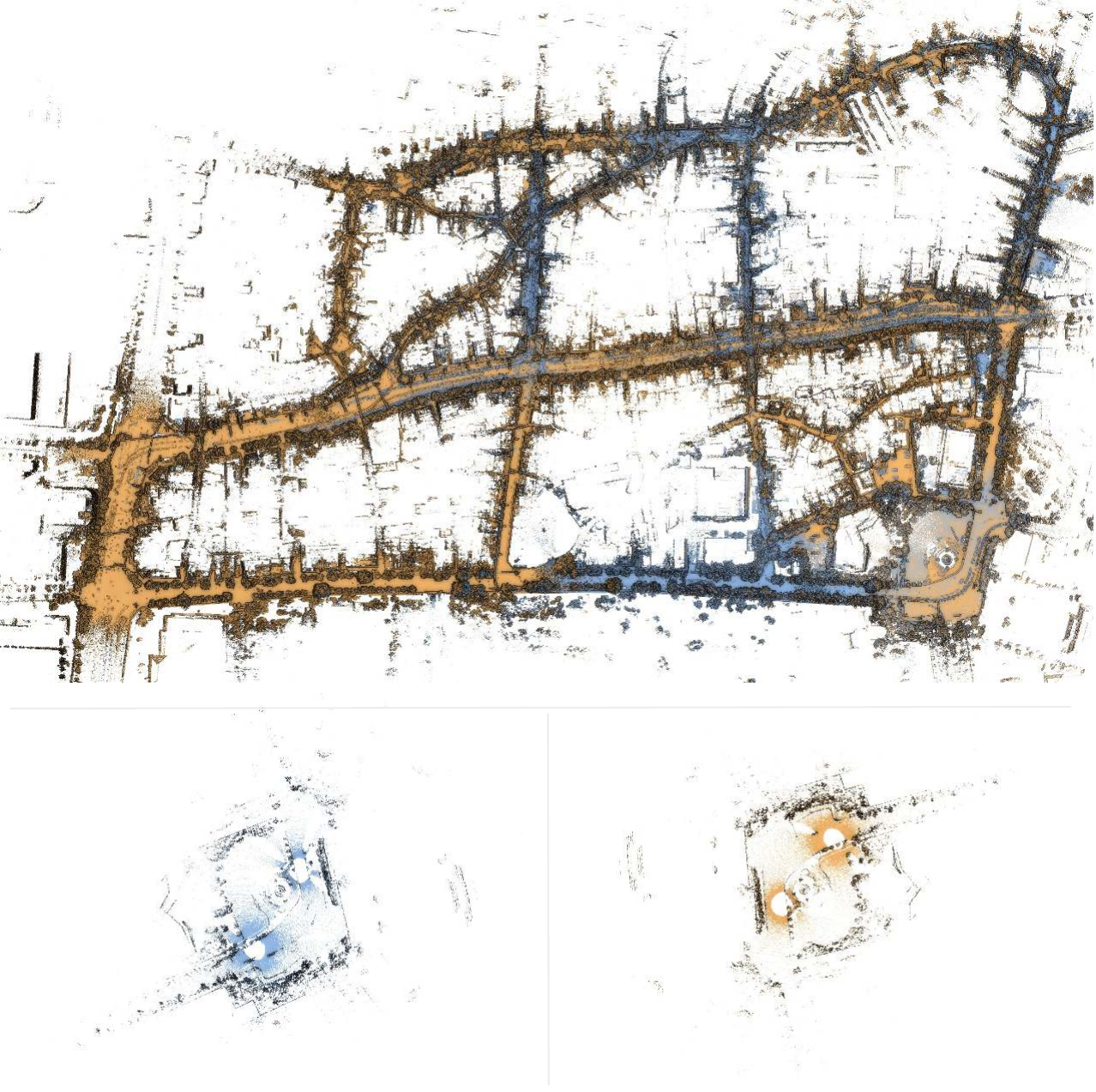}
        \vspace{-1.5em}
        \caption{}
    \end{subfigure}
    \includegraphics[trim={0 0 0 0},clip, width=0.7\columnwidth]{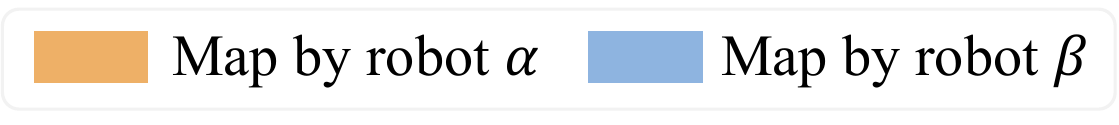}
     \vspace{-3mm}
    \caption{Temporal evolution of map merging quality on HeLiPR \texttt{Town 01-03} sequence at different mission stages. 
           Each scenario shows the merged global map constructed at the server (top) and the selected submaps transmitted from the robots to the server under the proposed exchange policy (bottom).
           (a) Both robots after covering the first quarter of the mission (10 min): successful alignment despite neither robot having established intra-robot loop closures yet.
           (b) Both robots, after completing roughly one-third of the mission (15 min), improved alignment as one robot began to accumulate intra-robot loop closures.
           (c) Robot $\alpha$ at roughly one third progress and Robot $\beta$ already fully mapped: an asymmetric situation similar to multi-session SLAM, where a fully optimized reference map is merged with an ongoing exploration session. 
           (d) Both robots have completed the mission with fully converged local maps: optimal alignment with the highest geometric consistency. 
           % Even scenario (a) without loop closures achieves robust inter-robot alignment through sequential matching.
           }  
     \label{fig:incremental}
     \vspace{-4mm}
\end{figure*}

\noindent
Furthermore, Table~\ref{tab:ablation_temporal} and \figref{fig:incremental} evaluate our method across four mission-duration scenarios~($\scenarioA$ to $\scenarioD$), ranging from early stages exploration to fully completed missions.
As the mission progresses, the growing intra-robot positional consistency enables our method to select increasingly coherent partial maps, progressively reducing $\rmsemetric$ while keeping $\commmetric$ and $\exchmetric$ bounded regardless of mission length.

Overall, Commerge remains reliable across short-, medium-, and long-term operations, achieving robust alignment even without intra-robot loop closures in early stages and maintaining stable communication overhead as missions extend.

% =============================================================
\begin{table}[t]
\captionsetup{width=.49\textwidth, justification=justified}
\caption{Temporal evolution of map merging performance on HeLiPR \texttt{Town 01-03} sequence across four mission scenarios. $\scenarioA$, $\scenarioB$, $\scenarioC$, and $\scenarioD$ indicate both robots at the first quarter (10\,min) without intra-robot loops, both at one-third (15\,min) with emerging loops, one robot at one-third and the other complete, and both missions complete with fully optimized local maps, respectively.}
\centering\resizebox{0.49\textwidth}{!}{\tiny
\begin{tabular}{l|l|ccc}
\toprule
\midrule
                  & Evaluation               & $\rmsemetric$         & $\commmetric$                   & $\exchmetric$          \\ \midrule 
\multirow{4}{*}{\rotatebox[origin=c]{90}{Method}} 
                  & $\scenarioA$             & 3.867\,m              & \firstc \textbf{13.555}\,sec    & \hspace{0.1cm} \firstc \textbf{11.4}\,MB \hspace{0.1cm}                           \\
                  & $\scenarioB$             & \thirdc 3.653\,m      & 25.803\,sec                     & \hspace{0.1cm} 21.7\,MB  \hspace{0.1cm}       \\
                  & $\scenarioC$             & \secondc 3.129\,m     & \thirdc 20.451\,sec             & \hspace{0.1cm} \thirdc 17.2\,MB  \hspace{0.1cm}      \\
                  & $\scenarioD$             & \hspace{0.1cm} \firstc \textbf{3.005}\,m          \hspace{0.1cm}
                                             & \hspace{0.1cm} \secondc 17.122\,sec       \hspace{0.1cm}        
                                             & \hspace{0.1cm} \secondc 14.4\,MB  \hspace{0.1cm}                          \\ \midrule
                                             \bottomrule
\end{tabular}}
\label{tab:ablation_temporal}
% \vspace{-3mm}
\end{table}
% =============================================================

% =============================================================
% =============================================================

% % =============================================================
\begin{figure*}[t]
    \centering
    \captionsetup{justification=justified}
    \captionsetup[subfigure]{justification=centering}
    \begin{subfigure}{0.975\columnwidth}
        \includegraphics[trim={0 20 0 0},clip, width=\columnwidth]{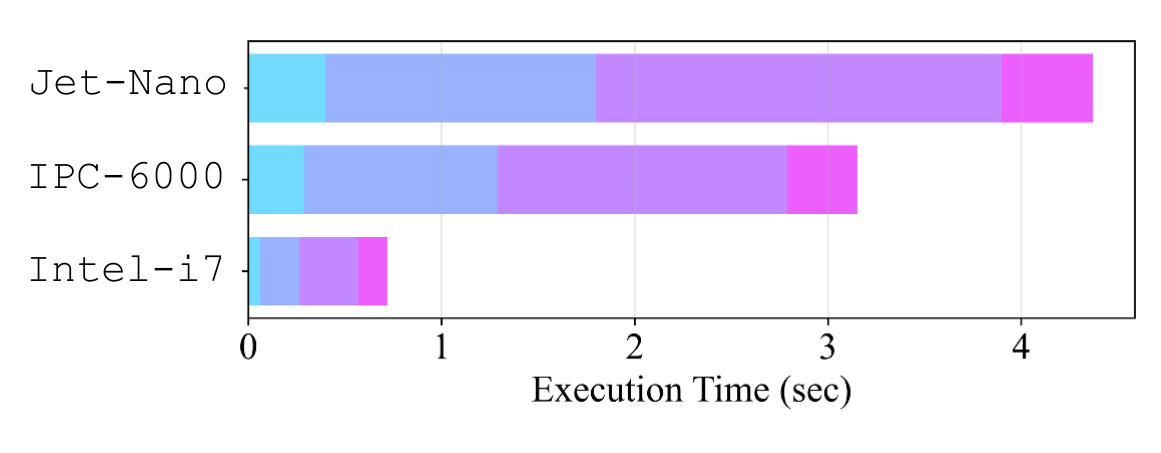}
        \vspace{-1.5em}
        \caption{}
    \end{subfigure}
    \begin{subfigure}{0.945\columnwidth}
        \includegraphics[trim={0 0 0 0},clip, width=\columnwidth]{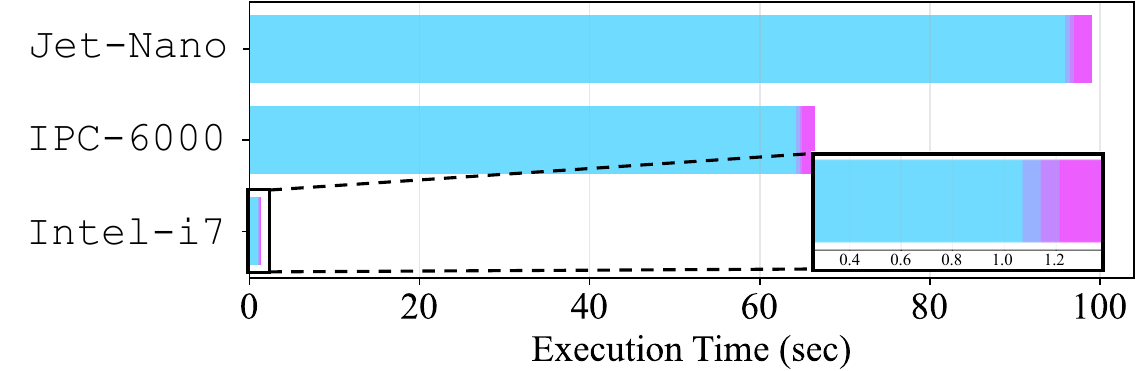}
        \vspace{-1.5em}
        \caption{}
    \end{subfigure}
    \vspace{-0.15cm}
    \includegraphics[trim={0 0 0 0},clip, width=1.8\columnwidth]{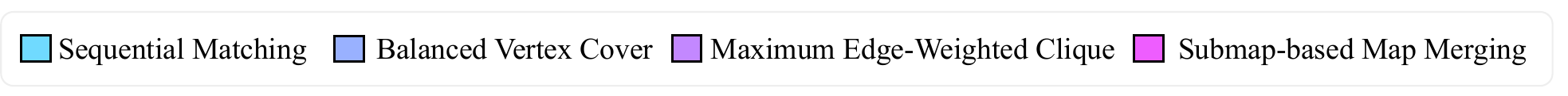}
  \caption{Execution time breakdown across hardware platforms ($\hardA$: Intel i7, $\hardB$: IPC-6000, $\hardC$: Jetson Nano) and operational scales: (a) small-scale \texttt{NTU 01-10} and (b) large-scale \texttt{Town 01-03} sequences.}
    \label{fig:execute_time}
    \vspace{-3mm}
\end{figure*}
% % =============================================================

\subsection{Module-Level Validation}
\label{sec:ablation_modules}
\noindent

\subsubsection{Place Recognition Performance}
\label{sec:ablation_pr}
\noindent
Additionally, as shown in Table~\ref{tab:place_recognition}, SOLiD achieves consistently strong performance relative to its descriptor size.
While RING++, OSK, and CVTNet also exhibit competitive performance, their descriptor sizes remain substantially larger than that of SOLiD, which occupies only 448 bytes.

Therefore, SOLiD offers the most favorable trade-off between descriptor compactness and recognition accuracy, justifying its selection as the place recognition module of our Commerge.

% =============================================================
\begin{table}[t]
\captionsetup{width=0.49\textwidth, justification=justified}
\caption{Place recognition performance comparison on MCD dataset~\citep{nguyen2024mcd} with non-repetitive scan pattern LiDAR sensor (Livox Mid-70). 
         Descriptors are evaluated on two sequences~(\texttt{NTU 01-02} and \texttt{NTU 01-10}).}
\centering\resizebox{0.49\textwidth}{!}{\tiny
\begin{tabular}{l|l|c|c|c|c}
\toprule \midrule
               &   Method                    & Size [B] $\downarrow$ & Recall@1 $\uparrow$    & AUC score $\uparrow$   & F1 max $\uparrow$  \\  \midrule
                    \multirow{9}{*}{\rotatebox[origin=c]{90}{\texttt{NTU 01-02}}} 
               &   Histogram                 & \secondc 928          & 0.422                  & 0.330                  & 0.376  \\
               &   M2DP                      & 1,664                 & 0.338                  & 0.136                  & 0.384  \\
               &   Scan Context              & 10,016                & 0.175                  & 0.017                  & 0.113  \\
               &   RING++                    & 345,728               & \firstc \textbf{0.787} & \thirdc 0.454          & 0.417  \\
               &   OSK                  & 1,600            & \thirdc 0.557     & 0.404             & \secondc 0.522  \\
               &   OverlapTransformer   & \thirdc 1,152    & 0.206             & 0.067             & 0.112  \\
               &   LoGG3D-Net           & 2,176            & 0.508             & 0.387             & 0.480  \\
               &   CVTNet               & 3,200            & 0.528             & \secondc 0.481    & \thirdc 0.486  \\
               &   SOLiD                     & \firstc \textbf{448}  & \secondc 0.587         & \firstc \textbf{0.587} & \firstc \textbf{0.608}  \\  \midrule
                   \multirow{9}{*}{\rotatebox[origin=c]{90}{\texttt{NTU 01-10}}} 
               &   Histogram                 & \secondc 928          & 0.299                  & 0.449                  & 0.502  \\
               &   M2DP                      & 1,664                 & 0.199                  & 0.214                  & 0.377  \\
               &   Scan Context              & 10,016                & 0.092                  & 0.026                  & 0.169  \\
               &   RING++                    & 345,728               & \firstc \textbf{0.471} & \firstc \textbf{0.801} & \firstc \textbf{0.712}  \\
               &   OSK                  & 1,600            & 0.347             & \thirdc 0.540     & \thirdc 0.605  \\
               &   OverlapTransformer   & \thirdc 1,152    & 0.206             & 0.067             & 0.112  \\
               &   LoGG3D-Net           & 2,176            & 0.288             & 0.475             & 0.512  \\
               &   CVTNet               & 3,200            & \secondc 0.354    & 0.479             & 0.504  \\
               &   SOLiD                     & \firstc \textbf{448}  & \thirdc 0.351          & \secondc 0.740         & \secondc 0.689  \\ 
\midrule \bottomrule
\end{tabular}}
\label{tab:place_recognition}
\vspace{-0.3cm}
\end{table}
% =============================================================
% =============================================================
\begin{table}[h]
\captionsetup{width=0.48\textwidth, justification=justified}
\caption{Scan-level registration performance evaluation on Kimera-Multi-\texttt{Tunnel}.
         STD fails to provide reliable correspondences (N/A) due to the lack of distinctive features and repetitive indoor structures.}
\centering\resizebox{0.48\textwidth}{!}{\tiny
\begin{tabular}{l|l|c|c|c}
\toprule \midrule
               &   Method                    & RTE [m] $\downarrow$      & RRE [\textdegree] $\downarrow$        & Success Rate [$\%$] $\uparrow$   \\  \midrule
                    \multirow{9}{*}{\rotatebox[origin=c]{90}{\texttt{A01-A02}}} 
               &   STD                       & N/A                        & N/A                                      & N/A                              \\
               &   RING++               & 0.461                 & 1.737                               & \thirdc 84.639              \\
               &   BEVPlace++           & 0.948                 & 1.431                               & 80.120                      \\
               &   G-ICP                     & \secondc 0.186             & \secondc 1.166                           & 43.173                           \\
               &   PLANE-ICP                 &  0.189                     & 1.429                                    & 38.253                           \\
               &   Predator             & \secondc 0.186        & 1.766                               & 76.205                      \\
               &   KISS-Matcher              & 0.266                      & 2.660                                    & 47.590                           \\
               &   KISS-Matcher + G-ICP      & \firstc\textbf{0.181}      & \thirdc 1.405                            & \secondc 88.755                  \\
               &   KISS-Matcher + PLANE-ICP  & 0.190                      & \firstc\textbf{1.128}                    & \firstc\textbf{89.056}           \\  \midrule
                    \multirow{9}{*}{\rotatebox[origin=c]{90}{\texttt{A01-AP01}}} 
               &   STD                       & N/A                        & N/A                                      & N/A                              \\
               &   RING++               & 0.554                 & 1.939                               & \firstc \textbf{74.301}     \\
               &   BEVPlace++           & 0.992                 & 1.732                               & 58.635                      \\
               &   G-ICP                     & \firstc\textbf{0.223}      & \firstc\textbf{1.183}                    & \thirdc 32.490                   \\
               &   PLANE-ICP                 & \secondc 0.233             & \thirdc 1.406                            & 30.493                           \\
               &   Predator             & 0.342                 & 2.506                               & 33.023                      \\
               &   KISS-Matcher              & 0.404                      & 2.899                                    & 18.375                           \\
               &   KISS-Matcher + G-ICP      & 0.377                      & 1.528                                    & \thirdc 65.113                   \\
               &   KISS-Matcher + PLANE-ICP  & \thirdc 0.313              & \secondc 1.284                           & \secondc 65.379                  \\  \midrule
                    \multirow{9}{*}{\rotatebox[origin=c]{90}{\texttt{A01-H01}}}  
               &   STD                       & N/A                        & N/A                                      & N/A                              \\
               &   RING++               & 0.668                 & 1.710                               & 73.360                      \\
               &   BEVPlace++           & 1.006                 & \thirdc 1.261                       & 70.738                      \\
               &   G-ICP                     & \firstc\textbf{0.160}      & \firstc\textbf{0.927}                    & 41.098                           \\
               &   PLANE-ICP                 & \secondc 0.193             & 1.414                                    & 31.593                           \\
               &   Predator             & \thirdc 0.203          & 1.662                               & \thirdc 73.628              \\
               &   KISS-Matcher              & 0.287                      & 2.720                                    & 45.382                           \\
               &   KISS-Matcher + G-ICP      & 0.220                      & 1.595                                    & \secondc 82.062                  \\
               &   KISS-Matcher + PLANE-ICP  & 0.209                      & \secondc 1.146                           & \firstc\textbf{84.605}           \\  \midrule
                    \multirow{9}{*}{\rotatebox[origin=c]{90}{\texttt{A01-S02}}} 
               &   STD                       & N/A                        & N/A                                      & N/A                              \\
               &   RING++               & 0.606                 & 1.914                               & \thirdc 63.600              \\
               &   BEVPlace++           & 1.027                 & \thirdc 1.516                       & 58.501                      \\
               &   G-ICP                     & \firstc\textbf{0.186}      & \firstc\textbf{1.256}                    & 32.800                           \\
               &   PLANE-ICP                 & \secondc 0.241             & 1.546                                    & 25.067                           \\
               &   Predator             & 0.286                 & 2.081                               & 55.333                      \\
               &   KISS-Matcher              & 0.463                      & 2.913                                    & 23.867                           \\
               &   KISS-Matcher + G-ICP      & \thirdc 0.268              & 1.840                                    & \secondc 67.867                  \\
               &   KISS-Matcher + PLANE-ICP  & 0.305                      & \secondc 1.355                           & \firstc\textbf{69.733}           \\
\midrule \bottomrule
\end{tabular}}
\label{tab:indoor_experiments}
\vspace{-3mm}
\end{table}
% =============================================================

\subsubsection{Registration Performance}
\label{sec:ablation_reg}
\noindent
Finally, as shown in Table~\ref{tab:indoor_experiments}, KISS-Matcher alone does not achieve the most robust scan-level registration performance.
However, owing to its fast runtime, KISS-Matcher can be complemented with G-ICP or PLANE-ICP in a coarse-to-fine manner while still maintaining comparable or faster runtime than the baselines.
Specifically, KISS-Matcher combined with PLANE-ICP attains the highest success rate on three out of four sequences, confirming the effectiveness of the coarse-to-fine strategy.

Furthermore, in our map-level registration setting, this runtime advantage becomes even more pronounced, and the robustness gain is further amplified, justifying the choice of KISS-Matcher as the registration module of our Commerge.

\section{Runtime Analysis}
\label{sec:hardware_ablation}
\noindent
Finally, we investigate the runtime of our \textit{Commerge} to demonstrate its computational efficiency across different hardware platforms: resource-constrained embedded system~($\hardC$), industrial mini PC ($\hardB$), and high-performance workstation ($\hardA$).

On small-scale sequences, our Commerge demonstrates real-time feasibility across all hardware platforms (\figref{fig:execute_time}(a)).
In contrast, large-scale HeLiPR sequences exhibit exponential time increases with decreasing hardware capability, indicating that real-time execution remains challenging on $\hardB$ and $\hardC$ platforms (\figref{fig:execute_time}(b)). 

It is noteworthy that despite extended execution times on resource-constrained platforms, all hardware tiers successfully complete map merging without OOM or T/O, demonstrating deployability even on embedded systems.

\section{Conclusion}
\noindent
This paper presents \emph{Commerge}, a communication-efficient multi-robot LiDAR map merging framework that achieves bandwidth reduction through graph-theoretic selective data exchange.
The key insight is that only a small subset of carefully selected scans is sufficient for robust map merging.
Building on this principle, our method transmits MB-scale data instead of complete GB-scale datasets while preserving alignment accuracy in large-scale and long-term missions.
Extensive evaluation on five public datasets demonstrates up to 99.98\,\% reduction in data exchange with competitive alignment performance across embedded to desktop platforms.
Furthermore, evaluations on our in-house datasets, covering subterranean cave, planetary-analog, and campus indoor/outdoor environments under both NetEm-emulated network degradation and real wireless infrastructure with non-line-of-sight conditions, confirm the practical deployability of Commerge in realistic multi-robot missions.

Despite these encouraging results, there is further space for improvement. Although our method operates across diverse hardware platforms, execution time on resource-constrained systems remains a practical limitation for large-scale scenarios.
Future work will focus on optimizing the sequential matching stage, which dominates computational cost, to enable real-time map merging on embedded platforms.
In addition, we plan to extend Commerge toward a fully decentralized system that removes the reliance on a central server in communication-restricted environments.
\section{Funding}
\noindent
This work was supported by National Research Foundation of Korea (NRF) grant (No. RS-2026-2555148 and RS-2025-02217000) and Institute of Information \& communications Technology Planning \& Evaluation (IITP) grant (RS-2022-II220448) funded by the Korea government(MSIT).

\end{document}